\documentclass[10pt]{article} 
\usepackage[github]{tmlr}       

\usepackage{amsmath,amsfonts,bm}

\def\eqref#1{equation~\ref{#1}}

\def\1{\bm{1}}

\DeclareMathAlphabet{\mathsfit}{\encodingdefault}{\sfdefault}{m}{sl}
\SetMathAlphabet{\mathsfit}{bold}{\encodingdefault}{\sfdefault}{bx}{n}

\usepackage{xcolor}
\usepackage[table]{xcolor}
\definecolor{citecolor}{HTML}{0071bc}
\usepackage{hyperref}
\makeatletter
\if@github
\hypersetup{
    colorlinks=true,
    linkcolor=red,
    citecolor=citecolor,
}
\fi
\makeatother

\usepackage{url}
\usepackage{booktabs}
\usepackage{tabularx}
\usepackage{enumitem}
\usepackage{graphicx}
\usepackage[dvipsnames]{xcolor}
\usepackage{colortbl}
\usepackage{multirow}
\usepackage{pifont}
\newcommand{\cmark}{{\color{green!60!black}\ding{52}}}
\newcommand{\xmark}{{\color{red}\ding{55}}}
\newcommand{\paperlink}[1]{\faBookOpen\,\href{#1}{\textcolor{magenta}{Paper}}}
\newcommand{\githublink}[1]{\faGithub\,\href{#1}{\textcolor{magenta}{Code}}}
\usepackage{hyperref}
\usepackage{tikz}
\usetikzlibrary{patterns,decorations.pathreplacing,fit,positioning,backgrounds}
\usepackage[edges]{forest}
\definecolor{darkpastelgreen}{rgb}{0.01, 0.75, 0.24}
\definecolor{BlueGreen}{rgb}{0.0, 0.64, 0.66}
\definecolor{Periwinkle}{rgb}{0.6, 0.6, 0.93}
\definecolor{Orchid}{rgb}{0.85, 0.44, 0.84}
\definecolor{Goldenrod}{rgb}{0.85, 0.65, 0.13}
\definecolor{Melon}{rgb}{0.99, 0.74, 0.71}
\definecolor{CornflowerBlue}{rgb}{0.39, 0.58, 0.93}
\usepackage{fontawesome6}
\usepackage[most]{tcolorbox}
\usepackage{mdframed}
\usepackage{epigraph}
\newtcolorbox{keypoint}[1][]{
  colback=blue!4,
  colframe=blue!35,
  colbacktitle=blue!35,
  coltitle=black,
  fonttitle=\bfseries,
  title={#1},
  boxrule=3pt,
  arc=5pt,
  drop shadow,
  parbox=false,
  breakable,
  left=5pt,
  right=15pt,
  top=5pt,
  bottom=5pt,
  before skip=8pt,
  after skip=8pt
}

\newtcolorbox{definition}[1][]{
  colback=green!4,
  colframe=green!35,
  colbacktitle=green!35,
  coltitle=black,
  fonttitle=\bfseries,
  title={#1},
  boxrule=3pt,
  arc=5pt,
  drop shadow,
  parbox=false,
  breakable,
  left=5pt,
  right=15pt,
  top=5pt,
  bottom=5pt,
  before skip=8pt,
  after skip=8pt
}

\newtcolorbox{takeaway}[1][]{
  colback=orange!4,
  colframe=orange!35,
  colbacktitle=orange!35,
  coltitle=black,
  fonttitle=\bfseries,
  title={#1},
  boxrule=3pt,
  arc=5pt,
  drop shadow,
  parbox=false,
  breakable,
  left=5pt,
  right=15pt,
  top=5pt,
  bottom=5pt,
  before skip=8pt,
  after skip=8pt
}

\title{Agentic World Modeling: Foundations, Capabilities, Laws, and Beyond}

\author{Meng~Chu$^{1\dagger}$,
Xuan~Billy~Zhang$^{2\dagger\text{\faCube}}$,
Kevin~Qinghong~Lin$^{3\dagger}$,
Lingdong~Kong$^{2\dagger}$,\\
Jize~Zhang$^{3\dagger}$,
Teng~Tu$^{2\dagger}$,
Weijian~Ma$^{2\dagger}$,
Ziqi~Huang$^{4}$,
Senqiao~Yang$^{5}$,
Wei~Huang$^{6}$,\\
Yeying~Jin$^{2}$,
Zhefan~Rao$^{1}$,
Jinhui~Ye$^{1}$,
Xinyu~Lin$^{2}$,
Xichen~Zhang$^{1}$,
Qisheng~Hu$^{4}$,\\
Shuai~Yang$^{1}$,
Leyang~Shen$^{2}$,
Wei Chow$^{2}$,
Yifei Dong$^{7}$,
Fengyi Wu$^{7}$,
Quanyu~Long$^{4}$,\\
Bin~Xia$^{5}$,
Shaozuo~Yu$^{5}$,
Mingkang~Zhu$^{5}$,
Wenhu~Zhang$^{1}$,
Jiehui~Huang$^{1}$,
Haokun~Gui$^{1}$,
Runyi~Li$^{8}$,
Chenyu~Tang$^{9}$,
Dong~Huang$^{2}$,
Xuhang~Chen$^{9}$,
Rui~Liu$^{5}$,
Chengzu~Li$^{9}$,\\
Shiyi~Du$^{10}$,
Xu~Huang$^{11}$,
Haoxuan~Che$^{1\S}$,
Long~Chen$^{1\S}$,
Qifeng~Chen$^{1\S}$,
Wenya~Wang$^{4\S}$,\\
Wenxuan~Zhang$^{12\S}$,
Xiaojuan~Qi$^{6\S}$,
Yang~Deng$^{13\S}$,
Yanwei~Li$^{5\S}$,
Mike~Zheng~Shou$^{2\S}$,\\
Zhi-Qi~Cheng$^{7\S}$,
See-Kiong~Ng$^{2\S}$,
Ziwei~Liu$^{4\S}$,
Philip~Torr$^{3\S}$,
Jiaya~Jia$^{1\S}$
\\[0.2em]
{\normalfont $^\dagger$Core Contributor.\;\;\faCube\;Project Lead.\;\;$^\S$Senior Author.}
\\[0.2em]
{\normalfont\texttt{$^{1}$Hong Kong University of Science and Technology
\\
$^{2}$National University of Singapore~$^{3}$University of Oxford
\\
$^{4}$Nanyang Technological University ~$^{5}$Chinese University of Hong Kong
\\
$^{6}$University of Hong Kong ~$^{7}$University of Washington
\\
$^{8}$University of Tokyo ~$^{9}$University of Cambridge
\\
$^{10}$Carnegie Mellon University 
~$^{11}$University of California, Berkeley
\\
~$^{12}$Singapore University of Technology and Design $^{13}$Singapore Management University}}}

\def\month{April}
\def\year{2026}
\def\openreview{\textit{OpenReview link to be added upon acceptance}}

\begin{document}

\maketitle
\makeatletter
\gdef\@github@title{\raisebox{-0.15em}{\includegraphics[height=1.1em]{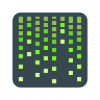}}\;\,Agentic World Modeling: \\ Foundations, Capabilities, Laws, and Beyond}
\makeatother

\makeatletter
\renewcommand{\aftertitskip}{0.8em}
\makeatother

\begin{abstract}

As AI systems move from generating text to accomplishing goals through sustained interaction, the ability to model environment dynamics becomes a central bottleneck. Agents that manipulate objects, navigate software, coordinate with others, or design experiments require predictive environment models, yet the term \emph{world model} carries different meanings across research communities. We introduce a ``levels $\times$ laws'' taxonomy organized along two axes. The first defines three capability levels: \textbf{L1 Predictor}, which learns one-step local transition operators; \textbf{L2 Simulator}, which composes them into multi-step, action-conditioned rollouts that respect domain laws; and \textbf{L3 Evolver}, which autonomously revises its own model when predictions fail against new evidence. The second identifies four governing-law regimes (physical, digital, social, and scientific) that determine what constraints a world model must satisfy and where it is most likely to fail. Using this framework, we synthesize over 400 works and summarize more than 100 representative systems spanning model-based reinforcement learning, video generation, web and GUI agents, multi-agent social simulation, and AI-driven scientific discovery. We analyze methods, failure modes, and evaluation practices across level--regime pairs, propose decision-centric evaluation principles and a minimal reproducible evaluation package, and outline architectural guidance, open problems, and governance challenges. The resulting roadmap connects previously isolated communities and charts a path from passive next-step prediction toward world models that can simulate, and ultimately reshape, the environments in which agents operate.

\end{abstract}

\begin{figure*}[tp]
\centering
\vspace{-36pt}
\resizebox{0.86\textwidth}{!}{\tikzset{
    my node/.style={
        draw,
        align=left,
        thin,
        text width=2.8cm,
        rounded corners=3,
    },
    my leaf/.style={
        draw,
        align=left,
        thin,
        text width=4cm,
        rounded corners=3,
    }
}

\forestset{
  every leaf node/.style={
    if n children=0{#1}{}
  },
  every tree node/.style={
    if n children=0{minimum width=1em}{#1}
  },
}
\begin{forest}
    for tree={%
        every leaf node={my leaf, font=\small\sffamily},
        every tree node={my node, font=\small\sffamily, l sep-=4.5pt, l-=1.pt},
        anchor=west,
        inner sep=2pt,
        l sep=10pt,
        s sep=4pt,
        fit=tight,
        grow'=east,
        edge={thick},
        parent anchor=east,
        child anchor=west,
        if n children=0{tier=last}{},
        edge path={
            \noexpand\path [draw, \forestoption{edge}] (!u.parent anchor) -- +(5pt,0) |- (.child anchor)\forestoption{edge label};
        },
        if={isodd(n_children())}{
            for children={
                if={equal(n,(n_children("!u")+1)/2)}{calign with current}{}
            }
        }{}
    }
    [{Agentic World Modeling}, draw=gray, color=gray!100, fill=gray!15, very thick, text=black, text width=2.2cm,
        [\S\ref{sec:introduction} Introduction, color=Cyan!100, fill=Cyan!15, very thick, text=black, text width=3.5cm
            [\S\ref{sec:motivation} Motivation, color=Cyan!100, fill=Cyan!15, very thick, text=black, text width=5cm]
            [\S\ref{sec:scope} Scope \& Organizing Principle, color=Cyan!100, fill=Cyan!15, very thick, text=black, text width=5cm]
            [\S\ref{sec:contributions} Contributions \& Positioning, color=Cyan!100, fill=Cyan!15, very thick, text=black, text width=5cm]
        ]
        [\S\ref{sec:preliminaries} Preliminaries, color=BlueGreen!100, fill=BlueGreen!15, very thick, text=black, text width=3.5cm
            [\S\ref{sec:philosophy} Epistemology to Capability Hierarchy, color=BlueGreen!100, fill=BlueGreen!15, very thick, text=black, text width=6cm]
            [\S\ref{sec:trends:representation} Representation in World Modeling, color=BlueGreen!100, fill=BlueGreen!15, very thick, text=black, text width=6cm]
            [\S\ref{subsec:notation_foundations} Notations, color=BlueGreen!100, fill=BlueGreen!15, very thick, text=black, text width=6cm]
            [\S\ref{subsec:formal_defs} Definitions of Capabilities, color=BlueGreen!100, fill=BlueGreen!15, very thick, text=black, text width=6cm]
            [\S\ref{subsec:scope_regimes} Scope of Laws, color=BlueGreen!100, fill=BlueGreen!15, very thick, text=black, text width=6cm]
        ]
        [\S\ref{sec:l1} L1 Predictor, color=Periwinkle!100, fill=Periwinkle!15, very thick, text=black, text width=3.5cm
            [\S\ref{subsec:l1_definition} Definition, color=Periwinkle!100, fill=Periwinkle!15, very thick, text=black, text width=4cm]
            [\S\ref{subsec:l1_methods} Approaches, color=Periwinkle!100, fill=Periwinkle!15, very thick, text=black, text width=4cm
                [State Inference, color=Periwinkle!100, fill=Periwinkle!15, very thick, text=black, tier=L1, text width=3.5cm]
                [Forward Dynamics, color=Periwinkle!100, fill=Periwinkle!15, very thick, text=black, tier=L1, text width=3.5cm]
                [Observation Decoding, color=Periwinkle!100, fill=Periwinkle!15, very thick, text=black, tier=L1, text width=3.5cm]
                [Inverse Dynamics, color=Periwinkle!100, fill=Periwinkle!15, very thick, text=black, tier=L1, text width=3.5cm]
            ]
            [\S\ref{subsec:l1_theory_boundaries} Discussion, color=Periwinkle!100, fill=Periwinkle!15, very thick, text=black, text width=4cm]
        ]
        [\S\ref{sec:l2} L2 Simulator, color=darkpastelgreen!100, fill=darkpastelgreen!15, very thick, text=black, text width=3.5cm
            [\S\ref{subsec:l2_requirements} Requirements for Elevation, color=darkpastelgreen!100, fill=darkpastelgreen!15, very thick, text=black, text width=4.5cm]
            [\S\ref{subsec:l2_app} Applications, color=darkpastelgreen!100, fill=darkpastelgreen!15, very thick, text=black, text width=3cm
                [\S\ref{subsec:l2_physical} Physical World, color=darkpastelgreen!100, fill=darkpastelgreen!15, very thick, text=black, tier=L2, text width=4.5cm]
                [\S\ref{subsec:l2_software} Digital World, color=darkpastelgreen!100, fill=darkpastelgreen!15, very thick, text=black, tier=L2, text width=4.5cm]
                [\S\ref{subsec:l2_social} Social World, color=darkpastelgreen!100, fill=darkpastelgreen!15, very thick, text=black, tier=L2, text width=4.5cm]
                [\S\ref{subsec:l2_science} Scientific World, color=darkpastelgreen!100, fill=darkpastelgreen!15, very thick, text=black, tier=L2, text width=4.5cm]
                [\S\ref{subsec:l2_crossdomain} Cross-Domain Analysis, color=darkpastelgreen!100, fill=darkpastelgreen!15, very thick, text=black, tier=L2, text width=4.5cm]
            ]
            [\S\ref{subsec:l2_failure_modes} Failure Modes, color=darkpastelgreen!100, fill=darkpastelgreen!15, very thick, text=black, text width=4.5cm]
        ]
        [\S\ref{sec:l3} L3 Evolver, color=Orchid!100, fill=Orchid!15, very thick, text=black, text width=3.5cm
            [\S\ref{subsec:l3_definition} Formal Definition, color=Orchid!100, fill=Orchid!15, very thick, text=black, text width=5cm]
            [\S\ref{subsec:l2_vs_l3} Distinction from L2, color=Orchid!100, fill=Orchid!15, very thick, text=black, text width=5cm]
            [\S\ref{subsec:l3_examples} Examples \& Applications, color=Orchid!100, fill=Orchid!15, very thick, text=black, text width=5cm]
            [\S\ref{subsec:l3_context} L3 in Context, color=Orchid!100, fill=Orchid!15, very thick, text=black, text width=5cm]
        ]
        [\S\ref{sec:evaluation} Evaluations, color=Goldenrod!100, fill=Goldenrod!20, very thick, text=black, text width=3.5cm
            [\S\ref{subsec:eval_decision} Prediction vs Decision-Centric, color=Goldenrod!100, fill=Goldenrod!20, very thick, text=black, text width=5cm]
            [\S\ref{subsec:eval_boundary} Three Boundary Conditions, color=Goldenrod!100, fill=Goldenrod!20, very thick, text=black, text width=5cm]
            [\S\ref{subsec:eval_levels} L1/L2/L3 Differentiation, color=Goldenrod!100, fill=Goldenrod!20, very thick, text=black, text width=5cm]
            [\S\ref{subsec:eval_benchmarks} Benchmarks \& Coverage, color=Goldenrod!100, fill=Goldenrod!20, very thick, text=black, text width=5cm]
            [\S\ref{subsec:eval_open} Open Challenges, color=Goldenrod!100, fill=Goldenrod!20, very thick, text=black, text width=5cm]
        ]
        [\S\ref{sec:implementation} Practice Instantiation, color=Melon!100, fill=Melon!20, very thick, text=black, text width=3.5cm
            [\S\ref{subsec:impl_blocks} Architectural Building Blocks, color=Melon!100, fill=Melon!20, very thick, text=black, text width=5cm]
            [\S\ref{subsec:impl_tradeoffs} Design Tradeoffs by Regime, color=Melon!100, fill=Melon!20, very thick, text=black, text width=5cm]
            [\S\ref{subsec:impl_roadmap} Implementation Roadmap, color=Melon!100, fill=Melon!20, very thick, text=black, text width=5cm]
        ]
        [\S\ref{sec:trends} Trends \& Open Problems, color=CornflowerBlue!100, fill=CornflowerBlue!15, very thick, text=black, text width=5cm
            [\S\ref{sec:trends:history} Historical Development, color=CornflowerBlue!100, fill=CornflowerBlue!15, very thick, text=black, text width=5cm]
            [\S\ref{sec:trends:open} Open Problems by Level, color=CornflowerBlue!100, fill=CornflowerBlue!15, very thick, text=black, text width=5cm]
            [\S\ref{sec:trends:security} Security \& Safety, color=CornflowerBlue!100, fill=CornflowerBlue!15, very thick, text=black, text width=5cm]
            [\S\ref{sec:trends:beyond} Beyond L3, color=CornflowerBlue!100, fill=CornflowerBlue!15, very thick, text=black, text width=5cm]
        ]
        [\S\ref{sec:conclusion} Conclusion, color=gray!100, fill=gray!15, very thick, text=black, text width=5cm
        ]
    ]
\end{forest}}
\caption{\textbf{Organizational structure of this survey.} The paper is organized around three capability levels (L1 Predictor, L2 Simulator, L3 Evolver) and four governing-law regimes (physical, digital, social, scientific worlds), with supporting sections on evaluation, implementation, and open problems.}

\label{fig:toc_tree}
\vspace{-6pt}
\end{figure*}
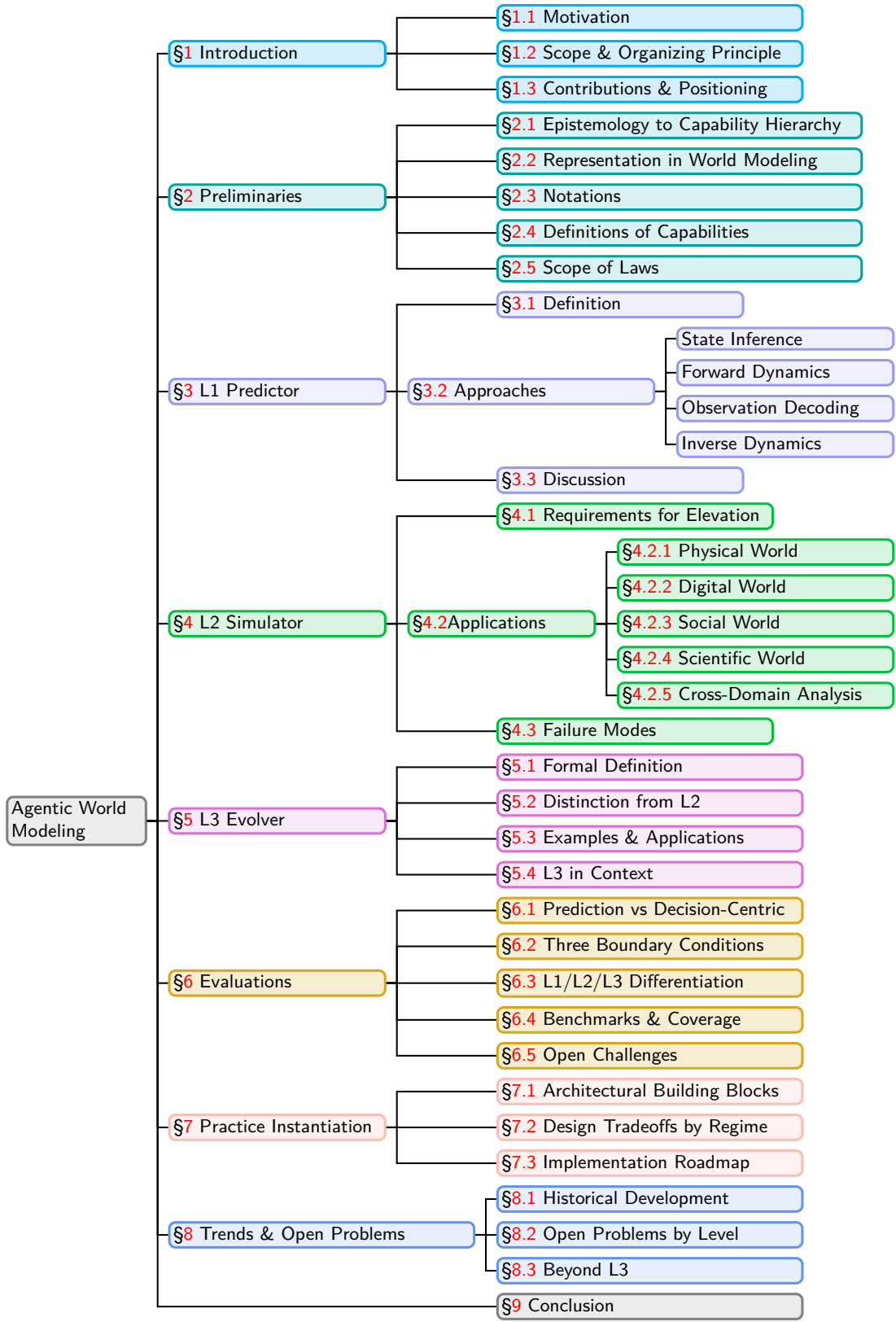

\newpage

\section{Introduction}
\label{sec:introduction}
\epigraph{\textit{One may say the eternal mystery of the world is its comprehensibility.}}{\citet{einstein1936physics}}

The ambition to build internal models of reality has a long intellectual history, appearing in philosophical accounts of mental models~\citep{craik1943nature,johnson1983mental} and in modern machine learning as learned latent dynamics that support prediction, control, simulation, and scientific reasoning~\citep{ha2018worldmodels,hafner2019dreamer,karniadakis2021pinn}. The phrase \textit{world model} is now widely used across research communities, but its precise technical meaning varies considerably~\citep{ding2024survey_wm,zhu2024sora_survey}. In reinforcement learning, agents learn transition structure to imagine futures before acting~\citep{sutton1991dyna,ha2018worldmodels,hafner2019dreamer,schrittwieser2020muzero}. In computer vision, \textit{world models} often denote video or 3D generators that maintain visual dynamics and temporal coherence~\citep{brooks2024sora,bruce2024genie,nvidia2025cosmos,worldlens,liang2026lidarcrafter,bian2025dynamiccity,kong2025survey_3d4d}. In language modeling and agent systems, the term can refer to text-grounded simulation for planning, web interaction, and social environments~\citep{wang2024worldsim,gu2024webdreamer,park2023generative,zhang2026scafgrpo,zhang2026searchgym}. In robotics, learned dynamics serve safe planning, data-efficient policy learning, and sim-to-real transfer~\citep{wu2023daydreamer,yang2024unisim,min2024driveworld}. For science, systems pair surrogate models with hypothesis-driven experimentation~\citep{karniadakis2021pinn,lu2024aiscientist}.

From a complementary perspective, world models and agents are closely coupled. At its core, a world model learns the state-transition dynamics of an environment: given a current state and an action, it predicts the resulting next state. An agent, conversely, selects actions given a task objective and its current observations. These two components are mutually supportive. Agents rely on world models to anticipate the consequences of candidate actions, enabling look-ahead planning and sample-efficient learning~\citep{hafner2023dreamerv3,schrittwieser2020muzero,dong2026lcvn,dong2026uniwm}. Conversely, world models benefit from agent-generated experience, which provides targeted, task-relevant trajectories that improve the model's accuracy in decision-critical regions of the state space~\citep{sutton1991dyna}. This close coupling motivates the capability-based perspective adopted in this survey: while world models serve many purposes, we operationally define their value by the quality of decisions they enable for downstream agents.

Because world models constitute a foundational component whose value extends beyond any single agent architecture, their growing importance makes conceptual clarity all the more urgent. Yet the diversity outlined above also creates conceptual fragmentation: a vision researcher may evaluate a world model by the visual fidelity of its generated frames, while a reinforcement learning practitioner judges the same term by whether it improves task performance. As a result, papers may report strong progress under one interpretation of \textit{world model} while remaining incomparable under another. This paper addresses that fragmentation by providing a common language that can align communities without erasing domain-specific differences.

\subsection{Motivation}
\label{sec:motivation}

\begin{enumerate}[leftmargin=*]
\item \textbf{Current survey landscape.}
Several recent surveys have attempted to organize this rapidly growing literature. \citet{ding2024survey_wm} propose a dual taxonomy of \emph{understanding} versus \emph{predicting}, mapping world models onto application domains such as autonomous driving, robotics, and social simulacra. \citet{zhu2024sora_survey} focus on the generative capabilities catalyzed by Sora, surveying world models for video generation, autonomous driving, and autonomous agents. \citet{yue2025simulating} provide a roadmap for 2D visual world modeling with a four-generation capability taxonomy (G1--G4) applied to robotics, autonomous driving, and gaming. Their G1--G4 taxonomy is useful for distinguishing increasingly interactive visual generation systems; our L1--L3 hierarchy is complementary rather than competing, because it abstracts away from the visual modality and asks whether a system supports local prediction, decision-usable simulation, or evidence-driven revision across physical, digital, social, and scientific regimes. Roughly, early G-levels emphasize appearance and action-conditioned prediction, whereas our L2/L3 boundary is determined by constraint-valid rollout and persistent model update. Domain-specific surveys have also proliferated: \citet{li2025embodied_wm_survey} provide a three-axis framework (functionality, temporal modeling, spatial representation) specifically for embodied AI; \citet{feng2025ad_wm_survey} and \citet{tu2025wm_ad_survey} survey world models for autonomous driving; \citet{kong2025survey_3d4d} examine 3D and 4D world modeling; \citet{zhang2025wm_manipulation_survey} survey world models for robotic manipulation; and a growing number of position papers question what it means for a learned model to ``understand'' physics~\citep{lecun2022path, kang2025howfar}. In AI for science, \citet{wei2025agenticscience} survey autonomous scientific discovery across life sciences, chemistry, materials, and physics, unifying process-oriented, autonomy-oriented, and mechanism-oriented perspectives. A parallel line of surveys addresses agent planning and reasoning: \citet{wei2025plangenllms} survey LLM planning capabilities across plan generation and verification, \citet{huang2024planningsurvey} taxonomize planning mechanisms into decomposition, selection, and reflection, \citet{cao2025llmplanning} provide a systematic comparison of fine-tuning versus search-based planning methods, \citet{zhao2025agenticreasoning} organize agentic reasoning into single-agent, tool-based, and multi-agent frameworks, and \citet{arunkumar2026agenticai} propose a unified agent taxonomy spanning perception, planning, action, and collaboration. These surveys complement ours: they focus on how agents \emph{decide and act}, whereas we focus on the predictive substrate (the world model) that makes those decisions informed. Despite their valuable contributions, existing surveys share a common organizational principle that we argue is fundamentally limiting: they partition the field by \textbf{modality} or by \textbf{application domain}. Our work differs by organizing the field through a capability-based taxonomy that cuts across modalities, covering decision-making domains from embodied manipulation and autonomous driving to web agents, multi-agent coordination, and scientific discovery pipelines.

\begin{figure}[t]
\centering
\includegraphics[width=\textwidth]{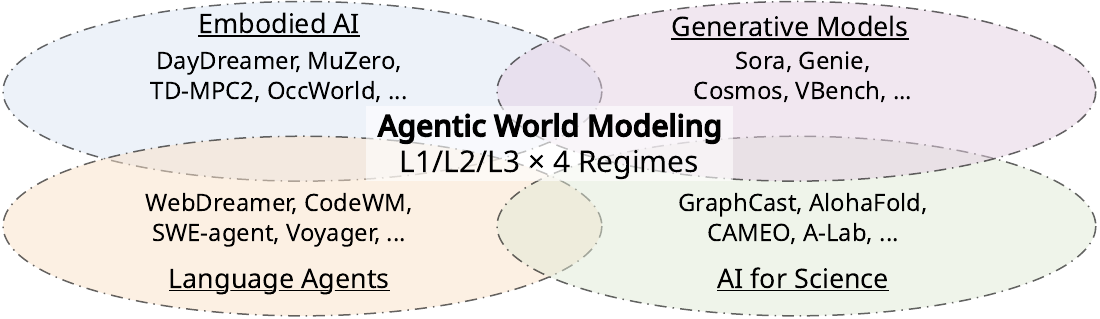}
\caption{\textbf{Positioning of this survey relative to existing world model and agent surveys.} Four clusters, Embodied World Models, Generative World Models, Language Agents, and AI for Science, each cover subsets of the field. Our survey (center) integrates cross domain coverage with a capability based taxonomy (L1/L2/L3 × four regimes), bridging largely isolated communities.}
\label{fig:survey_positioning}
\end{figure}

\paragraph{Gaps in existing surveys.}
The modality-centric and domain-centric taxonomies leave two critical gaps. First, they fail to capture the \emph{capability progression} that cuts across modalities. A key example is model-based reinforcement learning, where latent-space ``imagination'' rollouts can match or exceed model-free baselines across diverse domains such as Atari, continuous control, and Minecraft~\citep{hafner2023dreamerv3, schrittwieser2020muzero, hafner2019dreamer}. We formalize this progression as a three-level capability hierarchy: one-step prediction, long-horizon simulation, and evidence-driven model revision. A second motivation for our framework is the intensifying debate over whether large-scale generative models are merely plausible generators or genuine world simulators. Existing surveys have surfaced this tension~\citep{brooks2024sora, bruce2024genie, kang2025howfar, ding2024survey_wm}, but a capability-based taxonomy helps state the question more precisely in terms of rollout, intervention sensitivity, and constraint consistency. We identify four progressively stronger capabilities, namely rollout, intervention sensitivity, constraint consistency, and closed-loop use, that characterize world models and go beyond generic predictors (formalized in Section~\ref{sec:preliminaries}). Moreover, existing surveys underrepresent the role of world modeling in agentic AI applications, including web agents, tool-use agents, and multi-agent systems, where learned environment dynamics are essential for planning and action selection~\citep{gu2024webdreamer,earlyexp,wang2024worldsim,park2023generative}. The goal of this paper is to establish a capability-based taxonomy with clear boundary conditions, and to use it to connect research communities that currently evaluate world modeling systems with different assumptions, objectives, and metrics.
\end{enumerate}

Figure~\ref{fig:survey_positioning} positions this survey relative to existing work along two axes: scope (domain-specific to cross-domain) and organizing principle (modality-centric to capability-centric). Figure~\ref{fig:toc_tree} shows the organizational structure of the paper at a glance, grouping sections by the three capability levels (L1 Predictor, L2 Simulator, L3 Evolver) and the four governing-law regimes (physical, digital, social, and scientific worlds).

\subsection{Scope and Organizing Principle}
\label{sec:scope}

\begin{figure}[t]
\centering
\includegraphics[width=\textwidth]{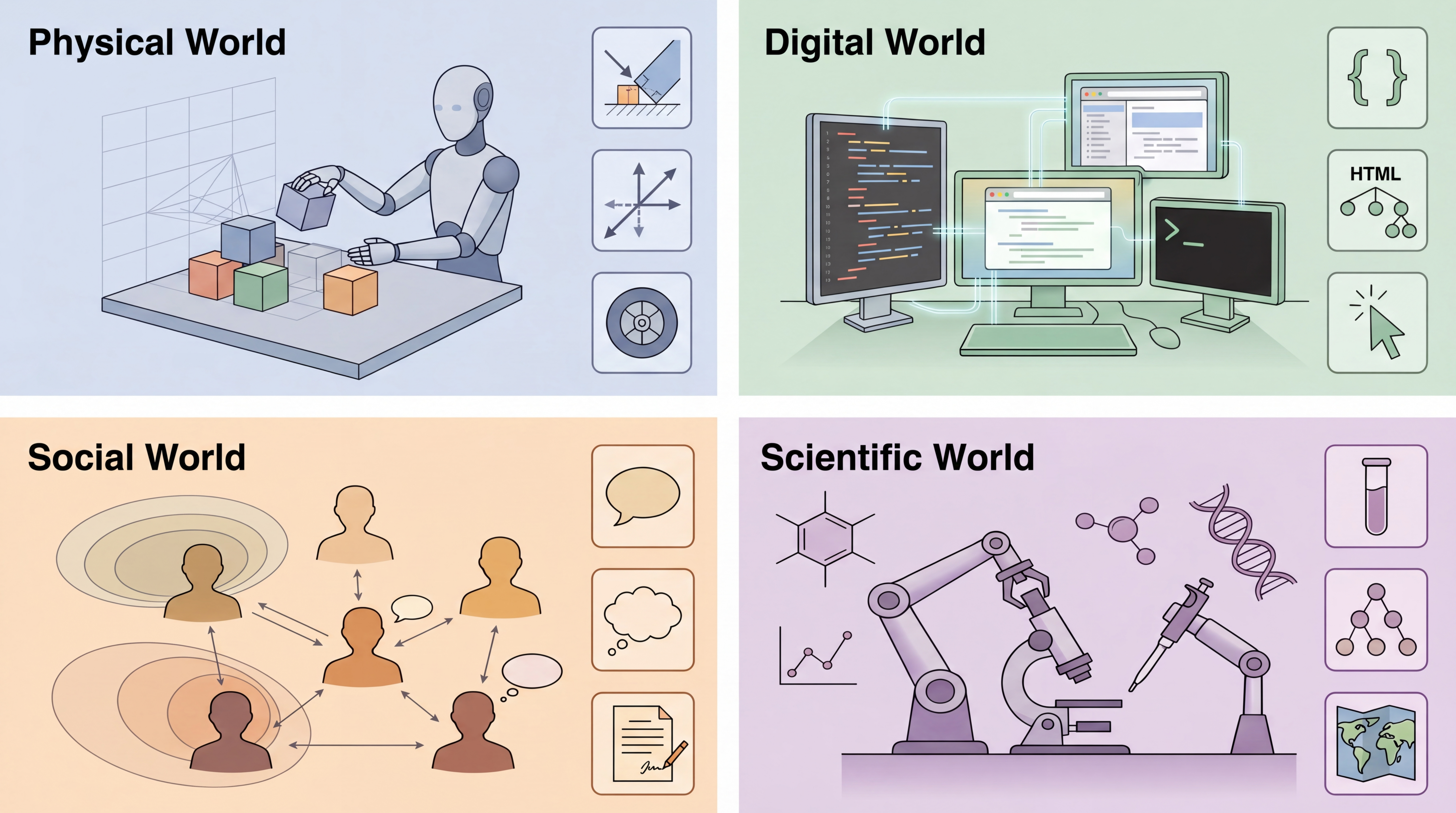}
\caption{\textbf{Schematic illustrations of the four governing-law regimes.}
Representative scenes for each regime: a humanoid agent manipulating blocks
(Physical World), code and UI surfaces (Digital World), a network of
interacting agents with speech acts (Social World), and instrumented
experimentation with robotic microscope and pipette (Scientific World).
Each regime's formal constraints are discussed in
Section~\ref{subsec:scope_regimes}.}
\label{fig:four_worlds}
\end{figure}

\paragraph{Governing principles across domains.}
We organize the paper along two orthogonal axes: (i)~\textbf{capability level} (L1/L2/L3, defined formally in Section~\ref{sec:preliminaries}), and (ii)~\textbf{governing-law regime}, the constraints that legitimate transitions must satisfy in a domain. These levels are stages of world-modeling capability rather than mutually exclusive model classes: the same system may invoke different levels at different moments depending on task demand. Figure~\ref{fig:four_worlds} provides a schematic overview of these four regimes.
\begin{itemize}[leftmargin=*]
  \item \textbf{Laws of the Physical World}: perception; physical interaction; robotic manipulation, navigation, autonomous driving, egocentric video prediction, action-conditioned video modeling, 3D world modeling..
  \item \textbf{Laws of the Digital World}: program semantics; web navigation, software tool use, GUI environments.
  \item \textbf{Laws of the Social World}: beliefs; goals; norms; social coordination, dialogue, multi-agent settings.
  \item \textbf{Laws of the Scientific World}: latent mechanisms; experimental observables; causal structure; scientific discovery pipelines, measurement-coupled prediction, hypothesis-driven experimentation.
\end{itemize}

In particular, the physical and scientific regimes are separated by how constraints are accessed: physical-world systems often admit analytic or simulator-based verification of transitions, whereas scientific-world systems typically require empirical validation because the governing mechanisms are only partially known.

Regimes are not ``orthogonal modalities'': real systems mix them. The value of the taxonomy is diagnostic; it clarifies \emph{which} invariants a method tries to preserve and \emph{which queries} it can answer reliably.

More generally, a world model can predict transitions along any organizing dimension, such as spatial scales, frequency bands, or causal depth, provided it maintains the capability criteria along that axis.
Throughout, we use \emph{world model} to denote learned (or hybrid) operators that support intervention-aware transition queries, and \emph{world modeling} to denote the staged process of strengthening those operators.

\paragraph{How an agent uses the three levels at runtime.}
The L1/L2/L3 taxonomy is not a static classification of systems but a description of the capability an agent invokes at any given moment. A single deployed system can operate at different levels depending on the task demand:
\begin{enumerate}[leftmargin=*, nosep]
  \item \textbf{L1 (Predictor).} The agent executes fast, reactive one-step predictions (such as perception, low-level motor control, or token-by-token generation) without maintaining a multi-step plan.
  \item \textbf{L2 (Simulator).} The agent upgrades to this level when the task requires comparing candidate action sequences, reasoning counterfactually about alternative futures, or verifying that a planned trajectory respects governing-law constraints; here the agent rolls out a multi-step simulation before committing.
  \item \textbf{L3 (Evolver).} The agent escalates to this level when its current model produces systematic prediction failures that cannot be resolved by re-planning within the existing model structure, that is, when the model itself must be revised, assets distilled, and updates validated before the next deployment.
\end{enumerate}
This runtime dispatch view clarifies why L3 is not a replacement for L1/L2 but a governance layer that improves the stack when evidence demands it. Within a full agentic stack, world models are only one component: tool use determines how the agent acts on the environment, memory determines what evidence persists across episodes, multi-agent coordination shapes the effective transition dynamics in social settings, and reflection determines when failures trigger revision rather than mere re-planning. Our focus is the world-model substrate, but its role is always in service of these broader agentic loops.

\begin{figure*}[t]
\centering
\includegraphics[width=\textwidth]{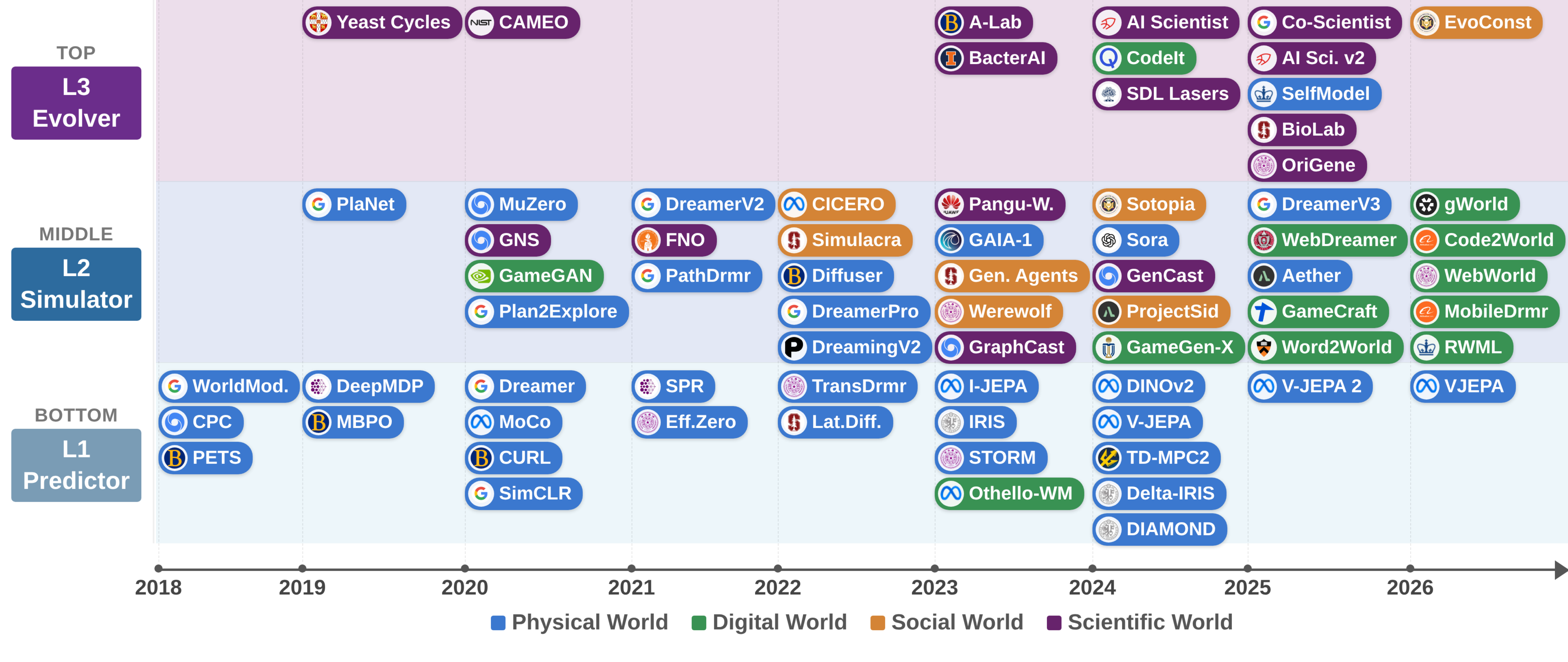}
\caption{\textbf{Timeline of representative world modeling systems (2018--2026) organized by capability level.} The roadmap shows 70 survey anchors, capped at five systems per year--level cell for readability. L1~Predictor denotes one-step dynamics, L2~Simulator denotes decision-usable multi-step rollout, and L3~Evolver denotes full evidence-driven model revision; partial L3 loops remain in Table~\ref{tab:l3_systems}. Each pill is colored by governing-law regime: \textcolor[HTML]{3B78CF}{\textbf{Physical}} (blue), \textcolor[HTML]{399253}{\textbf{Digital}} (green), \textcolor[HTML]{D48436}{\textbf{Social}} (orange), and \textcolor[HTML]{67236C}{\textbf{Scientific}} (purple).}
\label{fig:paper_roadmap}
\end{figure*}

\subsection{Contributions and Positioning}
\label{sec:contributions}

\begin{keypoint}[Key Contributions]
This paper makes three principal contributions (Figure~\ref{fig:paper_roadmap}):
\begin{enumerate}[leftmargin=*]
 \item \textbf{Capability-based roadmap for world modeling in agentic AI (L1$\to$L2$\to$L3).} We propose a three-level capability hierarchy with testable boundary conditions: L1 \textbf{Predict World} (one-step prediction), L2 \textbf{Simulate World} (long-horizon, action-conditioned rollout with constraint satisfaction), and L3 \textbf{Modify World} (evidence-driven model growth through autonomous data collection and dynamics revision). These are stages of capability, not types of models.
 \item \textbf{Cross-domain synthesis via governing laws.} We unify computer vision, language modeling, model-based RL and robotics, and AI for science into a single capability coordinate system. Different governing laws (Section~\ref{sec:preliminaries}) define the types or partitions of world models, \textbf{partially independent} of the L1$\to$L2$\to$L3 capability axis. This two-dimensional organization (capability level $\times$ law regime) reveals shared principles across communities that have developed in isolation, while clarifying domain-specific challenges that make direct transfer non-trivial.
 \item \textbf{L3 as a distinct capability level.} Evidence-driven model growth, where a system autonomously collects new evidence and revises its own dynamics model, has appeared in scattered forms across scientific discovery~\citep{lu2024aiscientist}, autonomous experimentation, and online adaptation. We argue this capability is qualitatively different from L2 rollout and formalize it as a distinct level, and identifying the open problems that must be resolved to realize this capability at scale.
\end{enumerate}
\end{keypoint}

\paragraph{Positioning.}
We present this paper as a \emph{position-driven survey proposing a capability taxonomy for world modeling}. It advances a specific conceptual framework, namely the L1/L2/L3 capability hierarchy paired with a governing-law regime taxonomy, and argues for its adoption across the world modeling community. Unlike a pure survey, it proposes testable boundary conditions and uses them to re-examine how existing systems are classified. Unlike a pure position paper, it substantiates each argument with a comprehensive literature review spanning computer vision, reinforcement learning, robotics, natural language processing, and AI for science. This paper does not introduce a new benchmark or leaderboard; instead, it offers a unifying conceptual framework for interpreting and comparing existing systems and evaluations.

\paragraph{Outline.}
Section~\ref{sec:preliminaries} establishes the conceptual and notational foundations: it motivates the three capability stages from epistemological intuition, gives each a formal definition with testable boundary conditions, and clarifies the distinctions between world modeling and generic prediction, world models and planners, and world modeling and commonsense.
Sections~\ref{sec:l1}--\ref{sec:l3} present the three capability levels in detail with representative methods and cross-domain analysis.
Section~\ref{sec:evaluation} discusses evaluation methodology, Section~\ref{sec:implementation} addresses architectural and computational considerations, and Section~\ref{sec:trends} identifies emerging trends and open problems.
Section~\ref{sec:conclusion} concludes.
We note that L3 is not a terminal stage; Section~\ref{sec:trends} introduces \emph{meta-world modeling}, in which the governing laws themselves become learnable, and identifies the open problems this entails.


\section{Preliminaries}
\label{sec:preliminaries}

This section establishes the conceptual and notational foundations used throughout the paper.
\textbf{(1)~From epistemology to a capability hierarchy} draws on philosophical traditions to propose a three-level decomposition of world modeling capability (L1 Predictor, L2 Simulator, L3 Evolver) and to motivate why the boundaries fall where they do.
\textbf{(2)~Notation and formal definitions} fixes a unified symbol system and uses it to give each stage (L1, L2, L3) a precise definition with testable boundary conditions.
\textbf{(3)~Conceptual boundaries} clarifies the distinctions between world modeling and generic prediction, between world models and planners, and relates world modeling to the broader notion of commonsense reasoning that underwrites the reliable everyday action that agents must exhibit beyond narrow predictive tasks.

\subsection{Philosophical Motivations} 
\label{sec:philosophy}

\begin{figure}[t]
\centering
\includegraphics[width=\textwidth]{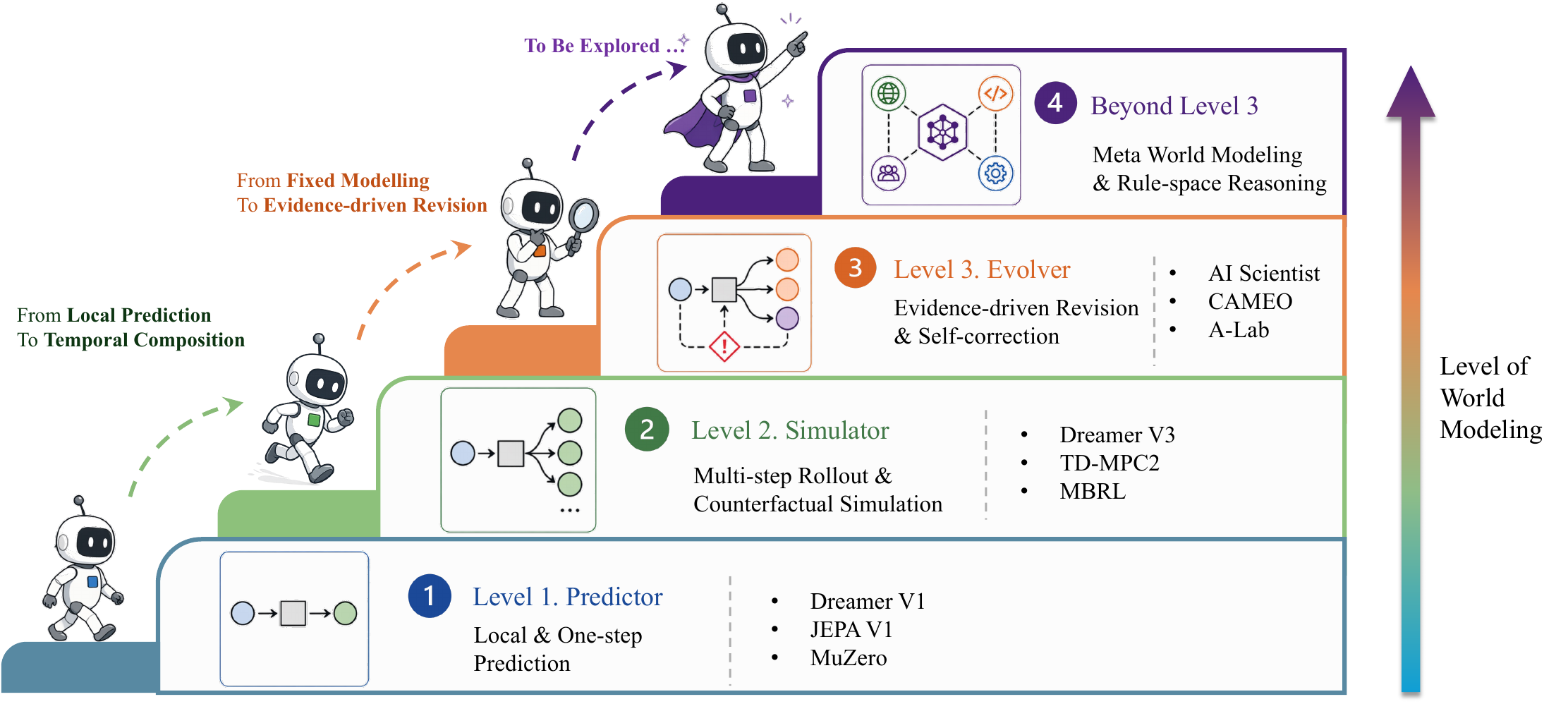}
\caption{\textbf{From local prediction to evidence-driven revision: a hierarchical view of world modeling.}
Level 1 models empirical regularities for prediction, Level 2 supports possible-world semantics and counterfactual simulation, and Level 3 introduces evidence-driven revision through continual interaction with the environment. This hierarchy frames world modeling as an ascending process from pattern recognition, to temporal rollout, to adaptive model evolution in real-world practice.}
\label{fig:philosophy_staircase}
\end{figure}

A natural question for any world-modeling survey is: \emph{what stages of understanding does a system pass through as it moves from pattern matching to genuine modeling?}
Epistemology, the study of what counts as knowledge and how knowledge grows, offers a useful lens.
Different philosophical traditions identify qualitatively different kinds of epistemic achievement; we draw on these traditions to propose a three-level capability hierarchy for world models.
These philosophical analogies are heuristic rather than historical or one-to-one. We do not claim that ML systems implement philosophical programmes, but that philosophical distinctions help us see why certain capability boundaries recur across domains and what design questions each stage foregrounds (Figure~\ref{fig:philosophy_staircase}).
Due to space constraints, detailed philosophical motivations and contemporary examples, together with extended historical context, are deferred to Appendix~\ref{app:philosophy_extended}.

\paragraph{L1 Predictor: from pattern to one-step forecast.}
\label{sec:philosophy_l1}
The simplest epistemic achievement is learning patterns from data: given past observations, predict the next one.
In philosophy, this is the terrain of Hume's \emph{constant conjunction}~\citep{hume1739treatise}; an agent records statistical co-occurrences without certifying why they hold.
When a model learns one-step latent transitions from trajectories, it occupies exactly this epistemic position: it extracts succession from data and bets that the pattern persists.
This view aligns with predictive coding framework in cognitive science~\citep{rao1999predictive, friston2010free} and the ``Bayesian brain'' hypothesis that perception is probabilistic inference~\citep{clark2015surfing}, motivating one-step latent forecasting as a computational primitive~\citep{lake2017building}.

We call this stage \textbf{L1 (Predictor)}.
This Humean stance has inherent fragility.
The i.i.d.\ assumption underlying most ML is effectively Hume's \emph{Uniformity Principle} (the premise that the future will resemble the past), so when the distribution shifts, L1 models that rely on learned regularities fail to generalize.
Nevertheless, this provides the most basic inductive bias, which is the foundation of modeling.

\paragraph{L2 Simulator: rollout and counterfactual.}
\label{sec:philosophy_l2}
Pattern matching alone does not answer \emph{what would happen if we acted differently}.
The next stage adds intervention and counterfactual reasoning: the ability to roll out coherent futures under chosen actions or hypothetical initial conditions and use the results for decision-making.
David Lewis's theory of \emph{closest possible worlds}~\citep{lewis1973counterfactuals} captures this jump: effective counterfactual reasoning explores worlds maximally similar to our own, where only a minimal intervention distinguishes actual from counterfactual outcomes, providing a principled basis for reasoning about what would have happened under alternative actions taken by the agent at decision points.

We call this stage \textbf{L2 (Simulator)}.
Because L2 rollouts are model-relative, their reliability depends on the learned model's own transition structure rather than on direct access to ground-truth dynamics.
They risk epistemic drift, which produces internally coherent trajectories for the training manifold.
Plato's Allegory of the Cave~\citep{plato1992republic} offers a vivid metaphor: a simulator excelling at predicting shadows on a wall may remain fundamentally bounded by the wall's dimensions, unable to access the fire casting those shadows.

\paragraph{L3 Evolver: model revision from evidence.}
\label{sec:philosophy_l3}
Even a powerful simulator eventually encounters situations where its predictions systematically fail, not because of parameter error but because the model class itself is too narrow.
Epistemology offers a rich vocabulary for this transition.
Lakatos's distinction between a \emph{hard core} (architecture, inductive biases) and a \emph{protective belt} (learned parameters)~\citep{lakatos1978methodology} provides a useful parallel.
Gradient steps mostly adjust the belt, while persistent structured errors may require changes to the core, such as new modules, parsers, constraints, or simulator hooks.

We call this stage \textbf{L3 (Evolver)}: the capacity to rebuild the laboratory when evidence demands it.
This extends the full design--execute--observe--reflect loop: the system not only simulates but actively designs experiments, executes them, observes outcomes, and reflects to revise its model stack.
Duhem--Quine holism~\citep{duhem1954aim, quine1951two} explains why blame-assignment is non-trivial. Errors redistribute across modules until diagnostics isolate the brittle component.
Proposed revisions should yield measurable improvements on held-out probes, regression suites, or experimental outcomes, rather than post-hoc adjustments that preserve the existing model despite contrary evidence from the environment.

\subsection{Representation in World Modeling: Lessons from Scientific Theories}
\label{sec:trends:representation}

The capability hierarchy in Section~\ref{sec:philosophy} addresses \emph{what a world model can do}, but leaves open a prior question: \emph{in what form should the world model actually be represented?} This question should no just solely treated as an implementation detail, yet it determines whether the capabilities defined above, especially L3 revision, are realizable in practice across the diverse application domains covered in later sections.


\begin{figure}[t]
\centering
\resizebox{16.5cm}{!}{%
\begin{tikzpicture}[
    x=1.1cm, y=1.4cm,
    ms/.style={circle, draw=black, thick, inner sep=2pt},
    msBlue/.style={ms, fill=black!6!blue!12},
    msOrange/.style={ms, fill=black!4!orange!12},
    msGreen/.style={ms, fill=black!4!green!10},
    msPurple/.style={ms, fill=black!4!purple!12},
    labA/.style={anchor=south, font=\footnotesize, align=center, inner sep=3pt},
    labB/.style={anchor=north, font=\footnotesize, align=center, inner sep=3pt},
    era/.style={font=\normalsize\bfseries, anchor=south},
]

\fill[black!6!blue!12]   (0,-0.6)    rectangle (3.6,2.2);
\fill[black!4!orange!12] (3.6,-0.6)  rectangle (9.5,2.2);
\fill[black!4!green!10]  (9.5,-0.6)  rectangle (16.0,2.2);
\fill[black!4!purple!12] (16.0,-0.6) rectangle (20.0,2.2);

\node[era] at (1.8, 2.3) {{\color{blue!60}\faCompass}\ Mathematical Principles};
\node[era] at (6.55, 2.3) {{\color{orange!70}\faPuzzlePiece}\ Symbolic Intelligence};
\node[era] at (12.75, 2.3) {{\color{green!60!black}\faBrain}\ Connectionist Resurgence};
\node[era] at (18.0, 2.3) {{\color{purple!60}\faRocket}\ Generative Revolution};

\draw[->, very thick] (-0.3, 0.7) -- (20.3, 0.7);

\fill[cyan!18] (8.1, -0.6) rectangle (8.8, 2.2);
\fill[pattern=north east lines, pattern color=blue!30]
    (8.1, -0.6) rectangle (8.8, 2.2);
\node[font=\small\bfseries, blue!70!black, rotate=90] at (8.45, 0.7)
    {\faSnowflake\ AI Winter I};

\fill[cyan!18] (10.3, -0.6) rectangle (11.0, 2.2);
\fill[pattern=north east lines, pattern color=blue!30]
    (10.3, -0.6) rectangle (11.0, 2.2);
\node[font=\small\bfseries, blue!70!black, rotate=90] at (10.65, 0.7)
    {\faSnowflake\ AI Winter II};

\node[msBlue] at (0.5, 0.7) {};
\node[labA] at (0.5, 0.9) {Newton\\[-2pt]1687};

\node[msBlue] at (1.5, 0.7) {};
\node[labB] at (1.5, 0.5) {Laplace\\[-2pt]1814};

\node[msBlue] at (2.6, 0.7) {};
\node[labA] at (2.6, 0.9) {Turing\\[-2pt]1950};

\node[msOrange] at (3.6, 0.7) {};
\node[labB] at (3.6, 0.5) {Dartmouth\\[-2pt]1956};

\node[msOrange] at (5.0, 0.7) {};
\node[labA] at (5.0, 0.9) {Frame Problem\\[-2pt]1969};

\node[msOrange] at (6.2, 0.7) {};
\node[labB] at (6.2, 0.5) {STRIPS\\[-2pt]1971};

\node[msOrange] at (7.2, 0.7) {};
\node[labA] at (7.2, 0.9) {Lighthill\\[-2pt]1973};

\node[msGreen] at (9.5, 0.7) {};
\node[labB] at (9.5, 0.5) {Backprop\\[-2pt]1986};

\node[msGreen] at (11.5, 0.7) {};
\node[labA] at (11.5, 0.9) {LeNet\\[-2pt]1998};

\node[msGreen] at (12.7, 0.7) {};
\node[labB] at (12.7, 0.5) {AlexNet\\[-2pt]2012};

\node[msGreen] at (13.9, 0.7) {};
\node[labA] at (13.9, 0.9) {Transformer\\[-2pt]2017};

\node[msGreen] at (15.1, 0.7) {};
\node[labB] at (15.1, 0.5) {World Models\\[-2pt]2018};

\node[msPurple] at (16.0, 0.7) {};
\node[labA] at (16.0, 0.9) {DDPM / GPT-3\\[-2pt]2020};

\node[msPurple] at (17.0, 0.7) {};
\node[labB] at (17.0, 0.5) {GraphCast\\[-2pt]2023};

\node[msPurple] at (17.8, 0.7) {};
\node[labA] at (17.8, 0.9) {DreamerV3\\[-2pt]2023};

\node[msPurple] at (18.7, 0.7) {};
\node[labB] at (18.7, 0.5) {Sora\\[-2pt]2024};

\node[msPurple] at (19.6, 0.7) {};
\node[labA] at (19.6, 0.9) {AlphaEvolve\\[-2pt]2025};

\end{tikzpicture}%
}
\caption{\textbf{Historical development of world modeling across four eras:} Mathematical Principles (--1956), Symbolic Intelligence (1956--1986), Connectionist Resurgence (1986--2020), and Generative Revolution (2020--present). Two AI winters (1974--1980, 1987--1993) mark transitions between paradigms. See discussions in Section~\ref{sec:trends:history}. This argues that \textit{a good representation of world model should be instantiation-agnostic}.}
\label{fig:historical_timeline}
\end{figure}
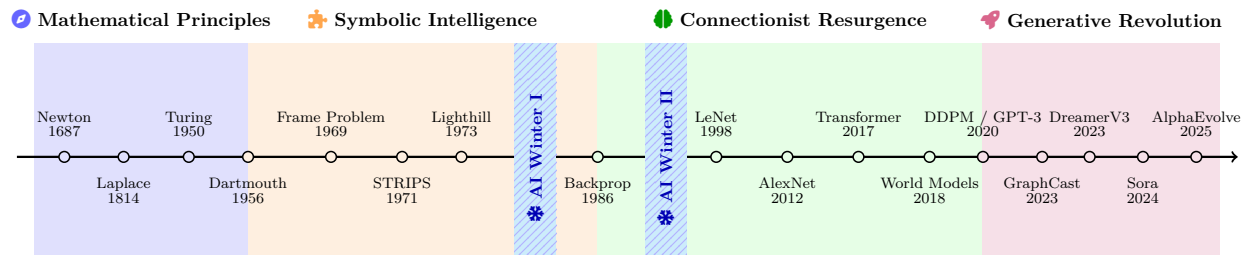

Historically, symbolic approaches to machine intelligence struggled to scale (see Section~\ref{sec:trends:history}), leading modern systems to adopt latent, implicit representations. Here scientific theories offer a telling contrast. Newton's laws, Maxwell's equations, and the Standard Model are instances of world models expressed in compact \emph{symbolic} form, and arguably represent the most successful human instances of L3 systems: explicit, revisable, and composable.
This contrast forces a question the field has largely avoided: {is the endpoint of world modeling symbolic discovery, with neural latents as a scaffold, or are latent dynamics themselves the goal?}

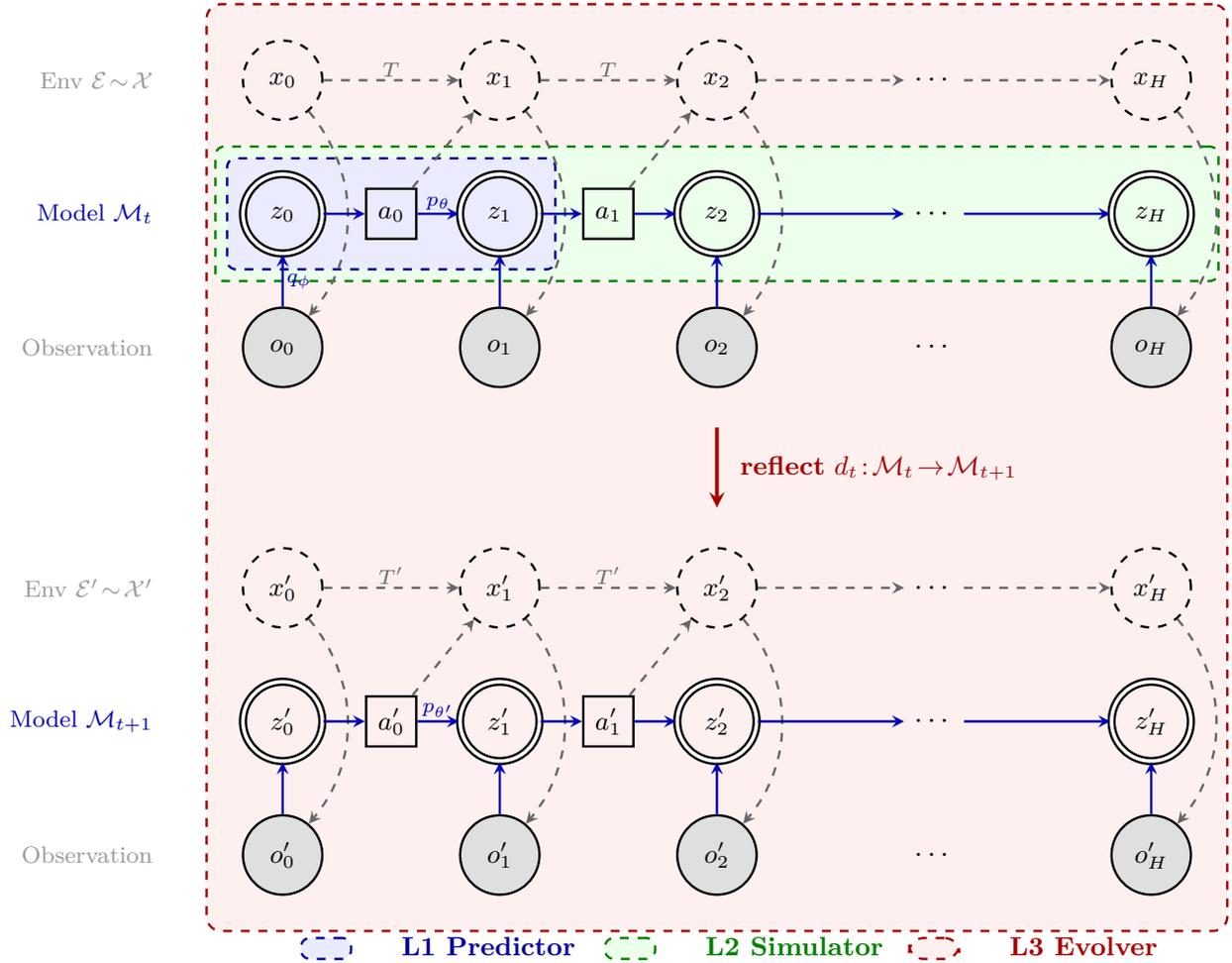
\begin{figure*}[t]
\centering
\resizebox{\textwidth}{!}{%
\begin{tikzpicture}[
    x=2.6cm, y=1.6cm,
    env/.style={circle, draw=black, thick, dashed, minimum size=0.95cm, font=\small},
    learned/.style={circle, draw=black, double, double distance=1.2pt, thick, minimum size=0.95cm, font=\small},
    obs/.style={circle, draw=black, thick, fill=gray!25, minimum size=0.95cm, font=\small},
    act/.style={rectangle, draw=black, thick, minimum size=0.6cm, font=\small},
    arr/.style={->, thick, >=stealth},
    dyn/.style={->, thick, >=stealth, blue!70!black},
    em/.style={->, thick, >=stealth, dashed, black!60},
    lbl/.style={font=\scriptsize, inner sep=1pt},
    rowlbl/.style={font=\footnotesize, anchor=east},
    l1box/.style={draw=blue!60!black, dashed, thick, rounded corners=3pt, inner sep=5pt, fill=blue!8},
    l2box/.style={draw=green!50!black, dashed, thick, rounded corners=3pt, inner sep=9pt, fill=green!8},
    l3box/.style={draw=red!65!black, dashed, thick, rounded corners=5pt, inner sep=12pt, fill=red!6},
]


\node[env] (x0) at (0, 3)   {$x_{0}$};
\node[env] (x1) at (1, 3)   {$x_{1}$};
\node[env] (x2) at (2, 3)   {$x_{2}$};
\node       (xd) at (3, 3)  {$\cdots$};
\node[env] (xH) at (4, 3)   {$x_{H}$};

\node[learned] (z0) at (0, 2)   {$z_{0}$};
\node[act]     (a0) at (0.5, 2) {$a_{0}$};
\node[learned] (z1) at (1, 2)   {$z_{1}$};
\node[act]     (a1) at (1.5, 2) {$a_{1}$};
\node[learned] (z2) at (2, 2)   {$z_{2}$};
\node           (zd) at (3, 2)  {$\cdots$};
\node[learned] (zH) at (4, 2)   {$z_{H}$};

\node[obs] (o0) at (0, 1) {$o_{0}$};
\node[obs] (o1) at (1, 1) {$o_{1}$};
\node[obs] (o2) at (2, 1) {$o_{2}$};
\node       (od) at (3, 1){$\cdots$};
\node[obs] (oH) at (4, 1) {$o_{H}$};

\draw[em] (x0) -- (x1) node[lbl, midway, above] {$T$};
\draw[em] (x1) -- (x2) node[lbl, midway, above] {$T$};
\draw[em] (x2) -- (xd);
\draw[em] (xd) -- (xH);

\draw[em] (x0) to[bend left=40] (o0);
\draw[em] (x1) to[bend left=40] (o1);
\draw[em] (x2) to[bend left=40] (o2);
\draw[em] (xH) to[bend left=40] (oH);

\draw[dyn] (o0) -- (z0) node[lbl, midway, right, text=blue!70!black] {$q_\phi$};
\draw[dyn] (o1) -- (z1);
\draw[dyn] (o2) -- (z2);
\draw[dyn] (oH) -- (zH);

\draw[dyn] (z0) -- (a0);
\draw[dyn] (a0) -- (z1) node[lbl, midway, above, text=blue!70!black] {$p_\theta$};
\draw[dyn] (z1) -- (a1);
\draw[dyn] (a1) -- (z2);
\draw[dyn] (z2) -- (zd);
\draw[dyn] (zd) -- (zH);

\draw[em] (a0) -- (x1);
\draw[em] (a1) -- (x2);

\node[rowlbl, gray!80]  at (-0.55, 3) {Env $\mathcal{E}\!\sim\!\mathcal{X}$};
\node[rowlbl, blue!70!black]  at (-0.55, 2) {Model $\mathcal{M}_{t}$};
\node[rowlbl, gray!80]  at (-0.55, 1) {Observation};


\draw[arr, red!65!black, line width=1.3pt]
  (2, 0.4) -- (2, -0.2)
  node[midway, right=4pt, font=\small, text=red!65!black]
  {\textbf{reflect} $d_t\!:\!\mathcal{M}_t\!\to\!\mathcal{M}_{t+1}$};

\node[env] (xp0) at (0, -0.8)  {$x'_{0}$};
\node[env] (xp1) at (1, -0.8)  {$x'_{1}$};
\node[env] (xp2) at (2, -0.8)  {$x'_{2}$};
\node       (xpd) at (3, -0.8) {$\cdots$};
\node[env] (xpH) at (4, -0.8)  {$x'_{H}$};

\node[learned] (zp0) at (0, -1.8)   {$z'_{0}$};
\node[act]     (ap0) at (0.5, -1.8) {$a'_{0}$};
\node[learned] (zp1) at (1, -1.8)   {$z'_{1}$};
\node[act]     (ap1) at (1.5, -1.8) {$a'_{1}$};
\node[learned] (zp2) at (2, -1.8)   {$z'_{2}$};
\node           (zpd) at (3, -1.8)  {$\cdots$};
\node[learned] (zpH) at (4, -1.8)   {$z'_{H}$};

\node[obs] (op0) at (0, -2.8) {$o'_{0}$};
\node[obs] (op1) at (1, -2.8) {$o'_{1}$};
\node[obs] (op2) at (2, -2.8) {$o'_{2}$};
\node       (opd) at (3, -2.8){$\cdots$};
\node[obs] (opH) at (4, -2.8) {$o'_{H}$};

\draw[em] (xp0) -- (xp1) node[lbl, midway, above] {$T'$};
\draw[em] (xp1) -- (xp2) node[lbl, midway, above] {$T'$};
\draw[em] (xp2) -- (xpd);
\draw[em] (xpd) -- (xpH);
\draw[em] (xp0) to[bend left=40] (op0);
\draw[em] (xp1) to[bend left=40] (op1);
\draw[em] (xp2) to[bend left=40] (op2);
\draw[em] (xpH) to[bend left=40] (opH);
\draw[dyn] (op0) -- (zp0);
\draw[dyn] (op1) -- (zp1);
\draw[dyn] (op2) -- (zp2);
\draw[dyn] (opH) -- (zpH);
\draw[dyn] (zp0) -- (ap0);
\draw[dyn] (ap0) -- (zp1) node[lbl, midway, above, text=blue!70!black] {$p_{\theta'}$};
\draw[dyn] (zp1) -- (ap1);
\draw[dyn] (ap1) -- (zp2);
\draw[dyn] (zp2) -- (zpd);
\draw[dyn] (zpd) -- (zpH);
\draw[em] (ap0) -- (xp1);
\draw[em] (ap1) -- (xp2);

\node[rowlbl, gray!80]  at (-0.55, -0.8) {Env $\mathcal{E}'\!\sim\!\mathcal{X}'$};
\node[rowlbl, blue!70!black]  at (-0.55, -1.8) {Model $\mathcal{M}_{t+1}$};
\node[rowlbl, gray!80]  at (-0.55, -2.8) {Observation};

\begin{scope}[on background layer]
\node[l3box, fit=(x0)(xH)(o0)(oH)(xp0)(xpH)(op0)(opH)] (L3) {};
\node[l2box, fit=(z0)(a0)(z1)(a1)(z2)(zd)(zH)] (L2) {};
\node[l1box, fit=(z0)(a0)(z1)] (L1) {};
\end{scope}

\node[l1box, minimum width=0.6cm, minimum height=0.25cm, inner sep=0pt] (legL1) at (0.2, -3.5) {};
\node[anchor=base west, font=\small\bfseries, text=blue!60!black]  at (0.5, -3.55) {L1 Predictor};

\node[l2box, minimum width=0.6cm, minimum height=0.25cm, inner sep=0pt] (legL2) at (1.6, -3.5) {};
\node[anchor=base west, font=\small\bfseries, text=green!50!black] at (1.9, -3.55) {L2 Simulator};

\node[l3box, minimum width=0.6cm, minimum height=0.25cm, inner sep=0pt] (legL3) at (3.0, -3.5) {};
\node[anchor=base west, font=\small\bfseries, text=red!65!black]   at (3.3, -3.55) {L3 Evolver};

\end{tikzpicture}%
}
\caption{\textbf{Unified POMDP graphical model of L1-L3.} Dashed circles denote hidden environment states $x$; double circles denote learned latent states $z$; shaded circles denote observations $o$; squares denote actions $a$. Blue solid arrows denote the learned model (inference $q_\phi$ and dynamics $p_\theta$); dashed gray arrows denote the environment transition $T$ and observation emission. The top block shows the agent's POMDP under the current environment $\mathcal{E}\!\sim\!\mathcal{X}$ with model $\mathcal{M}_t$; the bottom block shows the same structure under a revised environment $\mathcal{E}'\!\sim\!\mathcal{X}'$ with model $\mathcal{M}_{t+1}$, obtained via the red \textbf{reflect} arrow. Colored dashed boxes mark each level's scope: L1 covers a single-step latent transition $p_\theta(z_t\mid z_{t-1}, a_{t-1})$; L2 covers the full trajectory rollout $\hat{p}(\tau\mid z_0, a_{1:H}, c)$ under a fixed model; L3 covers evidence-driven model revision $\mathcal{M}_t\to\mathcal{M}_{t+1}$, which corresponds to moving from $\mathcal{X}$ to a revised environment $\mathcal{X}'$ when the current model systematically fails.}
\label{fig:l1l2l3_graphical}
\end{figure*}

In scientific discovery, model updates arise at multiple scales: small 
anomalies trigger local modifications, while persistent discrepancies 
such as the ``two dark clouds''~\citep{kelvin1901nineteenth} in late 
19th-century physics expose epistemic gaps that force revisions to a 
theory's \emph{invariance structure}. The shift from Newtonian to 
relativistic mechanics, for instance, replaced Galilean invariance with 
Lorentz invariance. Modern ML systems also encode invariances, such as 
translation equivariance in convolutions and shape bias in attention-based 
models~\citep{geirhos2018imagenet}, but do so \emph{implicitly}, through 
architecture and training, rather than as explicitly modifiable 
structures. This suits L1 prediction and L2 simulation under a fixed 
model, but at L3 (where the task is to revise the model structure 
itself) it becomes a liability. Symbolic representations, by contrast, 
expose governing principles as first-class objects that can be directly 
inspected and modified.

{We therefore take representation to be a foundational question 
about what a world model \emph{is}, not a choice among interchangeable 
designs. Latent dynamics are indispensable as a scaffold for L1 and L2, 
but the endpoint of L3, namely genuine revision of governing laws, 
requires a symbolic substrate. On this view, L1$\rightarrow$L2$\rightarrow$L3 
is a progression not only in rollout depth, but in how laws are 
discovered, composed, and revised.} Practical instantiations or implmentations across 
regimes are surveyed in Section~\ref{sec:implementation}.
In the next Section~\ref{subsec:formal_defs}, we introduce a foundational formalism that is instantiation-agnostic.

\begin{table}[!t]
\caption{\textbf{Notation summary used in this paper.}}
\label{tab:notation_summary}
\centering
\small
\setlength{\tabcolsep}{2mm}
\begin{tabularx}{\textwidth}{@{}p{0.28\textwidth}X@{}}
\toprule
\textbf{Symbol} & \textbf{Definition} \\
\midrule
\multicolumn{2}{c}{\textit{Environment}} \\
$\mathcal{E} = (\mathcal{X}, \mathcal{A}, \Omega, T, O, R, \gamma)$ & POMDP environment tuple \\
$x_t$ & Hidden environment state at time $t$ \\
$o_t$ & Observation at time $t$ (pixels, tokens, audio, etc.) \\
$a_t$ & Action at time $t$ \\
$T(x_{t+1}\mid x_t, a_t)$ & Environment transition kernel \\
$O(o_t \mid x_t)$ & Environment observation (emission) model \\
$R,\;\gamma$ & Reward function and discount factor \\
\midrule
\multicolumn{2}{c}{\textit{Learned World-Model Components}} \\
$z_t$ & Learned latent / internal state \\
$q_\phi(z_t \mid o_{\le t}, a_{\le t-1})$ & State inference (encoder / filter); parameters $\phi$ \\
$p_\theta(z_t \mid z_{t-1}, a_t)$ & Forward dynamics (one-step latent transition); parameters $\theta$ \\
$p_\psi(o_t \mid z_t)$ & Observation decoder; parameters $\psi$ \\
$\pi_\eta(a_t \mid z_{t-1}, z_t)$ & Inverse dynamics model; parameters $\eta$ \\
$\hat p(\cdot)$ & Trajectory-level (composed) distribution; hat marks approximate object \\
\midrule
\multicolumn{2}{c}{\textit{Trajectories and Planning}} \\
$a_{1:H}=(a_1,\ldots,a_H)$ & Action sequence of horizon length $H$ \\
$\tau=(z_1,\dots,z_H)$ & Future latent segment (anchored at $z_0$) \\
$\hat p(\tau \mid z_0, a_{1:H}, c)$ & L2 rollout query: trajectory distribution conditioned on anchor, actions, and constraints $c$ \\
$b_t;\;\mathrm{Bel}(b_t, a_t, o_{t+1})$ & Classical belief state and Bayesian belief update \\
$\pi$ & Policy (consumes world-model queries; not part of the world-model factorization) \\
\midrule
\multicolumn{2}{c}{\textit{L3 Model Revision}} \\
$\mathcal{M}_t$ & World-modeling stack at revision step $t$ \\
$d_t$ & Deployment evidence (trajectories, errors, tests) \\
$\mathcal{H}$ & Hypothesis space for model revision \\
\bottomrule
\end{tabularx}
\end{table}

\subsection{Notations}
\label{subsec:notation_foundations}

The preceding section proposed three capability stages from epistemological intuition.
We now fix a unified symbol system, and Section~\ref{subsec:formal_defs} uses it to give each stage a precise definition.
To cover model-based RL, predictive representation learning, video/world simulation, and generative modeling, we ground the notation in a Partially Observable Markov Decision Process (POMDP)~\citep{kaelbling1998pomdp, puterman1994mdp}. Figure~\ref{fig:l1l2l3_graphical} places this POMDP structure at the heart of the three-level taxonomy: each capability stage is visualized as a highlighted scope on the same graphical model.
The environment is denoted by the tuple
\[
\mathcal{E} = (\mathcal{X}, \mathcal{A}, \Omega, T, O, R, \gamma),
\]
where \(\mathcal{X}\) is the (unobserved) state space, \(\mathcal{A}\) the action space, and \(\Omega\) the observation space (pixels, tokens, audio, etc.).
Transitions and observations follow
\[
x_{t+1} \sim T(x_{t+1}\mid x_t, a_t),
\qquad
o_t \sim O(o_t \mid x_t).
\]
Under partial observability, agents maintain a belief \(b_t\) or a learned latent state \(z_t\).
Classical belief updates are written \(b_{t+1} = \mathrm{Bel}(b_t, a_t, o_{t+1})\); we reserve the symbol~\(\tau\) for latent trajectories below.
Learned systems infer latents from history:
\[
z_t = f_\phi(o_{\le t}, a_{\le t-1})
\quad\text{or}\quad
q_\phi(z_t \mid o_{\le t}, a_{\le t-1}).
\]

\begin{itemize}[leftmargin=*, nosep]
  \item \(T, O\): environment transition and observation mechanisms.
  \item \(q_\phi(\cdot)\): inference (history \(\rightarrow\) latent).
  \item \(p_\theta(\cdot)\): learned local predictive or generative factors (one-step dynamics, decoders, etc.), with parameters \(\theta\) (and analogously \(\phi,\psi\) for inference and rendering).
  \item \(\hat p(\cdot)\): trajectory-level (or otherwise composed) distributions; the hat marks an explicit approximate object, e.g.\ the rollout marginal induced by repeated application of \(p_\theta\).
  \item \(\pi,\,R,\,\gamma\): planner / policy, reward, and discount. These consume world-model queries but are not part of the world-model factorization \((q_\phi,p_\theta,p_\psi)\); the conceptual separation is discussed in Section~\ref{subsubsec:wm_vs_planner}.
\end{itemize}
\textbf{Convention:} \(\hat p\) is reserved for composed objects such as \(\hat p(\tau\mid z_0,a_{1:H},c)\); ordinary one-step dynamics are always written \(p_\theta(z_t\mid z_{t-1},a_t)\). Table~\ref{tab:notation_summary} provides a concise reference for the symbols used in this paper.

\(a_{1:H}=(a_1,\ldots,a_H)\) denotes an action sequence of length \(H\) applied starting immediately after an anchor state \(z_0\). The future segment is
\[
\tau \;=\; (z_1,z_2,\ldots,z_H),
\]
so that \(\hat p(\tau \mid z_0, a_{1:H}, c)\) matches the L2 formalism in Section~\ref{sec:l2}.
From an arbitrary time index \(t\), the same convention applies after a trivial shift: anchor at \(z_t\), condition on \(a_{t+1:t+H}\).

\subsection{Definitions of Capabilities}
\label{subsec:formal_defs}

With the symbol system established in Section~\ref{subsec:notation_foundations}, we now give each capability stage a precise definition with testable boundary conditions.
\begin{definition}[L1 Predictor]
An L1 world model provides local predictive operators that factorize into up to four components:
\begin{align}
\text{Inference / filtering:} \quad & q_\phi(z_t \mid o_{\le t}, a_{\le t-1}), \label{eq:l1_inf} \\
\text{Forward dynamics:} \quad & p_\theta(z_t \mid z_{t-1}, a_t)
\quad \text{or, without actions, } p_\theta(z_t\mid z_{t-1}), \label{eq:l1_dyn} \\
\text{Observation decoder:} \quad & p_\psi(o_t \mid z_t), \label{eq:l1_dec} \\
\text{Inverse dynamics:} \quad & \pi_\eta(a_t \mid z_{t-1}, z_t). \label{eq:l1_inv}
\end{align}
\end{definition}

These operators target \textbf{one-step} (or short-horizon) accuracy under the training distribution; no guarantee is made about the coherence of multi-step composition.
Section~\ref{sec:l1} presents representative methods in detail.

\begin{definition}[L2 Simulator]
An L2 world model extends L1 from local operators to \textbf{decision-usable multi-step simulation}.
It must support trajectory-level queries of the form
\[
\hat p(\tau \mid z_0, a_{1:H}, c), \qquad \tau=(z_1,\ldots,z_H),
\]
subject to three boundary conditions that collectively mark L1~$\to$~L2:
\begin{enumerate}[leftmargin=*]
  \item \textbf{Long-horizon coherence:} rollouts remain usable over \(H\) steps rather than degrading immediately via compounding error.
  \item \textbf{Intervention sensitivity:} counterfactual edits (action or premise changes) induce stable and directionally meaningful trajectory changes.
  \item \textbf{Constraint consistency:} generated futures respect the governing laws of the target regime (the physical, digital, social, or scientific world).
\end{enumerate}
\end{definition}

The key difference from L1 is not one-step quality but \textbf{rollout fidelity under composition}.

The three L2 boundary conditions are complementary rather than redundant. Long-horizon coherence concerns whether rollout quality survives composition over time; intervention sensitivity concerns whether changes in actions or premises induce stable and directionally meaningful changes in the predicted future; and constraint consistency concerns whether the resulting trajectories remain valid under the governing laws of the target regime. None of these implies the others in general: a model may generate coherent but action-insensitive rollouts, or action-sensitive rollouts that still violate domain constraints. In practice they can also trade off against one another, for example when aggressive constraint enforcement stabilizes trajectories at the cost of reduced responsiveness to interventions.

A fourth capability, \textbf{closed-loop use} (supporting planning, acting, and self-improvement through interaction with the modeled environment), further separates world modeling from generic prediction but is orthogonal to L1/L2/L3: a weather emulator can be an L2 world model with no embedded planner (see Appendix~\ref{app:boundaries} for extended discussion).
We reserve ``closed-loop'' for two different senses that must not be conflated: using a world model inside a control or planning loop is an orthogonal deployment property, whereas revising the world-model stack itself from deployment evidence is the defining hallmark of L3.

\begin{definition}[L3 Evolver]
An L3 world model extends L2 from rollout over a fixed scaffold to \textbf{evidence-driven model revision}.
In addition to simulation queries, an L3 system maintains an explicit update loop over model assets:
\[
(\mathcal{M}_t,\; d_t) \;\xrightarrow{\;\text{diagnose\,+\,distill\,+\,validate}\;}\; \mathcal{M}_{t+1},
\]
where \(\mathcal{M}_t\) is the current world-modeling stack at revision step \(t\) and \(d_t\) is new deployment evidence (trajectories, errors, counterexamples, tests).
Three boundary conditions mark L2~$\to$~L3:
\begin{enumerate}[leftmargin=*]
  \item \textbf{Evidence-grounded diagnosis:} failures are attributed to actionable causes using replayable evidence.
  \item \textbf{Persistent asset update:} fixes are promoted as reusable assets (skills, rules, parsers, tests), not only ephemeral in-context patches.
  \item \textbf{Governed validation:} updates pass regression and robustness gates (including rollback and canary policies) before default enablement.
\end{enumerate}
\end{definition}

The key difference from L2 is that the \textbf{model itself becomes an object of revision}, not merely a fixed scaffold to be queried~\citep{lu2024aiscientist, boiko2023autonomous}. Recapping the scopes in Figure~\ref{fig:l1l2l3_graphical}: \textbf{L1 (Predictor)} is a single-step transition $p_\theta(z_t\mid z_{t-1}, a_{t-1})$ with its supporting inference and decoding operators, acting locally on one edge of the latent chain; \textbf{L2 (Simulator)} composes those local operators into a trajectory $\hat p(\tau\mid z_0, a_{1:H}, c)$ under a fixed model $\mathcal{M}_t$ and governing-law constraint $c$; and \textbf{L3 (Evolver)} revises the model stack $\mathcal{M}_t\to\mathcal{M}_{t+1}$ from distilled evidence $d_t$, yielding a different latent graph (the bottom block of the figure) whose effective environment $\mathcal{E}'\!\sim\!\mathcal{X}'$ may differ from the original, whether because the world itself has shifted, because the agent has uncovered previously unmodeled structure, or because the hypothesis space has been expanded. The three levels form a containment hierarchy: L2 invokes L1 at each step, and L3 invokes L2 each time it probes the world for evidence before committing to a model update.

\paragraph{Agent-centered view: state, action, and task.}
The formal components above describe an agent whose decisions are determined by three elements: the state it believes the world to be in, the action it can execute, and the task (or constraint $c$) it must satisfy. This triple, not a flat observation-to-action mapping, defines the interface between world model and planner.
Building a useful $z_t$ involves two orthogonal challenges that structure Section~\ref{sec:l1}:
(i)~\emph{spatial representation}: compressing a high-dimensional observation $o_t$ into a compact latent that retains decision-relevant structure (geometry, semantics, affordances), and
(ii)~\emph{temporal fusion}: integrating history $(o_{\le t}, a_{\le t-1})$ so that $z_t$ approximates a Markov belief even in partially observable settings.
Actions are not flat variables: they can emerge from representation learning rather than being pre-defined, with the core dynamics captured by the latent representation and everything else serving as a decoder~\citep{lecun2022path}.
Real agent behavior decomposes across temporal scales and abstraction levels, including low-level motor primitives, mid-level skills, and high-level task plans. The world model must predict transitions at the granularity that matches the planner's query horizon. This action hierarchy interacts directly with the L1$\to$L2 boundary: local dynamics suffice for primitive-level prediction~\citep{sun2025learningprimitiveembodiedworld}, but skill- and task-level rollouts require the multi-step coherence that defines L2.
At the L3 level, the agent must not only predict transitions across temporal scales but also decide when its own transition model is inadequate and initiate model revision. L3 treats the world-modeling stack itself as an object of action. Diagnostic probes, architecture modifications, and regression tests become ``meta-actions'' that operate on the model rather than on the environment itself, reshaping how the system learns rather than merely how it acts.

\subsection{Scope of Laws}
\label{subsec:scope_regimes}

As introduced in Section~\ref{sec:scope}, we organize the survey along two orthogonal axes: capability level (L1/L2/L3) and governing-law regime. This subsection elaborates the four regimes and the constraints each imposes on the learned transition function.
We distinguish \textbf{Laws of the Physical World} (governing agents that perceive and act in physical environments),
\textbf{Laws of the Digital World} (governing deterministic program semantics: code, APIs, and state machines),
\textbf{Laws of the Social World} (governing the dynamics of minds and institutions: beliefs, goals, and norms),
and \textbf{Laws of the Scientific World} (governing systems that exist independently of human design, whose dynamics must be discovered from empirical observation).
These four regimes are representative, not exhaustive. Real-world systems often operate under multiple regimes simultaneously.
For example, autonomous driving involves both physical dynamics and social norms, while drug design couples natural mechanisms with digital simulation pipelines.

\textbf{Laws of the Physical World} constrain transitions through the physical dynamics that embodied agents must respect: contact mechanics, collision response, gravitational acceleration, friction, and kinematic feasibility.
In robotics manipulation, autonomous driving, and interactive 3D simulation, the learned transition $p_\theta(z_t \mid z_{t-1}, a_t)$ must encode these physical interactions faithfully~\citep{todorov2012mujoco, hu2023gaia1, wang2024drivedreamer}.
This regime is distinguished by analytically characterizable governing equations. A physics engine or analytic model can verify whether a predicted transition is consistent with rigid-body constraints and Newtonian mechanics.
Constraint violations appear as objects passing through each other, gravity reversing mid-rollout, or physically impossible deformations. Such failures are immediately detectable because the ground-truth dynamics admit closed-form or numerically exact reference solutions.

\textbf{Laws of the Digital World} constrain transitions through deterministic program semantics, including API contracts, UI state machines, file-system logic, and network protocols.
In web navigation, code generation, and software testing, the transition function $p_\theta(z_t \mid z_{t-1}, a_t)$ is largely \emph{deterministic} but branches heavily through error codes, permission checks, and edge cases~\citep{gu2024webdreamer, yao2022webshop}.
This regime is defined by transitions that are both \emph{specifiable} and \emph{verifiable}. The program can be executed and its output compared against the model's prediction.
Constraint violations appear as producing an API call that does not exist, ignoring returned error codes, or violating type constraints. Because the underlying system is a formal artifact, such errors are mechanically checkable.

\textbf{Laws of the Social World} constrain transitions through beliefs, goals, norms, social contracts, and institutional rules.
In social simulation, dialogue systems, and multi-agent interaction, $p_\theta(z_t \mid z_{t-1}, a_t)$ maps joint actions and mental states to new mental states and social outcomes~\citep{park2023generative, zhou2025socialwm}.
Two properties set this regime apart. Transitions are \emph{reflexive}, meaning that agents' beliefs about the state actively change the state itself. They are also \emph{normative}, governed not only by what will happen but by what should happen according to shared conventions.
Constraint violations appear as breaking a promise without consequence, forgetting a prior commitment, or ignoring established social norms. Such failures undermine coherence because social outcomes depend on mutual expectation.

\textbf{Laws of the Scientific World} constrain transitions through latent causal mechanisms that must be \emph{discovered} from empirical observation rather than specified \emph{a priori}.
In weather prediction, molecular dynamics, protein folding, and drug design, $p_\theta(z_t \mid z_{t-1}, a_t)$ encodes atmospheric dynamics, chemical kinetics, or biological processes whose exact functional forms are unknown or too complex to write analytically~\citep{karniadakis2021pinn, lam2023graphcast, abramson2024alphafold3}.
This regime differs in that the governing equations are \emph{not available in closed form}. The world model must learn them from data and be validated against experimental measurement.
Constraint violations appear as predicting physically impossible molecular configurations, violating conservation laws that hold empirically, or ignoring known causal dependencies. Detection typically requires comparison with laboratory or observational data rather than symbolic verification.

With these foundations in place, the following sections instantiate each capability level in turn: Section~\ref{sec:l1} surveys L1 methods, Section~\ref{sec:l2} addresses L2 simulation, and Section~\ref{sec:l3} examines L3 model revision. Appendix~\ref{app:boundaries} clarifies the distinctions between world modeling and generic prediction, world models and planners, and world modeling and the commonsense reasoning that agents rely on in unscripted settings.

\section{L1 Predictor: Local Markov Prediction}
\label{sec:l1}

The hierarchical structure begins with \textbf{L1}, which assesses a world model's local predictive ability by requiring it to sustain a meaningful internal state and use \emph{local} predictive mechanisms to anticipate the next state, including potential observations or actions. In the unified graphical model of Figure~\ref{fig:l1l2l3_graphical}, L1 is the scope of a single edge $z_{t-1}\to z_t$ conditioned on action $a_{t-1}$; everything in this section elaborates the operators that populate this one-step transition and examines how they are realized in contemporary world-model systems.

\subsection{Definition}
\label{subsec:l1_definition}
L1 concerns the local predictive ability of a world model for an agent acting in an environment to accomplish a task or goal. More precisely, an agent is a system that, given observations, makes decisions and takes actions in order to satisfy an objective. In this paper, the role of an L1 world model is therefore not merely to predict the next signal, but to provide local predictive operators that support such decision-making at the granularity of one step (or a short fixed horizon). This epistemic stance aligns with Hume's constant conjunction: regularities are extracted from observed data without claiming causal necessity (Section~\ref{sec:philosophy_l1}).

The POMDP formulation that underlies L1 originates from the reinforcement learning literature, where an agent must select actions under partial observability to maximize cumulative reward~\citep{kaelbling1998pomdp, puterman1994mdp}. In this setting, the agent maintains an internal belief over hidden states and crafts a policy $\pi(a_t \mid b_t)$ that maps beliefs to actions. This formulation constitutes the prototypical agent--environment loop~\citep{sutton1991dyna}. For an agent that interacts with the environment to accomplish a task, the POMDP decomposes into four local operators: state inference, forward dynamics, observation decoding, and inverse dynamics. Together these describe the foundational learning problems for world models at the L1 level.

Following this formulation (Section~\ref{sec:preliminaries}), \textbf{L1} is characterized by \textbf{local predictive operators} operating on a learned internal state $z_t$ (resembling a belief state), where the central modeling concept centers on a \textbf{one-step} (or short fixed-horizon) transition operator. In practical terms, $z_t$ is deduced from observations and actions and functions as a learned approximation to the latent environmental state and/or belief~\citep{hafner2023dreamerv3, schrittwieser2020muzero}. The concept of learning such latent dynamics can be traced back to locally linear latent models for control~\citep{watter2015e2c} and Gaussian-process dynamics~\citep{deisenroth2011pilco}, and has been enhanced by contemporary deep learning architectures~\citep{ha2018worldmodels, hafner2019dreamer}. The term ``Markov'' in \textbf{L1} denotes the \emph{Markovian property in the learned internal state} $z_t$, indicating that $z_t$ is adequate (or nearly adequate) for predicting the subsequent local step, rather than the direct observability of the environmental state~\citep{hafner2019recurrent, hafner2023dreamerv3, gelada2019deepmdp}.

At the model level, L1 factorizes into four local operators over $z_t$ (Table~\ref{tab:l1_architecture}). The \emph{core} operator is latent dynamics ($z_{t-1} \to z_t$); the others are common supporting operators:
\begin{itemize}[leftmargin=*, nosep]
    \item \textbf{State inference} (observation $\to$ state, Eq.~\eqref{eq:l1_inf}):
    $z_t = f_\phi(o_{\le t}, a_{\le t-1})$ or $q_\phi(z_t \mid o_{\le t}, a_{\le t-1})$.
    The learned belief-like state summarizes relevant history for prediction~\citep{hafner2019recurrent, lesort2018srl}.

    \item \textbf{Forward dynamics} (state $\to$ next state; core L1 operator, Eq.~\eqref{eq:l1_dyn}):
    $z_t \sim p_\theta(z_t \mid z_{t-1}, a_t)$ (action-conditioned) or $z_t \sim p_\theta(z_t \mid z_{t-1})$ (action-free).

    \item \textbf{Observation decoding} (state $\to$ observation, Eq.~\eqref{eq:l1_dec}):
    $p_\psi(o_t \mid z_t)$, mapping latent state back to observation space~\citep{kingma2014vae, rezende2014stochastic}.

    \item \textbf{Inverse dynamics} (Eq.~\eqref{eq:l1_inv}):
    $\pi_\eta(a_t \mid z_{t-1}, z_t)$, used as an auxiliary objective or for representation shaping~\citep{pathak2017curiosity, hafner2019dreamer}.
\end{itemize}

\begin{table}[!t]
\caption{\textbf{L1 component factorization.} The four local operators form the building blocks of L1 world models. The core operator is forward dynamics; the others are supporting operators.}
\label{tab:l1_architecture}
\centering
\renewcommand{\arraystretch}{1.3}
\setlength{\tabcolsep}{1mm}
\begin{tabularx}{\textwidth}{lllX}
\toprule
\textbf{Operator} & \textbf{Mapping} & \textbf{Formal Definition} & \textbf{Role} \\
\midrule
State Inference
& $o_t \to z_t$
& \shortstack[l]{$z_t = f_\phi(o_{\le t}, a_{\le t-1})$ \\
or $q_\phi(z_t \mid o_{\le t}, a_{\le t-1})$}
& \shortstack[l]{Compress observations into latent belief state} \\Forward Dynamics & $z_{t-1} \to z_t$ & $p_\theta(z_t \mid z_{t-1}, a_t)$ & Predict next latent state given action \\
Observation Decoding & $z_t \to o_t$ & $p_\psi(o_t \mid z_t)$ & \shortstack[l]{Reconstruct observations as training signals} \\
Inverse Dynamics & $(z_{t-1}, z_t) \to a_t$ & $\pi_\eta(a_t \mid z_{t-1}, z_t)$ & Infer actions; representation shaping \\
\bottomrule
\end{tabularx}%
\end{table}

\begin{table*}[!t]
\caption{\textbf{Representative L1 methods.} Columns indicate which local operators each method instantiates: state inference~(SI), forward dynamics~(FD), observation decoding~(OD), and inverse dynamics~(ID).}
\label{tab:l1_methods}
\centering
\setlength{\tabcolsep}{0.75mm}
\begin{tabular}{l|cc|ccccl}
\toprule
\textbf{Method} & \multicolumn{2}{c|}{\textbf{Links}} & \textbf{SI} & \textbf{FD} & \textbf{OD} & \textbf{ID} & \textbf{Architecture} \\
\midrule
\rowcolor{gray!15}
\multicolumn{8}{c}{\textit{Representation Learning}} \\
VAE~\citep{kingma2014vae} & \paperlink{https://arxiv.org/abs/1312.6114} & {--} & \cmark & \xmark & \cmark & \xmark & MLP encoder--decoder \\
$\beta$-VAE~\citep{higgins2017betavae} & \paperlink{https://openreview.net/forum?id=Sy2fzU9gl} & {--} & \cmark & \xmark & \cmark & \xmark & MLP encoder--decoder \\
VQ-VAE~\citep{oord2017vqvae} & \paperlink{https://arxiv.org/abs/1711.00937} & \githublink{https://github.com/MishaLaskin/vqvae} & \cmark & \xmark & \cmark & \xmark & CNN + discrete codebook \\
CPC~\citep{oord2018cpc} & \paperlink{https://arxiv.org/abs/1807.03748} & {--} & \cmark & \xmark & \xmark & \xmark & CNN + autoregressive \\
SimCLR~\citep{chen2020simclr} & \paperlink{https://arxiv.org/abs/2002.05709} & \githublink{https://github.com/google-research/simclr} & \cmark & \xmark & \xmark & \xmark & ResNet + projection head \\
MoCo~\citep{he2020moco} & \paperlink{https://arxiv.org/abs/1911.05722} & \githublink{https://github.com/facebookresearch/moco} & \cmark & \xmark & \xmark & \xmark & Momentum encoder \\
CURL~\citep{srinivas2020curl} & \paperlink{https://arxiv.org/abs/2004.04136} & \githublink{https://github.com/MishaLaskin/curl} & \cmark & \xmark & \xmark & \xmark & CNN+momentum encoder \\
SPR~\citep{schwarzer2021spr} & \paperlink{https://arxiv.org/abs/2007.05929} & \githublink{https://github.com/mila-iqia/spr} & \cmark & \xmark & \xmark & \xmark & CNN + prediction MLP \\
I-JEPA~\citep{assran2023ijepa} & \paperlink{https://arxiv.org/abs/2301.08243} & \githublink{https://github.com/facebookresearch/ijepa} & \cmark & \xmark & \xmark & \xmark & ViT + predictor \\
V-JEPA~\citep{bardes2024vjepa} & \paperlink{https://arxiv.org/abs/2404.08471} & \githublink{https://github.com/facebookresearch/jepa} & \cmark & \xmark & \xmark & \xmark & ViT + predictor \\
DINOv2~\citep{oquab2024dinov2} & \paperlink{https://arxiv.org/abs/2304.07193} & \githublink{https://github.com/facebookresearch/dinov2} & \cmark & \xmark & \xmark & \xmark & ViT self-distillation \\
\midrule
\rowcolor{gray!15}
\multicolumn{8}{c}{\textit{Model-Based RL}} \\
PILCO~\citep{deisenroth2011pilco} & \paperlink{https://dl.acm.org/doi/10.5555/3104482.3104541} & \githublink{https://github.com/UCL-SML/pilco-matlab} & \xmark & \cmark & \xmark & \xmark & Gaussian process \\
E2C~\citep{watter2015e2c} & \paperlink{https://arxiv.org/abs/1506.07365} & {--} & \cmark & \cmark & \cmark & \xmark & Locally linear latent \\
PETS~\citep{chua2018pets} & \paperlink{https://arxiv.org/abs/1805.12114} & \githublink{https://github.com/kchua/handful-of-trials} & \xmark & \cmark & \xmark & \xmark & Ensemble of NNs \\
World Models~\citep{ha2018worldmodels} & \paperlink{https://arxiv.org/abs/1803.10122} & \githublink{https://github.com/hardmaru/WorldModelsExperiments} & \cmark & \cmark & \cmark & \xmark & VAE + MDN-RNN \\
Dreamer~\citep{hafner2019dreamer} & \paperlink{https://arxiv.org/abs/1912.01603} & \githublink{https://github.com/danijar/dreamer} & \cmark & \cmark & \cmark & \xmark & RSSM (GRU + stoch.) \\
DreamerV2~\citep{hafner2020dreamerv2} & \paperlink{https://arxiv.org/abs/2010.02193} & \githublink{https://github.com/danijar/dreamerv2} & \cmark & \cmark & \cmark & \xmark & RSSM (discrete stoch.) \\
DreamerV3~\citep{hafner2023dreamerv3} & \paperlink{https://arxiv.org/abs/2301.04104} & \githublink{https://github.com/danijar/dreamerv3} & \cmark & \cmark & \cmark & \xmark & RSSM + symlog \\
MuZero~\citep{schrittwieser2020muzero} & \paperlink{https://arxiv.org/abs/1911.08265} & {--} & \cmark & \cmark & \xmark & \xmark & MLP dynamics + MCTS \\
EfficientZero~\citep{ye2021efficientzero} & \paperlink{https://arxiv.org/abs/2111.00210} & \githublink{https://github.com/YeWR/EfficientZero} & \cmark & \cmark & \xmark & \xmark & MuZero + self-sup.\ \\
TD-MPC2~\citep{hansen2024tdmpc2} & \paperlink{https://arxiv.org/abs/2310.16828} & \githublink{https://github.com/nicklashansen/tdmpc2} & \cmark & \cmark & \xmark & \xmark & MLP latent dynamics \\
DeepMDP~\citep{gelada2019deepmdp} & \paperlink{https://arxiv.org/abs/1906.02736} & {--} & \cmark & \cmark & \xmark & \xmark & Bellman-aligned latent \\
MBPO~\citep{janner2019mbpo} & \paperlink{https://arxiv.org/abs/1906.08253} & \githublink{https://github.com/jannerm/mbpo} & \xmark & \cmark & \xmark & \xmark & Ensemble of NNs \\
\midrule
\rowcolor{gray!15}
\multicolumn{8}{c}{\textit{Token / Diffusion-Based}} \\
IRIS~\citep{micheli2023iris} & \paperlink{https://arxiv.org/abs/2209.00588} & \githublink{https://github.com/eloialonso/iris} & \cmark & \cmark & \cmark & \xmark & VQ-VAE + Transformer \\
TransDreamer~\citep{chen2022transdreamer} & \paperlink{https://arxiv.org/abs/2202.09481} & \githublink{https://github.com/changchencc/TransDreamer} & \cmark & \cmark & \cmark & \xmark & Transformer-XL + stoch. \\
Latent Diffusion~\citep{rombach2022latentdiffusion} & \paperlink{https://arxiv.org/abs/2112.10752} & \githublink{https://github.com/CompVis/latent-diffusion} & \cmark & \xmark & \cmark & \xmark & Latent-space diffusion \\
STORM~\citep{zhang2023storm} & \paperlink{https://arxiv.org/abs/2310.09615} & \githublink{https://github.com/weipu-zhang/STORM} & \cmark & \cmark & \cmark & \xmark & Transformer + VAE \\
DIAMOND~\citep{alonso2024diamond} & \paperlink{https://arxiv.org/abs/2405.12399} & \githublink{https://github.com/eloialonso/diamond} & \xmark & \cmark & \xmark & \xmark & Pixel-space diffusion \\
Delta-IRIS~\citep{micheli2024deltairis} & \paperlink{https://arxiv.org/abs/2406.19320} & \githublink{https://github.com/vmicheli/delta-iris} & \cmark & \cmark & \cmark & \xmark & VQ-VAE + delta coding \\
\bottomrule
\end{tabular}%
\end{table*}

\subsection{Approaches}
\label{subsec:l1_methods}

We categorize notable \textbf{L1} techniques based on the four local operators delineated earlier: state inference (concerned with deriving $z_t$ from observations and historical data), forward dynamics (the fundamental transition model), observation decoding (associating $z_t$ with $o_t$), and inverse dynamics (deducing actions from successive states)~\citep{ding2024survey_wm, moerland2023mbrl}. Table~\ref{tab:l1_methods} summarizes representative methods and their key innovations. We devote the most space to forward dynamics because it is the operator that most directly determines whether an L1 system can later be elevated into an L2 simulator; the other components are still essential, but their role is primarily to make the latent state usable for that transition.

\subsubsection{State Inference}

State inference condenses high-dimensional observations into a compact latent representation $z_t$ that preserves crucial decision-making information, and it integrates temporal context to ensure that $z_t$ approximates a Markovian belief in partially observable scenarios~\citep{lesort2018srl}.

Contrastive Predictive Coding (CPC;~\citealt{oord2018cpc}) trains an encoder to maximize mutual information between present and future embeddings via the InfoNCE loss, which contrasts temporally adjacent positive pairs against negatives drawn from the same batch. SimCLR~\citep{chen2020simclr} and MoCo~\citep{he2020moco} established general-purpose contrastive frameworks through augmentation-based positive pairs and momentum-updated encoders, respectively, providing pretrained visual backbones that downstream world models build upon. However, general visual representations do not guarantee that $z_t$ preserves control-relevant information. CURL~\citep{srinivas2020curl} addressed this by extending contrastive learning to RL with temporal adjacency between consecutive frames as positive pairs, achieving sample efficiency comparable to model-based methods in Atari and continuous control. Self-Predictive Representations (SPR;~\citealt{schwarzer2021spr}) trains an encoder to forecast its own future representations, incorporating temporal structure and decision-relevant dynamics. Both methods show strong sample efficiency on Atari 100k, confirming that world model representations benefit from being tailored to decision-making objectives.

Rather than contrasting pairs, another family of methods predicts the embedding of masked regions directly in latent space. JEPA and its variants, I-JEPA~\citep{assran2023ijepa} and V-JEPA~\citep{bardes2024vjepa}, forecast hidden-region embeddings without decoding back to pixels~\citep{lecun2022path}. This approach encourages the encoder to grasp semantic and structural consistencies without being bound to intricate reconstruction at the pixel level. On another front, DINOv2~\citep{oquab2024dinov2} from the foundation-model domain generates versatile visual features through self-distillation, establishing robust state encoders for subsequent tasks. A complementary direction makes the inferred state explicitly object-centric and programmatic rather than purely continuous. Thinking with Blueprints converts an image into a JSON-style blueprint that records the positions, sizes, and attributes of question-relevant objects, and then reasons over this structured representation to answer spatial queries~\citep{ma2026thinking}. Although proposed for VLM spatial reasoning rather than sequential control, it is highly relevant to L1 state inference because it shows that useful internal state can take the form of a decision-oriented scene description, not only a dense latent embedding.

A third line of work shapes $z_t$ through control-oriented auxiliary objectives such as reward anticipation, inverse-model losses~\citep{pathak2017curiosity}, and value-function consistency, as formalized in DeepMDP~\citep{gelada2019deepmdp}. This framework articulates the necessity for the latent Markov chain to approximately adhere to the Bellman equations. Embed to Control (E2C;~\citealt{watter2015e2c}) acquired locally linear latent dynamics simultaneously with a VAE encoder-decoder for LQR-based planning within the latent space.

When a single observation is insufficient, the model must aggregate past information into $z_t$. The Recurrent State Space Model (RSSM) of \citet{hafner2019recurrent} splits the latent into a deterministic recurrent pathway $h_t = f(h_{t-1}, z_{t-1}, a_{t-1})$ and a stochastic component $z_t \sim q_\phi(z_t \mid h_t, o_t)$, compressing arbitrary-length histories while preserving stochastic uncertainty. This recurrent belief state $(h_t, z_t)$ serves as the internal state for all downstream prediction and control in the Dreamer family~\citep{hafner2019dreamer, hafner2020dreamerv2, hafner2023dreamerv3}.

Scientific applications illustrate how the same state-inference principle operates when raw observations are high-dimensional and the scientifically meaningful state is latent. In structural biology, protein structure prediction can be cast as L1 state inference: mapping an amino acid sequence (observation) to a dominant 3D coordinate state. The AlphaFold lineage progressed from learned distance-based potentials~\citep{senior2020alphafold} to end-to-end Evoformer architectures with near-experimental accuracy~\citep{jumper2021alphafold} to diffusion-based prediction of joint biomolecular complex structures~\citep{abramson2024alphafold3}. Parallel efforts showed that strong structure prediction is also achievable through three-track networks~\citep{baek2021rosettafold} and protein language models enabling single-sequence inference~\citep{lin2023esmfold}.
In neuroscience, HMM~\citep{baker2014hmm}, RNN~\citep{gohil2022dynemo} and Transformer~\citep{khan2023dynemoc} are used to map electrophysiological recordings to a set of latent network modes, following a state-inference paradigm conceptually similar to \citet{ha2018worldmodels}.
Analysis of the learned interpretable latent representations reveals various findings: cortical activity at rest can be described by transient, intermittently recurring events~\citep{vidaurre2018spontaneous} organized into cycles on 300–1,000 ms timescales~\citep{van2025large}.

\subsubsection{Forward Dynamics: The Core L1 Operator}

These approaches directly establish $p_\theta(z_t \mid z_{t-1}, a_t)$ and form the \emph{core} of L1. The precision of the dynamics network is crucial for producing valuable one-step forecasts and must be expressive enough to be aggregated over numerous steps, a critical requirement that is more stringent at L2~\citep{moerland2023mbrl}.

In model-based Reinforcement Learning (RL), the action-conditioned latent transition model plays a pivotal role. PILCO~\citep{deisenroth2011pilco} employed Gaussian-process dynamics coupled with analytical uncertainty propagation for efficient continuous control. MuZero~\citep{schrittwieser2020muzero} adopts a deterministic dynamics function $z_t = f_\theta(z_{t-1}, a_t)$ trained end-to-end for value prediction and Monte Carlo Tree Search without observation reconstruction. EfficientZero~\citep{ye2021efficientzero} further enhanced this approach by incorporating self-supervised consistency losses, achieving superhuman performance in Atari games with just two hours of experience. Conversely, the Recurrent State Space Model (RSSM) in Dreamer~\citep{hafner2019dreamer, hafner2020dreamerv2, hafner2023dreamerv3} leverages stochastic dynamics to facilitate uncertainty-aware rollouts. PETS~\citep{chua2018pets} illustrated that an \emph{ensemble} of dynamics models offers reliable epistemic uncertainty estimation crucial for robust planning. TD-MPC~\citep{hansen2022tdmpc} learns latent dynamics through temporal difference objectives, directly aligning the dynamics model with value estimation; TD-MPC2~\citep{hansen2024tdmpc2} scales up to a single 317 million-parameter agent proficient in mastering 104 tasks across various domains.

A recent trend is the shift from continuous latent dynamics to discrete-token or diffusion-based transitions. For instance, IRIS~\citep{micheli2023iris} tokenizes observations using a VQ-VAE codebook~\citep{oord2017vqvae} and models the resulting sequence with an autoregressive Transformer. Meanwhile, TransDreamer~\citep{chen2022transdreamer} swaps the GRU of RSSM with Transformer-XL to enhance long-range attention. \citet{zhang2023storm} combine Transformer sequence modeling with stochastic VAE dynamics, whereas \citet{micheli2024deltairis} encode stochastic deltas between frames instead of entire frames. On the diffusion front, DIAMOND~\citep{alonso2024diamond} employs diffusion denoising as the one-step transition operator to preserve visual intricacies that may be overlooked by low-capacity latent dynamics.

Building on the principles of predictive coding~\citep{rao1999predictive, friston2010free}, methodologies such as CPC~\citep{oord2018cpc}, SPR~\citep{schwarzer2021spr}, and JEPA~\citep{bardes2024vjepa, assran2023ijepa} forecast the \emph{latent embedding} of the forthcoming observation rather than the observation itself. While SimPLe~\citep{kaiser2019simple} showcased the viability of pixel-level video prediction as a world model for efficient Atari RL, the discrepancy in prediction fidelity between pixel-space and latent-space underscores the progression towards abstract dynamics and the transition from L1 to L2.

Beyond using dreaming for policy optimization, one-step dynamics models also function as generators of experience: MBPO~\citep{janner2019mbpo} integrates short model rollouts into a replay buffer to enhance sample efficiency; \citet{nagabandi2018mbmf} demonstrated that model-based pre-training combined with model-free fine-tuning leverages the strengths of both approaches. DayDreamer~\citep{wu2023daydreamer} transfers latent imagination to physical robots, \citet{wang2024coworld} extend this to transfer reinforcement learning (RL) knowledge across visual domains, and \citet{hao2025mosim} propel world models towards acquiring long-horizon physical skills through neural motion simulation. These applications underscore that L1 dynamics serve as a data catalyst, a planning foundation~\citep{schrittwieser2020muzero, hafner2023dreamerv3, hansen2024tdmpc2}, and a mechanism for compressing raw interactions into high-level behaviors~\citep{moerland2023mbrl}.

\subsubsection{Observation Decoding}

The decoder implements $p_\psi(o_t \mid z_t)$ and has three key functions. It provides a training signal to ensure that $z_t$ preserves sufficient information, serves as a diagnostic interface to examine the model's learned representations, and acts as a rendering engine for generating envisioned observations during dreaming~\citep{ha2018worldmodels, hafner2019dreamer}.

The Variational Autoencoder (VAE;~\citealt{kingma2014vae}, independently~\citealt{rezende2014stochastic}) offers a standard probabilistic framework: the encoder $q_\phi(z_t \mid o_t)$ maps observations to a latent posterior, while the decoder $p_\psi(o_t \mid z_t)$ reconstructs observations from latents, trained simultaneously through the ELBO. In world-modeling workflows, the VAE compresses raw pixel inputs into a concise code transmitted to a dynamics model~\citep{ha2018worldmodels, hafner2019recurrent}. $\beta$-VAE~\citep{higgins2017betavae} amplifies the KL divergence term to encourage disentangled factors, while VQ-VAE~\citep{oord2017vqvae} replaces continuous latents with a discrete codebook, foundational for token-based world models~\citep{micheli2023iris}. The \emph{World Models} concept by \citet{ha2018worldmodels} introduced the fusion of a VAE encoder, an LSTM dynamics model (MDN-RNN), and a distinct controller. Subsequently, the Dreamer lineage~\citep{hafner2019dreamer, hafner2020dreamerv2, hafner2023dreamerv3} expanded this paradigm into an end-to-end latent imagination framework, where policy and value functions are trained solely on imagined latent trajectories, with the decoder acting mainly as a regularizer and auxiliary fidelity check rather than as a generative objective in its own right.

Large-scale video generation models like Sora~\citep{brooks2024sora} leverage high-capacity observation decoders to produce photorealistic frames from latent trajectories. Latent Diffusion Models~\citep{rombach2022latentdiffusion} compress images into a lower-dimensional latent space and apply diffusion processes more efficiently in that space, while the Diffusion Transformer (DiT;~\citealt{peebles2023dit}) enhances scalability by replacing the U-Net backbone with a standard Transformer. Recent image-generation backbones also explore information-adaptive tokenization and generation, such as Dynamic Generative Image Transformer~\citep{mao2026dgit}, which suggests a useful design direction for observation-level decoders even though it is not itself a multi-step world model. These models illustrate the feasibility of achieving high-quality $p_\psi(o_t \mid z_t)$ at scale. However, the quality of the \emph{latent dynamics} that steer the decoder, and the ability of $z_t$ to facilitate coherent multi-step prediction, remain challenges that motivate the transition to the L2 discussion.

\subsubsection{Inverse Dynamics}

The inverse dynamics operator $\pi_\eta(a_t \mid z_{t-1}, z_t)$ deduces the action taken between two consecutive latent states. This operator serves multiple roles in modern world-model systems. \citet{pathak2017curiosity} utilized inverse dynamics as a curiosity-driven exploration strategy, refining the representation to capture only the controllable aspects of the environment. By training the encoder to predict actions between consecutive states, the inverse model filters out exogenous visual noise (e.g., moving clouds, flickering backgrounds) that is irrelevant to the agent's decisions. In a broader context, inverse dynamics acts as an additional training signal that prompts $z_t$ to preserve action-relevant characteristics, complementing forward dynamics and reconstruction as a mechanism for learning decision-useful representations~\citep{lesort2018srl}.

A particularly impactful application of inverse dynamics is retrospective action labeling for large-scale imitation learning. \citet{baker2022video} trained an inverse dynamics model on a small corpus of action-labeled Minecraft gameplay and then applied it to label a vastly larger set of unlabeled internet videos, enabling Video PreTraining (VPT) to learn complex behaviors such as diamond mining from passive observation alone. This pipeline demonstrates that inverse dynamics can bridge the gap between abundant unlabeled video and the action annotations required for behavioral cloning, effectively transforming observation-only data into a usable training signal for policy learning. Inverse dynamics also underpins goal-conditioned policy architectures. Given a current state $z_{t-1}$ and a desired goal state $z_g$, the inverse model predicts the action that would transition toward the goal, providing a natural interface for hierarchical planning where a high-level planner selects subgoals and a low-level inverse model executes them~\citep{ghosh2021learning}. \citet{agrawal2016learning} demonstrated that a robot can learn intuitive physics by ``poking'' objects and training an inverse model to predict the poke parameters from observed state changes, illustrating how inverse dynamics can ground physical understanding through interaction.

Many effective world models exclude inverse dynamics entirely, while others use it as a light regularization technique. A practical but under-discussed issue is action-label quality: when actions are inferred retrospectively rather than logged directly, inverse-model errors accumulate precisely at distribution edges where world models most need reliable supervision. Furthermore, the inverse operator assumes a unique or near-unique action between state pairs, an assumption that breaks down in stochastic environments or when multiple actions lead to the same outcome, limiting its reliability in such settings.

\subsection{Discussion}
\label{subsec:l1_theory_boundaries}

Although this paper focuses on temporal local prediction indexed by time t, the same local-operator view
can apply along non-temporal axes such as diffusion steps, refinement steps, or hierarchical update stages.
We treat these as edge cases of L1 rather than as a distinct capability level, because the key property is still
local transition prediction rather than decision-usable multi-step rollout.

L1 alone does not ensure coherent behavior over long horizons. Challenges such as compounding one-step errors~\citep{janner2019mbpo, chua2018pets}, maintaining consistency across numerous steps, and lacking methods for intervention or counterfactual reasoning highlight the need for L2. Likewise, neither L1 nor L2 inherently adapt the model based on new evidence; this capability is the focus of L3. The fundamental distinction between L1 and L2 lies in whether the system is formulated and assessed for \emph{multi-step rollout accuracy and constraint adherence}, rather than solely focusing on one-step prediction precision~\citep{hafner2023dreamerv3, ding2024survey_wm, moerland2023mbrl}. The limitation of L1 is not that one-step prediction is unimportant, but that local predictive quality alone does not guarantee decision-usable behavior under composition. The practical question is thus when short-horizon operators stop meeting the planner's needs.

\section{L2 Simulator: Decision-Usable Multi-Step Simulation}
\label{sec:l2}

Where L1 answers \textit{``what is the next local state given the current state and action?''}, L2 answers a decision-relevant question: \textit{``if the agent executes a candidate action sequence under task constraints, what future trajectory is likely to unfold?''} This elevation turns one-step operators into a simulator that an agent can query before committing to action, thus providing an \emph{imagination} of the future without requiring real-environment interaction. Model-based planning exploits exactly this capability: by rolling out candidate plans inside a learned model, the agent compares outcomes and selects the most promising course of action~\citep{sutton1991dyna,hafner2023dreamerv3,schrittwieser2020muzero}. An important corollary is that any system used to generate synthetic training data for an agent implicitly serves as a world model, since it must produce state transitions realistic enough to support policy improvement~\citep{gu2024webdreamer,webevolver2025}. It is worth noting that, decision-usable simulation focuses on plausible dynamics, which holds its distinction between L1 where state changes are arbitrary. For example, a cup passing through a solid table, a car drifting through lane boundaries without consequence, or a social commitment silently vanishing each represent failures to preserve the governing invariants of the target regime.

Table~\ref{tab:l2_boundary_conditions} maps the three L2 boundary conditions to concrete instantiations in each governing-law regime. More precisely, an L2 system supports trajectory-level queries of the form
\[
\hat p(\tau \mid z_0, a_{1:H}, c), \quad \tau=(z_1,\ldots,z_H),
\]
where $a_{1:H}$ denotes an action sequence and $c$ denotes optional constraints imposed by the governing-law regime. Intervention-structured rollouts align with the interventional rung of Pearl's causal hierarchy (Section~\ref{sec:philosophy_l2}). What separates L2 from L1 is not one-step predictive quality alone, but \textbf{coherent multi-step rollout under the governing laws}. L2 thus stitches per-edge L1 operators into a full trajectory $z_0\to z_1\to\cdots\to z_H$ (top block of Figure~\ref{fig:l1l2l3_graphical}).

\begin{table*}[t]
\caption{\textbf{L2 boundary conditions instantiated by governing-law regime.} Each cell specifies what the abstract condition means concretely in that domain.}
\label{tab:l2_boundary_conditions}
\centering
\footnotesize
\setlength{\tabcolsep}{0.9mm}
\begin{tabularx}{\textwidth}{@{}l *{4}{>{\raggedright\arraybackslash}X}@{}}
\toprule
& \textbf{Physical World} & \textbf{Digital World} & \textbf{Social World} & \textbf{Scientific World} \\
\midrule
\textbf{Coherence}
& Object persistence and stable contacts over $H$-step manipulation sequences
& DOM/file-system consistency across multi-step UI/code interactions
& Commitment and relationship stability across multi-turn dialogue
& Causal chain validity across experimental sequences \\
\addlinespace
\hline
\addlinespace
\textbf{Sensitivity}
& Force/placement perturbation alters grasp outcome proportionally
& UI failure injection (pop-ups, timeouts) causes appropriate replan
& Changing one agent's strategy shifts negotiation outcome
& Parameter change produces directionally correct measurement shift \\
\addlinespace
\hline
\addlinespace
\textbf{Consistency}
& No interpenetration, energy conservation, kinematic feasibility
& API contract adherence, type constraints, state-machine validity
& Norm compliance, belief consistency, reflexive social dynamics
& Conservation laws, causal graph consistency, evidence-chain validity \\
\bottomrule
\end{tabularx}
\end{table*}

\subsection{Requirements for Elevation}
\label{subsec:l2_requirements}
Composing L1's local operators over multiple steps does not automatically yield a decision-usable simulator: compounding errors, action-insensitive rollouts, and violated domain invariants can each render the resulting trajectories misleading for planning. This echoes the classical frame problem~\citep{mccarthy1969some,shanahan1997frame}, in particular, local transition rules alone do not specify which properties should remain invariant under action, while the concern here is operational rather than logical. The interface between a planner and an L2 world model is the \emph{query}: given an action sequence $a_{1:H}$ from state $z_0$ under constraints $c$, the model returns rollouts that the planner uses to compare candidates and select the one maximizing an objective. We treat closed-loop use (planning, acting, or control in interaction with an environment) as an orthogonal deployment property: the level boundary is determined by the depth and reliability of the world-model query, not by whether the system operates in a feedback loop. We use three boundary conditions to mark the elevation from L1 to L2:

\begin{enumerate}[leftmargin=*,nosep]
    \item \textbf{Long-horizon coherence:} rollouts remain usable over multiple steps, rather than degrading immediately through compounding error.
    \item \textbf{Intervention sensitivity:} counterfactual edits, for example, changing actions, premises, or controllable inputs. These induce stable and directionally meaningful trajectory changes.
    \item \textbf{Constraint consistency:} generated futures respect the governing-law constraints of the target regime, whether physical, digital, social, or scientific.
\end{enumerate}

These are not merely conceptual distinctions; together they induce a practical test for whether a system
deserves to be called an L2 simulator. A candidate system should be evaluated not only on one-step
prediction quality, but also on whether performance remains decision-usable as rollout horizon increases,
whether counterfactual interventions produce coherent and policy-relevant divergence, and whether generated
trajectories continue to satisfy regime-specific validity constraints. A model that predicts the next step
accurately yet collapses under composition, ignores action edits, or violates domain rules remains better
understood as L1 with strong local prediction, rather than as a full L2 simulator.

\paragraph{From L1 to L2.}
At L1, composing one-step operators yields a trajectory distribution that factorizes
as $\hat p(\tau \mid z_0, a_{1:H}) = \prod_{t=1}^{H} p_\theta(z_t \mid z_{t-1}, a_t)$, with each step optimized
independently; the trajectory is an unregulated byproduct. At L2, the governing-law constraint c couples
steps together: conceptually,
\[
\hat p(\tau \mid z_0, a_{1:H}, c) \propto \prod_{t=1}^{H} p_\theta(z_t \mid z_{t-1}, a_t)\,\phi_c(\tau),
\]
where $\phi_c(\tau)$ is a governing-law compatibility term over the full rollout. The hard-indicator case
$\mathbf{1}[c(\tau)]$ is a special case of $\phi_c(\tau)$ when violations are treated as strictly inadmissible.
Because $\phi_c(\tau)$ depends on the entire trajectory, the L2 distribution does not factorize into independent
per-step terms.

Each requirement maps to a diagnostic signal and a mitigation strategy. \emph{Long-horizon fidelity} is diagnosed by a success cliff at a specific horizon $H$. The primary mitigation of this is task segmentation with frequent replanning. \emph{Action controllability} is diagnosed by action insensitivity across rollouts (changing $a_t$ produces no meaningful trajectory change), where mitigation requires explicit action-consistency evaluation. \emph{Constraint consistency} is measured by the constraint-violation rate. In such cases, mitigations include hard constraint layers and verification gates. A fourth property, \emph{calibration}, requires that confidence aligns with actual accuracy under distribution shift; in such cases, overconfident wrong predictions signal failure, and distribution-shift detection is the main remedy.

\paragraph{Residual frame-problem manifestations.}
Modern neural world models sidestep the classical frame problem's representational burden by learning implicitly from data what persists and what changes~\citep{goodfellow2016deep,hafner2023dreamerv3}, enabling scalable model-based RL~\citep{hafner2019dreamer,schrittwieser2020muzero,moerland2023mbrl} and video prediction~\citep{babaeizadeh2018sv2p,brooks2024sora} without explicit frame axioms. Yet the problem resurfaces at rollout time: context-window limits and hallucination cause models to lose track of relevant past information, violating long-horizon coherence, while rare preconditions under-represented in training data undermine constraint consistency~\citep{ding2024survey_wm,shanahan1997frame}. These failure modes motivate the techniques surveyed below (see Appendix~\ref{app:boundaries}).

\subsection{Applications}
\label{subsec:l2_app}

In this section, we categorize L2 systems into four governing-law regimes. Tables~\ref{tab:l2_systems_phys_dig} and~\ref{tab:l2_systems_soc_sci} provide anchor systems for cross-regime comparison and summarize how each domain instantiates the boundary conditions.

\begin{table*}[ht]
\caption{\textbf{Representative L2 anchor systems:} Physical and Digital Worlds.
Columns indicate long-horizon coherence (\textbf{LH}), intervention sensitivity (\textbf{IS}),
and constraint consistency (\textbf{CC}). The table is a compact comparison set, not an exhaustive inventory of all systems discussed in the prose.}
\label{tab:l2_systems_phys_dig}
\centering
\small
\setlength{\tabcolsep}{0.9mm}
\begin{tabular}{l|cc|cccl}
\toprule
\textbf{Method} & \multicolumn{2}{c|}{\textbf{Links}} & \textbf{LH} & \textbf{IS} & \textbf{CC} & \textbf{Architecture} \\
\midrule
\rowcolor{gray!15}
\multicolumn{7}{c}{\textit{Physical World}} \\
MuZero~\citep{schrittwieser2020muzero} & \paperlink{https://arxiv.org/abs/1911.08265} & {---} & \cmark & \cmark & \cmark & MLP dynamics + MCTS \\
Plan2Explore~\citep{sekar2020plan2explore} & \paperlink{https://arxiv.org/abs/2005.05960} & \githublink{https://github.com/ramanans1/plan2explore} & \cmark & \cmark & \xmark & Dreamer + self-supervised exploration \\
PathDreamer~\citep{koh2021pathdreamer} & \paperlink{https://arxiv.org/abs/2105.08756} & \githublink{https://github.com/google-research/pathdreamer} & \cmark & \xmark & \xmark & Autoregressive visual VLN \\
DreamerPro~\citep{deng2022dreamerpro} & \paperlink{https://arxiv.org/abs/2110.14565} & \githublink{https://github.com/fdeng18/dreamer-pro} & \cmark & \cmark & \xmark & RSSM + prototypical representations \\
DreamingV2~\citep{okada2022dreamingv2} & \paperlink{https://arxiv.org/abs/2203.00494} & {---} & \cmark & \cmark & \xmark & DreamerV2 + reconstruction-free \\
Diffuser~\citep{janner2022diffuser} & \paperlink{https://arxiv.org/abs/2205.09991} & \githublink{https://github.com/jannerm/diffuser} & \cmark & \xmark & \xmark & Diffusion trajectory planning \\
DreamerV3~\citep{hafner2023dreamerv3} & \paperlink{https://arxiv.org/abs/2301.04104} & \githublink{https://github.com/danijar/dreamerv3} & \cmark & \cmark & \cmark & RSSM + symlog loss \\
DayDreamer~\citep{wu2023daydreamer} & \paperlink{https://arxiv.org/abs/2206.14176} & \githublink{https://github.com/danijar/daydreamer} & \cmark & \cmark & \cmark & RSSM real-world robots \\
GAIA-1~\citep{hu2023gaia1} & \paperlink{https://arxiv.org/abs/2309.17080} & {---} & \cmark & \cmark & \xmark & Transformer video generation \\
DIAMOND~\citep{alonso2024diamond} & \paperlink{https://arxiv.org/abs/2405.12399} & \githublink{https://github.com/eloialonso/diamond} & \cmark & \cmark & \cmark & U-Net diffusion \\
Sora~\citep{brooks2024sora} & \paperlink{https://openai.com/index/video-generation-models-as-world-simulators/} & {---} & \cmark & \xmark & \xmark & DiT video diffusion \\
Genie~\citep{bruce2024genie} & \paperlink{https://arxiv.org/abs/2402.15391} & {---} & \cmark & \cmark & \xmark & ST-transformer + VQ actions \\
iVideoGPT~\citep{wu2024ivideogpt} & \paperlink{https://arxiv.org/abs/2405.15223} & \githublink{https://github.com/thuml/iVideoGPT} & \cmark & \cmark & \xmark & Transformer + VQ-VAE \\
OccWorld~\citep{zheng2024occworld} & \paperlink{https://arxiv.org/abs/2311.16038} & \githublink{https://github.com/wzzheng/OccWorld} & \cmark & \cmark & \cmark & GPT 3D occupancy prediction \\
Vista~\citep{gao2024vista} & \paperlink{https://arxiv.org/abs/2405.17398} & \githublink{https://github.com/OpenDriveLab/Vista} & \cmark & \cmark & \xmark & Diffusion driving generation \\
DriveDreamer~\citep{wang2024drivedreamer} & \paperlink{https://arxiv.org/abs/2309.09777} & \githublink{https://github.com/JeffWang987/DriveDreamer} & \cmark & \cmark & \xmark & Diffusion AD generation \\
Copilot4D~\citep{zhang2024copilot4d} & \paperlink{https://arxiv.org/abs/2311.01017} & {---} & \cmark & \cmark & \cmark & VQ-VAE + point diffusion \\
LWM~\citep{liu2024lwm} & \paperlink{https://arxiv.org/abs/2402.08268} & \githublink{https://github.com/LargeWorldModel/LWM} & \cmark & \xmark & \xmark & RingAttention long-context LLM \\
DreMa~\citep{wu2024drema} & \paperlink{https://arxiv.org/abs/2412.14957} & \githublink{https://github.com/leobarcellona/drema_code} & \cmark & \cmark & \xmark & Compositional 3DGS twins \\
Cosmos~\citep{nvidia2025cosmos} & \paperlink{https://arxiv.org/abs/2501.03575} & \githublink{https://github.com/NVIDIA/Cosmos} & \cmark & \cmark & \xmark & Autoregressive + diffusion hybrid \\
Aether~\citep{zhu2025aether} & \paperlink{https://arxiv.org/abs/2503.18945} & \githublink{https://github.com/OpenRobotLab/Aether} & \cmark & \cmark & \cmark & CogVideoX geometry fine-tune \\
PIN-WM~\citep{li2025pinwm} & \paperlink{https://arxiv.org/abs/2504.16693} & \githublink{https://github.com/XuAdventurer/PIN-WM} & \cmark & \cmark & \cmark & Differentiable rigid-body + 3DGS \\
Yume~\citep{mao2025yume} & \paperlink{https://arxiv.org/abs/2507.17744} & \githublink{https://github.com/stdstu12/YUME} & \cmark & \cmark & \xmark & Video diffusion world generation \\
GAIA-2~\citep{nvidia2025gaia2} & \paperlink{https://arxiv.org/abs/2503.20523} & {---} & \cmark & \cmark & \xmark & Latent diffusion multi-view AD \\
RoboScape~\citep{chen2025roboscape} & \paperlink{https://arxiv.org/abs/2506.23135} & \githublink{https://github.com/tsinghua-fib-lab/RoboScape} & \cmark & \cmark & \xmark & Physics-informed robot video \\
BridgeV2W~\citep{wang2026bridgev2w} & \paperlink{https://arxiv.org/abs/2602.03793} & {---} & \cmark & \cmark & \xmark & Action-conditioned embodied video \\
HWM~\citep{zhang2026hwm} & \paperlink{https://arxiv.org/abs/2604.03208} & \githublink{https://github.com/kevinghst/HWM_PLDM} & \cmark & \cmark & \xmark & Hierarchical latent + MCTS \\
\midrule
\rowcolor{gray!15}
\multicolumn{7}{c}{\textit{Digital World}} \\
GameGAN~\citep{kim2020gamegan} & \paperlink{https://arxiv.org/abs/2005.12126} & \githublink{https://github.com/nv-tlabs/GameGAN_code} & \cmark & \cmark & \xmark & GAN neural game engine \\
WebDreamer~\citep{gu2024webdreamer} & \paperlink{https://arxiv.org/abs/2411.06559} & \githublink{https://github.com/OSU-NLP-Group/WebDreamer} & \cmark & \cmark & \xmark & LLM web state simulation \\
CodeWM~\citep{dainese2024codewm} & \paperlink{https://arxiv.org/abs/2405.15383} & \githublink{https://github.com/nicoladainese96/code-world-models} & \cmark & \cmark & \cmark & LLM + MCTS code generation \\
WorldCoder~\citep{tang2024worldcoder} & \paperlink{https://arxiv.org/abs/2402.12275} & \githublink{https://github.com/ma-labo/worldcoder} & \cmark & \cmark & \cmark & LLM incremental code synthesis \\
GameGen-X~\citep{che2024gamegenx} & \paperlink{https://arxiv.org/abs/2411.00769} & \githublink{https://github.com/GameGen-X/GameGen-X} & \cmark & \cmark & \xmark & Diffusion transformer, InstructNet control \\
GameNGen~\citep{valevski2025gamengin} & \paperlink{https://arxiv.org/abs/2408.14837} & {---} & \cmark & \cmark & \xmark & U-Net diffusion \\
WMA~\citep{chae2025wma} & \paperlink{https://arxiv.org/abs/2410.13232} & \githublink{https://github.com/kyle8581/WMA-Agents} & \cmark & \cmark & \xmark & LLM web transition prediction \\
WebSynthesis~\citep{gao2025websynthesis} & \paperlink{https://arxiv.org/abs/2507.04370} & \githublink{https://github.com/LucusFigoGao/WebSynthesis} & \cmark & \cmark & \xmark & LLM + MCTS planning \\
NeuralOS~\citep{rivard2025neuralos} & \paperlink{https://arxiv.org/abs/2507.08800} & \githublink{https://github.com/yuntian-group/neural-os} & \cmark & \cmark & \xmark & RNN + pixel diffusion \\
GameFactory~\citep{yu2025gamefactory} & \paperlink{https://arxiv.org/abs/2501.08325} & \githublink{https://github.com/KwaiVGI/GameFactory} & \cmark & \cmark & \xmark & Action-controlled video generation \\
GameCraft~\citep{li2025gamecraft} & \paperlink{https://arxiv.org/abs/2506.17201} & \githublink{https://github.com/Tencent-Hunyuan/Hunyuan-GameCraft-1.0} & \cmark & \cmark & \xmark & Diffusion game video generation \\
MobileDreamer~\citep{cao2026mobiledreamer} & \paperlink{https://arxiv.org/abs/2601.04035} & {---} & \cmark & \cmark & \xmark & LLM GUI sketch prediction \\
Word2World~\citep{li2025wordtoworld} & \paperlink{https://arxiv.org/abs/2512.18832} & \githublink{https://github.com/X1AOX1A/Word2World} & \cmark & \cmark & \xmark & LLM text-based WM \\
Code2World~\citep{zheng2026code2world} & \paperlink{https://arxiv.org/abs/2602.09856} & \githublink{https://github.com/AMAP-ML/Code2World} & \cmark & \cmark & \cmark & VLM code rendering \\
gWorld~\citep{gworld2026} & \paperlink{https://arxiv.org/abs/2602.01576} & \githublink{https://github.com/trillion-labs/gWorld} & \cmark & \cmark & \cmark & VLM code rendering \\
WebWorld~\citep{xiao2026webworld} & \paperlink{https://arxiv.org/abs/2602.14721} & {---} & \cmark & \cmark & \xmark & Fine-tuned VLM web simulator \\
RWML~\citep{yu2026rwml} & \paperlink{https://arxiv.org/abs/2602.05842} & {---} & \cmark & \cmark & \xmark & LLM + RL sim-to-real \\
\bottomrule
\end{tabular}%
\end{table*}

\begin{table*}[ht]
\caption{\textbf{Representative L2 anchor systems (continued):} Social and Scientific Worlds.
Columns indicate long-horizon coherence (\textbf{LH}), intervention sensitivity (\textbf{IS}),
and constraint consistency (\textbf{CC}). 
Links are provided where available; the table is a compact comparison set, not an exhaustive inventory of all systems discussed in the prose.
}
\label{tab:l2_systems_soc_sci}
\centering
\small
\setlength{\tabcolsep}{0.9mm}
\begin{tabular}{l|cc|cccl}
\toprule
\textbf{Method} & \multicolumn{2}{c|}{\textbf{Links}} & \textbf{LH} & \textbf{IS} & \textbf{CC} & \textbf{Architecture} \\
\midrule
\rowcolor{gray!15}
\multicolumn{7}{c}{\textit{Social World}} \\
Deal or No Deal~\citep{lewis2017dealornodeal} & \paperlink{https://arxiv.org/abs/1706.05125} & \githublink{https://github.com/facebookresearch/end-to-end-negotiator} & \cmark & \cmark & \cmark & RNN + RL self-play \\
Social Simulacra~\citep{park2022socialsimulacra} & \paperlink{https://dl.acm.org/doi/10.1145/3526113.3545616} & {---} & \cmark & \cmark & \xmark & GPT prompt-chain community simulation \\
CICERO~\citep{meta2022cicero} & \paperlink{https://doi.org/10.1126/science.ade9097} & \githublink{https://github.com/facebookresearch/diplomacy_cicero} & \cmark & \cmark & \cmark & LLM + strategic planning \\
Generative Agents~\citep{park2023generative} & \paperlink{https://arxiv.org/abs/2304.03442} & \githublink{https://github.com/joonspk-research/generative_agents} & \cmark & \cmark & \cmark & LLM reflective memory \\
Sotopia~\citep{zhou2024sotopia} & \paperlink{https://arxiv.org/abs/2310.11667} & \githublink{https://github.com/sotopia-lab/sotopia} & \cmark & \cmark & \cmark & LLM social evaluation \\
AvalonBench~\citep{light2023avalonbench} & \paperlink{https://arxiv.org/abs/2310.05036} & \githublink{https://github.com/jonathanmli/Avalon-LLM} & \cmark & \cmark & \cmark & LLM deductive reasoning \\
Werewolf~\citep{xu2024werewolf} & \paperlink{https://arxiv.org/abs/2310.18940} & \githublink{https://github.com/xuyuzhuang11/Werewolf} & \cmark & \cmark & \cmark & LLM + RL strategic policy \\
ProjectSid~\citep{altera2024projectsid} & \paperlink{https://arxiv.org/abs/2411.00114} & \githublink{https://github.com/altera-al/project-sid} & \cmark & \cmark & \cmark & LLM multi-agent civilization simulation \\
OASIS~\citep{yang2024oasis} & \paperlink{https://arxiv.org/abs/2411.11581} & \githublink{https://github.com/camel-ai/oasis} & \cmark & \cmark & \cmark & LLM social simulation \\
MASim~\citep{zhang2025masim} & \paperlink{https://arxiv.org/abs/2512.07195} & {---} & \cmark & \cmark & \xmark & Multilingual agent simulation \\
SWM-AP~\citep{zhang2025swmap} & \paperlink{https://arxiv.org/abs/2510.19270} & {---} & \cmark & \cmark & \xmark & Social WM mechanism design \\
AIvilization~\citep{fan2026aivilization} & \paperlink{https://arxiv.org/abs/2602.10429} & {---} & \cmark & \cmark & \cmark & Sandbox economy simulation \\
PolicySim~\citep{huang2026policysim} & \paperlink{https://arxiv.org/abs/2603.19649} & \githublink{https://github.com/renH2/PolicySim} & \cmark & \cmark & \xmark & LLM platform policy sandbox \\
\midrule
\rowcolor{gray!15}
\multicolumn{7}{c}{\textit{Scientific World}} \\
GNS~\citep{sanchez2020gns} & \paperlink{https://arxiv.org/abs/2002.09405} & \githublink{https://github.com/deepmind/deepmind-research} & \cmark & \xmark & \cmark & GNN message passing \\
ChemBO~\citep{pmlr-v108-korovina20a} & \paperlink{https://proceedings.mlr.press/v108/korovina20a.html} & \githublink{https://github.com/kamikaze0923/ChemBo} & \cmark & \xmark & \cmark & GP + synthesis-graph BO \\
P3BO~\citep{angermueller2020population} & \paperlink{https://proceedings.mlr.press/v119/angermueller20a.html} & {---} & \cmark & \cmark & \xmark & Adaptive population-based optimization \\
FNO~\citep{li2021fno} & \paperlink{https://arxiv.org/abs/2010.08895} & \githublink{https://github.com/neuraloperator/neuraloperator} & \cmark & \xmark & \cmark & Fourier neural operator \\
Pangu-Weather~\citep{bi2023panguweather} & \paperlink{https://arxiv.org/abs/2211.02556} & \githublink{https://github.com/198808xc/Pangu-Weather} & \cmark & \xmark & \cmark & 3D Earth transformer \\
ClimaX~\citep{nguyen2023climax} & \paperlink{https://arxiv.org/abs/2301.10343} & \githublink{https://github.com/microsoft/ClimaX} & \cmark & \xmark & \cmark & ViT climate foundation \\
GraphCast~\citep{lam2023graphcast} & \paperlink{https://arxiv.org/abs/2212.12794} & \githublink{https://github.com/google-deepmind/graphcast} & \cmark & \xmark & \cmark & GNN autoregressive weather \\
GenCast~\citep{price2024gencast} & \paperlink{https://arxiv.org/abs/2312.15796} & \githublink{https://github.com/google-deepmind/graphcast} & \cmark & \xmark & \cmark & Spherical ensemble diffusion \\
NeuralGCM~\citep{kochkov2024neuralgcm} & \paperlink{https://arxiv.org/abs/2311.07222} & \githublink{https://github.com/google-research/neuralgcm} & \cmark & \xmark & \cmark & Hybrid physics--NN core \\
BAX~\citep{chitturi2024targeted} & \paperlink{https://www.nature.com/articles/s41524-024-01326-2} & \githublink{https://github.com/sathya-chitturi/multibax-sklearn} & \cmark & \cmark & \cmark & GP + user-directed acquisition \\
Aurora~\citep{bodnar2025aurora} & \paperlink{https://arxiv.org/abs/2405.13063} & \githublink{https://github.com/microsoft/aurora} & \cmark & \xmark & \cmark & 3D Swin weather foundation \\
Lingshu-Cell~\citep{zhang2026lingshucell} & \paperlink{https://arxiv.org/abs/2603.25240} & {---} & \cmark & \xmark & \cmark & Masked diffusion cellular WM \\
\bottomrule
\end{tabular}%
\end{table*}

\subsubsection{Laws of the Physical World}
\label{subsec:l2_physical}

In the physical domain, L2 models should respect geometry, kinematics, and conservation laws.
The governing constraints are contact, reachability, stability, and energy conservation; violations of any of these will mislead a planner into proposing actions that fail catastrophically in real execution.

\paragraph{Physics simulation.}
\emph{Rigid-body control simulators.}
Classical physics simulators remain the foundation layer for executable
transition validity in embodied world modeling.
MuJoCo provides articulated rigid-body dynamics and contact-rich control, with
dm\_control packaging these capabilities into a standardized continuous-control
suite~\citep{todorov2012mujoco,tassa2020dmcontrol}.
Brax pushes differentiable rigid-body simulation toward accelerator-scale
throughput~\citep{freeman2021brax}, while Isaac Gym and Isaac Lab emphasize
massive GPU-parallel robotics simulation~\citep{makoviychuk2021isaacgym,nvidia2025isaaclab}.

\emph{Scalable and general-purpose simulation platforms.}
Genesis positions itself as a generative and universal physics engine~\citep{genesis2024}, reflecting the broader trend toward higher-throughput simulators that can jointly support both control and large-scale synthetic-data generation.

\emph{Interaction-centric embodied simulators.}
At the graphics-and-robotics interface, SAPIEN provides part-aware,
interaction-centric simulation, and ManiSkill3 scales GPU-parallel rendering
for generalizable embodied AI~\citep{xiang2020sapien,tao2024maniskill3}.
These systems are not learned simulators; they are explicit law executors whose
value lies in precise contact handling, articulated constraints, and
reproducible rollouts.

\paragraph{Video generation models.}
\emph{Appearance-first long-horizon video generation.}
A scalable route to physical-world simulation is the \emph{video interface}:
given current observations and optional actions, the model returns imagined
future frames.
This line begins with appearance-first rollout, where systems such as Sora,
Lumiere, and VideoPoet demonstrate coherent visual dynamics over extended
horizons~\citep{brooks2024sora,bartal2024lumiere,kondratyuk2024videopoet},
with geometry-aware structure increasingly emerging beyond pixel-level
realism~\citep{li2024sora_geometry}.
FramePack~\citep{zhang2025framepack} and
Self-Forcing~\citep{huang2025selfforcing} reduce long-horizon drift through frame-context packing.

\emph{Action-conditioned and interactive video worlds.}
A second direction moves from passive continuation toward intervention-aware generation.
Genie learns latent action spaces from unlabeled Internet video~\citep{bruce2024genie}, while GAIA-1 conditions future generation on explicit control signals for counterfactual evaluation~\citep{hu2023gaia1}.
More recent systems push this line toward real-time, long-horizon, and streaming interaction:
Oasis explores open-ended interactive generation in a unified transformer world~\citep{oasis2024};
WorldPlay emphasizes long-term geometric consistency for real-time interactive world modeling~\citep{sun2025worldplay};
Matrix-Game~3.0 extends interactive generation to streaming settings with explicit long-horizon memory~\citep{matrixgame3};
Yume-1.5 studies text-controlled interactive world generation~\citep{mao2025yume};
and LongLive targets real-time interactive long video generation~\citep{yang2025longlive}.
Taken together, these systems mark a shift from passive video prediction toward controllable, intervention-aware, and temporally persistent video worlds.

\emph{Decision-oriented video world models.}
In model-based RL, SimPLe~\citep{kaiser2019simple} and
DIAMOND~\citep{alonso2024diamond} make the decision-theoretic role of video
world models explicit.
In robotics, DreamZero~\citep{ye2026dreamzero} and
DreamDojo~\citep{gao2026dreamdojo} demonstrate zero-shot and generalist policy
learning via video world models, while
FutureVLA~\citep{xu2026futurevla} couples visuomotor prediction directly with Vision-Language-Action policies to unify perception and control.

\emph{Evaluation and limitations.}
Within our L2 framing, however, visual plausibility does not equal
decision-usability.
Intervention sensitivity remains fragile, long-horizon coherence is easily
overstated when judged by perceptual quality alone~\citep{guo2025logic}, and
constraint consistency is difficult to verify from rendered frames.
Standard metrics such as FVD~\citep{unterthiner2018fvd} capture distributional
realism; VBench-style suites~\citep{huang2023vbench,huang2025vbench++} better
decompose controllability; VBench-2.0~\citep{zheng2025vbench2} extends
evaluation to physics consistency and commonsense reasoning; and
VChain~\citep{huang2025vchain} introduces visual chain-of-thought for causal
coherence.
Video interfaces are the most scalable observation-layer entry point, but planner-critical structure remains implicit in pixels; Appendix~\ref{app:l2_extended} surveys geometry-carrying alternatives that make such structure explicit.

\paragraph{Robotics and sim-to-real transfer.}
\emph{World models transferred to real robots.}
DayDreamer~\citep{wu2023daydreamer} showed that Dreamer-family world models can
transfer from simulation to physical robots while handling sensor noise,
contact dynamics, and actuation delays.
DreamZero~\citep{ye2026dreamzero} achieves zero-shot policy learning via world
action models that predict both next states and actions, and
FutureVLA~\citep{xu2026futurevla} embeds visuomotor prediction within
Vision-Language-Action models to improve action grounding.

\emph{Physics-grounded bridges for sim-to-real robustness.}
PIN-WM~\citep{li2025pinwm} integrates differentiable physics with learned visual world modeling, creating ``digital cousins'' via physics-aware randomization.

\paragraph{Spatial Reasoning.}
\citet{mindcube} shows that agents can construct spatial mental models from limited views; \citet{tos} asks whether agents can \emph{actively} explore to construct spatial beliefs rather than reasoning passively over given views.

\emph{Representation requirement.}
Across these systems, the key question is not whether richer representations are
possible, but what is the weakest representation that still preserves
planner-critical structure, such as object persistence, free space, contact onset,
support relations, and action-conditioned change over useful horizons.
Extended details on 3D-structured world models and autonomous driving appear in
Appendix~\ref{app:l2_extended}.

\subsubsection{Laws of the Digital World}
\label{subsec:l2_software}

The Laws of the Digital World govern transitions in systems defined by formal specifications, from finite automata (UI state machines) and context-free grammars (structured data formats) to Turing-complete programs (general software). Unlike the Laws of the Physical World or the Laws of the Social World, these constraints are \emph{explicitly specified and mechanically verifiable}: a transition either satisfies the program's semantics or it does not. Because software transitions approximate deterministic state machines and failures are loggable (error codes, popups, permission denials, timeouts), the core challenge for a Simulator in code worlds is \emph{structured state prediction} (DOM trees, program state, game state) rather than visual fidelity.

\paragraph{Coding agents.}
An emerging paradigm represents world models as executable programs rather than neural networks. CodeWM~\citep{dainese2024codewm} uses LLMs guided by Monte Carlo Tree Search to generate Python programs that serve as explicit, interpretable world models for reinforcement learning across 18 environments. WorldCoder~\citep{tang2024worldcoder} takes a complementary approach, with a model-based LLM agent building a Python world model through environment interaction, gaining sample efficiency and transferring across environments by editing its code. WKM~\citep{qiao2024wkm} provides both global task knowledge and dynamic state knowledge to guide LLM agent planning, while CWM~\citep{copet2025cwm}, a 32B open-weights LLM released to advance research on code generation with world models, achieves 65.8\% on SWE-bench Verified with test-time scaling. A conceptually distinct variant pushes further: rather than using an LLM to \emph{generate} a code world model, the world model \emph{is} a running software system. Web World Models~\citep{feng2025wwm} implement world state as ordinary web code with latent state exposed as typed web interfaces, delegating logical consistency to deterministic execution of the web stack while LLMs generate context and high-level decisions. In a similar spirit, SWE-World~\citep{sweworld} replaces containerized execution with a learned surrogate that predicts intermediate execution outcomes and final test feedback, preserving the standard agent-environment loop without running real environments and lifting a 32B coder from 6.2\% to 52.0\% on SWE-bench Verified.
These code-based approaches yield interpretable, composable, and verifiable world models that neural dynamics can only approximate.

\paragraph{Web agents.}
Web agents usually browse websites; therefore, modeling and simulating state transitions within a website is crucial for building effective web world models.
WebDreamer~\citep{gu2024webdreamer} introduced the idea of using an LLM as an implicit world model of the internet, but subsequent work showed that off-the-shelf LLMs are insufficient: dedicated training with transition-focused abstraction is needed~\citep{chae2025wma}. A growing body of work addresses the co-evolution of agent and world model. WebEvolver~\citep{webevolver2025} tightly couples the two in a mutual improvement loop, while DreamGym~\citep{chen2025dreamgym} builds experience models with chain-of-thought reasoning, achieving over 30\% improvement on WebArena. At larger scale, WebSynthesis~\citep{gao2025websynthesis} combines world models with MCTS-based planning using entirely synthetic data, and WebWorld~\citep{xiao2026webworld} trains an open-web simulator on over one million trajectories supporting 30+ step simulation.
Early Experience~\citep{earlyexp} proposes a paradigm that leverages agent-generated trajectories as reward-free supervision, enabling implicit world modeling and self-reflection while bridging imitation learning and reinforcement learning.
AUI~\citep{lin2025computer} takes a different approach, employing a Coder to optimize websites by leveraging feedback from a Computer-Use Agent in an iterative collaboration loop.
Orthogonal design choices concern what data to learn from and when to query the model: generating training trajectories from tool specifications alone (Simia;~\citealt{li2025simia}), adding a metacognitive layer that decides \emph{whether} to consult the world model at each step (WAC;~\citealt{wac2026}), and collecting on-policy agent trajectories to cover the out-of-distribution behaviors that statically harvested corpora tend to miss, so the learned simulator keeps pace with the agent's shifting behavior over training.

\paragraph{GUI agents.}
GUI agents~\citep{qin2025ui,lin2025showui,xu2024aguvis} typically execute actions in real environments. However, in scenarios where actions may be dangerous or lead to undesired outcomes, it is beneficial to estimate them beforehand. A GUI world model can simulate and evaluate these actions, thereby providing a more reliable assessment.
Therefore, MobileDreamer~\citep{cao2026mobiledreamer} transforms GUI images into task-related sketches for structured state prediction, while MobileWorldBench~\citep{li2025mobileworldbench} provides systematic evaluation with 1.4 million (state, action, future state) triplets. Complementary to explicit GUI world models, UI-AGILE shows that effective reinforcement learning and precise inference-time grounding remain equally important for strong downstream GUI-agent performance \citep{lian2025ui}. A central design question is the output representation: ViMo~\citep{luo2025vimo} generates future observations as images using symbolic text representation, while gWorld~\citep{gworld2026} generates renderable web code as the predicted next state, suggesting that generating the code that renders the GUI can be more faithful than generating pixels directly. At the OS level, NeuralOS~\citep{rivard2025neuralos} simulates desktop GUIs by predicting screen frames from user inputs, while CUWM~\citep{cuwm2026} targets desktop software where persistent document state must be preserved across long-horizon workflows. 
Code2World~\citep{zheng2026code2world} further extends this line by treating code as a renderable world, where generated programs directly produce visual states (e.g., HTML) upon execution. This enables modeling environment dynamics as executable code generation, tightly coupling perception, action, and state transition in interactive domains such as GUIs.

\paragraph{Game agents.}
Game agents must act over long horizons in open-ended environments whose rules and win conditions are formally checkable, so a world model can serve both as a reliable surrogate environment for training and as a learned predictor the agent consults to roll out and evaluate candidate moves before committing to them. The most explicit realization treats the world model as a verifiable rollout engine: Code World Models for General Game Playing~\citep{lehrach2025cwmgame} prompts an LLM to translate natural-language rules and observed trajectories into executable Python implementing state transitions, legal-move enumeration, and termination checks, and a planner such as MCTS then queries this program to search only over genuinely legal continuations, matching or surpassing Gemini~2.5~Pro on nine of ten games. MineWorld~\citep{guo2025mineworld} instead learns environment dynamics directly, training a visual-action autoregressive Transformer over interleaved image and action tokens whose parallel decoding sustains real-time interaction at four to seven frames per second, introducing action-following metrics beyond visual quality and significantly outperforming state-of-the-art open-source diffusion world models. A complementary line internalizes the world model inside the policy: Game-TARS~\citep{wang2025game} pretrains a generalist multimodal game agent that roughly doubles the success rate of the prior state of the art on open-world Minecraft. RAGEN~\citep{wang2025ragen} frames agent learning as multi-turn reinforcement learning over interactive tasks, establishing that it works for self-evolving LLM agents while characterizing the training-stability failure modes that arise. VAGEN~\citep{wang2025vagen} extends this paradigm to VLM agents and shows that explicitly reinforcing world-model reasoning, such as state estimation and transition prediction within the reasoning trace, yields further gains. RAGEN-2~\citep{wang2026ragen} continues in this direction by diagnosing a reasoning-collapse failure mode in agentic RL and proposing a fix. Across these systems the recurring theme is structured prediction over program and game state, where transitions are explicitly specified and, by construction, mechanically verifiable rather than merely visually plausible.

\paragraph{Tool-calling agents.}
Tool-calling agents invoke external functions such as APIs, databases, and Model Context Protocol (MCP) servers, where each call mutates hidden environment state that the agent cannot observe directly, so a world model is most useful as a verifiable rollout engine that the agent queries to predict the outcome of a call before committing to it, enabling look-ahead planning, self-verification, and cheap training environments. Because tool transitions are explicitly specified by function signatures and database schemas and their failures are mechanically loggable as error codes or rejected parameters, the simulator's task here is structured prediction of program and database state rather than visual fidelity. GTM~\citep{ren2025gtm} instantiates this as a universal tool simulator that, from a prompt-level configuration, generates outputs faithfully mimicking real execution, with training data synthesized across more than 20{,}000 tools, and in agent reinforcement learning it runs at significantly faster simulation speed than real tools while preserving comparable output quality. AWM~\citep{wang2026awm} instead grounds the simulator in executable code, generating code-driven, database-backed environments whose state transitions are more reliable and consistent than purely LLM-simulated feedback (e.g., Simia~\citep{li2025simia}), and whose accessible database states supply reliable reward functions for large-scale RL of multi-turn tool-use agents. MCP-Cosmos~\citep{ganapavarapu2026mcpcosmos} pushes the simulator into the MCP ecosystem under a ``Bring Your Own World Model'' strategy, letting agents simulate state transitions and refine plans in latent space before execution and improving tool success rate and parameter accuracy in experiments spanning more than twenty MCP-Bench tasks. At a finer granularity, DyMo~\citep{guo2025dymo} equips the model with state prediction alongside function calling during post-training, so it predicts the future state of its actions through an internal environment model; on the Berkeley Function Calling Leaderboard V2 this raises success rates and significantly reduces hallucinations, and when wrapped in self-verification sampling it lets the model refuse unreliable outputs. Yet a foresight study~\citep{qian2026foresight} cautions that simply exposing such simulators is not enough, since agents rarely invoke simulation, frequently misuse predicted rollouts, and can even degrade in performance, leaving when to simulate and how to interpret and integrate foresight into downstream reasoning as the central open challenges for reliable anticipatory tool use.

\subsubsection{Laws of the Social World}
\label{subsec:l2_social}

Societal world models extend L2 to human interaction, where governing laws are beliefs, desires, intentions, norms, and institutions rather than physics. Social worlds exhibit three distinctive properties, in particular, \emph{opacity} (agents cannot directly observe each other's mental states), \emph{reflexivity} (beliefs about social state create feedback loops), and \emph{normativity} (transitions are governed partly by shared norms). Such traits make the transition function partially constituted by collective agreement rather than natural law~\citep{zheng2023mcu}. A usable social simulator separates surface language from underlying social state: dialogue can vary, but core states (goals, beliefs, relations, norms) must remain consistent and yield interpretable transitions, as formalized by the Rational Speech Acts framework~\citep{goodman2016rsa,degen2023rsa}. Concretely, a social compatibility term $\phi_c(\tau)$ can encode commitment consistency: if agent \(i\) promises action \(b\) at time \(t\), later states receive low compatibility when \(i\) violates \(b\) without explanation, renegotiation, or sanction. Similar terms can score norm compliance, role consistency, or belief-state coherence over the trajectory.

\paragraph{Theory of mind as social state.}
The computational foundation was laid by Bayesian ToM (BToM), which formalizes mental state inference as probabilistic inverse planning over rational agents~\citep{baker2011bayesian}. Neural approaches began with ToMnet~\citep{rabinowitz2018tomnet}, whose character, mental state, and prediction networks jointly infer traits and beliefs, and recent work such as LaBToM~\citep{ying2025labtom} bridges Bayesian inverse planning with formal epistemic language. However, current models lack robust mental state reasoning: FANToM~\citep{kim2023fantom} reveals ``illusory ToM'' across all state-of-the-art LLMs, and ExploreToM~\citep{sclar2024exploretom} achieves accuracy as low as 9\% for GPT-4o~\citep{chen2025tomsurvey}. A complementary challenge is the \emph{dual-structure} problem: a social agent must simultaneously model others' mental states (theory of mind) and maintain its own persistent internal state across long interactions, in particular, goals, persona, memory, and knowledge. Cognitive Architectures for Language Agents (CoALA)~\citep{sumers2024coala} formalizes this dual structure as separate memory and action spaces that must remain mutually consistent, and provides a principled framework for understanding how current LLM agents do and do not achieve stable self-representation.

\paragraph{Strategic interaction.}
CICERO~\citep{meta2022cicero} integrates a language model with piKL planning for Diplomacy, jointly optimizing game actions and dialogue while modeling second-order beliefs, achieving more than $2\times$ the average human score. Deal or No Deal~\citep{lewis2017dealornodeal} pioneered dialogue rollouts for forward simulation of negotiation dynamics. Werewolf and Avalon games serve as concentrated testbeds for deception, trust, and belief manipulation~\citep{xu2024werewolf,light2023avalonbench}, revealing that deceivers consistently prevail by exploiting cognitive limitations.

\paragraph{Sandbox simulation.}
Generative Agents demonstrated emergent social dynamics: a 25-agent simulation~\citep{park2023generative} used memory-based state tracking and periodic reflection, while Sotopia~\citep{zhou2024sotopia} formalized social simulation evaluation across seven dimensions. Scale has increased dramatically: Project Sid~\citep{altera2024projectsid} deployed 1,000 agents exhibiting emergent specialization and governance, and OASIS~\citep{yang2024oasis} scaled to one million agents reproducing information spreading and group polarization. At the individual level, \citet{argyle2023silicon} demonstrate ``silicon sampling'', which conditions LLMs on specific demographic profiles to simulate survey responses from targeted subpopulations and shows strong alignment with American National Election Studies data, opening a path toward individual social world modeling. Generative Social Choice~\citep{fish2023generativesocialchoice} extends this to democratic aggregation, using LLMs to generate representative statements from diverse synthetic participants and enabling deliberation.

\paragraph{Challenges and design principles.}
Social simulation remains premature: LLMs degrade sharply beyond second-order belief reasoning~\citep{wu2023hitom}, agents suffer from role drift and goal forgetting~\citep{park2023generative,zhou2024sotopia}, and formal commitment tracking~\citep{telang2023commitments} remains unintegrated into any LLM architecture. A practical design pattern separates a compact social state representation (commitments, constraints, relations), a dialogue generator, and a transition updater that enforces consistency and makes state transitions loggable and replayable. Flexible persona generation is essential for populating social simulators with diverse, controllable agents; PersonaGym~\citep{samuel2024personagym} provides a benchmark for evaluating how faithfully LLMs enact specified personas across complex social tasks, revealing systematic failures in maintaining persona consistency under adversarial probing. For personalization at the individual level, LaMP~\citep{salemi2023lamp} introduces a benchmark of seven tasks requiring LLMs to generate outputs consistent with a specific user's history, and shows that retrieval-augmented approaches significantly close the gap. Extended details on ToM prompting, sandbox architectures, emergent phenomena, digital twins, and institutional approaches appear in Appendix~\ref{app:l2_extended}.

\subsubsection{Laws of the Scientific World}
\label{subsec:l2_science}

In AI for Science, the transition from L1 to L2 shifts the focus from modeling local states or structures to simulating dynamics over multiple steps.
These dynamics arise along two axes.
The first concerns the temporal evolution of a system, where the model predicts how a natural system unfolds over time under given conditions or interventions.
The second concerns the scientific research itself, where the model simulates sequences of hypotheses, experiments, and outcomes to support reasoning and action.
These two forms define the corresponding forms of simulation in scientific world models: forward simulation of system dynamics, and decision simulation based on surrogate evaluation of candidate experiments.

\paragraph{Forward simulation.}
World models approximate the evolution of scientific systems by replacing expensive numerical solvers with learned transition operators.
GNS~\citep{sanchez2020gns} showed that message passing on particle graphs can simulate fluids, rigid bodies, and deformable materials with generalizable dynamics.
The Fourier Neural Operator~\citep{li2021fno} established resolution-invariant operator learning via spectral convolutions, achieving 1000$\times$ speedup over traditional solvers and underpinning subsequent weather and fluid surrogates.
At planetary scale, Pangu-Weather~\citep{bi2023panguweather} and GraphCast~\citep{lam2023graphcast} outperform the ECMWF operational system on 90\% of verification targets.
GenCast~\citep{price2024gencast} extends these to probabilistic forecasting via a diffusion architecture, outperforming the ensemble system on 97.2\% of targets.
NeuralGCM~\citep{kochkov2024neuralgcm} integrates learned parameterizations within a differentiable general circulation model, producing emergent phenomena such as tropical cyclones and illustrating the value of coupling mechanistic structure with learned components.
Aurora~\citep{bodnar2025aurora} further scales this paradigm to a foundation model of the Earth system, achieving strong performance across multiple forecasting tasks at substantially reduced computational cost.
In molecular science, neural network potentials pioneered by \citet{behler2007nnpotentials} enabled orders-of-magnitude speedup over density functional theory for molecular dynamics, establishing the foundation for all subsequent ML fields \citet{deng2023chgnet, chen2025multi, huang2025cross}.

%


\paragraph{Decision simulation.}
World models reduce the cost of scientific discovery by simulating the experimental decision loop in-silico.
Representative systems span molecular design (ChemBO;~\citealt{pmlr-v108-korovina20a}), biological sequence optimization with population-based model ensembles and meta-level search reallocation (P3BO;~\citealt{angermueller2020population}), and materials discovery guided by user-defined algorithmic objectives (BAX;~\citealt{chitturi2024targeted}).
Across these systems, the model simulates not only individual outcomes but the sequential process of experiment selection, maintaining and updating beliefs over candidates while identifying inconsistencies during optimization.
However, these capabilities remain confined to a fixed data regime: the model cannot actively design and execute experiments to acquire \emph{new information} that challenges its current assumptions.
As a result, while such systems can correct optimization errors, they cannot resolve uncertainty arising from incomplete knowledge, leading to accumulated bias over long horizons.
L3 world models (Section~\ref{sec:l3}) overcome this by actively gathering evidence to revise the model.

\subsubsection{Cross-Domain Analysis}
\label{subsec:l2_crossdomain}

\begin{figure}[t]
\centering
\includegraphics[width=\textwidth]{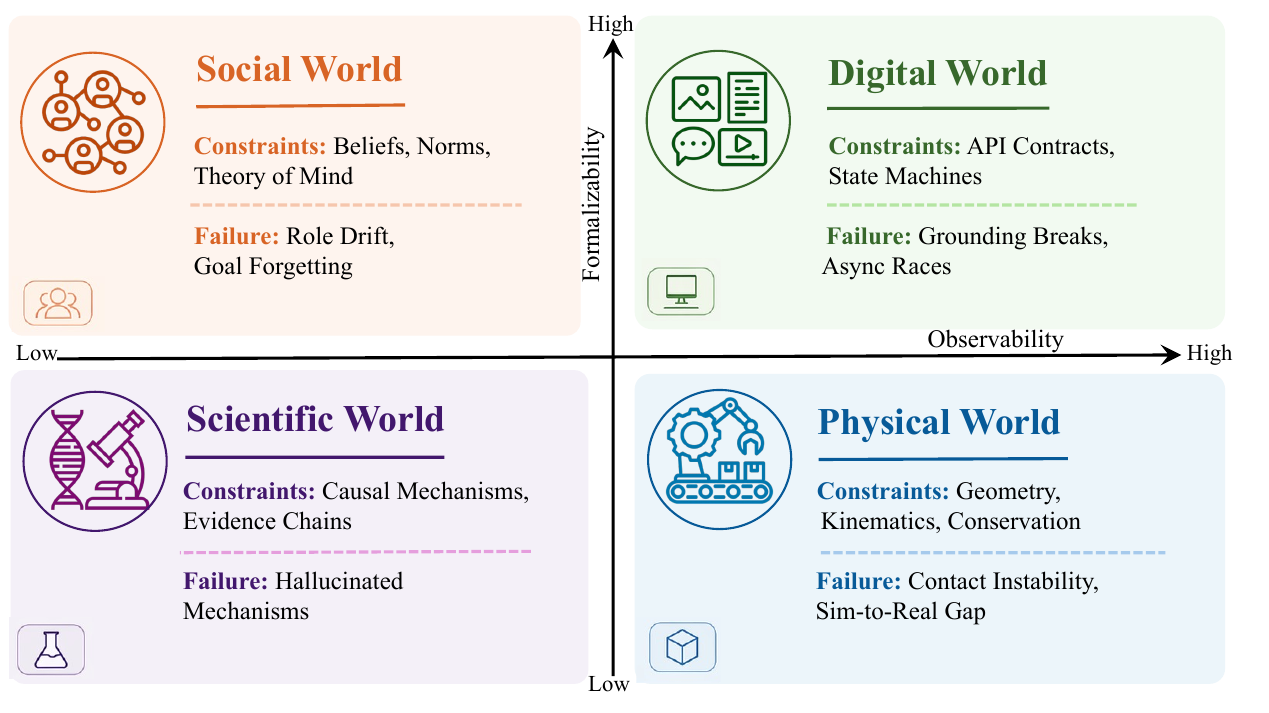}
\caption{\textbf{Diagnostic map of the four governing-law regimes.} The axes are schematic rather than metric:
the horizontal axis reflects how formally specifiable and mechanically verifiable the transition rules are,
while the vertical axis reflects how directly the relevant state and constraints are observable. The purpose
of the figure is comparative rather than classificatory: it highlights why different regimes demand different
forms of rollout validation even when all are instances of L2 simulation. Real systems are often mixed-regime
and may sit between regions rather than inside a single box.}
\label{fig:crossdomain_regime_map}
\end{figure}

\begin{table}[ht]
\caption{Cross-domain comparison of L2 simulators. Each governing-law regime imposes different constraints, failure modes, and evaluation priorities.}
\label{tab:l2_crossdomain}
\centering
\small
\setlength{\tabcolsep}{0.9mm}
\renewcommand{\arraystretch}{1.3}
\resizebox{\textwidth}{!}{%
\begin{tabular}{l|llll}
\toprule
\textbf{Domain} & \textbf{Governing Laws} & \textbf{State Type} & \textbf{Common Failures} & \textbf{Evaluation Focus} \\
\midrule
Physical & Geometry, kinematics & Continuous & Contact instability, drift & Stability; failure clustering \\
Social & Beliefs, norms, ToM & Goals, relations & Role drift, goal forgetting & Counterfactual sensitivity \\
Digital & API contracts, UIs & DOM, permissions & Grounding breaks, races & Error-branch coverage \\
Scientific & Mechanisms, evidence & Hypotheses & Hallucinated mechanisms & Evidence-chain repair \\
\bottomrule
\end{tabular}%
}
\end{table}

Figure~\ref{fig:crossdomain_regime_map} positions the four regimes along two diagnostic axes: formalizability and observability of the governing constraints. Across all four regimes, a recurring pattern emerges: a good Simulator does not have to look more like the world; it must look more like the constraints. Physics uses geometry/contact constraints; software uses state machines and structured feedback channels; social worlds use role/norm consistency; science uses evidence chains and falsifiability. Making constraints explicit (loggable, replayable, regressable) often improves long-horizon stability more than increasing perceptual fidelity. Table~\ref{tab:l2_crossdomain} summarizes the governing laws, state types, common failure modes, and evaluation focus for each regime.

\paragraph{Cross-regime systems.}
Many real-world deployments do not fall neatly into a single governing-law regime; instead, they require an L2 simulator to maintain coherent rollouts across \emph{multiple} constraint families simultaneously. When regimes interact, a violation in one domain can cascade into another: a physically implausible vehicle maneuver may render a social-intent prediction meaningless, or a software bug may invalidate an otherwise sound experimental plan. Designing and evaluating cross-regime systems therefore demands joint constraint satisfaction rather than per-regime evaluation in isolation.

\begin{itemize}[leftmargin=*,nosep]
  \item \textbf{Autonomous driving:} physical (vehicle dynamics, contact mechanics) + social (pedestrian intent prediction, traffic norm compliance)~\citep{hu2023gaia1,wang2024drivedreamer}.
  \item \textbf{Minecraft agents (Voyager):} physical (3D navigation, combat dynamics) + digital (crafting recipes, inventory management, game-state logic)~\citep{wang2023voyager}.
  \item \textbf{Diplomacy (CICERO):} social (negotiation, trust modeling, alliance formation) + digital (game-state management, rule enforcement)~\citep{meta2022cicero}.
  \item \textbf{Autonomous laboratories (A-Lab):} scientific (experiment design, hypothesis evaluation) + physical (sample manipulation, instrument constraints)~\citep{szymanski2023alab, huang2025cascade, fei2026agentic, boiko2023autonomous}.
\end{itemize}

\subsection{Failure Modes}
\label{subsec:l2_failure_modes}

Across all four domains, five recurring failure modes constrain L2 systems:
\begin{enumerate}[leftmargin=*,nosep]
\item \textbf{Compounding error.} Small per-step deviations are amplified over time, pushing imagined trajectories into branches increasingly unrelated to reality. The most effective mitigations are not making one-step predictions look better, but shortening effective planning windows (decomposing long tasks into verifiable short segments and replanning frequently with real feedback), using multi-timescale structure~\citep{shaj2023multitimescale}, and baking evidence-gathering actions into policies.
\item \textbf{State aliasing and drift.} In complex environments, distinct real states can look highly similar (two UI pages, slightly different kitchen layouts, or a one-word change in social tone). When representations collapse these states, agents can take irreversible wrong actions. Effective practices include explicit verification at key nodes, memory and retrieval augmentation, and explicit failure attribution labels~\citep{xie2024osworld,yang2025macosworld,nasiriany2024robocasa}.
\item \textbf{Controllability failure.} A visually rich model that is weakly action-conditioned is less useful for planning than a rough model that responds to actions. When the model is action-insensitive, comparing ``do A vs.\ do B'' becomes meaningless~\citep{wu2024ivideogpt,liu2024lwm,brooks2024sora,deepmind2025genie3}.
\item \textbf{Exploitability and simulator escape.} If a simulator or evaluation harness has loopholes, search/planning will exploit them systematically. This is especially common in software worlds and automated evaluation~\citep{xie2024osworld,yang2025macosworld,zheng2023mcu}.
\item \textbf{Calibration failure under distribution shift.} Environment changes (UI versions, layouts, accents, object properties) often trigger overconfident wrong predictions. In practice, confident but wrong should be treated as a strong signal for evolution~\citep{xie2024osworld,yang2025macosworld,nasiriany2024robocasa}.
\end{enumerate}

These failures are not merely model shortcomings; they are system-level pathologies produced by the interaction between representation, rollout horizon, control procedure, and evidence quality. The takeaway is that improving average-case predictions is insufficient unless systems can (i) localize failures via evidence and (ii) change behavior under shift and exploit pressure.

This constraints-first lens is also a useful guide for choosing what to log and what to regress-test. If the core constraint is violated (e.g., impossible action succeeds, or a structured feedback channel disappears), the agent will learn the wrong lessons. Conversely, when constraints are explicit and stable, even simple agents can improve reliably via Evolver-style asset distillation~\citep{xie2024osworld,yang2025macosworld,nasiriany2024robocasa,zheng2023mcu,ghugare2025builderbench}.

\section{L3 Evolver: Evidence-Driven Model Revision}
\label{sec:l3}

Scientific discovery is naturally organized as a loop: a researcher \emph{designs} experiments, \emph{executes} them, \emph{observes} outcomes, and \emph{reflects} to guide the next step.
Recent systems realize one or more components of this process, but many operate without a fully autonomous loop.
Decision simulators (Section~\ref{subsec:l2_science}) operate within the \emph{design} step: they simulate experimental outcomes and update beliefs, but all updates occur within a fixed information regime without active data collection.
At the other end, RL-based self-reflection systems such as VL-Rethinker~\citep{wang2025vlrethinker} realize the \emph{reflect} step through explicit verify-and-rethink behavior, but do not maintain a persistent world-model stack.
%
Many automated research pipelines~\citep{yang2024moose,li2024ldc} chain \emph{design} and \emph{reflect} (hypothesis generation and literature synthesis) but lack the \emph{execute} and \emph{observe} components that would close the full loop.


The key distinction of L3 lies in how new information is acquired and used.
Rather than passively fitting incoming data or exploiting a fixed model for planning, an L3 system actively designs interventions to reduce uncertainty in its own world model.
In the unified view of Figure~\ref{fig:l1l2l3_graphical}, this revision is the vertical \emph{reflect} arrow that connects the top block (model $\mathcal{M}_t$ operating over environment $\mathcal{X}$) to the bottom block (revised model $\mathcal{M}_{t+1}$ operating over an effective environment $\mathcal{X}'$): each iteration of the loop modifies the latent graph that L2 rolls out on.
Each iteration of the loop targets discrepancies between prediction and observation, using them to refine parameters, extend model structure, or revise underlying assumptions.
In this sense, L3 is driven not by reward maximization alone, but by the systematic reduction of model uncertainty through evidence accumulated across many iterations of design, execution, and reflection.

The meta-learning paradigm, notably ``learning to learn''~\citep{andrychowicz2016learning}, which casts optimizer design itself as a learning problem, foreshadows this self-revision capability: a system improves its own learning procedure rather than merely fitting data. L3 extends this principle from parameter optimization to world-model revision. The result is a paradigm for the next stage of world modeling: systems that not only simulate but also evolve by themselves, acquiring continual learning and self-revision capabilities. In an agent context, this means that the agent's world model is no longer a static artifact consumed at inference time; instead, it becomes a curious living component that diagnoses its own failures, designs targeted experiments to resolve ambiguities, and distills the resulting evidence into persistent model updates. Such self-evolving agents represent a qualitative shift from L2, where the model is a fixed tool for planning, to L3, where the model itself is the object of continual improvement driven by its own deployment experience.

\subsection{Formal Definition}
\label{subsec:l3_definition}

L3 extends L2 from simulation within a fixed information regime to \textbf{closed-loop, evidence-driven model revision}. 
In summary, L3 systems are defined by realizing the entire design--execute--observe--reflect loop, in which new evidence is actively acquired to challenge and revise the model across iterations.

\begin{figure}[t]
\centering
\includegraphics[width=0.85\textwidth]{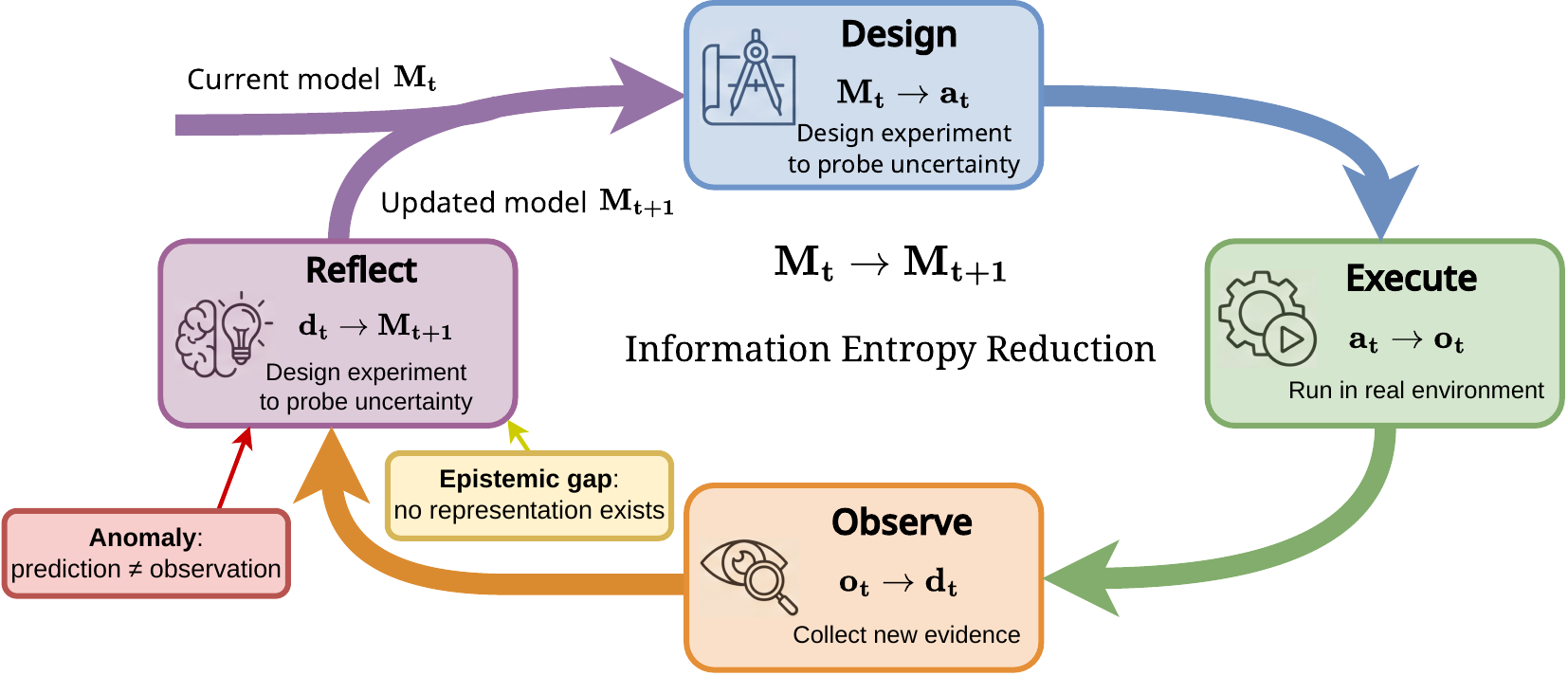}
\caption{\textbf{The L3 evolution loop.} A full cycle proceeds through four stages: design, execute, observe, and reflect, producing a revised world-modeling stack $\mathcal{M}_{t+1}$.}
\label{fig:l3_evolution_loop}
\end{figure}

Formally, an L3 system maintains and updates a world-modeling stack:
\[
\mathcal{M}_t
\xrightarrow{\text{design}} a_t
\xrightarrow{\text{execute}} o_t
\xrightarrow{\text{observe}} d_t
\xrightarrow{\text{reflect}}
\mathcal{M}_{t+1}
\]
where $\mathcal{M}_t$ denotes the current world-modeling stack at iteration $t$, $a_t$ the designed experiment or action, $o_t$ the raw outcome, and $d_t$ the distilled evidence used to update the stack (Figure~\ref{fig:l3_evolution_loop}).

Importantly, the presence of this loop alone is not sufficient to establish L3 capability. What distinguishes L3 systems is that evidence is translated into \emph{persistent, reusable model updates} that are validated under regression checks, rather than remaining as transient, in-context adjustments. The model itself becomes the object of improvement, not merely a fixed substrate for planning.

This formulation closely mirrors the structure of scientific practice.
A scientific community can be viewed as an L3 system operating over a shared model $\mathcal{M}_t$, consisting of established theories together with known anomalies.
Under ``normal science''~\citep{kuhn1962structure}, the community continuously updates $\mathcal{M}_t \rightarrow \mathcal{M}_{t+1}$ through incremental refinements that preserve the underlying model class.
When accumulated anomalies exceed the explanatory capacity of the current model, the same update process produces a more substantial transition, in which the structure of $\mathcal{M}_t$ itself is revised, corresponding to a paradigm shift.
In this view, both gradual refinement and paradigm shifts are instances of the same evidence-driven update, differing only in scale.

In practice, the bottleneck of an L3 system is typically not generating candidate fixes but validating them safely. Multimodal critic models~\citep{zhang2025critic} and regression-gated update pipelines~\citep{ren2026aligning,jimenez2023swebench,yang2024sweagent} provide practical infrastructure for the observe and reflect stages. We cite these systems not as full L3 exemplars, but as components of an evolver design space once persistent update and validation mechanisms are in place.

\paragraph{Revision triggers and evolution policy.}
The reflect stage is responsible for deciding \emph{when} and \emph{how} the world model should be revised, in particular distinguishing between incremental improvement and structural change. In practice, this decision is driven by two types of signals.
An \emph{anomaly} denotes a mismatch between prediction and observation. While many anomalies can be absorbed through local adjustments, persistent anomalies that resist resolution within the current model class reveal an \emph{epistemic gap}, indicating that the underlying representation or hypothesis space is insufficient. Resolving such gaps typically requires structural changes to the model, corresponding to a paradigm shift.
In the philosophical framing of Section~\ref{sec:philosophy_l3}, anomalies that can be absorbed correspond to adjustments within Lakatos's ``protective belt'' (learned parameters), while persistent anomalies that expose epistemic gaps demand changes to the ``hard core'' (architecture, inductive biases). In operational terms, this induces a hierarchy of responses: small anomalies are handled via online adaptation within an episode, persistent anomalies trigger parameter updates that are distilled into the model, and epistemic gaps require structural modifications such as introducing new modules or expanding the hypothesis space.
Duhem--Quine holism (Section~\ref{sec:philosophy_l3}) highlights that this attribution is inherently ambiguous: a mismatch between prediction and observation can often be explained by multiple components of the model, including representation, dynamics, or auxiliary assumptions. As a result, it is non-trivial to determine whether an anomaly can be resolved through local updates within the current model class or reflects a deeper epistemic gap that requires structural change to the underlying representation.

This difficulty is further shaped by the choice of representation.
Epistemic gaps often involve missing or incorrect inductive biases or invariances in the model.
While such gaps can, in principle, be addressed through structural changes in learned models, these changes are typically indirect: modifying architecture or training procedures does not guarantee a predictable change in the underlying invariances the model captures.
In contrast, symbolic representations allow these invariances to be expressed and manipulated explicitly, as seen in scientific laws and principles.
This suggests that while latent representations provide a flexible substrate for absorbing anomalies through parameter updates, resolving epistemic gaps may require representations that expose and manipulate invariances explicitly, as is standard in symbolic scientific models.
We return to this tension between latent and symbolic representations in Section~\ref{sec:trends:representation}.

\subsection{Distinction from L2}
\label{subsec:l2_vs_l3}

We use three boundary conditions to mark L2 $\rightarrow$ L3, each corresponding to a transition in the design--execute--observe--reflect loop:
\begin{enumerate}
    \item \textbf{Active information expansion ($\mathcal{M}_t \rightarrow e_t$)}: the system designs experiments that actively probe uncertainty or challenge its current belief, rather than only optimizing within existing knowledge.
    \item \textbf{Autonomous execution and observation ($e_t \rightarrow d_t$)}: the system carries out experiments and acquires evidence through interaction, rather than relying on simulated or pre-existing data.
    \item \textbf{Belief revision under challenge ($d_t \rightarrow \mathcal{M}_{t+1}$)}: observations are used to reflect on and revise the model, including updating parameters, structure, or assets, enabling correction of prior assumptions.
\end{enumerate}

The boundary conditions above can be unified as a single principle: whether the world model remains fixed or becomes plastic during deployment.
This transition from L2 to L3 manifests in three aspects: whether the model can update its parameters and structure after deployment, how it accumulates new capabilities over time, and whether it passively consumes data or actively generates it through experimentation.

\paragraph{Fixed vs.\ adaptive.}
An L2 simulator is typically fixed post-training. It can generate infinite rollouts based on its training data, but its core transition function $p_\theta(z_t \mid z_{t-1}, a_t)$ does not evolve; it explores the implications of its frozen knowledge. In contrast, an L3 system is adaptive post-deployment: it treats its own parameters or structure as a hypothesis to be updated, i.e. $\mathcal{M}_{t+1} \leftarrow \mathcal{M}_t + \text{Evidence}$.

\paragraph{Modes of growth.}
L3 growth goes beyond simple data buffering and encompasses three different modes:
\begin{itemize}[nosep]
    \item \textbf{Parameter update}: modifying weights via gradient descent or Bayesian updates on new evidence, e.g., online learning, continual RL fine-tuning, and Bayesian model updates.
    \item \textbf{Architecture update}: dynamically adding new modules, experts, or capacity to handle complexity, for example, expanding the context window or allocating new memory slots.
    \item \textbf{Hypothesis-space expansion}: extending the model class to represent explanations that were previously inexpressible. This corresponds to introducing new variables, mechanisms, or abstractions, shifting from ``I don't know which of these $k$ options is true'' to ``the correct explanation is not among the current $k$ options.'' This is the most challenging mode and is closely tied to abduction and genuine scientific discovery.
\end{itemize}

\paragraph{Passive vs.\ active.}
While L2 systems may support passive online learning (updating weights on a stream of incoming data) or decision simulating (Section~\ref{subsec:l2_science}), L3 is characterized by \emph{active} \emph{trial-and-error loop}.
It does not just wait for data; it acts to generate data that maximizes information gain regarding a specific hypothesis or area of uncertainty. This active stance transforms the agent from a consumer of experience to a designer of experiments, a qualitative shift that connects directly to the philosophy of abduction and scientific method (Section~\ref{sec:philosophy_l3}). L3 should not be defined by closed-loop use in the generic planning sense; rather, it is defined by closing the evidence-to-revision loop, so that deployment outcomes are used to diagnose, update, and validate the world-modeling stack itself over successive iterations of use.

\subsection{Examples and Applications}
\label{subsec:l3_examples}

\begin{figure*}[t]
\centering
\includegraphics[width=\textwidth]{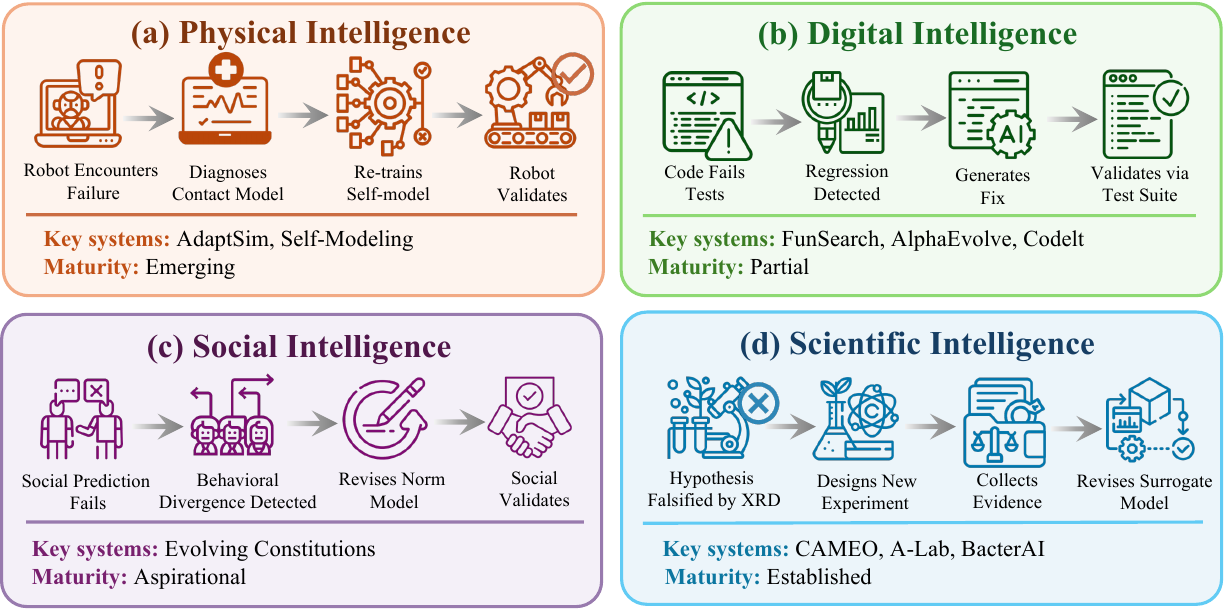}
\caption{\textbf{L3 evolution across four governing-law regimes.} Each panel illustrates the design--execute--observe--reflect loop in a representative domain: (a)~Physical intelligence---adaptive probing revises contact dynamics; (b)~Social intelligence---norm drift triggers social-model revision; (c)~Digital intelligence---evaluator-driven program search with regression gates; (d)~Scientific intelligence---closed-loop autonomous experimentation at a synchrotron beamline.}
\label{fig:l3_domains}
\end{figure*}

L3 is most tractable in domains that are highly instrumented, offer rapid feedback, and provide well-defined evaluation criteria.
Empirical support for L3 is uneven across domains: autonomous science and other highly instrumented settings provide the clearest demonstrations, whereas social, code, and embodied environments remain partly empirical and partly prospective design space.
We illustrate this landscape, together with the characteristic evidence signals and failure modes in each, across four governing-law regimes in Figure~\ref{fig:l3_domains}.


\paragraph{Physical intelligence.}
In embodied settings, L3 manifests as adaptive probing to infer and update dynamics models. When a robot encounters unexpected contact dynamics, such as a slippery surface or a deformable object, the system can actively execute diagnostic actions (small perturbations designed to disambiguate between hypotheses about the contact model) and use the resulting evidence to update its dynamics model. The anomaly signals in this regime are inherently physical: force/torque deviations, unexpected contact events, and discrepancies between predicted and observed end-effector trajectories provide quantitative evidence for model updates. Recent work demonstrates that robots can autonomously detect physical damage and re-train persistent self-models: \citet{hu2025selfmodeling} show that an egocentric visual self-model detects morphology changes via prediction-versus-observation mismatch and re-trains to recover locomotion. AdaptSim~\citep{ren2023adaptsim} meta-learns an adaptation policy that iteratively revises simulation parameters from small amounts of real-world task performance data, closing the sim-to-real gap through evidence-driven simulation revision rather than fixed domain randomization, with each real-world deployment informing the next round of simulation updates (see Appendix~\ref{app:l3_examples} for a worked physical-intelligence example).

\paragraph{Digital intelligence.}
Software and web environments are naturally suited to L3 because state is fully observable, actions are deterministically replayable, and regression testing provides a built-in validation gate. Evaluator-driven discovery loops exemplify this regime. \citet{romera2024funsearch} pair a pretrained LLM with an automated evaluator in an evolutionary loop: the LLM generates candidate programs, the evaluator scores them against a formal specification, and high-scoring solutions are fed back for further refinement. This loop discovered new constructions for the cap set problem (a long-standing open problem in combinatorics) and new bin-packing heuristics that outperform known baselines. The evaluator serves as an automated regression gate, a key L3 property, although the system realizes only the design and observe components (program generation and automated scoring) without active information expansion or persistent model revision. \citet{alphaevolve2025} extend this evolutionary coding paradigm: by pairing LLM-generated program mutations with automated correctness evaluators, the system improved on Strassen's matrix multiplication algorithm after 56 years and solved 20\% of open mathematical problems beyond the prior state of the art, illustrating the power of formal verification as an L3 gatekeeper in algorithmic domains. CodeIt~\citep{butt2024codeit} closes a tighter loop: the LLM is fine-tuned from its own search trajectories via prioritized hindsight replay, so that the generative model itself (serving as an implicit world model of program space) persistently improves across tasks. The AI Scientist-v2~\citep{yamada2025aiscientistv2} pushes further into computational experiments by employing agentic tree search for experiment selection: the system autonomously formulates hypotheses, designs and executes experiments, analyzes results, and writes complete manuscripts. A VLM feedback loop iteratively refines figures and content. In 2025, this system produced an entirely AI-generated paper that passed peer review at an ICLR workshop. However, the system's experiments are computational (running ML training jobs), and its revision loop operates on paper quality rather than mechanistic understanding, illustrating the gap between L3 in well-instrumented computational domains and the harder challenge of genuine scientific discovery.
In AUI~\citep{lin2025computer}, a Coder--Computer-Use Agent loop instantiates this principle in website: the Coder iteratively revises website implementations, while the CUA acts as an automated evaluator by executing task trajectories and verifying functional correctness (e.g., navigation success and task completion). The resulting feedback, grounded in executable interactions rather than static inspection, serves as a regression signal that guides subsequent code updates, forming a closed-loop optimization process aligned with L3 properties.

\paragraph{Social intelligence.}
L3 in social domains requires revising the agent's social model when predicted behavior of other agents deviates from observed behavior, for example, when Theory-of-Mind predictions fail systematically or when social norms drift over time. This is currently the hardest regime for L3 because attribution is inherently ambiguous (a failed social prediction may reflect incorrect beliefs about the other agent's goals, an outdated norm model, or stochastic behavior) and because social experiments are ethically constrained. Early work on norm emergence and convention formation in multi-agent populations (Section~\ref{subsec:l2_social}) represents a preliminary step toward social L3, but persistent, validated revision of social world models from deployment evidence remains largely open. A preliminary step toward social L3 is the evolutionary synthesis of multi-agent governance rules: \citet{kumar2026constitutions} use LLM-driven genetic programming to evolve interpretable constitutions from societal stability scores, surpassing human-designed rules by 123\%.

\paragraph{Scientific intelligence.}
The most complete current examples of L3 come from autonomous science, where the full design--execute--observe--reflect loop is closed by instrumentation.
The paradigm of autonomous closed-loop scientific discovery was established by Robot Scientist Adam~\citep{sparkes2010robotscientist}, the first machine to autonomously design experiments about gene function, execute them, observe the outcomes, and revise its model. Its successor system demonstrated closed-loop cycles of experiment design, execution, and model revision in yeast systems biology, accelerating biological model development~\citep{coutant2019yeast}.
CAMEO~\citep{kusne2020cameo} implements closed-loop materials discovery via Bayesian active learning at a synchrotron beamline: the system predicts which phase a candidate composition will form, synthesizes it, characterizes the product via X-ray diffraction, updates its Bayesian belief model, and actively selects the next experiment to maximize information gain. Each experimental cycle takes seconds to minutes, and the system discovered a novel phase-change memory material without additional human training.
A-Lab~\citep{szymanski2023alab} extends this to fully autonomous synthesis: three robotic arms automate powder dosing, heating, and XRD characterization, with an active-learning algorithm generating improved recipes when targets fail. In 17 days of closed-loop operation, A-Lab performed 353 experiments and realized 36 compounds from 57 targets. Crucially, analysis of failed syntheses provided structured evidence to refine future synthesis strategies; the failures were not discarded but distilled into persistent knowledge.
\citet{striethkalthoff2024sdl} extend the self-driving laboratory paradigm to distributed, multi-site operation: a delocalized SDL autonomously discovers novel organic laser emitters by iteratively updating a Bayesian surrogate from synthesis and characterization data across geographically separated facilities. BacterAI~\citep{dama2023bacterai} demonstrates that L3 can operate with zero prior biological knowledge: the system iteratively designs and executes experiments to map microbial amino acid requirements, revising its metabolic model purely from experimental evidence.
In computational chemistry, MOOSE-Chem~\citep{yang2025moosechem} demonstrates that an LLM-based framework can rediscover chemistry hypotheses published in Nature and Science in 2024 using only pre-2024 literature, providing evidence that the hypothesis-generation component of the L3 loop is already feasible for natural-science domains. Its successor, MOOSE-Chem2~\citep{yang2025moosechem2}, introduces hierarchical search over fine-grained hypothesis components to improve both precision and novelty of generated discoveries. CASCADE~\citep{huang2025cascade} provides a self-evolving multi-agent framework for materials science and chemistry that cumulatively acquires reusable executable skills via continuous learning and self-reflection meta-skills, complemented by memory systems, benchmarked across hundreds of functions in over 20 databases, packages, and software tools, with applications in hypothesis-driven discovery and autonomous laboratory workflows. Appendix~\ref{app:l3_examples} presents worked examples spanning all four regimes.
Broader agentic systems are pushing the L3 loop further into biomedicine. Biomni~\citep{huang2025biomni} provides a general-purpose biomedical AI agent that integrates over 100 tools and 59 databases spanning 25 subfields, enabling autonomous execution of tasks from causal gene prioritization to drug repurposing. BioLab~\citep{jin2025biolab} extends this to end-to-end autonomous life-sciences research via a multi-agent system built on biological foundation models. OriGene~\citep{zhang2025origene} demonstrates a self-evolving virtual disease biologist that autonomously discovers therapeutic targets through iterative hypothesis refinement. The AI co-scientist system~\citep{gottweis2025coscientist} employs a generate--debate--evolve approach to hypothesis generation, with multi-agent tournament processes that have been validated in drug repurposing and epigenetic target discovery. Complementing these systems, \citet{yang2026llmknowledge} introduce a dynamic benchmark revealing that current LLMs still fall short on genuine biological knowledge derivation, underscoring the persistent gap between literature retrieval and true L3 revision that actually updates the underlying model.

\begin{table*}[!t]
\caption{\textbf{Representative L3 systems by governing-law regime.} Loop steps indicate which stages of the design, execute, observe, and reflect cycle each system realizes.}
\label{tab:l3_systems}
\centering
\small
\setlength{\tabcolsep}{0.9mm}
\renewcommand{\arraystretch}{1.05}
\begin{tabular}{l|cc|cccc}
\toprule
\textbf{System} & \multicolumn{2}{c|}{\textbf{Links}} & \textbf{Design} & \textbf{Execute} & \textbf{Observe} & \textbf{Reflect} \\
\midrule

\rowcolor{gray!15}
\multicolumn{7}{c}{\textit{Physical World}} \\
\shortstack[l]{AdaptSim~\citep{ren2023adaptsim}}
& \paperlink{https://arxiv.org/abs/2302.04903}
& \githublink{https://github.com/irom-princeton/AdaptSim}
& \cmark & \cmark & \cmark & \xmark \\
\shortstack[l]{Self-Modeling~\citep{hu2025selfmodeling}}
& \paperlink{https://arxiv.org/abs/2207.03386}
& \githublink{https://github.com/H-Y-H-Y-H/Egocentric_VSM}
& \cmark & \cmark & \cmark & \cmark \\
\midrule

\rowcolor{gray!15}
\multicolumn{7}{c}{\textit{Digital World}} \\
\shortstack[l]{FunSearch~\citep{romera2024funsearch}}
& \paperlink{https://doi.org/10.1038/s41586-023-06924-6}
& \githublink{https://github.com/google-deepmind/funsearch}
& \cmark & \cmark & \cmark & \xmark \\
\shortstack[l]{CodeIt~\citep{butt2024codeit}}
& \paperlink{https://arxiv.org/abs/2402.04858}
& \githublink{https://github.com/Qualcomm-AI-research/codeit}
& \cmark & \cmark & \cmark & \cmark \\
\shortstack[l]{SWE-agent~\citep{yang2024sweagent}}
& \paperlink{https://arxiv.org/abs/2405.15793}
& \githublink{https://github.com/princeton-nlp/SWE-agent}
& \cmark & \cmark & \cmark & \xmark \\
\shortstack[l]{AUI~\citep{lin2025computer}}
& \paperlink{https://arxiv.org/abs/2511.15567}
& \githublink{https://github.com/showlab/AUI}
& \cmark & \cmark & \cmark & \xmark \\
\shortstack[l]{AlphaEvolve~\citep{alphaevolve2025}}
& \paperlink{https://arxiv.org/abs/2506.13131}
& \githublink{https://github.com/google-deepmind/alphaevolve_results}
& \cmark & \cmark & \cmark & \xmark \\
\midrule

\rowcolor{gray!15}
\multicolumn{7}{c}{\textit{Social World}} \\
\shortstack[l]{Evolving Const.~\citep{kumar2026constitutions}}
& \paperlink{https://arxiv.org/abs/2602.00755}
& {---}
& \cmark & \cmark & \cmark & \cmark \\
\shortstack[l]{AgentSociety~\citep{piao2025agentsociety}}
& \paperlink{https://arxiv.org/abs/2502.08691}
& \githublink{https://github.com/tsinghua-fib-lab/AgentSociety}
& \cmark & \cmark & \cmark & \xmark \\
\midrule

\rowcolor{gray!15}
\multicolumn{7}{c}{\textit{Scientific World}} \\
\shortstack[l]{Robot Scientist~\citep{sparkes2010robotscientist}}
& \paperlink{https://doi.org/10.1186/1759-4499-2-1}
& {---}
& \cmark & \cmark & \cmark & \cmark \\
\shortstack[l]{CAMEO~\citep{kusne2020cameo}}
& \paperlink{https://arxiv.org/abs/2006.06141}
& \githublink{https://github.com/KusneNIST/CAMEO_NComm}
& \cmark & \cmark & \cmark & \cmark \\
\shortstack[l]{Yeast Cycles~\citep{coutant2019yeast}}
& \paperlink{https://doi.org/10.1073/pnas.1900548116}
& {---}
& \cmark & \cmark & \cmark & \cmark \\
\shortstack[l]{BacterAI~\citep{dama2023bacterai}}
& \paperlink{https://doi.org/10.1038/s41564-023-01376-0}
& \githublink{https://github.com/jensenlab/BacterAI}
& \cmark & \cmark & \cmark & \cmark \\
\shortstack[l]{A-Lab~\citep{szymanski2023alab}}
& \paperlink{https://doi.org/10.1038/s41586-023-06734-w}
& {---}
& \cmark & \cmark & \cmark & \cmark \\
\shortstack[l]{SDL Lasers~\citep{striethkalthoff2024sdl}}
& \paperlink{https://doi.org/10.1126/science.adk9227}
& \githublink{https://github.com/aspuru-guzik-group/acdc_laser}
& \cmark & \cmark & \cmark & \cmark \\
\shortstack[l]{AI Scientist~\citep{lu2024aiscientist}}
& \paperlink{https://arxiv.org/abs/2408.06292}
& \githublink{https://github.com/SakanaAI/AI-Scientist}
& \cmark & \cmark & \cmark & \cmark \\
\shortstack[l]{CASCADE~\citep{huang2025cascade}}
& \paperlink{https://arxiv.org/abs/2512.23880}
& \githublink{https://github.com/CederGroupHub/CASCADE}                   & \cmark & \cmark & \cmark & \cmark \\
\shortstack[l]{Biomni~\citep{huang2025biomni}}
& \paperlink{https://doi.org/10.1101/2025.05.30.656746}
& \githublink{https://github.com/snap-stanford/Biomni}
& \cmark & \cmark & \cmark & \xmark \\
\shortstack[l]{BioLab~\citep{jin2025biolab}}
& \paperlink{https://doi.org/10.1101/2025.09.03.674085}
& {---}
& \cmark & \cmark & \cmark & \cmark \\
\shortstack[l]{OriGene~\citep{zhang2025origene}}
& \paperlink{https://doi.org/10.1101/2025.06.03.657658}
& \githublink{https://github.com/GENTEL-lab/OriGene}
& \cmark & \cmark & \cmark & \cmark \\
\shortstack[l]{Co-Scientist~\citep{gottweis2025coscientist}}
& \paperlink{https://arxiv.org/abs/2502.18864}
& {---}
& \cmark & \cmark & \cmark & \cmark \\
\shortstack[l]{AI Scientist v2~\citep{yamada2025aiscientistv2}}
& \paperlink{https://arxiv.org/abs/2504.08066}
& \githublink{https://github.com/SakanaAI/AI-Scientist-v2}
& \cmark & \cmark & \cmark & \cmark \\
\bottomrule
\end{tabular}
\end{table*}

Table~\ref{tab:l3_systems} summarizes representative L3 systems across the four governing-law regimes, indicating which stages of the design--execute--observe--reflect loop each system realizes.

\paragraph{Evidence quality and falsifiability.}
The quality of evolution depends on the quality of evidence. Table~\ref{tab:l3_evidence} organizes the revision signals that trigger L3 model updates in each governing-law regime: what the agent detects, why it indicates the current model is wrong, and how falsifiable the signal is.
\begin{table}[t]
\caption{\textbf{Revision signals for L3 evolution by governing-law regime.} Row color encodes \emph{within-domain} falsifiability (not comparable across regimes): \colorbox{green!20}{High}, \colorbox{yellow!30}{Medium}, \colorbox{red!25}{Low}.}
\label{tab:l3_evidence}
\centering
\begin{tabular}{@{}ll@{}}
\toprule
\textbf{Revision Signal} & \textbf{Trigger Condition} \\
\midrule
\rowcolor{gray!15} \multicolumn{2}{l}{Physical World} \\
\addlinespace
\rowcolor{green!20} Kinematic infeasibility~\citep{ren2023adaptsim} & Trajectory violates joint limits or collision bounds \\
\rowcolor{yellow!30} Contact dynamics mismatch~\citep{hu2025selfmodeling} & Force/torque deviates from predicted contact model \\
\rowcolor{red!25} Morphology change~\citep{hu2025selfmodeling} & Visual self-model diverges from observed body state \\
\midrule
\rowcolor{gray!15} \multicolumn{2}{l}{Social World} \\
\addlinespace
\rowcolor{green!20} Interventional inconsistency~\citep{piao2025agentsociety} & \shortstack[l]{Policy change fails to elicit proportional behavioral \\ shift} \\
\rowcolor{yellow!30} Global behavioral drift~\citep{kumar2026constitutions} & \shortstack[l]{Parameter perturbation yields inconsistent aggregate \\ response} \\
\rowcolor{red!25} \shortstack[l]{Individual faithfulness violation\\~\citep{taubenfeld2024biases}} & Agent behavior deviates from demographic priors \\
\midrule
\rowcolor{gray!15} \multicolumn{2}{l}{Digital World} \\
\addlinespace
\rowcolor{green!20} Regression detection~\citep{romera2024funsearch} & Previously passing test fails post-update \\
\rowcolor{yellow!30} Execution outcome mismatch~\citep{lin2025computer} & Predicted state differs from actual execution result \\
\rowcolor{red!25} Task completion failure~\citep{butt2024codeit} & Action sequence fails to achieve specified goal \\
\midrule
\rowcolor{gray!15} \multicolumn{2}{l}{Scientific World} \\
\addlinespace
\rowcolor{green!20} Hypothesis falsification~\citep{szymanski2023alab} & Experiment contradicts predicted outcome \\
\rowcolor{yellow!30} Prediction--measurement gap~\citep{kusne2020cameo} & Surrogate output diverges from measurement \\
\rowcolor{red!25} Epistemic gap detection~\citep{dama2023bacterai} & \shortstack[l]{Observation falls outside model's representational \\ scope} \\
\bottomrule
\end{tabular}
\end{table}
A useful principle is to prefer falsifiable evidence (Section~\ref{sec:philosophy_l3}). A screenshot combined with a DOM snapshot, error code, and action sequence is reproducible and refutable; ``I think the page didn't load'' is not. Human feedback should not be treated as a single falsifiability class: subjective or preference feedback is weakly falsifiable, whereas expert diagnostic feedback can be strongly falsifiable when its claims are subsequently checked by tests, experiments, or structured evaluation. Evolver's progress depends on making lessons verifiable, and reversible when wrong. This requirement connects directly to the anomaly and epistemic-gap triggers defined in Section~\ref{subsec:l3_definition}: an anomaly is actionable only when the deviation between prediction and observation can be quantified from recorded evidence, and an epistemic gap is recognizable only when the system can demonstrate that no existing hypothesis adequately accounts for the observation. In large-scale deployments, evidence must also be compressible and indexable. Practical systems maintain multi-resolution evidence: a compact error category combined with a state fingerprint and diff summary for fast retrieval, together with pointers to heavier artifacts (screenshots, DOM snapshots, full logs) for deep audits. Evidence quality is also tightly coupled to privacy and safety constraints: an Evolver pipeline must separate what is stored persistently (sanitized logs, hashed fingerprints) from what is kept transient or behind access controls, protecting sensitive data while retaining an audit trail~\citep{xie2024osworld,yang2025macosworld}.

Continuous self-improvement also introduces governance challenges, including benchmark overfitting, knowledge contamination, and misattribution of failures to wrong components. These risks and the practical measures to mitigate them (versioning, rollback, regression gates) are discussed as open problems in Section~\ref{sec:trends}.

\subsection{L3 in Context: Maturity, Governance, and Outlook}
\label{subsec:l3_context}

Having established the L3 evolution loop, its domain instantiations, and the role of evidence quality, we now examine its practical status and implications. This subsection addresses two complementary questions: \emph{maturity}, i.e., where L3 systems have been successfully realized across governing-law regimes; and \emph{governance}, i.e., what risks arise from persistent, automated model revision. Together, these perspectives characterize L3 both as a modeling paradigm and as a deployed system that must evolve reliably under real-world constraints.

\paragraph{Maturity across different domains.}
We summarize maturity across the four governing-law regimes:
\begin{enumerate}[leftmargin=*]
  \item \textbf{Scientific (Established).} The most mature regime, offering fast, structured feedback, unambiguous anomaly signals (hypothesis falsification), and well-defined revision targets (surrogate model parameters, synthesis recipes)~\citep{kusne2020cameo, szymanski2023alab, sparkes2010robotscientist, dama2023bacterai}. Primary bottleneck: instrument access and real-data budget.
  \item \textbf{Digital (Partial).} Regression testing provides an automated validation gate, but many systems still lack the active information-expansion boundary condition~\citep{romera2024funsearch, alphaevolve2025, butt2024codeit}. Primary bottleneck: active experiment design is often absent.
  \item \textbf{Physical (Emerging).} Promising but limited by attribution difficulty: a failed manipulation can stem from perception, dynamics, actuation, or environmental change, and isolating the brittle component requires careful experimental design~\citep{ren2023adaptsim, hu2025selfmodeling}. Primary bottleneck: failure attribution across perception, dynamics, and actuation.
  \item \textbf{Social (Aspirational).} Social experiments are ethically constrained, attribution is inherently ambiguous, and behavioral ground truth is noisy~\citep{kumar2026constitutions}. Primary bottleneck: attribution ambiguity and ethical constraints on social experimentation.
\end{enumerate}

\paragraph{Governance challenges.}
Three governance risks arise specifically from persistent, automated model revision. \emph{Benchmark overfitting} occurs when the regression gate is too close to the training distribution; the system learns to pass tests rather than improve genuinely. \emph{Knowledge contamination} occurs when the revision loop incorporates evidence that is itself biased or adversarially constructed, silently degrading the model on OOD inputs. \emph{Misattribution cascades} occur when a fix for one failure mode inadvertently degrades another component; without comprehensive regression suites, the net effect of an update can be negative. Mitigations include held-out probe sets that are refreshed independently of training data, canary deployment that surfaces regressions before full rollout, and causal ablations that isolate the contribution of each update.

\paragraph{Relationship to Sections~\ref{sec:evaluation} and~\ref{sec:implementation}.}
From an evaluation perspective (Section~\ref{sec:evaluation}), assessing L3 requires protocols that go beyond single-episode accuracy: the key metric is whether the system improves across revision cycles $k$ without regressing on held-out probes. From an implementation perspective (Section~\ref{sec:implementation}), L3 places the heaviest demands on the system stack (persistent storage, replay infrastructure, regression harnesses, and rollback mechanisms) that are often underspecified in current architectures. Building toward L3 therefore means investing in evaluation infrastructure as much as in model capacity.

\section{Evaluations}
\label{sec:evaluation}

Evaluating world models for agentic AI requires moving beyond standard generative metrics toward decision-centric protocols organized around three boundary conditions: long-horizon coherence, intervention sensitivity, and constraint consistency. This section first motivates this shift (Section~\ref{subsec:eval_decision}), then maps the benchmark landscape by governing-law regime and provides detailed evaluation protocols for each condition (Section~\ref{subsec:eval_boundary}), and finally shows how the \emph{same} benchmark can test L1, L2, or L3 depending on the evaluation protocol (Section~\ref{subsec:eval_levels}). World-model-specific evaluations show that even frontier models still suffer from substantial capability gaps, while no single benchmark fully captures the space of interest. For further clarification, Appendix~\ref{app:eval_extended} provides a capability coverage matrix and a Minimal Reproducible Evaluation Package (MREP).

\subsection{From prediction-centric to decision-centric evaluation}
\label{subsec:eval_decision}

While standard generative metrics such as Fr\'{e}chet Inception Distance (FID), Fr\'{e}chet Video Distance (FVD), SSIM, and per-pixel reconstruction loss capture perceptual quality, they are at best weak indicators~\citep{brooks2024sora,deepmind2025genie3} of agentic capability and offer limited predictive power for the downstream decisions an agent must ultimately make once embedded in the real-world environment.

As a result, a world model can generate visually convincing rollouts while still breaking down during planning because of hallucinated object dynamics, action-insensitive transitions, or subtle physics violations. These errors are often invisible to distribution-level metrics but devastating for downstream decision-making.

The root cause is a mismatch between what is measured and what matters. The object of evaluation should not be a single-step prediction $p_\theta(z_t\mid z_{t-1},a_t)$ in isolation, but the \textbf{trajectory-level rollout}
\[
\hat p(\tau \mid z_0, a_{1:H}, c), \qquad \tau=(z_1,\ldots,z_H),
\]
and specifically whether this rollout is reliable enough for a planner to act on (Section~\ref{sec:l2}). Aggregate measures such as mean success rates further obscure the picture by masking high variance across task instances~\citep{agarwal2021statistical,henderson2018deep}.

We therefore organize evaluation around the three \textbf{boundary conditions} that mark L1$\to$L2 (Section~\ref{subsec:l2_requirements}):
\begin{enumerate}[leftmargin=*]
  \item \textbf{Long-horizon coherence:} rollouts remain decision-usable over $H$ steps rather than degrading via error.
  \item \textbf{Intervention sensitivity:} counterfactual edits (action or premise changes) induce stable and directionally meaningful trajectory changes.
  \item \textbf{Constraint consistency:} generated futures respect the governing laws of the target regime (Section~\ref{sec:l2}).
\end{enumerate}
These three conditions hold across the four governing-law regimes introduced in Section~\ref{sec:l2}, and together they give a common framework within which we can organize the evaluation protocols, benchmark analyses, and reporting standards described in the remainder of this section.

World-model evaluation is ultimately meaningful only insofar as it reflects downstream decision quality. Benchmarks that capture long-horizon coherence, intervention sensitivity, or constraint consistency are valuable not merely as diagnostics, but because these properties should translate into better plan selection, fewer costly invalid actions, and greater task success under distribution shift. The relevant bridge is therefore not ``Does the model look realistic?'' but ``Does improved model validity change what the agent chooses, and does that shift in choice in turn improve real-world task outcomes?''

Two aggregate metrics operationalize these conditions for downstream decision-making. The \textbf{Action Success Rate} (ASR) measures how often a planner that uses the world model's rollouts to select actions achieves the task goal in the real environment:
\[
\mathrm{ASR} = \frac{1}{N}\sum_{i=1}^{N} \mathbf{1}\bigl[\text{task}_i \text{ succeeds under policy derived from } \hat{p}\bigr].
\]
The \textbf{Counterfactual Outcome Deviation} (COD) measures intervention sensitivity by comparing rollout outcomes under two policies $a^{(1)}_{1:H}$ and $a^{(2)}_{1:H}$ that differ at a single intervention step $k$:
\[
\mathrm{COD}(k) = \mathbb{E}\bigl[d\bigl(\hat{z}^{(1)}_H,\, \hat{z}^{(2)}_H\bigr)\bigr],
\]
where $d$ is a task-relevant distance (e.g., goal-state distance in physical tasks, edit distance in software tasks). When COD is low, a world model is largely unresponsive to changes in action, which makes it uninformative for counterfactual planning. Together, ASR and COD provide a more direct link between world-model quality and downstream agentic performance: ASR assesses whether the model supports good decisions, whereas COD assesses whether the model responds in a meaningful way to action-level interventions.

\subsection{Evaluating the three boundary conditions}
\label{subsec:eval_boundary}

No single benchmark can evaluate an agent's mastery of all world rules. Because benchmark selection heavily shapes which boundary conditions are actually tested, we first sketch the landscape by governing-law regime.

\paragraph{Benchmark landscape by laws.}
In physical-world domains, the \textbf{Atari 100k} benchmark~\citep{kaiser2019simple} tests sample-efficient world-model learning under a strict 100k-step budget across 26 games, while \textbf{Meta-World}~\citep{yu2020metaworld} provides 50 distinct robotic manipulation tasks for multi-task and meta-RL evaluation. \textbf{CALVIN}~\citep{mees2022calvin} evaluates language-conditioned long-horizon manipulation with 24 hours of teleoperated play data and 20K language directives. \textbf{RoboCasa}~\citep{nasiriany2024robocasa} and its successor RoboCasa365~\citep{nasiriany2026robocasa365} test long-horizon manipulation and physical stability, while \textbf{BuilderBench}~\citep{ghugare2025builderbench} tests structural stability under physical load. \textbf{ManiSkill3}~\citep{tao2024maniskill3} and \textbf{RLBench}~\citep{james2020rlbench} provide large-scale demonstrations for generalizable manipulation. In autonomous driving, \textbf{nuScenes}~\citep{caesar2020nuscenes} provides the standard multimodal benchmark with full 360-degree sensor suite across 1000 scenes. The \textbf{Habitat} series~\citep{savva2019habitat,szot2021habitat20,puig2023habitat3,yokoyama2024hm3d} remains the standard for 3D navigation and rearrangement, \textbf{iGibson~2.0} and \textbf{BEHAVIOR-1K}~\citep{li2021igibson2,li2024behavior1k} extend to household activities depending on object states and long-horizon semantics, and \textbf{VBench}~\citep{huang2023vbench} evaluates video generation for physical compliance.
In digital-world domains, \textbf{OSWorld}~\citep{xie2024osworld} and \textbf{macOSWorld}~\citep{yang2025macosworld} test GUI grounding and receipt parsing on desktop operating systems; \textbf{SWE-bench}~\citep{jimenez2023swebench} evaluates cross-file software engineering; \textbf{WebArena}~\citep{zhou2023webarena} and \textbf{Mind2Web}~\citep{deng2023mind2web} test web interactions; and \textbf{AppAgent}~\citep{zhang2025appagent} and \textbf{AndroidWorld}~\citep{rawles2024androidworld} extend digital constraints to mobile operating systems. \textbf{GameWorld}~\citep{ouyang2026gameworld} introduces verfiable evaluation for multimodal agents.
In social-world domains, Theory of Mind benchmarks have systematically mapped LLM capabilities: \textbf{ToMi}~\citep{le2019tomi} established properly balanced false-belief testing, \textbf{BigToM}~\citep{gandhi2023bigtom} introduced causal templates for belief inference, and \textbf{OpenToM}~\citep{xu2024opentom} expanded to psychological states where LLMs fall notably short. \textbf{Sotopia}~\citep{zhou2024sotopia} provides multi-dimensional social simulation with negotiation and norm compliance; \textbf{AgentBench}~\citep{liu2023agentbench} offers broad cross-domain assessment including role-playing; and game-based environments such as Werewolf and Avalon test deception, trust, and strategic social reasoning.
In scientific-world domains, \textbf{ScienceWorld}~\citep{wang2022scienceworld} tests elementary scientific reasoning; \textbf{DiscoveryBench}~\citep{majumder2024discoverybench} evaluates hypothesis generation and verification; \textbf{ChemCrow}~\citep{bran2024augmenting} assesses chemical synthesis under strict validity constraints; and \textbf{FutureX}~\citep{zeng2025futurex} tests evidence-based prediction from dynamic information streams.
Finally, open-world environments such as \textbf{Minecraft} (via MCU or Voyager)~\citep{zheng2023mcu,wang2023voyager}, \textbf{Crafter}~\citep{stanic2023learning}, and \textbf{NetHack}~\citep{kurenkov2023katakomba} evaluate the composition of skills across multiple governing laws simultaneously; e.g. combining physical combat with resource economics and long-horizon planning in procedurally generated worlds.

Table~\ref{tab:benchmark_summary} maps compact benchmark anchors to their primary governing-law regime, capability-level coverage, and key evaluation metrics; the prose above gives the broader landscape. Detailed evaluation protocols for each boundary condition (including counterfactual divergence testing for intervention sensitivity, degradation curves for long-horizon coherence, and regime-specific constraint verification) appear in Appendix~\ref{app:eval_extended}.


\begin{table*}[!t]
\caption{\textbf{Representative benchmark anchors by governing-law regime with capability-level coverage and core evaluation metrics.} The table is a compact comparison set; Section~\ref{sec:evaluation} discusses additional benchmarks in prose. \cmark\ = supported; \xmark\ = not supported.}
\label{tab:benchmark_summary}
\centering
\small
\setlength{\tabcolsep}{0.9mm}
\renewcommand{\arraystretch}{1.05}
\begin{tabular}{l|cc|ccc|l}
\toprule
\textbf{Benchmark} & \multicolumn{2}{c|}{\textbf{Links}} & \textbf{L1} & \textbf{L2} & \textbf{L3} & \textbf{Core Metrics} \\
\midrule

\rowcolor{gray!15}
\multicolumn{7}{l}{\textit{Physical World}} \\
\shortstack[l]{Atari 100k~\citep{kaiser2019simple}}
& \paperlink{https://arxiv.org/abs/1903.00374}
& {---}
& \cmark & \cmark & \xmark
& Human-norm. score \\
\shortstack[l]{Meta-World~\citep{yu2020metaworld}} 
& \paperlink{https://arxiv.org/abs/1910.10897} 
& \githublink{https://github.com/Farama-Foundation/Metaworld} 
& \cmark & \cmark & \xmark 
& Success rate \\
\shortstack[l]{CALVIN~\citep{mees2022calvin}} 
& \paperlink{https://arxiv.org/abs/2112.03227} 
& \githublink{https://github.com/mees/calvin} 
& \cmark & \cmark & \xmark 
& Lang-cond. success \\
\shortstack[l]{RoboCasa~\citep{nasiriany2024robocasa}} 
& \paperlink{https://arxiv.org/abs/2406.02523} 
& \githublink{https://github.com/robocasa/robocasa} 
& \cmark & \cmark & \xmark 
& Task completion \\
\shortstack[l]{nuScenes~\citep{caesar2020nuscenes}} 
& \paperlink{https://arxiv.org/abs/1903.11027} 
& \githublink{https://github.com/nutonomy/nuscenes-devkit} 
& \cmark & \cmark & \xmark 
& mAP, NDS \\
\midrule

\rowcolor{gray!15}
\multicolumn{7}{l}{\textit{Digital World}} \\
\shortstack[l]{OSWorld~\citep{xie2024osworld}} 
& \paperlink{https://arxiv.org/abs/2404.07972} 
& \githublink{https://github.com/xlang-ai/OSWorld} 
& \cmark & \cmark & \xmark 
& Task success \\
\shortstack[l]{SWE-bench~\citep{jimenez2023swebench}} 
& \paperlink{https://arxiv.org/abs/2310.06770} 
& \githublink{https://github.com/princeton-nlp/SWE-bench} 
& \cmark & \cmark & \cmark 
& Resolve rate \\
\shortstack[l]{WebArena~\citep{zhou2023webarena}} 
& \paperlink{https://arxiv.org/abs/2307.13854} 
& \githublink{https://github.com/web-arena-x/webarena} 
& \cmark & \cmark & \xmark 
& Task success \\
\midrule

\rowcolor{gray!15}
\multicolumn{7}{l}{\textit{Social World}} \\
\shortstack[l]{Sotopia~\citep{zhou2024sotopia}} 
& \paperlink{https://arxiv.org/abs/2310.11667} 
& \githublink{https://github.com/sotopia-lab/sotopia} 
& \cmark & \cmark & \xmark 
& Social score \\
\shortstack[l]{FANToM~\citep{kim2023fantom}} 
& \paperlink{https://arxiv.org/abs/2310.15421} 
& \githublink{https://github.com/skywalker023/fantom} 
& \cmark & \xmark & \xmark 
& False-belief acc. \\
\shortstack[l]{Hi-ToM~\citep{wu2023hitom}}
& \paperlink{https://arxiv.org/abs/2310.16755}
& \githublink{https://github.com/ying-hui-he/Hi-ToM_dataset}
& \cmark & \xmark & \xmark 
& Belief acc. \\
\midrule

\rowcolor{gray!15}
\multicolumn{7}{l}{\textit{Scientific World}} \\
\shortstack[l]{ScienceWorld~\citep{wang2022scienceworld}} 
& \paperlink{https://arxiv.org/abs/2203.07540} 
& \githublink{https://github.com/allenai/ScienceWorld} 
& \cmark & \cmark & \xmark 
& Task completion \\
\shortstack[l]{DiscoveryBench~\citep{majumder2024discoverybench}} 
& \paperlink{https://arxiv.org/abs/2407.01725} 
& \githublink{https://github.com/allenai/discoverybench} 
& \cmark & \cmark & \cmark 
& Hypothesis acc. \\
\bottomrule
\end{tabular}
\end{table*}

\subsection{Differentiating L1, L2, and L3 via evaluation protocol}
\label{subsec:eval_levels}

Importantly, the \textit{same} benchmark can evaluate different capability levels depending on protocol. The level is fixed not by the benchmark but by what the protocol demands: L1 tests local prediction, L2 tests decision-usable simulation under the three boundary conditions, and L3 tests whether the system can revise itself from evidence (Section~\ref{subsec:l3_definition}). The examples below make this concrete, one per governing-law regime.

\paragraph{RoboCasa (Physical World).}
At L1, the benchmark reduces to predicting the next end-effector position given current state and action, measured by single-step position error. Elevating the protocol to L2 requires executing a full kitchen task (e.g., a pick-place-heat sequence) under mid-task perturbations such as object displacement or drawer obstruction; the relevant metrics shift to long-horizon success rate, catastrophic action fraction, and recovery rate after perturbation. An L3 protocol would further demand that the agent, after repeated failures on a novel kitchen layout, distills a persistent grasp strategy (e.g., ``this handle requires a top-down approach'') whose benefit carries over to subsequent trials. In practice, most robotic manipulation systems still report L1-style single-task success rates, and perturbation injection remains nonstandard.

\paragraph{OSWorld and SWE-bench (Digital World).}
These benchmarks span the L1--L2 boundary. Single-step click prediction (OSWorld) or single-line code completion (SWE-bench) constitutes L1 evaluation. L2 demands cross-file issue resolution under injected failures (network timeouts, unexpected pop-ups, out-of-distribution states), tracked via long-horizon consistency and catastrophic action fraction~\citep{xie2024osworld,jimenez2023swebench}. The leap to L3 would require the system to generate durable artifacts: a reproduction script that becomes a regression test (SWE-bench), or a reusable installation procedure distilled from a failed attempt (OSWorld). Today, leaderboard systems primarily operate at L1--L2; L3-style asset generation is reported anecdotally across isolated case studies but not yet evaluated systematically or at scale.

\paragraph{Sotopia (Social World).}
Social simulation introduces a distinctive challenge: the ``perturbation'' is another agent's strategy shift rather than a physics disturbance. An L1 evaluation measures next-turn prediction accuracy or perplexity. For L2, one agent's strategy is changed mid-conversation (counterfactual injection), and the protocol tracks goal completion under perturbation, commitment consistency, and whether social outcomes shift appropriately~\citep{zhou2024sotopia}. L3 evaluation would require the agent to distill, after repeated negotiation failures, a new social strategy or norm-handling rule that persists and transfers to structurally similar scenarios. Existing social benchmarks rarely inject such counterfactual perturbations, leaving L3-style social strategy evolution largely unexplored in current negotiation and role-playing settings.

\paragraph{ScienceWorld and DiscoveryBench (Scientific World).}
Of the four regimes, autonomous science is closest to a genuine L3 evaluation paradigm. At L1, the task is predicting the outcome of a single experimental action (e.g., ``what happens when acid is added to base?''). L2 protocols require designing and executing a multi-step experimental sequence while maintaining causal coherence, measured by sequence validity, hypothesis-consistent action rate, and robustness to misleading observations~\citep{wang2022scienceworld,majumder2024discoverybench}. The critical L3 step is hypothesis revision: when experimental evidence falsifies a prediction, the system must update its belief structure and avoid previously falsified paths. Closed-loop systems such as CAMEO already demonstrate this kind of evidence-driven model revision in laboratory settings.

\paragraph{Evaluation gaps and coverage.}
The vast majority of current systems are evaluated only at L1 (single-step accuracy or end-to-end success rate without perturbation injection). L2-style evaluation protocols (counterfactual injection, degradation curves, constraint-violation detection) exist in principle and are demonstrated by individual benchmarks, but are not yet standard practice across the field. L3 evaluation infrastructure (regression suites, asset validation gates, cross-episode improvement tracking) is essentially nonexistent outside autonomous science. Closing this gap is a prerequisite for claims about world-modeling capability.

\subsection{Benchmarks and coverage analysis}
\label{subsec:eval_benchmarks}

A growing line of work asks whether current systems genuinely learn world models or merely surface correlations. WorldSimBench~\citep{qin2025worldsimbench} and WorldModelBench~\citep{fan2025worldmodelbench} reveal that perceptual realism and action-conditioned fidelity can diverge sharply, while \citet{vafa2024evaluating} and \citet{kang2025howfar} demonstrate fundamental coherence and consistency failures that persist at scale. No single benchmark covers all capabilities; a capability coverage matrix mapping benchmarks to boundary conditions and regimes is provided in Appendix~\ref{app:eval_extended}.
We use four qualitative coverage labels in the capability matrix. \textbf{Strong (S)} means the benchmark directly and intentionally tests the capability through explicit task design and scoring. \textbf{Medium (M)} means the capability is exercised in a substantial but partial or indirect way. \textbf{Weak (W)} means the benchmark offers only incidental evidence about the capability. \textbf{--} means the capability is not meaningfully tested. These labels are judgment-based but are assigned according to task design, scoring visibility, and whether failure on the capability can be unambiguously detected from benchmark traces.
Broader multimodal evaluation resources such as HSSBench can complement this landscape by probing humanities- and social-science reasoning, although they do not directly evaluate decision-usable social-state rollouts or other world-model-specific boundary conditions \citep{kang2025hssbench}. We also propose the Minimal Reproducible Evaluation Package (MREP), a community standard for version locking, trace logging, failure taxonomy, tail statistics, and boundary condition mapping, detailed in Appendix~\ref{app:eval_extended}.
Seen this way, MREP is not only an evaluation proposal for individual papers but also the minimal evidence infrastructure required to sustain any credible L3 gatekeeping loop in practice.

\subsection{Open Challenges in Evaluation}
\label{subsec:eval_open}

Evaluation challenges remain even when benchmark suites and logging infrastructure improve. Some are methodological, including benchmark saturation and evaluation gaming; others are infrastructural, including the cost and variability of human evaluation and the lack of systematic meta-evaluation.
\emph{Benchmark saturation}: as top systems converge on near-ceiling performance, the discriminative power of existing benchmarks decreases. \emph{Evaluation gaming}: systems can optimize for benchmark-specific artifacts rather than genuine capability. \emph{Human evaluation}: for social-world and open-ended scenarios where automated metrics are unreliable, human judgment remains necessary. \emph{Meta-evaluation}: whether an evaluation protocol itself is valid is a question that is rarely addressed systematically across current world-model benchmarks.

\section{Architectural and Computational Considerations}
\label{sec:implementation}

The value of a taxonomy is not categorization for its own sake, but guiding system design. This section decomposes world-model implementations along three architectural axes, namely representation, dynamics, and control interface (Section~\ref{subsec:impl_blocks}), and examines how the governing-law regime constrains which combinations are viable in practice (Section~\ref{subsec:impl_tradeoffs}). Deploying these systems raises cross-cutting engineering challenges: the choice between end-to-end and modular training, latency-compute tradeoffs, sim-to-real transfer, and graceful degradation under model uncertainty. A learned world model amortizes simulation cost into a fixed computation graph at inference time, whereas explicit simulation typically scales more directly with the number of entities, interactions, solver steps, or horizon length. This does not mean neural inference is literally $O(1)$ in every relevant variable: its cost still depends on model size, input resolution, sequence length, and rollout depth. The practical advantage is instead that learned dynamics can offer near-constant-cost approximations with respect to aspects of system complexity that would otherwise require increasingly expensive explicit simulation. Efficiency techniques matter here not as generic deployment tricks but because they interact differently with the three capability levels. For L1 systems, compression mainly trades off against one-step predictive accuracy. For L2 systems, memory and rollout efficiency directly affect achievable horizon, counterfactual branching, and thus long-horizon coherence. For L3 systems, the same efficiency choices affect whether regression-gated update loops are cheap enough to run continuously in deployment. Scaling further demands efficiency techniques: few-step distillation for real-time planning, quantization and pruning under the constraint that compounding errors amplify even minor per-step degradation, and KV cache compression for long-horizon autoregressive dynamics. A more extended treatment of these deployment and efficiency topics, together with concrete compute and latency measurements, appears in Appendix~\ref{app:impl_extended}.

\subsection{Architectural building blocks: representation, dynamics, and control}
\label{subsec:impl_blocks}

Building a world-model system requires choosing components along three axes (Table~\ref{tab:arch_building_blocks}).
Each choice carries distinct tradeoffs that determine which capability level (L1/L2/L3) the resulting system can reach and in which governing-law regime the resulting design will be most effective along each of these three axes.

\paragraph{Representation.}
At one extreme, symbolic or programmatic states (e.g., VirtualHome~\citep{puig2018virtualhome}) offer interpretability and enable hard constraint enforcement, but demand heavy manual engineering and cover only pre-specified state spaces; they are best evaluated by success rate and error-branch coverage.
At the other extreme, latent continuous representations, such as the RSSM in DreamerV3~\citep{hafner2023dreamerv3} and V-JEPA2~\citep{meta2025vjepa2}, handle high-dimensional multimodal inputs with relatively little hand-designed structure. Their weakness is that, over long horizons, they are more susceptible to semantic drift and state aliasing, making long-horizon consistency and failure attribution especially important for evaluation.
VL-JEPA~\cite{chen2025vl} develop a joint embedding predictive architecture which predicts the continuous embeddings of the target text.
VLog~\citep{lin2025vlog} use a learnable token to retrieve the narration then serve as video-centric vocabulary in the long video understanding.
Between these two extremes lie structured 3D representations, including occupancy models such as RoboOccWorld~\citep{zhang2025robooccworld} and point-flow models such as PointWorld~\citep{huang2026pointworld}. They are appealing because they fit physical constraints more naturally, but this advantage often comes with reconstruction and computational bottlenecks. As a result, reachability and stability become particularly important in evaluation.
Finally, discrete token representations (e.g., VQ-VAE codebooks in IRIS~\citep{micheli2023iris}) enforce compositionality and enable exact likelihood training via cross-entropy, bridging continuous perception with autoregressive dynamics.

\paragraph{Dynamics.}
Stochastic latent dynamics, exemplified by DreamerV3~\citep{hafner2023dreamerv3}, express uncertainty and multimodality through principled ELBO training and uncertainty-aware rollouts, but may degrade or become miscalibrated over long horizons.
Where uncertainty modeling is less critical, deterministic value-aware dynamics (MuZero~\citep{schrittwieser2020muzero}, TD-MPC2~\citep{hansen2024tdmpc2}) optimize the transition function directly for downstream value prediction, trading generative flexibility for tighter integration with the control objective.
Autoregressive token dynamics (iVideoGPT~\citep{wu2024ivideogpt}, LWM~\citep{liu2024lwm}) offer a unified scalable interface that handles multiple modalities through a shared vocabulary, though long-horizon logical consistency remains a weak point.
Diffusion-based dynamics (the Sora technical line~\citep{brooks2024sora}, DIAMOND~\citep{alonso2024diamond}, and interactive environments such as Genie~\citep{deepmind2025genie3}) deliver photorealistic observation-level transitions, but the multi-step denoising they require at inference time often comes with weak action controllability.

\paragraph{Control interface.}
Online MPC-style approaches (TD-MPC2~\citep{hansen2024tdmpc2}, PETS~\citep{chua2018pets}) replan at every step using short-horizon rollouts, providing fast correction at the cost of compute and latency pressure.
Tree search and expansion (MuZero~\citep{schrittwieser2020muzero}, EfficientZero~\citep{ye2021efficientzero}) enable counterfactual branching and systematic look-ahead, though they amplify model errors and can exploit benchmark loopholes.
Rather than planning in the environment at all, imagined-rollout policy optimization (the Dreamer family~\citep{hafner2019dreamer,hafner2020dreamerv2,hafner2023dreamerv3}) trains a policy entirely on model-generated trajectories, avoiding real interaction during learning but requiring highly accurate dynamics.
At the deployment end, offline policy distillation (GR-1~\citep{wu2023gr1}) enables cheap inference yet is fragile under distribution shift, motivating OOD stress tests.
A distinct strategy altogether, replayable-environment interfaces (OSWorld~\citep{xie2024osworld}, SWE-agent~\citep{yang2024sweagent}) sidestep learned dynamics entirely, treating the real environment as its own simulator and relying on receipt parsing and state fingerprinting. More broadly, part of the control problem is deciding when external computation should be invoked at all, rather than treating tool use as either mandatory or absent; adaptive tool-integration work provides a useful planner-side example of this distinction \citep{wang2025tocode}.

\begin{table*}[ht]
\caption{\textbf{Architectural building blocks for world models.} Three design axes are cross-referenced with concrete options, representative systems, strengths, and dominant failure modes.}
\label{tab:arch_building_blocks}
\centering
\small
\renewcommand{\arraystretch}{1.1}
\begin{tabular}{p{2.8cm}|p{13.2cm}}
\hline
\textbf{Design Axis} & \textbf{Options, Systems, Strengths, and Failure Modes} \\
\hline
\rowcolor{gray!15} \multicolumn{2}{c}{\textbf{Representation}} \\
\hline
\textbf{Representation} &
$\square$ \textbf{Symbolic / Programmatic:} VirtualHome. Interpretable; hard constraint enforcement. Failure: heavy manual engineering; limited state space.\\[0.3em]
& $\square$ \textbf{Latent Continuous:} DreamerV3 (RSSM); V-JEPA~2. Scalable; absorbs high-dim multimodal input. Failure: semantic drift; state aliasing over long horizons.\\[0.3em]
& $\square$ \textbf{Structured 3D:} RoboOccWorld; PointWorld. Natural physical-constraint alignment. Failure: reconstruction bottleneck; high compute cost.\\[0.3em]
& $\square$ \textbf{Discrete Tokens:} IRIS (VQ-VAE codebook). Compositional; exact cross-entropy training. Failure: codebook collapse; lossy quantization. \\
\hline
\rowcolor{gray!15} \multicolumn{2}{c}{\textbf{Dynamics}} \\
\hline
\textbf{Dynamics} &
$\square$ \textbf{Stochastic Latent:} DreamerV3. Principled uncertainty via ELBO; multimodal. Failure: miscalibration over long horizons.\\[0.3em]
& $\square$ \textbf{Deterministic Value-Aware:} MuZero; TD-MPC2. Tight value integration; planning-optimized. Failure: no explicit uncertainty; brittle under novelty.\\[0.3em]
& $\square$ \textbf{Autoregressive Token:} iVideoGPT; LWM. Unified multi-modal interface; scalable. Failure: weak long-horizon logical consistency.\\[0.3em]
& $\square$ \textbf{Diffusion-Based:} Sora; DIAMOND; Genie~2. Photorealistic observation-level transitions. Failure: multi-step denoising latency; weak action control. \\
\hline
\rowcolor{gray!15} \multicolumn{2}{c}{\textbf{Control Interface}} \\
\hline
\textbf{Control Interface} &
$\square$ \textbf{Online MPC:} TD-MPC2; PETS. Fast closed-loop correction; reactive. Failure: high per-step compute; latency pressure.\\[0.3em]
& $\square$ \textbf{Tree Search:} MuZero; EfficientZero. Counterfactual branching; systematic look-ahead. Failure: amplifies model errors; benchmark exploitation.\\[0.3em]
& $\square$ \textbf{Imagined-Rollout Policy:} Dreamer family. No real interaction during training. Failure: requires highly accurate dynamics.\\[0.3em]
& $\square$ \textbf{Offline Distillation:} GR-1. Cheap and fast deployment. Failure: distribution shift.\\[0.3em]
& $\square$ \textbf{Replayable Environment:} OSWorld; SWE-agent. Real env as simulator; attributable failures. Failure: grounding breaks under UI/API changes. \\
\hline
\end{tabular}
\end{table*}

\subsection{Design tradeoffs across governing-law regimes}
\label{subsec:impl_tradeoffs}

The building blocks above are not interchangeable; the governing-law regime determines which combinations are viable and which failure modes dominate. Table~\ref{tab:compute_latency} summarizes how deployment-regime latency budgets constrain the viable dynamics model classes and their control interfaces.

\paragraph{Physical-world systems.}
Everything hinges on contact, reachability, and stability under continuous actions.
Representations must preserve geometry and contact relations; dynamics must be stable over short-to-medium horizons; and the control interface must be fast enough for closed-loop correction.
Latent or structured 3D representations paired with MPC or imagined-rollout policies dominate this regime~\citep{hafner2023dreamerv3,hansen2024tdmpc2,huang2026pointworld}.
Short-horizon rollouts reduce compounding error~\citep{janner2019mbpo}, and MPC provides an online correction mechanism.
The main pitfalls are the preassumed existence of de facto 3D scenes, degraded 3D reconstruction capabilities, semantic drift in latent space, constraint violations that remain plausible in the learned representation, and the sim-to-real gap for contact-rich interactions. It is useful to distinguish at least three transfer curves in practice: transfer across input modalities, transfer across sensor suites, and transfer across environments, since each exposes a distinct failure mode of the learned dynamics and demands its own diagnostic instrumentation.

\paragraph{Digital-world systems.}
State-machine and branch consistency, rather than learned dynamics, are the primary bottleneck.
Symbolic or DOM-based states paired with replayable environments are the dominant design in this setting~\citep{xie2024osworld,jimenez2023swebench,yang2024sweagent}. Because they expose explicit state machines and support strong evidence logging, they make failures easier to trace and thus support Evolver-style asset distillation. This transparency, however, is not without cost: grounding may break under UI changes, loading variability and race conditions introduce non-deterministic noise, and benchmark artifacts remain vulnerable to reward gaming and to subtle shifts in the underlying software stack.

\paragraph{Social-world systems.}
The dominant bottleneck is maintaining coherent agent identity and relational state across extended interactions.
Persona state must persist over hundreds of turns without drift, yet Theory-of-Mind (ToM) inference, which updates beliefs about other agents' goals, knowledge, and intentions, imposes per-step costs that grow with the number of modeled agents.
Multi-agent communication compounds the problem: $n$-agent interactions generate $O(n^2)$ pairwise belief updates per step, making na\"ive scaling infeasible for the 10,000$+$-agent simulations now appearing in the literature~\citep{piao2025agentsociety}.
Norm-consistency checking adds a further constraint: valid social rollouts must respect evolving norms (politeness conventions, negotiation protocols, institutional rules), and violations must be detectable at rollout time rather than post hoc~\citep{zhou2024sotopia}.
The overarching challenge is that agent identity is not a fixed state vector but an emergent property of interaction history; maintaining stable identity under multi-turn dynamics while still allowing genuine belief revision remains an open architectural problem that current LLM-based agents address only superficially through system-prompt conditioning.

\paragraph{Generative simulation systems.}
The central tension is between visual fidelity and action controllability.
High-fidelity diffusion or autoregressive models~\citep{brooks2024sora,bruce2024genie,alonso2024diamond} excel at producing photorealistic outputs useful for demonstration and synthetic data generation, but action-conditioning is often unstable and long-horizon consistency is difficult.
A system can be mistakenly treated as planning-ready when it is not decision-usable; evaluation should prioritize action-response consistency and long-horizon stability over raw perceptual realism~\citep{wu2024ivideogpt,liu2024lwm}.

\paragraph{Scientific-world systems.}
Evidence-chain validity and falsifiability matter more than perceptual quality in this regime (cf.\ the Popperian reading in Section~\ref{sec:philosophy_l3}).
Representations must be interpretable and traceable to experimental evidence; dynamics must respect known mechanism boundaries; and the control interface should support experiment selection and belief updates rather than action execution~\citep{wang2022scienceworld}.
The distinctive risks are hallucinated mechanisms that appear plausible but lack grounding, correlation mistaken for causation, and negative results that are silently discarded rather than propagated through the model.

\paragraph{VLA vs.\ native world models.}
A crosscutting architectural question is whether to embed world-model capabilities inside a Vision-Language-Action (VLA) model or to build a dedicated world-model module. VLAs inherit the scaling infrastructure and pretraining data of large language models, but their world-modeling capacity is implicit and difficult to isolate or evaluate. Recent efforts to make this capacity more explicit include spatially guided training that injects geometric structure into VLA policy learning~\citep{chen2025internvla}, aiming to bridge the gap between implicit visual knowledge and the explicit physical state awareness that world models require. Related work makes this implicit capacity more procedural than geometric: Pixel Reasoner equips VLMs with explicit visual operations such as zoom-in and select-frame for curiosity-driven evidence gathering, while Visual Rationale Learning treats such visual actions as core reasoning primitives rather than optional tools, together highlighting a broader shift toward explicit perceptual control inside VLM-like agents even when no standalone transition model is exposed \citep{wang2025pixelreasoner,wang2025virl}. Native world models expose an explicit transition function that can be queried, composed, and stress-tested independently. Competition between these paradigms is partly a sociotechnical question: the massive investment in LLM infrastructure creates path dependencies that favor VLA-style integration even when a dedicated module might be technically superior~\citep{hooker2021hardware}. From an evaluation standpoint, the litmus test is whether the system’s predictions can be decoupled from its language generation and tested against the three boundaries (Section~\ref{subsec:l2_requirements}). Some architectural choices are also sociotechnical rather than purely algorithmic: whether the field converges on native world models or VLA-style surrogates may partly depend on tool ecosystems, available datasets, and hardware compatibility, besides intrinsic modeling power.

These regimes are not mutually exclusive.
In practice, mature systems often stack multiple design patterns: symbolic or workflow planning at the top for high-level task decomposition, replayable environments in the middle for receipt validation and failure attribution, and short-horizon continuous control at the bottom for real-time correction~\citep{moerland2023mbrl}. This suggests that \textbf{the relevant unit of analysis is the composed system, not any single module in isolation}. Representation, dynamics, and control should therefore be evaluated together, in light of the constraints they impose and the evidence they make available. Many apparent disagreements in the literature then look less like fundamental disputes about whether world models work, and more like differences in where systems land along these design axes.

\begin{table*}[ht]
\caption{\textbf{Deployment latency budgets and engineering bottlenecks by regime.} Inference latency budgets range from sub-100\,ms for real-time robotics to minutes for offline scientific planning; the table maps each regime to viable dynamics model classes and primary engineering bottlenecks. These are deployment budget ranges rather than measured benchmark results; empirical throughput depends on model size, hardware, batching, simulator implementation, and verification overhead.}
\label{tab:compute_latency}
\centering
\small
\renewcommand{\arraystretch}{1.1}
\begin{tabular}{p{3.5cm}|p{12cm}}
\hline
\textbf{Regime} & \textbf{Latency, Dynamics Class, and Bottleneck} \\
\hline
\rowcolor{gray!15} \multicolumn{2}{c}{\textbf{Physical World}} \\
\hline
\textbf{Real-time robotics} &
$\square$ \textbf{Latency:} $<$100\,ms.\\[0.3em]
& $\square$ \textbf{Dynamics:} Latent dynamics + MPC; lightweight RSSM; neural ODE.\\[0.3em]
& $\square$ \textbf{Bottleneck:} Per-step inference latency; compounding error within control loop. \\
\hline
\textbf{Autonomous driving} &
$\square$ \textbf{Latency:} $<$200\,ms.\\[0.3em]
& $\square$ \textbf{Dynamics:} Occupancy flow; latent diffusion; BEV prediction.\\[0.3em]
& $\square$ \textbf{Bottleneck:} Sensor-fusion latency; safety-critical constraint verification. \\
\hline
\textbf{Embodied navigation} &
$\square$ \textbf{Latency:} 100--500\,ms.\\[0.3em]
& $\square$ \textbf{Dynamics:} RSSM; object-centric GNN; point-cloud dynamics.\\[0.3em]
& $\square$ \textbf{Bottleneck:} 3D reconstruction cost; memory for large-scale maps. \\
\hline
\rowcolor{gray!15} \multicolumn{2}{c}{\textbf{Digital World}} \\
\hline
\textbf{Web / GUI agents} &
$\square$ \textbf{Latency:} $\sim$1--5\,s.\\[0.3em]
& $\square$ \textbf{Dynamics:} LLM-as-world-model; DOM-based prediction; state-machine rollout.\\[0.3em]
& $\square$ \textbf{Bottleneck:} LLM inference cost; UI non-determinism and race conditions. \\
\hline
\textbf{Software engineering} &
$\square$ \textbf{Latency:} $\sim$5--30\,s.\\[0.3em]
& $\square$ \textbf{Dynamics:} LLM + MCTS rollout; code-graph traversal.\\[0.3em]
& $\square$ \textbf{Bottleneck:} Context window limits; cross-file dependency resolution. \\
\hline
\textbf{Game AI (real-time)} &
$\square$ \textbf{Latency:} $<$50\,ms.\\[0.3em]
& $\square$ \textbf{Dynamics:} Tree search (MCTS); value-aware latent dynamics; EfficientZero.\\[0.3em]
& $\square$ \textbf{Bottleneck:} Branching factor; search-depth vs.\ latency tradeoff. \\
\hline
\rowcolor{gray!15} \multicolumn{2}{c}{\textbf{Social World}} \\
\hline
\textbf{Social / multi-agent} &
$\square$ \textbf{Latency:} $\sim$1--10\,s.\\[0.3em]
& $\square$ \textbf{Dynamics:} ToM network; multi-agent rollout; commitment-graph update.\\[0.3em]
& $\square$ \textbf{Bottleneck:} $O(n^2)$ pairwise belief updates; persona-state drift. \\
\hline
\rowcolor{gray!15} \multicolumn{2}{c}{\textbf{Scientific World}} \\
\hline
\textbf{Scientific planning} &
$\square$ \textbf{Latency:} Minutes to hours.\\[0.3em]
& $\square$ \textbf{Dynamics:} Full diffusion ensemble; Bayesian surrogate; PINN; active learning.\\[0.3em]
& $\square$ \textbf{Bottleneck:} Experiment budget; surrogate calibration; data scarcity. \\
\hline
\end{tabular}
\end{table*}

\subsection{Implementation Roadmap}
\label{subsec:impl_roadmap}

Table~\ref{tab:impl_roadmap} distills the architectural guidance from the preceding sections into a concise roadmap organized by capability level and governing-law regime. For each cell, we list the representation format that best preserves the regime's planner-critical structure, the dynamics model class that is most tractable at that capability level, and the single most important engineering bottleneck that must be addressed to reach the next level.

\begin{table*}[ht]
\caption{\textbf{Design roadmap across governing-law regimes.} For each regime, we summarize the representation, dynamics, and bottleneck at L1--L3.}
\label{tab:impl_roadmap}
\centering
\small
\renewcommand{\arraystretch}{1.08}
\begin{tabular}{llll}
\hline
 & \textbf{Representation} & \textbf{Dynamics} & \textbf{Bottleneck} \\
\hline

\multicolumn{4}{c}{\textbf{Physical}} \\
\hline
\textbf{L1} & Latent state, point-cloud input & RSSM, latent transitions & Long-horizon prediction error \\
\textbf{L2} & 3D, object-centric state & Latent MBRL, neural ODE rollout & Contact instability, constraints \\
\textbf{L3} & Physics prior, residual model & Hybrid sim-to-real adaptation & Failure attribution across modules \\
\hline

\multicolumn{4}{c}{\textbf{Digital}} \\
\hline
\textbf{L1} & DOM tree, UI state & LLM-based state prediction & Grounding on unseen layouts \\
\textbf{L2} & State-machine abstraction & LLM rollout, MCTS planning & Exploits, race conditions \\
\textbf{L3} & Versioned tests, execution traces & Regression-gated updates & Safe deployment, rollback \\
\hline

\multicolumn{4}{c}{\textbf{Social}} \\
\hline
\textbf{L1} & Belief state, dialogue history & ToM, recurrent updates & Hidden mental states \\
\textbf{L2} & Commitment graph, norm state & Multi-agent rollout & Role drift, forgetting \\
\textbf{L3} & Social model, update gates & Bayesian revision & Attribution ambiguity, ethics \\
\hline

\multicolumn{4}{c}{\textbf{Scientific}} \\
\hline
\textbf{L1} & Molecular graph, field state & GNN surrogate, FNO dynamics & OOD generalization \\
\textbf{L2} & Hypothesis-evidence chain & Bayesian surrogate, PINN rollout & Hallucinated calibration \\
\textbf{L3} & Protocol, surrogate model & Active Bayesian learning & Data budget, instruments \\
\hline
\end{tabular}
\end{table*}

Three cross-cutting engineering principles hold across all cells. First, \textbf{separate what is learned from what is enforced}: hard constraint layers (collision checkers, state-machine validators, regression gates) should be applied at inference time rather than learned implicitly, because soft enforcement through training loss cannot guarantee zero-violation rollouts. Second, \textbf{instrument before you iterate}: logging, replay, and failure attribution infrastructure should be built into the system from the start; without replay, L3 revision becomes anecdotal and ungovernable. Third, \textbf{match the representation to the planner's query}: a representation that looks realistic but does not expose the variables the planner needs (free space, permission state, reaction rate) is worse than a lower-fidelity representation that does.

\section{Trends \& Open Problems}
\label{sec:trends}

The preceding sections have established L1--L3 as a capability ladder for world models.
We now situate this ladder historically (Figure~\ref{fig:historical_timeline}), survey the research frontier that is pushing each rung forward, and catalog the open problems whose resolution will determine whether world models mature from impressive demonstrations into reliable scientific and engineering tools.

\subsection{Historical Development}
\label{sec:trends:history}


\paragraph{Mathematical Principles (--1956).}
The impulse to build predictive models of reality long predates artificial intelligence.
Newton's \emph{Principia}~\citep{newton1687principia} provided the first unified mathematical world model: given initial positions and velocities, his laws of motion and gravitation could, in principle, predict arbitrary future states of a mechanical system.
\citet{laplace1814essai} distilled this ambition into the thought experiment now known as ``Laplace's demon,'' an intelligence that, given complete knowledge of the present, could compute the entire future of the universe.
\citet{turing1950computing} then posed the question of whether machines could think, establishing the conceptual bridge from mathematical modeling to artificial intelligence.
These developments established the core tension that persists today: the tradeoff between model fidelity and tractable horizon.
The philosophical foundations of this progression, from Hume's empiricism through Kant's structural priors to Lakatos's governed revision, are discussed in Section~\ref{sec:philosophy}.

\paragraph{Symbolic Intelligence (1956--1986).}
Early AI attempted to hand-code world models as logical rules and constraints. STRIPS~\citep{fikes1971strips} introduced the first action-schema representation for robotic planning, but the Frame Problem~\citep{mccarthy1969some} (see Section~\ref{subsec:l2_requirements}) revealed that every action requires explicit axioms specifying what does \emph{not} change, a burden that grows combinatorially.
The Lighthill Report~\citep{lighthill1973artificial} catalyzed the first AI winter (1974--1980) by exposing the gap between laboratory demonstrations and real-world competence.
The second winter (1987--1993) followed the brittleness of expert systems and the collapse of the Lisp-machine market: hand-crafted knowledge bases such as CYC could not gracefully handle uncertainty and commonsense exceptions~\citep{lenat1995cyc}.
The overarching lesson was clear: purely symbolic world models do not scale to open-world domains.

\paragraph{Connectionist Resurgence (1986--2020).}
The revival of neural networks, from backpropagation~\citep{rumelhart1986learning} through deep convolutional networks~\citep{lecun1998gradient,krizhevsky2012imagenet} and Transformers~\citep{vaswani2017attention}, shifted the paradigm from hand-coded rules to learned representations.
World models re-emerged in model-based reinforcement learning, from latent dynamics models to general pixel-based control (see Section~\ref{sec:l1} for details).

\paragraph{Generative Revolution (2020--present).}
Diffusion models~\citep{ho2020ddpm} and large-scale language models such as GPT-3~\citep{brown2020gpt3} have catalyzed a qualitative shift, building on the Transformer backbone established in the preceding era.
Video generation models~\citep{brooks2024sora,nvidia2025cosmos,bruce2024genie} and LLM-based agents~\citep{hao2023reasoning,wang2023voyager} are blurring the boundary between prediction and simulation, though systematic physics violations persist~\citep{gu2025phyworldbench} (Sections~\ref{sec:l1}--\ref{sec:l2}).
More broadly, the field is converging toward a \emph{neuro-symbolic} frontier~\citep{marra2024neurosymbolic_survey,zhao2026neurosymbolic_synergy} that combines neural dynamics modules for learning transition functions (L1/L2) with symbolic components for constraint enforcement and hypothesis-space expansion (L3).

Across all four eras, representation learning serves as shared infrastructure: the quality of the learned state $z_t$ determines the ceiling for prediction (L1), simulation (L2), and revision (L3) alike. Whether the representation is a latent vector, a discrete token sequence, a 3D point cloud, or a program, the governing-law regime determines which invariants the representation must preserve.

This historical arc suggests a consistent lesson: progress in world modeling has come not from scale alone,
but from changing what is represented, what is compositional over horizon, and what can be revised from
evidence. The open problems below are organized around remaining bottlenecks at L1, L2, and L3.

\subsection{Open Problems by Capability Level}
\label{sec:trends:open}

The preceding sections reveal a clear trajectory: world models are progressing from isolated one-step predictors toward integrated, agent-facing simulators that must respect domain-specific governing laws over extended horizons.
Across all four regimes, this progression exposes a common pattern. In embodied domains, visual plausibility is outpacing physical faithfulness: models generate convincing video but violate conservation laws and object permanence under rollout, with the best systems achieving only 0.262 success rate on physical-consistency tests~\citep{gu2025phyworldbench,li2025embodied_wm_survey}. In social domains, large-scale agent simulations reproduce emergent phenomena such as opinion polarization and governance formation~\citep{piao2025agentsociety,dai2024artificialleviathan}, but LLM agents exhibit systematic biases toward consensus that diverge from human behavioral patterns~\citep{taubenfeld2024biases,chuang2024opinion}. In code domains, agents treat software as deterministic state machines while real systems are partially observable, asynchronous, and multi-tenant~\citep{xu2025warex}. In scientific domains, neural surrogates trained on simulation data degrade when applied to real experimental measurements, exposing a surrogate-to-reality gap analogous to sim-to-real in robotics~\citep{minami2025simtorealsciml}. The overarching theme is that the bottleneck has shifted from generating plausible futures to ensuring those futures are \emph{decision-usable}: faithful to governing constraints, responsive to interventions, and calibrated against real-world evidence.

We organize ten concrete open problems by the capability level at which they most directly arise.

\paragraph{Representation and Local Prediction.}

\begin{enumerate}[leftmargin=*]
\item \textbf{Physical faithfulness beyond visual plausibility.} Current video and 3D world models achieve perceptual realism but fail physical-consistency tests: PhyWorldBench~\citep{gu2025phyworldbench} reports that the best of twelve frontier models attains only a 0.262 success rate on conservation-law and object-permanence probes, with long-horizon error accumulation as the core structural weakness.
Closing this gap requires physically grounded representations that enforce constraints under counterfactual rollout, not merely pixel fidelity, where representation choices discussed in Section~\ref{sec:trends:representation} may offer one possible direction.
Spatially guided training strategies that inject geometric supervision into vision-language-action models~\citep{ye2026st4vla} offer one promising direction.

\item \textbf{Metric-aware video world modeling.} Extending geometry-grounded editing from image pairs to temporally coherent video demands four coupled abilities: metric estimation across time, temporal composition of short-step predictions, identity and appearance preservation across frames, and instruction grounding that aligns predicted motion with semantic specifications~\citep{ho2022videodiffusion,xing2023dynamicrafter}. Moving to video provides denser temporal supervision and stronger identity constraints than image-pair approaches. Subject-faithful controllable editing methods such as RealCustom++~\citep{mao2026realcustompp} may supply useful interface components. Evaluation must measure metric controllability directly, not just perceptual quality~\citep{huang2023vbench,ge2024fvdbias}.

\item \textbf{Programmable visual representation.} Current visual world models represent state as raw pixels or latent embeddings, neither of which is compositional or precisely editable. Code offers a structured alternative: VCode~\citep{lin2025vcode} reconstructs images as SVG programs preserving symbolic semantics over pixel fidelity, Code2Video~\citep{chen2025code2video} shows that executable Manim scripts outperform pixel-generation models on structured content by making every spatial and temporal element directly addressable, and VIGA~\citep{yin2026vision} extends the paradigm to 3D by reconstructing scenes and simulating physical interactions through generated Blender code. The open problem is unifying these code-based representations into a single world-model interface for both 2D and 3D compositional editing.
\end{enumerate}

\paragraph{Simulation Fidelity and Intervention.}

\begin{enumerate}[leftmargin=*,resume]
\item \textbf{Partially observable software as a POMDP.} No existing code world model maintains belief distributions over hidden backend state (server sessions, database rows, in-flight requests, and background processes), nor reasons about asynchronous transitions with variable latency. Injecting realistic asynchronous failures into standard benchmarks causes significant drops in task success across all state-of-the-art agents~\citep{xu2025warex}. Solving this requires temporal-belief architectures that jointly model what has happened, what is in progress, and what the agent cannot yet observe.

\item \textbf{Concurrent multi-user state.} Real software is multi-tenant; world models must predict state under a Dec-POMDP where concurrent users' actions are unobservable~\citep{wang2025decpomdp}. Conflict-free replicated data types~\citep{kleppmann2017crdt} provide the formal substrate for merging concurrent updates, but no current world model integrates distributed-systems semantics with learned belief tracking over hidden users and pending writes. This problem sits at the intersection of the Digital World and Social World regimes, requiring joint reasoning about software state and multi-agent intent.

\item \textbf{Agent-human behavioral alignment at scale.} LLM agents exhibit systematic biases toward moderation and consensus~\citep{taubenfeld2024biases,chuang2024opinion}, producing two failure modes: mode collapse, where diverse simulated populations converge to homogeneous behavior, and calibration inadequacy, where single-turn persona alignment fails under multi-turn dynamics. Language and cultural priors can inject diversity, but this effect diminishes as cultural distance between populations shrinks. Systematic methods for grounding simulated behavior in real human behavioral distributions is lacking.
\end{enumerate}

\paragraph{Evidence-Driven Revision and Self-Evolution.}
Existing autonomous science flagships such as CAMEO~\citep{kusne2020cameo} and A-Lab~\citep{szymanski2023alab} demonstrate that closed-loop model revision is feasible in highly instrumented domains, while evaluator-guided algorithmic discovery systems such as FunSearch~\citep{romera2024funsearch} and AlphaEvolve~\citep{alphaevolve2025} demonstrate partial L3 loops with strong validation (Section~\ref{sec:l3}). Several open problems must be solved before L3 generalizes.

\begin{enumerate}[leftmargin=*,resume]
\item \textbf{Continual learning of societal transition functions.} Large-scale simulations with 10,000+ agents across millions of interactions reproduce emergent phenomena such as opinion polarization and governance formation~\citep{piao2025agentsociety,dai2024artificialleviathan}, yet cannot autonomously detect when social dynamics have shifted. The core challenge is to identify an outdated transition model, acquire corrective evidence, and revise without catastrophic forgetting of stable patterns~\citep{vandeven2024continuallearning}. This problem connects to L3, where model revision must be triggered by distributional evidence rather than supervised labels.

\item \textbf{Closing the surrogate-to-reality gap.} Scientific surrogates validated on simulation data degrade on real measurements; prediction error decreases as a power law with computational data but plateaus without real-data calibration, mirroring the sim-to-real gap in robotics~\citep{minami2025simtorealsciml}. In the scientific regime, this surrogate-to-reality gap is the direct analogue of sim-to-real transfer in robotics; the implementation-side mitigations discussed in Section~\ref{subsec:impl_tradeoffs} therefore provide a useful template, even though the measurement bottlenecks and evidence budgets differ. Notably, L2 scientific simulators such as GraphCast, NeuralGCM, and Aurora (Section~\ref{sec:l2}) provide the prediction substrate on which L3 revision operates; their fidelity sets the ceiling for downstream evidence-driven diagnosis. The OPAL-surrogate framework~\citep{singh2024opal} provides hierarchical Bayesian credibility gates that formalize when a surrogate is trustworthy. The central open question is how to allocate scarce real experimental observations optimally between model calibration and scientific discovery.

\item \textbf{Modeling laws that themselves evolve.} In biology, ecology, and climate, governing dynamics are non-stationary: viral fitness landscapes shift~\citep{lassig2017predicting}, climate forcing alters atmospheric dynamics~\citep{beucler2024climate}, and evolutionary pressures create tipping points~\citep{evangelou2024coevolving}. World models must learn second-order meta-transition operators governing how $p_\theta$ itself drifts, together with revision triggers that detect law change from observational evidence. Causal discovery under non-stationarity~\citep{huang2020cdnod,song2023nctrl} provides identifiability results but treats change as variation within a fixed meta-model rather than as structural law replacement.

\item \textbf{Harness designs for agentic world modeling.} Agent performance has evolved through three successive abstractions: prompt engineering optimizes what the model is told, context engineering~\cite{anthropic2025context} curates the information state across turns, and harness engineering designs the executable environment surrounding the model: tools, memory, feedback loops, and inter-agent topology~\citep{rajasekaran2026harness, pan2026nlah}. This progression implies that agent behavior is governed not by the model alone but by transition dynamics of its execution environment, making harness design a form of world modeling for software agents. The problem is how to learn and synthesize harnesses from interaction data, treating the execution environment itself as the object of modeling rather than a fixed engineering assumption.
\end{enumerate}

\paragraph{Cross-regime shared challenges.}
Despite the diversity of governing-law regimes, three open problems recur across all four domains and constitute the deepest bottlenecks for world modeling in agentic AI. \emph{Deployment shift}: world models trained on offline data or simulation systematically underperform when the environment drifts. UI layouts change, physical contact properties shift, social norms evolve, and scientific instruments recalibrate. Robust world modeling requires online mechanisms that detect distribution shift early and trigger targeted revision rather than waiting for catastrophic failure.
\emph{Constraint enforcement}: all four regimes have governing laws that valid trajectories must satisfy (contact stability, state-machine consistency, norm compliance, evidence-chain validity), yet current models enforce these constraints only softly through training objectives; hard enforcement at inference time, via symbolic layers, constrained rollout, or verification gates, remains an open architectural problem.
\emph{Persistent update governance}: L3 systems that revise themselves from evidence face a trilemma of stability (avoid regressing on past capabilities), plasticity (incorporate new evidence quickly), and auditability (trace every update to its evidence source); no current system resolves all three, and the governance infrastructure (versioning, canary deployment, rollback policies, and regression harnesses) is underspecified in most published architectures. The MREP framework proposed in Section~\ref{sec:evaluation} offers a starting point for standardizing evaluation across these shared challenges by providing version-locked, reproducible evaluation packages that make cross-regime comparison tractable.

\subsection{Security and Safety of World Models}
\label{sec:trends:security}

The survey's governing-law axis offers a unifying lens on world-model safety: a security failure is, at bottom, a violation of one of the laws a world model should respect, whether physical, digital, social, or scientific, and because agents act on the model's \emph{imagined} rollouts, corrupting it can propagate to the law-governed decisions that depend on its rollouts, making the world model itself a potentially high-leverage attack surface. \citet{zeng2024wmsafety} review world models from a trustworthiness and safety standpoint, while \citet{li2026embodiedsafety} synthesize over 500 works into a multi-level taxonomy of adversarial, backdoor, and jailbreak attacks and their defenses across the embodied pipeline. Concretely, PhysCond-WMA~\citep{guo2026physcond} perturbs physical-condition inputs such as HDMap and 3D-box features of a driving world model to a reported 55\% targeted attack success rate while preserving perceptual fidelity; CtrlAttack~\citep{xu2026ctrlattack} injects a low-dimensional velocity field into diffusion image-to-video dynamics with high attack success in both white- and black-box settings; and the backdoor TRAP~\citep{duan2026trap} reorders a few decision-critical imagined trajectories to hijack planning on DreamerV3 and TD-MPC2 while leaving clean inputs intact, with WMAttack~\citep{guo2026wmattack} automating such adversarial evaluation as a finite-budget attack search. These attack studies are recent preprints from a small set of overlapping author groups whose findings await independent replication; analogous probes target the social regime through adversarial Theory-of-Mind and persona stress-testing~\citep{sclar2024exploretom,samuel2024personagym}, and JailWAM~\citep{liu2026jailwam} jailbreaks world-action models in robot control (an 84.2\% attack success rate on LingBot-VA), where a digital breach extends to the personal, property, and environmental safety that its accompanying JailWAM-Bench benchmark is built to quantify, as strong physical-interaction capability is itself a potent attack lever.

Defensively, the same generative machinery is dual-use: SafeDream~\citep{yan2026safedream} runs a lightweight safety-state world model that contrastively imagines attack-versus-benign futures to flag multi-turn jailbreaks before the model complies, while CounterScene~\citep{jing2026counterscene} uses a counterfactual BEV world model to surface safety-critical driving scenarios (raising the long-horizon collision rate from 12.3\% to 22.7\% over the strongest baseline), the same machinery that can equally fabricate adversarial cases. Such monitors and generators complement harder inference-time constraint enforcement, symbolic layers and verification gates that reject law-violating rollouts before they reach a planner, and connect to L3 governance (Section~\ref{subsec:l3_context}), where revision loops that ingest adversarially constructed evidence threaten scientific law through knowledge contamination. Overall, security evaluation of world-model agents remains nascent and rests largely on not-yet-peer-reviewed preprints, so the reported figures should be read as preliminary; progress will require regime-specific threat models and benchmarks rather than a single notion of robustness assumed to hold uniformly across the physical, digital, social, and scientific regimes, which fail in distinct ways.

\subsection{Beyond L3}
\label{sec:trends:beyond}

The L1$\to$L2$\to$L3 hierarchy (Sections~\ref{sec:l1}--\ref{sec:l3}) assumes that the world operates under a fixed set of governing laws: L1 learns local regularities, L2 composes them into constraint-consistent rollouts, and L3 revises the model when evidence contradicts predictions.
At L3, the system can update the laws governing its world model, but these updates remain grounded in explaining a single underlying reality.

A natural extension is \textbf{meta-world modeling}: systems that reason not only about a particular transition function, but about the space of possible transition functions itself. Rather than refining a model of the observed world, such systems would explore alternative rule systems that define different possible environments, for example by varying, extending, or constructing new assumptions, constraints, or governing principles.

As discussed in Section~\ref{sec:trends:representation}, the ability to explicitly represent and manipulate such principles becomes increasingly important in this setting. In particular, symbolic representations may provide a more natural interface for meta-world modeling, as they allow governing rules to be directly modified, composed, and compared across alternative worlds.

What forms this capability might take, whether through program synthesis, open-ended evolution, procedural world generation, or other mechanisms, remains an open question. More broadly, this raises the question of whether the endpoint of world modeling lies in increasingly accurate models of a world, or in systems that can systematically explore and reason over multiple worlds defined by different governing laws. We expect different research communities to arrive at different formulations of this capability, depending on whether the focus lies on predictive performance, scientific understanding, or generative modeling, and we leave the question of what constitutes the ultimate world model as an open invitation to the broader community.
\section{Conclusion}
\label{sec:conclusion}

This paper has proposed a capability-based taxonomy for world modeling organized along two axes: three capability levels (Predictor, Simulator, Evolver) and four perspectives (physical, digital, social, scientific).

An \textbf{L1 Predictor} learns local transition operators $p_\theta(z_t \mid z_{t-1}, a_t)$ whose quality is measured by one-step calibration, robustness, and identifiability (Section~\ref{sec:l1}).
An \textbf{L2 Simulator} composes these operators into long-horizon, action-conditioned rollouts that must satisfy three boundary conditions (long-horizon coherence, intervention sensitivity, and constraint consistency) under the governing laws of the target domain (Section~\ref{sec:l2}).
An \textbf{L3 Evolver} closes the loop by autonomously designing experiments, collecting evidence, and revising its dynamics model when predictions fail (Section~\ref{sec:l3}). Current systems such as CAMEO~\citep{kusne2020cameo} and A-Lab~\citep{szymanski2023alab} in autonomous science provide the strongest evidence that closed-loop model revision is already feasible in well-instrumented domains, while evaluator-guided algorithmic discovery systems such as FunSearch~\citep{romera2024funsearch} and AlphaEvolve~\citep{alphaevolve2025} show how automated scoring and regression gates can support partial L3 loops.

Organizing domains by \textbf{governing-law} regime rather than by modality has revealed both shared principles and irreducible differences (Section~\ref{sec:l2}). \textit{Physical}-world systems benefit from geometric and conservation-law priors; \textit{digital}-world systems exploit deterministic program semantics; \textit{social}-world systems demand Theory-of-Mind representations; and \textit{scientific}-world systems couple models to experimental evidence streams. The L1$\to$L2$\to$L3 taxonomy applies uniformly across these regimes, but the content of each level (what constitutes a valid rollout, what counts as a law violation, what evidence is available) varies fundamentally.

To make capability claims testable, we proposed decision-centric evaluation principles and a minimal reproducible evaluation package (Section~\ref{sec:evaluation}), and provided an architectural roadmap mapping representation, dynamics, and control choices to each capability level and deployment regime (Section~\ref{sec:implementation}). The open problems in Section~\ref{sec:trends} trace a research agenda: causal representation learning at L1, law-consistent rollout and compositional generalization at L2, and safe autonomous experiment design and model revision at L3.

A cross-cutting theme emerging from this survey is the question of representation substrate. The L1$\to$L2$\to$L3 progression describes what a world model can do, but leaves open what form it should take.
Latent continuous representations have proven indispensable for learning transition operators at scale, yet the history of scientific discovery suggests that law revision, the hallmark of L3, has typically relied on symbolic substrates: Newton's laws, Maxwell's equations, and the Standard Model are all world models whose governing principles are explicit, composable, and directly revisable.
Current neural world models encode invariances implicitly through architecture and training, which suits L1 and L2 but becomes a liability at L3, where the task is to revise model structure itself.
We therefore view the development of world models that can discover and manipulate symbolic governing laws from data, rather than merely absorbing them into latent representations, as one of the most important open problems in the field.
How to extend this to physical, digital, social, and scientific worlds remains a fundamental challenge.

\begin{takeaway}[Key Takeaway]
The future of agentic AI lies not in larger predictors, but in models that internalize the governing laws of the world, simulate its dynamics, and continuously evolve themselves through active trial-and-error loops, enabling them to navigate, interpret, and ultimately reshape the world.
\end{takeaway}

\clearpage
\begingroup
\makeatletter\if@github\small\fi\makeatother
\bibliography{main}

\begin{thebibliography}{518}
\providecommand{\natexlab}[1]{#1}
\providecommand{\url}[1]{\texttt{#1}}
\expandafter\ifx\csname urlstyle\endcsname\relax
  \providecommand{\doi}[1]{doi: #1}\else
  \providecommand{\doi}{doi: \begingroup \urlstyle{rm}\Url}\fi

\bibitem[Abramson et~al.(2024)Abramson, Adler, Dunger, Evans, Green, Pritzel, Ronneberger, Willmore, Ballard, Bambrick, Bodenstein, Evans, Hung, O'Neill, Reiman, Tunyasuvunakool, Wu, {\v{Z}}emguly{\.{t}}{\.{e}}, Arvaniti, Beattie, Bertolli, Bridgland, Cherepanov, Congreve, Cowen-Rivers, Cowie, Figurnov, Fuchs, Gladman, Jain, Khan, Low, Perlin, Potapenko, Savy, Singh, Stecula, Thillaisundaram, Tong, Yakneen, Zhong, Zielinski, {\v{Z}}{\'\i}dek, Bapst, Kohli, Jaderberg, Hassabis, and Jumper]{abramson2024alphafold3}
J.~Abramson, J.~Adler, J.~Dunger, R.~Evans, T.~Green, A.~Pritzel, O.~Ronneberger, L.~Willmore, A.~J. Ballard, J.~Bambrick, S.~W. Bodenstein, D.~A. Evans, C.-C. Hung, M.~O'Neill, D.~Reiman, K.~Tunyasuvunakool, Z.~Wu, A.~{\v{Z}}emguly{\.{t}}{\.{e}}, E.~Arvaniti, C.~Beattie, O.~Bertolli, A.~Bridgland, A.~Cherepanov, M.~Congreve, A.~I. Cowen-Rivers, A.~Cowie, M.~Figurnov, F.~B. Fuchs, H.~Gladman, R.~Jain, Y.~A. Khan, C.~M.~R. Low, K.~Perlin, A.~Potapenko, P.~Savy, S.~Singh, A.~Stecula, A.~Thillaisundaram, C.~Tong, S.~Yakneen, E.~D. Zhong, M.~Zielinski, A.~{\v{Z}}{\'\i}dek, V.~Bapst, P.~Kohli, M.~Jaderberg, D.~Hassabis, and J.~M. Jumper.
\newblock Accurate structure prediction of biomolecular interactions with {AlphaFold} 3.
\newblock \emph{Nature}, 630:\penalty0 493--500, 2024.

\bibitem[Agarwal et~al.(2025)Agarwal, Ali, Bala, Balaji, Barker, Cai, Chattopadhyay, Chen, Cui, Ding, Dworakowski, Fan, Fenzi, Ferroni, Fidler, Fox, Ge, Ge, Gu, Gururani, He, Huang, Huffman, Jannaty, Jin, Kim, Klár, Lam, Lan, Leal-Taixe, Li, Li, Lin, Lin, Ling, Liu, Liu, Luo, Ma, Mao, Mo, Mousavian, Nah, Niverty, Page, Paschalidou, Patel, Pavao, Ramezanali, Reda, Ren, Sabavat, Schmerling, Shi, Stefaniak, Tang, Tchapmi, Tredak, Tseng, Varghese, Wang, Wang, Wang, Wang, Wei, Wei, Wu, Xu, Yang, Yen-Chen, Zeng, Zeng, Zhang, Zhang, Zhang, Zhao, and Zolkowski]{nvidia2025cosmos}
N.~Agarwal, A.~Ali, M.~Bala, Y.~Balaji, E.~Barker, T.~Cai, P.~Chattopadhyay, Y.~Chen, Y.~Cui, Y.~Ding, D.~Dworakowski, J.~Fan, M.~Fenzi, F.~Ferroni, S.~Fidler, D.~Fox, S.~Ge, Y.~Ge, J.~Gu, S.~Gururani, E.~He, J.~Huang, J.~Huffman, P.~Jannaty, J.~Jin, S.~W. Kim, G.~Klár, G.~Lam, S.~Lan, L.~Leal-Taixe, A.~Li, Z.~Li, C.-H. Lin, T.-Y. Lin, H.~Ling, M.-Y. Liu, X.~Liu, A.~Luo, Q.~Ma, H.~Mao, K.~Mo, A.~Mousavian, S.~Nah, S.~Niverty, D.~Page, D.~Paschalidou, Z.~Patel, L.~Pavao, M.~Ramezanali, F.~Reda, X.~Ren, V.~R.~N. Sabavat, E.~Schmerling, S.~Shi, B.~Stefaniak, S.~Tang, L.~Tchapmi, P.~Tredak, W.-C. Tseng, J.~Varghese, H.~Wang, H.~Wang, H.~Wang, T.-C. Wang, F.~Wei, X.~Wei, J.~Z. Wu, J.~Xu, W.~Yang, L.~Yen-Chen, X.~Zeng, Y.~Zeng, J.~Zhang, Q.~Zhang, Y.~Zhang, Q.~Zhao, and A.~Zolkowski.
\newblock Cosmos world foundation model platform for physical {AI}.
\newblock \emph{arXiv preprint arXiv:2501.03575}, 2025.

\bibitem[Agarwal et~al.(2021)Agarwal, Schwarzer, Castro, Courville, and Bellemare]{agarwal2021statistical}
R.~Agarwal, M.~Schwarzer, P.~S. Castro, A.~C. Courville, and M.~G. Bellemare.
\newblock Deep reinforcement learning at the edge of the statistical precipice.
\newblock In \emph{Advances in Neural Information Processing Systems}, volume~34, 2021.

\bibitem[Agrawal et~al.(2016)Agrawal, Nair, Abbeel, Malik, and Levine]{agrawal2016learning}
P.~Agrawal, A.~V. Nair, P.~Abbeel, J.~Malik, and S.~Levine.
\newblock Learning to poke by poking: Experiential learning of intuitive physics.
\newblock In \emph{Advances in Neural Information Processing Systems}, volume~29, pages 5092--5100, 2016.

\bibitem[AL et~al.(2024)AL, Ahn, Becker, Carroll, Christie, Cortes, Demirci, Du, Li, Luo, Wang, Willows, Yang, and Yang]{altera2024projectsid}
A.~AL, A.~Ahn, N.~Becker, S.~Carroll, N.~Christie, M.~Cortes, A.~Demirci, M.~Du, F.~Li, S.~Luo, P.~Y. Wang, M.~Willows, F.~Yang, and G.~R. Yang.
\newblock Project {Sid}: Many-agent simulations toward {AI} civilization.
\newblock \emph{arXiv preprint arXiv:2411.00114}, 2024.

\bibitem[Alonso et~al.(2024)Alonso, Jelley, Micheli, Kanervisto, Storkey, Pearce, and Fleuret]{alonso2024diamond}
E.~Alonso, A.~Jelley, V.~Micheli, A.~Kanervisto, A.~Storkey, T.~Pearce, and F.~Fleuret.
\newblock Diffusion for world modeling: Visual details matter in atari.
\newblock \emph{Advances in Neural Information Processing Systems}, 37:\penalty0 58757--58791, 2024.

\bibitem[Andrychowicz et~al.(2016)Andrychowicz, Denil, Gomez, Hoffman, Pfau, Schaul, Shillingford, and de~Freitas]{andrychowicz2016learning}
M.~Andrychowicz, M.~Denil, S.~Gomez, M.~W. Hoffman, D.~Pfau, T.~Schaul, B.~Shillingford, and N.~de~Freitas.
\newblock Learning to learn by gradient descent by gradient descent.
\newblock In \emph{Advances in Neural Information Processing Systems}, volume~29, pages 3988--3996, 2016.

\bibitem[Angermueller et~al.(2020)Angermueller, Belanger, Gane, Mariet, Dohan, Murphy, Colwell, and Sculley]{angermueller2020population}
C.~Angermueller, D.~Belanger, A.~Gane, Z.~Mariet, D.~Dohan, K.~Murphy, L.~Colwell, and D.~Sculley.
\newblock Population-based black-box optimization for biological sequence design.
\newblock In \emph{International Conference on Machine Learning}, pages 324--334. PMLR, 2020.

\bibitem[{Anthropic}(2025)]{anthropic2025context}
{Anthropic}.
\newblock Effective context engineering for {AI} agents.
\newblock Anthropic Engineering Blog, 2025.
\newblock URL \url{https://www.anthropic.com/engineering/effective-context-engineering-for-ai-agents}.

\bibitem[Argyle et~al.(2023)Argyle, Busby, Fulda, Gubler, Rytting, and Wingate]{argyle2023silicon}
L.~P. Argyle, E.~C. Busby, N.~Fulda, J.~R. Gubler, C.~Rytting, and D.~Wingate.
\newblock Out of one, many: Using language models to simulate human samples.
\newblock \emph{Political Analysis}, 31\penalty0 (3):\penalty0 337--351, 2023.

\bibitem[Arunkumar et~al.(2026)Arunkumar, Gangadharan, and Buyya]{arunkumar2026agenticai}
V.~Arunkumar, G.~R. Gangadharan, and R.~Buyya.
\newblock Agentic artificial intelligence ({AI}): Architectures, taxonomies, and evaluation of large language model agents.
\newblock \emph{arXiv preprint arXiv:2601.12560}, 2026.

\bibitem[Ashery et~al.(2025)Ashery, Aiello, and Baronchelli]{ashery2025conventions}
A.~F. Ashery, L.~M. Aiello, and A.~Baronchelli.
\newblock Emergent social conventions and collective bias in {LLM} populations.
\newblock \emph{Science Advances}, 11, 2025.

\bibitem[Ashkboos et~al.(2024)Ashkboos, Mohtashami, Croci, Li, Cameron, Jaggi, Alistarh, Hoefler, and Hensman]{ashkboos2024quarot}
S.~Ashkboos, A.~Mohtashami, M.~L. Croci, B.~Li, P.~Cameron, M.~Jaggi, D.~Alistarh, T.~Hoefler, and J.~Hensman.
\newblock {QuaRot}: Outlier-free 4-bit inference in rotated {LLMs}.
\newblock In \emph{Advances in Neural Information Processing Systems}, volume~37, pages 100213--100240, 2024.

\bibitem[Assran et~al.(2023)Assran, Duval, Misra, Bojanowski, Vincent, Rabbat, LeCun, and Ballas]{assran2023ijepa}
M.~Assran, Q.~Duval, I.~Misra, P.~Bojanowski, P.~Vincent, M.~Rabbat, Y.~LeCun, and N.~Ballas.
\newblock Self-supervised learning from images with a joint-embedding predictive architecture.
\newblock In \emph{IEEE/CVF Conference on Computer Vision and Pattern Recognition}, pages 15619--15629, 2023.

\bibitem[Assran et~al.(2025)Assran, Bardes, Fan, Garrido, Howes, Mojtaba, Komeili, Muckley, Rizvi, Roberts, Sinha, Zholus, Arnaud, Gejji, Martin, Hogan, Dugas, Bojanowski, Khalidov, Labatut, Massa, Szafraniec, Krishnakumar, Li, Ma, Chandar, Meier, LeCun, Rabbat, and Ballas]{meta2025vjepa2}
M.~Assran, A.~Bardes, D.~Fan, Q.~Garrido, R.~Howes, Mojtaba, Komeili, M.~Muckley, A.~Rizvi, C.~Roberts, K.~Sinha, A.~Zholus, S.~Arnaud, A.~Gejji, A.~Martin, F.~R. Hogan, D.~Dugas, P.~Bojanowski, V.~Khalidov, P.~Labatut, F.~Massa, M.~Szafraniec, K.~Krishnakumar, Y.~Li, X.~Ma, S.~Chandar, F.~Meier, Y.~LeCun, M.~Rabbat, and N.~Ballas.
\newblock {V-JEPA 2}: Self-supervised video models enable understanding, prediction and planning.
\newblock \emph{arXiv preprint arXiv:2506.09985}, 2025.

\bibitem[Babaeizadeh et~al.(2018)Babaeizadeh, Finn, Erhan, Campbell, and Levine]{babaeizadeh2018sv2p}
M.~Babaeizadeh, C.~Finn, D.~Erhan, R.~H. Campbell, and S.~Levine.
\newblock Stochastic variational video prediction.
\newblock In \emph{International Conference on Learning Representations}, 2018.

\bibitem[Baek et~al.(2021)Baek, DiMaio, Anishchenko, Dauparas, Ovchinnikov, Lee, Wang, Cong, Kinch, Schaeffer, Mill{\'a}n, Park, Adams, Glassman, DeGiovanni, Pereira, Rodrigues, van Dijk, Ebrecht, Opperman, Sagmeister, Buhlheller, Pavkov-Keller, Rathinaswamy, Dalwadi, Yip, Burke, Garcia, Grishin, Adams, Read, and Baker]{baek2021rosettafold}
M.~Baek, F.~DiMaio, I.~Anishchenko, J.~Dauparas, S.~Ovchinnikov, G.~R. Lee, J.~Wang, Q.~Cong, L.~N. Kinch, R.~D. Schaeffer, C.~Mill{\'a}n, H.~Park, C.~Adams, C.~R. Glassman, A.~DeGiovanni, J.~H. Pereira, A.~V. Rodrigues, A.~A. van Dijk, A.~C. Ebrecht, D.~J. Opperman, T.~Sagmeister, C.~Buhlheller, T.~Pavkov-Keller, M.~K. Rathinaswamy, U.~Dalwadi, C.~K. Yip, J.~E. Burke, K.~C. Garcia, N.~V. Grishin, P.~D. Adams, R.~J. Read, and D.~Baker.
\newblock Accurate prediction of protein structures and interactions using a three-track neural network.
\newblock \emph{Science}, 373\penalty0 (6557):\penalty0 871--876, 2021.

\bibitem[Baker et~al.(2014)Baker, Brookes, Rezek, Smith, Behrens, Probert~Smith, and Woolrich]{baker2014hmm}
A.~P. Baker, M.~J. Brookes, I.~A. Rezek, S.~M. Smith, T.~Behrens, P.~J. Probert~Smith, and M.~Woolrich.
\newblock Fast transient networks in spontaneous human brain activity.
\newblock \emph{eLife}, 3:\penalty0 e01867, 2014.

\bibitem[Baker et~al.(2022)Baker, Akkaya, Zhokhov, Huizinga, Tang, Ecoffet, Houghton, Sampedro, and Clune]{baker2022video}
B.~Baker, I.~Akkaya, P.~Zhokhov, J.~Huizinga, J.~Tang, A.~Ecoffet, B.~Houghton, R.~Sampedro, and J.~Clune.
\newblock Video {P}re{T}raining ({VPT}): Learning to act by watching unlabeled online videos.
\newblock In \emph{Advances in Neural Information Processing Systems}, volume~35, pages 24639--24654, 2022.

\bibitem[Baker et~al.(2011)Baker, Saxe, and Tenenbaum]{baker2011bayesian}
C.~Baker, R.~Saxe, and J.~Tenenbaum.
\newblock Bayesian theory of mind: Modeling joint belief-desire attribution.
\newblock In \emph{Annual Meeting of the Cognitive Science Society}, volume~33, 2011.

\bibitem[Bakhtin et~al.(2022)Bakhtin, Brown, Dinan, Farina, Flaherty, Fried, Goff, Gray, Hu, Jacob, Komeili, Konath, et~al.]{meta2022cicero}
A.~Bakhtin, N.~Brown, E.~Dinan, G.~Farina, C.~Flaherty, D.~Fried, A.~Goff, J.~Gray, H.~Hu, A.~P. Jacob, M.~Komeili, K.~Konath, et~al.
\newblock Human-level play in the game of diplomacy by combining language models with strategic reasoning.
\newblock \emph{Science}, 378\penalty0 (6624):\penalty0 1067--1074, 2022.

\bibitem[Ball et~al.(2025)Ball, Bauer, Belletti, Brownfield, Ephrat, Fruchter, Gupta, Holsheimer, Holynski, Hron, Kaplanis, Limont, McGill, Oliveira, Parker-Holder, Perbet, Scully, Shar, Spencer, Tov, Villegas, Wang, Yung, Baetu, Berbel, Bridson, Bruce, Buttimore, Chakera, Chandra, Collins, Cullum, Damoc, Dasagi, Gazeau, Gbadamosi, Han, Hirst, Kachra, Kerley, Kjems, Knoepfel, Koriakin, Lo, Lu, Mehring, Moufarek, Nandwani, Oliveira, Pardo, Park, Pierson, Poole, Ran, Salimans, Sanchez, Saprykin, Shen, Sidhwani, Smith, Stanton, Tomlinson, Vijaykumar, Wang, Wingfield, Wong, Xu, Yew, Young, Zubov, Eck, Erhan, Kavukcuoglu, Hassabis, Gharamani, Hadsell, van~den Oord, Mosseri, Bolton, Singh, and Rockt{\"a}schel]{deepmind2025genie3}
P.~J. Ball, J.~Bauer, F.~Belletti, B.~Brownfield, A.~Ephrat, S.~Fruchter, A.~Gupta, K.~Holsheimer, A.~Holynski, J.~Hron, C.~Kaplanis, M.~Limont, M.~McGill, Y.~Oliveira, J.~Parker-Holder, F.~Perbet, G.~Scully, J.~Shar, S.~Spencer, O.~Tov, R.~Villegas, E.~Wang, J.~Yung, C.~Baetu, J.~Berbel, D.~Bridson, J.~Bruce, G.~Buttimore, S.~Chakera, B.~Chandra, P.~Collins, A.~Cullum, B.~Damoc, V.~Dasagi, M.~Gazeau, C.~Gbadamosi, W.~Han, E.~Hirst, A.~Kachra, L.~Kerley, K.~Kjems, E.~Knoepfel, V.~Koriakin, J.~Lo, C.~Lu, Z.~Mehring, A.~Moufarek, H.~Nandwani, V.~Oliveira, F.~Pardo, J.~Park, A.~Pierson, B.~Poole, H.~Ran, T.~Salimans, M.~Sanchez, I.~Saprykin, A.~Shen, S.~Sidhwani, D.~Smith, J.~Stanton, H.~Tomlinson, D.~Vijaykumar, L.~Wang, P.~Wingfield, N.~Wong, K.~Xu, C.~Yew, N.~Young, V.~Zubov, D.~Eck, D.~Erhan, K.~Kavukcuoglu, D.~Hassabis, Z.~Gharamani, R.~Hadsell, A.~van~den Oord, I.~Mosseri, A.~Bolton, S.~Singh, and T.~Rockt{\"a}schel.
\newblock Genie 3: A new frontier for world models, 2025.
\newblock URL \url{https://deepmind.google/discover/blog/genie-3-a-new-frontier-for-world-models/}.

\bibitem[Bar-Tal et~al.(2024)Bar-Tal, Chefer, Tov, Herrmann, Paiss, Zada, Ephrat, Hur, Liu, Raj, Li, Rubinstein, Michaeli, Wang, Sun, Dekel, and Mosseri]{bartal2024lumiere}
O.~Bar-Tal, H.~Chefer, O.~Tov, C.~Herrmann, R.~Paiss, S.~Zada, A.~Ephrat, J.~Hur, G.~Liu, A.~Raj, Y.~Li, M.~Rubinstein, T.~Michaeli, O.~Wang, D.~Sun, T.~Dekel, and I.~Mosseri.
\newblock Lumiere: A space-time diffusion model for video generation.
\newblock In \emph{SIGGRAPH Asia}, pages 1--11, 2024.

\bibitem[Barcellona et~al.(2024)Barcellona, Zadaianchuk, Allegro, Papa, Ghidoni, and Gavves]{wu2024drema}
L.~Barcellona, A.~Zadaianchuk, D.~Allegro, S.~Papa, S.~Ghidoni, and E.~Gavves.
\newblock Dream to manipulate: Compositional world models empowering robot imitation learning with imagination.
\newblock \emph{arXiv preprint arXiv:2412.14957}, 2024.

\bibitem[Bardes et~al.(2024)Bardes, Garrido, Ponce, Chen, Rabbat, LeCun, Assran, and Ballas]{bardes2024vjepa}
A.~Bardes, Q.~Garrido, J.~Ponce, X.~Chen, M.~G. Rabbat, Y.~LeCun, M.~Assran, and N.~Ballas.
\newblock Revisiting feature prediction for learning visual representations from video.
\newblock \emph{arXiv preprint arXiv:2404.08471}, 2024.

\bibitem[Behler and Parrinello(2007)]{behler2007nnpotentials}
J.~Behler and M.~Parrinello.
\newblock Generalized neural-network representation of high-dimensional potential-energy surfaces.
\newblock \emph{Physical Review Letters}, 98\penalty0 (14):\penalty0 146401, 2007.

\bibitem[Beucler et~al.(2024)Beucler, Gentine, Yuval, Gupta, Peng, Lin, Yu, Rasp, Ahmed, O'Gorman, Neelin, Lutsko, and Pritchard]{beucler2024climate}
T.~Beucler, P.~Gentine, J.~Yuval, A.~Gupta, L.~Peng, J.~Lin, S.~Yu, S.~Rasp, F.~Ahmed, P.~A. O'Gorman, J.~D. Neelin, N.~J. Lutsko, and M.~Pritchard.
\newblock Climate-invariant machine learning.
\newblock \emph{Science Advances}, 10\penalty0 (6):\penalty0 eadj7250, 2024.

\bibitem[Bi et~al.(2023)Bi, Xie, Zhang, Chen, Gu, and Tian]{bi2023panguweather}
K.~Bi, L.~Xie, H.~Zhang, X.~Chen, X.~Gu, and Q.~Tian.
\newblock Accurate medium-range global weather forecasting with {3D} neural networks.
\newblock \emph{Nature}, 619:\penalty0 533--538, 2023.

\bibitem[Bian et~al.(2025)Bian, Kong, Xie, Pan, Qiao, and Liu]{bian2025dynamiccity}
H.~Bian, L.~Kong, H.~Xie, L.~Pan, Y.~Qiao, and Z.~Liu.
\newblock {DynamicCity}: Large-scale {4D} occupancy generation from dynamic scenes.
\newblock In \emph{International Conference on Learning Representations}, 2025.

\bibitem[Bianchi et~al.(2024)Bianchi, Chia, Yuksekgonul, Tagliabue, Jurafsky, and Zou]{bianchi2024negotiationarena}
F.~Bianchi, P.~J. Chia, M.~Yuksekgonul, J.~Tagliabue, D.~Jurafsky, and J.~Zou.
\newblock How well can {LLMs} negotiate? negotiationarena platform and analysis.
\newblock \emph{arXiv preprint arXiv:2402.05863}, 2024.

\bibitem[Bodnar et~al.(2025)Bodnar, Bruinsma, Lucic, Stanley, Allen, Brandstetter, Garvan, Riechert, Weyn, Dong, Gupta, Thambiratnam, Archibald, Wu, Heider, Welling, Turner, and Perdikaris]{bodnar2025aurora}
C.~Bodnar, W.~P. Bruinsma, A.~Lucic, M.~Stanley, A.~Allen, J.~Brandstetter, P.~Garvan, M.~Riechert, J.~A. Weyn, H.~Dong, J.~K. Gupta, K.~Thambiratnam, A.~T. Archibald, C.-C. Wu, E.~Heider, M.~Welling, R.~E. Turner, and P.~Perdikaris.
\newblock A foundation model for the earth system.
\newblock \emph{Nature}, 641\penalty0 (8065):\penalty0 1180--1187, 2025.

\bibitem[Boella and van~der Torre(2007)]{boella2007norms}
G.~Boella and L.~van~der Torre.
\newblock A game-theoretic approach to normative multi-agent systems.
\newblock In \emph{Normative Multi-agent Systems}. Schloss Dagstuhl, 2007.

\bibitem[Boffi et~al.(2025)Boffi, Albergo, and Vanden-Eijnden]{boffi2024flow}
N.~M. Boffi, M.~S. Albergo, and E.~Vanden-Eijnden.
\newblock Flow map matching with stochastic interpolants: A mathematical framework for consistency models.
\newblock \emph{Transactions on Machine Learning Research}, 2025.

\bibitem[Boiko et~al.(2023)Boiko, MacKnight, Kline, and Gomes]{boiko2023autonomous}
D.~A. Boiko, R.~MacKnight, B.~Kline, and G.~Gomes.
\newblock Autonomous chemical research with large language models.
\newblock \emph{Nature}, 624:\penalty0 570--578, 2023.

\bibitem[Bolya and Hoffman(2023)]{bolya2023token}
D.~Bolya and J.~Hoffman.
\newblock Token merging for fast stable diffusion.
\newblock In \emph{IEEE/CVF Conference on Computer Vision and Pattern Recognition Workshops}, pages 4599--4603, 2023.

\bibitem[Bolya et~al.(2023)Bolya, Fu, Dai, Zhang, Feichtenhofer, and Hoffman]{bolya2022token}
D.~Bolya, C.-Y. Fu, X.~Dai, P.~Zhang, C.~Feichtenhofer, and J.~Hoffman.
\newblock Token merging: Your vit but faster.
\newblock In \emph{International Conference on Learning Representations}, 2023.

\bibitem[Bran et~al.(2024)Bran, Cox, Schilter, Baldassari, White, and Schwaller]{bran2024augmenting}
A.~M. Bran, S.~Cox, O.~Schilter, C.~Baldassari, A.~D. White, and P.~Schwaller.
\newblock Augmenting large language models with chemistry tools.
\newblock \emph{Nature Machine Intelligence}, 6\penalty0 (5):\penalty0 525--535, 2024.

\bibitem[Brooks et~al.(2024)Brooks, Peebles, Holmes, DePue, Guo, Jing, Schnurr, Taylor, Luhman, Luhman, Ng, Wang, and Ramesh]{brooks2024sora}
T.~Brooks, B.~Peebles, C.~Holmes, W.~DePue, Y.~Guo, L.~Jing, D.~Schnurr, J.~Taylor, T.~Luhman, E.~Luhman, C.~Ng, R.~Wang, and A.~Ramesh.
\newblock Video generation models as world simulators.
\newblock Technical report, OpenAI, 2024.
\newblock URL \url{https://openai.com/research/video-generation-models-as-world-simulators}.

\bibitem[Brown et~al.(2020)Brown, Mann, Ryder, Subbiah, Kaplan, Dhariwal, Neelakantan, Shyam, Sastry, Askell, Agarwal, Herbert-Voss, Krueger, Henighan, Child, Ramesh, Ziegler, Wu, Winter, Hesse, Chen, Sigler, Litwin, Gray, Chess, Clark, Berner, McCandlish, Radford, Sutskever, and Amodei]{brown2020gpt3}
T.~B. Brown, B.~Mann, N.~Ryder, M.~Subbiah, J.~D. Kaplan, P.~Dhariwal, A.~Neelakantan, P.~Shyam, G.~Sastry, A.~Askell, S.~Agarwal, A.~Herbert-Voss, G.~Krueger, T.~Henighan, R.~Child, A.~Ramesh, D.~M. Ziegler, J.~Wu, C.~Winter, C.~Hesse, M.~Chen, E.~Sigler, M.~Litwin, S.~Gray, B.~Chess, J.~Clark, C.~Berner, S.~McCandlish, A.~Radford, I.~Sutskever, and D.~Amodei.
\newblock Language models are few-shot learners.
\newblock In \emph{Advances in Neural Information Processing Systems}, volume~33, pages 1877--1901, 2020.

\bibitem[Bruce et~al.(2024)Bruce, Dennis, Edwards, Parker-Holder, Shi, Hughes, Lai, Mavalankar, Steigerwald, Apps, Aytar, Bechtle, Behbahani, Chan, Heess, Gonzalez, Osindero, Ozair, Reed, Zhang, Zolna, Clune, de~Freitas, Singh, and Rockt{\"a}schel]{bruce2024genie}
J.~Bruce, M.~Dennis, A.~Edwards, J.~Parker-Holder, Y.~Shi, E.~Hughes, M.~Lai, A.~Mavalankar, R.~Steigerwald, C.~Apps, Y.~Aytar, S.~Bechtle, F.~Behbahani, S.~Chan, N.~Heess, L.~Gonzalez, S.~Osindero, S.~Ozair, S.~Reed, J.~Zhang, K.~Zolna, J.~Clune, N.~de~Freitas, S.~Singh, and T.~Rockt{\"a}schel.
\newblock Genie: Generative interactive environments.
\newblock In \emph{International Conference on Machine Learning}, pages 4603--4623, 2024.

\bibitem[Butt et~al.(2024)Butt, Manczak, Wiggers, Rainone, Zhang, Defferrard, and Cohen]{butt2024codeit}
N.~Butt, B.~Manczak, A.~Wiggers, C.~Rainone, D.~W. Zhang, M.~Defferrard, and T.~Cohen.
\newblock {CodeIt}: Self-improving language models with prioritized hindsight replay.
\newblock In \emph{International Conference on Machine Learning}, pages 5013--5034. PMLR, 2024.

\bibitem[Caesar et~al.(2020)Caesar, Bankiti, Lang, Vora, Liong, Xu, Krishnan, Pan, Baldan, and Beijbom]{caesar2020nuscenes}
H.~Caesar, V.~Bankiti, A.~H. Lang, S.~Vora, V.~E. Liong, Q.~Xu, A.~Krishnan, Y.~Pan, G.~Baldan, and O.~Beijbom.
\newblock {nuScenes}: A multimodal dataset for autonomous driving.
\newblock In \emph{IEEE/CVF Conference on Computer Vision and Pattern Recognition}, pages 11621--11631, 2020.

\bibitem[Cao et~al.(2025{\natexlab{a}})Cao, Men, Liu, Zhang, Li, Lin, Sui, Cao, Liu, and Zhao]{cao2025llmplanning}
P.~Cao, T.~Men, W.~Liu, J.~Zhang, X.~Li, X.~Lin, D.~Sui, Y.~Cao, K.~Liu, and J.~Zhao.
\newblock Large language models for planning: A comprehensive and systematic survey.
\newblock \emph{arXiv preprint arXiv:2505.19683}, 2025{\natexlab{a}}.

\bibitem[Cao et~al.(2026)Cao, Zhong, Zeng, Zheng, Huang, Qiu, Shi, Mao, and Wan]{cao2026mobiledreamer}
Y.~Cao, Y.~Zhong, Z.~Zeng, L.~Zheng, J.~Huang, H.~Qiu, P.~Shi, W.~Mao, and G.~Wan.
\newblock {MobileDreamer}: Generative sketch world model for {GUI} agent.
\newblock \emph{arXiv preprint arXiv:2601.04035}, 2026.

\bibitem[Cao et~al.(2025{\natexlab{b}})Cao, Hong, Chen, Pan, and Liu]{cao2025physxanything}
Z.~Cao, F.~Hong, Z.~Chen, L.~Pan, and Z.~Liu.
\newblock {PhysX-Anything}: Simulation-ready physical {3D} assets from single image.
\newblock \emph{arXiv preprint arXiv:2511.13648}, 2025{\natexlab{b}}.

\bibitem[Chae et~al.(2025)Chae, Kim, Ong, Gwak, Song, Kim, Kim, Lee, and Yeo]{chae2025wma}
H.~Chae, N.~Kim, K.~T.-i. Ong, M.~Gwak, G.~Song, J.~Kim, S.~Kim, D.~Lee, and J.~Yeo.
\newblock Web agents with world models: Learning and leveraging environment dynamics in web navigation.
\newblock In \emph{International Conference on Learning Representations}, 2025.

\bibitem[Chai et~al.(2025)Chai, Deng, Shao, Zhang, Lv, Xing, Li, Zhang, and Liu]{chai2025gaf}
Y.~Chai, L.~Deng, R.~Shao, J.~Zhang, K.~Lv, L.~Xing, X.~Li, H.~Zhang, and Y.~Liu.
\newblock Gaf: Gaussian action field as a 4d representation for dynamic world modeling in robotic manipulation.
\newblock \emph{arXiv preprint arXiv:2506.14135}, 2025.

\bibitem[Che et~al.(2024)Che, He, Liu, Jin, and Chen]{che2024gamegenx}
H.~Che, X.~He, Q.~Liu, C.~Jin, and H.~Chen.
\newblock {GameGen-X}: Interactive open-world game video generation.
\newblock \emph{arXiv preprint arXiv:2411.00769}, 2024.

\bibitem[Chen et~al.(2022)Chen, Wu, Yoon, and Ahn]{chen2022transdreamer}
C.~Chen, Y.-F. Wu, J.~Yoon, and S.~Ahn.
\newblock {TransDreamer}: Reinforcement learning with transformer world models.
\newblock \emph{arXiv preprint arXiv:2202.09481}, 2022.

\bibitem[Chen et~al.(2025{\natexlab{a}})Chen, Shukor, Moutakanni, Chung, Yu, Kasarla, Bang, Bolourchi, LeCun, and Fung]{chen2025vl}
D.~Chen, M.~Shukor, T.~Moutakanni, W.~Chung, J.~Yu, T.~Kasarla, Y.~Bang, A.~Bolourchi, Y.~LeCun, and P.~Fung.
\newblock {VL-JEPA}: Joint embedding predictive architecture for vision-language.
\newblock \emph{arXiv preprint arXiv:2512.10942}, 2025{\natexlab{a}}.

\bibitem[Chen et~al.(2025{\natexlab{b}})Chen, Huang, Hua, He, and Schwaller]{chen2025multi}
J.~Chen, X.~Huang, C.~Hua, Y.~He, and P.~Schwaller.
\newblock A multi-modal transformer for predicting global minimum adsorption energy.
\newblock \emph{Nature Communications}, 16\penalty0 (1):\penalty0 3232, 2025{\natexlab{b}}.

\bibitem[Chen et~al.(2025{\natexlab{c}})Chen, Meng, Tang, Ma, Jiang, Wang, Wang, and Zhu]{chen2025q}
L.~Chen, Y.~Meng, C.~Tang, X.~Ma, J.~Jiang, X.~Wang, Z.~Wang, and W.~Zhu.
\newblock {Q-DiT}: Accurate post-training quantization for diffusion transformers.
\newblock In \emph{IEEE/CVF Conference on Computer Vision and Pattern Recognition}, pages 28306--28315, 2025{\natexlab{c}}.

\bibitem[Chen et~al.(2025{\natexlab{d}})Chen, Jiang, Qin, and Tan]{chen2025tomsurvey}
R.~Chen, W.~Jiang, C.~Qin, and C.~Tan.
\newblock Theory of mind in large language models: Assessment and enhancement.
\newblock In \emph{Annual Meeting of the Association for Computational Linguistics}, pages 31539--31558, 2025{\natexlab{d}}.

\bibitem[Chen et~al.(2020)Chen, Kornblith, Norouzi, and Hinton]{chen2020simclr}
T.~Chen, S.~Kornblith, M.~Norouzi, and G.~Hinton.
\newblock A simple framework for contrastive learning of visual representations.
\newblock In \emph{International Conference on Machine Learning}, pages 1597--1607. PMLR, 2020.

\bibitem[Chen et~al.(2025{\natexlab{e}})Chen, Chen, Fu, Gao, Jia, Jin, Li, Mu, Pang, Qiao, Tian, Wang, Wang, Wang, Wang, Wang, Wang, Wei, Wu, Yang, Ye, Yu, Zeng, Zhang, Zhang, Zhang, Zheng, Zhou, and Zhu]{chen2025internvla}
X.~Chen, Y.~Chen, Y.~Fu, N.~Gao, J.~Jia, W.~Jin, H.~Li, Y.~Mu, J.~Pang, Y.~Qiao, Y.~Tian, B.~Wang, B.~Wang, F.~Wang, H.~Wang, T.~Wang, Z.~Wang, X.~Wei, C.~Wu, S.~Yang, J.~Ye, J.~Yu, J.~Zeng, J.~Zhang, J.~Zhang, S.~Zhang, F.~Zheng, B.~Zhou, and Y.~Zhu.
\newblock {InternVLA-M1}: A spatially guided vision-language-action framework for generalist robot policy.
\newblock \emph{arXiv preprint arXiv:2510.13778}, 2025{\natexlab{e}}.

\bibitem[Chen et~al.(2025{\natexlab{f}})Chen, Lin, and Shou]{chen2025code2video}
Y.~Chen, K.~Q. Lin, and M.~Z. Shou.
\newblock {Code2Video}: A code-centric paradigm for educational video generation.
\newblock \emph{arXiv preprint arXiv:2510.01174}, 2025{\natexlab{f}}.

\bibitem[Chen et~al.(2026)Chen, Li, Yang, He, Wu, Xu, Wang, Liu, Liu, Huang, and Wang]{wang2026bridgev2w}
Y.~Chen, P.~Li, J.~Yang, K.~He, X.~Wu, Y.~Xu, K.~Wang, J.~Liu, N.~Liu, Y.~Huang, and L.~Wang.
\newblock {BridgeV2W}: Bridging video generation models to embodied world models via embodiment masks.
\newblock \emph{arXiv preprint arXiv:2602.03793}, 2026.

\bibitem[Chen et~al.(2025{\natexlab{g}})Chen, Zhao, Zhang, Liu, Qi, Wu, Kalluri, Cao, Xiong, Tong, Yao, Li, Zhu, Li, Song, Li, Weston, and Huynh]{chen2025dreamgym}
Z.~Chen, Z.~Zhao, K.~Zhang, B.~Liu, Q.~Qi, Y.~Wu, T.~Kalluri, S.~Cao, Y.~Xiong, H.~Tong, H.~Yao, H.~Li, J.~Zhu, X.~Li, D.~Song, B.~Li, J.~Weston, and D.~Huynh.
\newblock Scaling agent learning via experience synthesis.
\newblock \emph{arXiv preprint arXiv:2511.03773}, 2025{\natexlab{g}}.

\bibitem[Chitturi et~al.(2024)Chitturi, Ramdas, Wu, Rohr, Ermon, Dionne, Jornada, Dunne, Tassone, Neiswanger, and Ratner]{chitturi2024targeted}
S.~R. Chitturi, A.~Ramdas, Y.~Wu, B.~Rohr, S.~Ermon, J.~Dionne, F.~H.~d. Jornada, M.~Dunne, C.~Tassone, W.~Neiswanger, and D.~Ratner.
\newblock Targeted materials discovery using bayesian algorithm execution.
\newblock \emph{NPJ Computational Materials}, 10\penalty0 (1):\penalty0 156, 2024.

\bibitem[Chua et~al.(2018)Chua, Calandra, McAllister, and Levine]{chua2018pets}
K.~Chua, R.~Calandra, R.~McAllister, and S.~Levine.
\newblock Deep reinforcement learning in a handful of trials using probabilistic dynamics models.
\newblock In \emph{Advances in Neural Information Processing Systems}, volume~31, pages 4759--4770, 2018.

\bibitem[Chuang et~al.(2024)Chuang, Goyal, Harlalka, Suresh, Hawkins, Yang, Shah, Hu, and Rogers]{chuang2024opinion}
Y.-S. Chuang, A.~Goyal, N.~Harlalka, S.~Suresh, R.~Hawkins, S.~Yang, D.~Shah, J.~Hu, and T.~Rogers.
\newblock Simulating opinion dynamics with networks of llm-based agents.
\newblock In \emph{Findings of the association for computational linguistics: NAACL 2024}, pages 3326--3346, 2024.

\bibitem[Clark(2015)]{clark2015surfing}
A.~Clark.
\newblock \emph{Surfing Uncertainty: Prediction, Action, and the Embodied Mind}.
\newblock Oxford University Press, 2015.

\bibitem[Community(2026)]{starvla2025}
S.~Community.
\newblock Starvla: A lego-like codebase for vision-language-action model developing.
\newblock \emph{arXiv preprint arXiv:2604.05014}, 2026.

\bibitem[Copet et~al.(2025)Copet, Carbonneaux, Cohen, Gehring, Kahn, Kossen, Kreuk, McMilin, Meyer, Wei, Zhang, Zheng, Armengol-Estap{\'e}, Bashiri, Beck, et~al.]{copet2025cwm}
J.~Copet, Q.~Carbonneaux, G.~Cohen, J.~Gehring, J.~Kahn, J.~Kossen, F.~Kreuk, E.~McMilin, M.~Meyer, Y.~Wei, D.~Zhang, K.~Zheng, J.~Armengol-Estap{\'e}, P.~Bashiri, M.~Beck, et~al.
\newblock {CWM}: An open-weights {LLM} for research on code generation with world models.
\newblock \emph{arXiv preprint arXiv:2510.02387}, 2025.

\bibitem[Coutant et~al.(2019)Coutant, Roper, Trejo-Banos, Bouthinon, Carpenter, Grzebyta, Santini, Soldano, Elati, Ramon, Rouveirol, Soldatova, and King]{coutant2019yeast}
A.~Coutant, K.~Roper, D.~Trejo-Banos, D.~Bouthinon, M.~Carpenter, J.~Grzebyta, G.~Santini, H.~Soldano, M.~Elati, J.~Ramon, C.~Rouveirol, L.~N. Soldatova, and R.~D. King.
\newblock Closed-loop cycles of experiment design, execution, and learning accelerate systems biology model development in yeast.
\newblock \emph{Proceedings of the National Academy of Sciences}, 116\penalty0 (36):\penalty0 18142--18147, 2019.

\bibitem[Craik(1943)]{craik1943nature}
K.~J.~W. Craik.
\newblock \emph{The Nature of Explanation}.
\newblock Cambridge University Press, 1943.

\bibitem[Curvo(2025)]{curvo2025traitors}
P.~M. Curvo.
\newblock The traitors: Deception and trust in multi-agent language model simulations.
\newblock \emph{arXiv preprint arXiv:2505.12923}, 2025.

\bibitem[Dai et~al.(2024)Dai, Zhang, Li, Yang, lbe, Rao, Caetano, and Sra]{dai2024artificialleviathan}
G.~Dai, W.~Zhang, J.~Li, S.~Yang, C.~O. lbe, S.~Rao, A.~Caetano, and M.~Sra.
\newblock Artificial leviathan: Exploring social evolution of llm agents through the lens of hobbesian social contract theory.
\newblock \emph{arXiv preprint arXiv:2406.14373}, 2024.

\bibitem[Dainese et~al.(2024)Dainese, Merler, Alakuijala, and Marttinen]{dainese2024codewm}
N.~Dainese, M.~Merler, M.~Alakuijala, and P.~Marttinen.
\newblock Generating code world models with large language models guided by {Monte Carlo} tree search.
\newblock In \emph{Advances in Neural Information Processing Systems}, volume~37, pages 60429--60474, 2024.

\bibitem[Dama et~al.(2023)Dama, Kim, Leyva, Lunkes, Schmid, Jijakli, and Jensen]{dama2023bacterai}
A.~C. Dama, K.~S. Kim, D.~M. Leyva, A.~P. Lunkes, N.~S. Schmid, K.~Jijakli, and P.~A. Jensen.
\newblock {BacterAI} maps microbial metabolism without prior knowledge.
\newblock \emph{Nature Microbiology}, 8:\penalty0 1018--1025, 2023.

\bibitem[Decart et~al.(2024)Decart, Quevedo, McIntyre, Campbell, Chen, and Wachen]{oasis2024}
Decart, J.~Quevedo, Q.~McIntyre, S.~Campbell, X.~Chen, and R.~Wachen.
\newblock Oasis: A universe in a transformer.
\newblock Blog post, 2024.
\newblock URL \url{https://oasis-model.github.io}.

\bibitem[Degen(2023)]{degen2023rsa}
J.~Degen.
\newblock The rational speech act framework.
\newblock \emph{Annual Review of Linguistics}, 9\penalty0 (1):\penalty0 519--540, 2023.

\bibitem[Deisenroth and Rasmussen(2011)]{deisenroth2011pilco}
M.~P. Deisenroth and C.~E. Rasmussen.
\newblock {PILCO}: A model-based and data-efficient approach to policy search.
\newblock In \emph{International Conference on Machine Learning}, pages 465--472, 2011.

\bibitem[Deng et~al.(2023{\natexlab{a}})Deng, Zhong, Jun, Riebesell, Han, Bartel, and Ceder]{deng2023chgnet}
B.~Deng, P.~Zhong, K.~Jun, J.~Riebesell, K.~Han, C.~J. Bartel, and G.~Ceder.
\newblock Chgnet as a pretrained universal neural network potential for charge-informed atomistic modelling.
\newblock \emph{Nature Machine Intelligence}, 5\penalty0 (9):\penalty0 1031--1041, 2023{\natexlab{a}}.

\bibitem[Deng et~al.(2022)Deng, Jang, and Ahn]{deng2022dreamerpro}
F.~Deng, I.~Jang, and S.~Ahn.
\newblock {DreamerPro}: Reconstruction-free model-based reinforcement learning with prototypical representations.
\newblock In \emph{International Conference on Machine Learning}, pages 4956--4975. PMLR, 2022.

\bibitem[Deng et~al.(2023{\natexlab{b}})Deng, Gu, Zheng, Chen, Stevens, Wang, Sun, and Su]{deng2023mind2web}
X.~Deng, Y.~Gu, B.~Zheng, S.~Chen, S.~Stevens, B.~Wang, H.~Sun, and Y.~Su.
\newblock {Mind2Web}: Towards a generalist agent for the web.
\newblock In \emph{Advances in Neural Information Processing Systems}, volume~36, pages 28091--28114, 2023{\natexlab{b}}.

\bibitem[Dettmers et~al.(2022)Dettmers, Lewis, Belkada, and Zettlemoyer]{dettmers2022gpt3int8}
T.~Dettmers, M.~Lewis, Y.~Belkada, and L.~Zettlemoyer.
\newblock {LLM.int8()}: 8-bit matrix multiplication for transformers at scale.
\newblock In \emph{Advances in Neural Information Processing Systems}, volume~35, pages 30318--30332, 2022.

\bibitem[Dignum and Dignum(2025)]{dignum2026agentifying}
V.~Dignum and F.~Dignum.
\newblock Agentifying agentic {AI}.
\newblock \emph{arXiv preprint arXiv:2511.17332}, 2025.

\bibitem[Ding et~al.(2025{\natexlab{a}})Ding, Zhang, Shang, Feng, Zhang, Zong, Yuan, Su, Li, Piao, Deng, Sukiennik, Gao, Xu, and Li]{ding2024survey_wm}
J.~Ding, Y.~Zhang, Y.~Shang, J.~Feng, Y.~Zhang, Z.~Zong, Y.~Yuan, H.~Su, N.~Li, J.~Piao, Y.~Deng, N.~Sukiennik, C.~Gao, F.~Xu, and Y.~Li.
\newblock Understanding world or predicting future? {A} comprehensive survey of world models.
\newblock \emph{ACM Computing Surveys}, 2025{\natexlab{a}}.

\bibitem[Ding et~al.(2019)Ding, Ding, Guo, and Han]{ding2019centripetal}
X.~Ding, G.~Ding, Y.~Guo, and J.~Han.
\newblock Centripetal sgd for pruning very deep convolutional networks with complicated structure.
\newblock In \emph{IEEE/CVF Conference on Computer Vision and Pattern Recognition}, pages 4943--4953, 2019.

\bibitem[Ding et~al.(2025{\natexlab{b}})Ding, Jin, Liu, Zheng, Singh, Zhang, Kang, Lin, and Liu]{ding2025dollar}
Z.~Ding, C.~Jin, D.~Liu, H.~Zheng, K.~K. Singh, Q.~Zhang, Y.~Kang, Z.~Lin, and Y.~Liu.
\newblock Dollar: Few-step video generation via distillation and latent reward optimization.
\newblock In \emph{IEEE/CVF International Conference on Computer Vision}, pages 17961--17971, 2025{\natexlab{b}}.

\bibitem[Dockhorn et~al.(2022)Dockhorn, Vahdat, and Kreis]{dockhorn2022genie}
T.~Dockhorn, A.~Vahdat, and K.~Kreis.
\newblock Genie: Higher-order denoising diffusion solvers.
\newblock In \emph{Advances in Neural Information Processing Systems}, volume~35, pages 30150--30166, 2022.

\bibitem[Dong et~al.(2017)Dong, Chen, and Pan]{dong2017learning}
X.~Dong, S.~Chen, and S.~Pan.
\newblock Learning to prune deep neural networks via layer-wise optimal brain surgeon.
\newblock \emph{Advances in Neural Information Processing Systems}, 30:\penalty0 4860--4874, 2017.

\bibitem[Dong et~al.(2025)Dong, Wu, Chen, Kong, Zhu, Hu, Zhou, Sun, He, Dai, Hauptmann, and Cheng]{dong2026uniwm}
Y.~Dong, F.~Wu, G.~Chen, L.~Kong, X.~Zhu, Q.~Hu, Y.~Zhou, J.~Sun, J.-Y. He, Q.~Dai, A.~G. Hauptmann, and Z.-Q. Cheng.
\newblock Towards unified world models for visual navigation via memory-augmented planning and foresight.
\newblock \emph{arXiv preprint arXiv:2510.08713}, 2025.

\bibitem[Dong et~al.(2026)Dong, Wu, Dai, Kong, Chen, Zhu, Hu, Wang, Garnica, Liu, Huang, Dai, and Cheng]{dong2026lcvn}
Y.~Dong, F.~Wu, Y.~Dai, L.~Kong, G.~Chen, X.~Zhu, Q.~Hu, T.~Wang, J.~Garnica, F.~Liu, S.~Huang, Q.~Dai, and Z.-Q. Cheng.
\newblock Language-conditioned world modeling for visual navigation.
\newblock \emph{arXiv preprint arXiv:2603.26741}, 2026.

\bibitem[Duan et~al.(2026)Duan, Zhang, and Luo]{duan2026trap}
S.~Duan, K.~Zhang, and X.~Luo.
\newblock {TRAP}: Tail-aware ranking attack for world-model planning.
\newblock \emph{arXiv preprint arXiv:2605.01950}, 2026.

\bibitem[Duhem(1954)]{duhem1954aim}
P.~Duhem.
\newblock \emph{The Aim and Structure of Physical Theory}.
\newblock Princeton University Press, 1954.

\bibitem[Einstein(1936)]{einstein1936physics}
A.~Einstein.
\newblock Physics and reality.
\newblock \emph{Journal of the Franklin Institute}, 221\penalty0 (3):\penalty0 349--382, 1936.

\bibitem[Esteva et~al.(2001)Esteva, Rodriguez-Aguilar, Sierra, Garcia, and Arcos]{esteva2001electronic}
M.~Esteva, J.-A. Rodriguez-Aguilar, C.~Sierra, P.~Garcia, and J.~L. Arcos.
\newblock On the formal specification of electronic institutions.
\newblock In \emph{Agent Mediated Electronic Commerce: The European AgentLink Perspective}, pages 126--147. Springer, 2001.

\bibitem[Evangelou et~al.(2024)Evangelou, Cui, Bello-Rivas, Makeev, and Kevrekidis]{evangelou2024coevolving}
N.~Evangelou, T.~Cui, J.~M. Bello-Rivas, A.~Makeev, and I.~G. Kevrekidis.
\newblock Tipping points of evolving epidemiological networks: Machine learning-assisted, data-driven effective modeling.
\newblock \emph{Chaos: An Interdisciplinary Journal of Nonlinear Science}, 34\penalty0 (6), 2024.

\bibitem[Fan et~al.(2026)Fan, Zhang, Wang, Yang, Chan, Chen, Bi, Zhou, Liu, and Chen]{fan2026aivilization}
W.~Fan, S.~Zhang, X.~Wang, H.~Yang, T.~W. Chan, X.~Chen, J.~Bi, Z.~Zhou, J.~Liu, and K.~Chen.
\newblock {AIvilization v0}: Toward large-scale artificial social simulation with a unified agent architecture and adaptive agent profiles.
\newblock \emph{arXiv preprint arXiv:2602.10429}, 2026.

\bibitem[Fang et~al.(2025)Fang, Zhang, Zhang, Ma, Yu, Mi, and Yu]{webevolver2025}
T.~Fang, H.~Zhang, Z.~Zhang, K.~Ma, W.~Yu, H.~Mi, and D.~Yu.
\newblock {WebEvolver}: Enhancing web agent self-improvement with coevolving world model.
\newblock In \emph{Conference on Empirical Methods in Natural Language Processing}, pages 8959--8975, 2025.

\bibitem[Fei et~al.(2026)Fei, Rendy, Yang, Woo, Huang, Li, Wang, Milsted, Zeng, and Ceder]{fei2026agentic}
Y.~Fei, B.~Rendy, X.~Yang, J.~Woo, X.~Huang, C.~Li, S.~Wang, D.~Milsted, Y.~Zeng, and G.~Ceder.
\newblock Agentic llm reasoning in a self-driving laboratory for air-sensitive lithium halide spinel conductors.
\newblock \emph{arXiv preprint arXiv:2604.11957}, 2026.

\bibitem[Feng et~al.(2025{\natexlab{a}})Feng, Zhang, Zhang, Lu, Liu, and Wang]{feng2025wwm}
J.~Feng, Y.~Zhang, C.~Zhang, Y.~Lu, S.~Liu, and M.~Wang.
\newblock Web world models.
\newblock \emph{arXiv preprint arXiv:2512.23676}, 2025{\natexlab{a}}.

\bibitem[Feng et~al.(2025{\natexlab{b}})Feng, Li, Yang, Xi, Li, Li, Zhang, Yang, Peng, Han, Agrawala, Keutzer, Kodaira, and Xu]{feng2025streamdiffusionv2}
T.~Feng, Z.~Li, S.~Yang, H.~Xi, M.~Li, X.~Li, L.~Zhang, K.~Yang, K.~Peng, S.~Han, M.~Agrawala, K.~Keutzer, A.~Kodaira, and C.~Xu.
\newblock {StreamDiffusionV2}: A streaming system for dynamic and interactive video generation.
\newblock \emph{arXiv preprint arXiv:2511.07399}, 2025{\natexlab{b}}.

\bibitem[Feng et~al.(2025{\natexlab{c}})Feng, Wang, and Yang]{feng2025ad_wm_survey}
T.~Feng, W.~Wang, and Y.~Yang.
\newblock A survey of world models for autonomous driving.
\newblock \emph{arXiv preprint arXiv:2501.11260}, 2025{\natexlab{c}}.

\bibitem[Fikes and Nilsson(1971)]{fikes1971strips}
R.~E. Fikes and N.~J. Nilsson.
\newblock {STRIPS}: A new approach to the application of theorem proving to problem solving.
\newblock \emph{Artificial Intelligence}, 2\penalty0 (3--4):\penalty0 189--208, 1971.

\bibitem[Fish et~al.(2024)Fish, G{\"o}lz, Parkes, Procaccia, Rusak, Shapira, and W{\"u}thrich]{fish2023generativesocialchoice}
S.~Fish, P.~G{\"o}lz, D.~C. Parkes, A.~D. Procaccia, G.~Rusak, I.~Shapira, and M.~W{\"u}thrich.
\newblock Generative social choice.
\newblock In \emph{ACM Conference on Economics and Computation}, pages 985--985, 2024.

\bibitem[Frans et~al.(2025)Frans, Hafner, Levine, and Abbeel]{frans2024one}
K.~Frans, D.~Hafner, S.~Levine, and P.~Abbeel.
\newblock One step diffusion via shortcut models.
\newblock In \emph{International Conference on Learning Representations}, 2025.

\bibitem[Frantar et~al.(2023)Frantar, Ashkboos, Hoefler, and Alistarh]{frantar2023gptq}
E.~Frantar, S.~Ashkboos, T.~Hoefler, and D.~Alistarh.
\newblock {GPTQ}: Accurate post-training quantization for generative pre-trained transformers.
\newblock In \emph{International Conference on Learning Representations}, 2023.

\bibitem[Freeman et~al.(2021)Freeman, Frey, Raichuk, Girgin, Mordatch, and Bachem]{freeman2021brax}
C.~D. Freeman, E.~Frey, A.~Raichuk, S.~Girgin, I.~Mordatch, and O.~Bachem.
\newblock Brax -- a differentiable physics engine for large scale rigid body simulation.
\newblock In \emph{Advances in Neural Information Processing Systems}, 2021.

\bibitem[Friston(2010)]{friston2010free}
K.~Friston.
\newblock The free-energy principle: A unified brain theory?
\newblock \emph{Nature Reviews Neuroscience}, 11\penalty0 (2):\penalty0 127--138, 2010.

\bibitem[Friston et~al.(2017)Friston, FitzGerald, Rigoli, Schwartenbeck, and Pezzulo]{friston2017active}
K.~Friston, T.~FitzGerald, F.~Rigoli, P.~Schwartenbeck, and G.~Pezzulo.
\newblock Active inference: A process theory.
\newblock \emph{Neural Computation}, 29\penalty0 (1):\penalty0 1--49, 2017.

\bibitem[Ganapavarapu and Patel(2026)]{ganapavarapu2026mcpcosmos}
G.~Ganapavarapu and D.~Patel.
\newblock {MCP-Cosmos}: World model-augmented agents for complex task execution in {MCP} environments.
\newblock \emph{arXiv preprint arXiv:2605.09131}, 2026.

\bibitem[Gandhi et~al.(2023)Gandhi, Fr{\"a}nken, Gerstenberg, and Goodman]{gandhi2023bigtom}
K.~Gandhi, J.-P. Fr{\"a}nken, T.~Gerstenberg, and N.~Goodman.
\newblock Understanding social reasoning in language models with language models.
\newblock In \emph{Advances in Neural Information Processing Systems}, volume~36, pages 13518--13529, 2023.

\bibitem[Gao et~al.(2023)Gao, Lan, Lu, Mao, Piao, Wang, Jin, and Li]{gao2023s3}
C.~Gao, X.~Lan, Z.~Lu, J.~Mao, J.~Piao, H.~Wang, D.~Jin, and Y.~Li.
\newblock S3: Social-network simulation system with large language model-empowered agents.
\newblock \emph{arXiv preprint arXiv:2307.14984}, 2023.

\bibitem[Gao et~al.(2024)Gao, Yang, Chen, Chitta, Qiu, Geiger, Zhang, and Li]{gao2024vista}
S.~Gao, J.~Yang, L.~Chen, K.~Chitta, Y.~Qiu, A.~Geiger, J.~Zhang, and H.~Li.
\newblock Vista: A generalizable driving world model with high fidelity and versatile controllability.
\newblock In \emph{Advances in Neural Information Processing Systems}, volume~37, pages 91560--91596, 2024.

\bibitem[Gao et~al.(2026)Gao, Liang, Zheng, Malik, Ye, Yu, Tseng, Dong, Mo, Lin, Ma, Nah, Magne, Xiang, Xie, Zheng, Niu, Tan, Zentner, Kurian, Indupuru, Jannaty, Gu, Zhang, Malik, Abbeel, Liu, Zhu, Jang, and Fan]{gao2026dreamdojo}
S.~Gao, W.~Liang, K.~Zheng, A.~Malik, S.~Ye, S.~Yu, W.-C. Tseng, Y.~Dong, K.~Mo, C.-H. Lin, Q.~Ma, S.~Nah, L.~Magne, J.~Xiang, Y.~Xie, R.~Zheng, D.~Niu, Y.~L. Tan, K.~R. Zentner, G.~Kurian, S.~Indupuru, P.~Jannaty, J.~Gu, J.~Zhang, J.~Malik, P.~Abbeel, M.-Y. Liu, Y.~Zhu, J.~Jang, and L.~Fan.
\newblock {DreamDojo}: A generalist robot world model from large-scale human videos.
\newblock \emph{arXiv preprint arXiv:2602.06949}, 2026.

\bibitem[Gao et~al.(2025)Gao, Ye, Wang, and Sang]{gao2025websynthesis}
Y.~Gao, J.~Ye, J.~Wang, and J.~Sang.
\newblock {WebSynthesis}: World-model-guided {MCTS} for efficient {WebUI}-trajectory synthesis.
\newblock \emph{arXiv preprint arXiv:2507.04370}, 2025.

\bibitem[Ge et~al.(2024)Ge, Mahapatra, Parmar, Zhu, and Huang]{ge2024fvdbias}
S.~Ge, A.~Mahapatra, G.~Parmar, J.-Y. Zhu, and J.-B. Huang.
\newblock On the content bias in {Fr\'{e}chet} video distance.
\newblock In \emph{IEEE/CVF Conference on Computer Vision and Pattern Recognition}, pages 7277--7288, 2024.

\bibitem[Geirhos et~al.(2018)Geirhos, Rubisch, Michaelis, Bethge, Wichmann, and Brendel]{geirhos2018imagenet}
R.~Geirhos, P.~Rubisch, C.~Michaelis, M.~Bethge, F.~A. Wichmann, and W.~Brendel.
\newblock Imagenet-trained cnns are biased towards texture; increasing shape bias improves accuracy and robustness.
\newblock In \emph{International Conference on Learning Representations}, 2018.

\bibitem[Gelada et~al.(2019)Gelada, Kumar, Buckman, Nachum, and Bellemare]{gelada2019deepmdp}
C.~Gelada, S.~Kumar, J.~Buckman, O.~Nachum, and M.~G. Bellemare.
\newblock {DeepMDP}: Learning continuous latent space models for representation learning.
\newblock In \emph{International Conference on Machine Learning}, pages 2170--2179. PMLR, 2019.

\bibitem[{Genesis Authors}(2024)]{genesis2024}
{Genesis Authors}.
\newblock Genesis: A generative and universal physics engine for robotics and beyond.
\newblock {GitHub Repository}, 2024.
\newblock URL \url{https://github.com/Genesis-Embodied-AI/Genesis}.

\bibitem[Geng et~al.(2025{\natexlab{a}})Geng, Deng, Bai, Kolter, and He]{geng2025mean}
Z.~Geng, M.~Deng, X.~Bai, J.~Z. Kolter, and K.~He.
\newblock Mean flows for one-step generative modeling.
\newblock \emph{arXiv preprint arXiv:2505.13447}, 2025{\natexlab{a}}.

\bibitem[Geng et~al.(2025{\natexlab{b}})Geng, Pokle, Luo, Lin, and Kolter]{geng2024consistency}
Z.~Geng, A.~Pokle, W.~Luo, J.~Lin, and J.~Z. Kolter.
\newblock Consistency models made easy.
\newblock In \emph{International Conference on Learning Representations}, 2025{\natexlab{b}}.

\bibitem[Ghosh et~al.(2021)Ghosh, Gupta, Reddy, Fu, Devin, Eysenbach, and Levine]{ghosh2021learning}
D.~Ghosh, A.~Gupta, A.~Reddy, J.~Fu, C.~Devin, B.~Eysenbach, and S.~Levine.
\newblock Learning to reach goals via iterated supervised learning.
\newblock In \emph{International Conference on Learning Representations}, 2021.

\bibitem[Ghugare et~al.(2025)Ghugare, Castanyer, Ji, Wantlin, Schofield, Narasimhan, and Eysenbach]{ghugare2025builderbench}
R.~Ghugare, R.~C. Castanyer, C.~Ji, K.~Wantlin, J.~Schofield, K.~Narasimhan, and B.~Eysenbach.
\newblock Builderbench: The building blocks of intelligent agents.
\newblock \emph{arXiv preprint arXiv:2510.06288}, 2025.

\bibitem[Gohil et~al.(2022)Gohil, Roberts, Timms, Skates, Higgins, Quinn, Pervaiz, van Amersfoort, Notin, Gal, Adaszewski, and Woolrich]{gohil2022dynemo}
C.~Gohil, E.~Roberts, R.~Timms, A.~Skates, C.~Higgins, A.~Quinn, U.~Pervaiz, J.~van Amersfoort, P.~Notin, Y.~Gal, S.~Adaszewski, and M.~Woolrich.
\newblock Mixtures of large-scale dynamic functional brain network modes.
\newblock \emph{NeuroImage}, 263:\penalty0 119595, 2022.
\newblock ISSN 1053-8119.
\newblock \doi{10.1016/j.neuroimage.2022.119595}.

\bibitem[Goodfellow et~al.(2016)Goodfellow, Bengio, and Courville]{goodfellow2016deep}
I.~Goodfellow, Y.~Bengio, and A.~Courville.
\newblock \emph{Deep learning}, volume~1.
\newblock MIT Press, 2016.

\bibitem[Goodman and Frank(2016)]{goodman2016rsa}
N.~D. Goodman and M.~C. Frank.
\newblock Pragmatic language interpretation as probabilistic inference.
\newblock \emph{Trends in cognitive sciences}, 20\penalty0 (11):\penalty0 818--829, 2016.

\bibitem[Goswami et~al.(2023)Goswami, Bora, Yu, and Karniadakis]{goswami2022pideepone}
S.~Goswami, A.~Bora, Y.~Yu, and G.~E. Karniadakis.
\newblock Physics-informed deep neural operator networks.
\newblock In \emph{Machine Learning in Modeling and Simulation: Methods and Applications}, pages 219--254. Springer, 2023.

\bibitem[Gottweis et~al.(2025)Gottweis, Weng, Daryin, Tu, Palepu, Sirkovic, Myaskovsky, Weissenberger, Rong, Tanno, Saab, Popovici, Blum, Zhang, Chou, Hassidim, Gokturk, Vahdat, Kohli, Matias, Carroll, Kulkarni, Tomasev, Guan, Dhillon, Vaishnav, Lee, Costa, Penad{\'e}s, Peltz, Xu, Pawlosky, Karthikesalingam, and Natarajan]{gottweis2025coscientist}
J.~Gottweis, W.-H. Weng, A.~Daryin, T.~Tu, A.~Palepu, P.~Sirkovic, A.~Myaskovsky, F.~Weissenberger, K.~Rong, R.~Tanno, K.~Saab, D.~Popovici, J.~Blum, F.~Zhang, K.~Chou, A.~Hassidim, B.~Gokturk, A.~Vahdat, P.~Kohli, Y.~Matias, A.~Carroll, K.~Kulkarni, N.~Tomasev, Y.~Guan, V.~Dhillon, E.~D. Vaishnav, B.~Lee, T.~R.~D. Costa, J.~R. Penad{\'e}s, G.~Peltz, Y.~Xu, A.~Pawlosky, A.~Karthikesalingam, and V.~Natarajan.
\newblock Towards an {AI} co-scientist.
\newblock \emph{arXiv preprint arXiv:2502.18864}, 2025.

\bibitem[Gu et~al.(2025{\natexlab{a}})Gu, Liu, Zeng, Nagarajan, Zhu, Hong, Fan, Yan, Zhou, Liu, and Wang]{gu2025phyworldbench}
J.~Gu, X.~Liu, Y.~Zeng, A.~Nagarajan, F.~Zhu, D.~Hong, Y.~Fan, Q.~Yan, K.~Zhou, M.-Y. Liu, and X.~E. Wang.
\newblock {PhyWorldBench}: A comprehensive evaluation of physical realism in text-to-video models.
\newblock \emph{arXiv preprint arXiv:2507.13428}, 2025{\natexlab{a}}.

\bibitem[Gu et~al.(2025{\natexlab{b}})Gu, Zhang, Ning, Zheng, Gou, Xue, Chang, Srivastava, Xie, Qi, Sun, and Su]{gu2024webdreamer}
Y.~Gu, K.~Zhang, Y.~Ning, B.~Zheng, B.~Gou, T.~Xue, C.~Chang, S.~Srivastava, Y.~Xie, P.~Qi, H.~Sun, and Y.~Su.
\newblock Is your {LLM} secretly a world model of the internet? {M}odel-based planning for web agents.
\newblock \emph{Transactions on Machine Learning Research}, 2025{\natexlab{b}}.

\bibitem[Guan et~al.(2026)Guan, Yu, Zhang, Wang, Zhang, Li, Qiao, Qin, Huang, Yang, Zhao, Wutschitz, Kessler, Inan, Sim, Rajmohan, Lin, and Zhang]{cuwm2026}
Y.~Guan, R.~Yu, J.~Zhang, L.~Wang, C.~Zhang, L.~Li, B.~Qiao, S.~Qin, H.~Huang, F.~Yang, P.~Zhao, L.~Wutschitz, S.~Kessler, H.~A. Inan, R.~Sim, S.~Rajmohan, Q.~Lin, and D.~Zhang.
\newblock Computer-using world model.
\newblock \emph{arXiv preprint arXiv:2602.17365}, 2026.

\bibitem[Guo et~al.(2025{\natexlab{a}})Guo, Ye, He, Wu, Jiang, Pearce, and Bian]{guo2025mineworld}
J.~Guo, Y.~Ye, T.~He, H.~Wu, Y.~Jiang, T.~Pearce, and J.~Bian.
\newblock {MineWorld}: a real-time and open-source interactive world model on minecraft.
\newblock \emph{arXiv preprint arXiv:2504.08388}, 2025{\natexlab{a}}.

\bibitem[Guo et~al.(2025{\natexlab{b}})Guo, Darwiche~Domingues, Avalos, Courville, and Strub]{guo2025dymo}
S.~Guo, O.~Darwiche~Domingues, R.~Avalos, A.~Courville, and F.~Strub.
\newblock World modelling improves language model agents.
\newblock \emph{arXiv preprint arXiv:2506.02918}, 2025{\natexlab{b}}.

\bibitem[Guo et~al.(2025{\natexlab{c}})Guo, Chen, Wang, He, Xu, Ye, Sun, and Xiong]{guo2025logic}
W.~Guo, Z.~Chen, S.~Wang, J.~He, Y.~Xu, J.~Ye, Y.~Sun, and H.~Xiong.
\newblock Logic-in-frames: Dynamic keyframe search via visual semantic-logical verification for long video understanding.
\newblock \emph{arXiv preprint arXiv:2503.13139}, 2025{\natexlab{c}}.

\bibitem[Guo et~al.(2026{\natexlab{a}})Guo, Liang, Balogh, Lunberry, Tu, Jelasity, and Tao]{guo2026physcond}
Z.~Guo, S.~Liang, A.~Balogh, N.~Lunberry, R.-C. Tu, M.~Jelasity, and D.~Tao.
\newblock When world models dream wrong: Physical-conditioned adversarial attacks against world models.
\newblock \emph{arXiv preprint arXiv:2602.18739}, 2026{\natexlab{a}}.

\bibitem[Guo et~al.(2026{\natexlab{b}})Guo, Liang, Fu, Guo, Balogh, Jelasity, and Tao]{guo2026wmattack}
Z.~Guo, S.~Liang, S.~Fu, C.~Guo, A.~Balogh, M.~Jelasity, and D.~Tao.
\newblock {WMAttack}: Automated attack search for adversarial evaluation of world-model agents.
\newblock \emph{arXiv preprint arXiv:2605.23220}, 2026{\natexlab{b}}.

\bibitem[Ha and Schmidhuber(2018)]{ha2018worldmodels}
D.~Ha and J.~Schmidhuber.
\newblock Recurrent world models facilitate policy evolution.
\newblock In \emph{Advances in Neural Information Processing Systems}, volume~31, pages 2455--2467, 2018.

\bibitem[Hafner et~al.(2019)Hafner, Lillicrap, Fischer, Villegas, Ha, Lee, and Davidson]{hafner2019recurrent}
D.~Hafner, T.~Lillicrap, I.~Fischer, R.~Villegas, D.~Ha, H.~Lee, and J.~Davidson.
\newblock Learning latent dynamics for planning from pixels.
\newblock In \emph{International Conference on Machine Learning}, pages 2555--2565. PMLR, 2019.

\bibitem[Hafner et~al.(2020)Hafner, Lillicrap, Ba, and Norouzi]{hafner2019dreamer}
D.~Hafner, T.~Lillicrap, J.~Ba, and M.~Norouzi.
\newblock Dream to control: Learning behaviors by latent imagination.
\newblock In \emph{International Conference on Learning Representations}, 2020.

\bibitem[Hafner et~al.(2021)Hafner, Lillicrap, Norouzi, and Ba]{hafner2020dreamerv2}
D.~Hafner, T.~Lillicrap, M.~Norouzi, and J.~Ba.
\newblock Mastering {A}tari with discrete world models.
\newblock In \emph{International Conference on Learning Representations}, 2021.

\bibitem[Hafner et~al.(2025)Hafner, Pasukonis, Ba, and Lillicrap]{hafner2023dreamerv3}
D.~Hafner, J.~Pasukonis, J.~Ba, and T.~Lillicrap.
\newblock Mastering diverse control tasks through world models.
\newblock \emph{Nature}, 640\penalty0 (8059):\penalty0 647--653, 2025.

\bibitem[Hansen et~al.(2022)Hansen, Su, and Wang]{hansen2022tdmpc}
N.~Hansen, H.~Su, and X.~Wang.
\newblock Temporal difference learning for model predictive control.
\newblock In \emph{International Conference on Machine Learning}, pages 8387--8406. PMLR, 2022.

\bibitem[Hansen et~al.(2024)Hansen, Su, and Wang]{hansen2024tdmpc2}
N.~Hansen, H.~Su, and X.~Wang.
\newblock {TD-MPC2}: Scalable, robust world models for continuous control.
\newblock In \emph{International Conference on Learning Representations}, 2024.

\bibitem[Hao et~al.(2025)Hao, Lu, Xu, and Chen]{hao2025mosim}
C.~Hao, W.~Lu, Y.~Xu, and Y.~Chen.
\newblock Neural motion simulator: Pushing the limit of world models in reinforcement learning.
\newblock In \emph{IEEE/CVF Conference on Computer Vision and Pattern Recognition}, pages 27608--27617, 2025.

\bibitem[Hao et~al.(2023)Hao, Gu, Ma, Hong, Wang, Wang, and Hu]{hao2023reasoning}
S.~Hao, Y.~Gu, H.~Ma, J.~J. Hong, Z.~Wang, D.~Z. Wang, and Z.~Hu.
\newblock Reasoning with language model is planning with world model.
\newblock In \emph{Conference on Empirical Methods in Natural Language Processing}, pages 8154--8173, 2023.

\bibitem[He et~al.(2018)He, Chen, Balakrishnan, and Liang]{he2018craigslist}
H.~He, D.~Chen, A.~Balakrishnan, and P.~Liang.
\newblock Decoupling strategy and generation in negotiation dialogues.
\newblock In \emph{Conference on Empirical Methods in Natural Language Processing}, pages 2333--2343, 2018.

\bibitem[He et~al.(2020)He, Fan, Wu, Xie, and Girshick]{he2020moco}
K.~He, H.~Fan, Y.~Wu, S.~Xie, and R.~Girshick.
\newblock Momentum contrast for unsupervised visual representation learning.
\newblock In \emph{IEEE/CVF Conference on Computer Vision and Pattern Recognition}, pages 9729--9738, 2020.

\bibitem[He et~al.(2023)He, Liu, Wu, Zhou, and Zhuang]{he2023efficientdm}
Y.~He, J.~Liu, W.~Wu, H.~Zhou, and B.~Zhuang.
\newblock Efficientdm: Efficient quantization-aware fine-tuning of low-bit diffusion models.
\newblock \emph{arXiv preprint arXiv:2310.03270}, 2023.

\bibitem[Heek et~al.(2024)Heek, Hoogeboom, and Salimans]{heek2024multistep}
J.~Heek, E.~Hoogeboom, and T.~Salimans.
\newblock Multistep consistency models.
\newblock \emph{arXiv preprint arXiv:2403.06807}, 2024.

\bibitem[Helfrich et~al.(2014)Helfrich, Schneider, Rach, Trautmann-Lengsfeld, Engel, and Herrmann]{helfrich2014entrainment}
R.~F. Helfrich, T.~R. Schneider, S.~Rach, S.~A. Trautmann-Lengsfeld, A.~K. Engel, and C.~S. Herrmann.
\newblock Entrainment of brain oscillations by transcranial alternating current stimulation.
\newblock \emph{Current biology}, 24\penalty0 (3):\penalty0 333--339, 2014.

\bibitem[Henderson et~al.(2018)Henderson, Islam, Bachman, Pineau, Precup, and Meger]{henderson2018deep}
P.~Henderson, R.~Islam, P.~Bachman, J.~Pineau, D.~Precup, and D.~Meger.
\newblock Deep reinforcement learning that matters.
\newblock In \emph{AAAI Conference on Artificial Intelligence}, volume~32, pages 3207--3214, 2018.

\bibitem[Higgins et~al.(2017)Higgins, Matthey, Pal, Burgess, Glorot, Botvinick, Mohamed, and Lerchner]{higgins2017betavae}
I.~Higgins, L.~Matthey, A.~Pal, C.~Burgess, X.~Glorot, M.~Botvinick, S.~Mohamed, and A.~Lerchner.
\newblock $\beta$-{VAE}: Learning basic visual concepts with a constrained variational framework.
\newblock In \emph{International Conference on Learning Representations}, 2017.

\bibitem[Ho et~al.(2020)Ho, Jain, and Abbeel]{ho2020ddpm}
J.~Ho, A.~Jain, and P.~Abbeel.
\newblock Denoising diffusion probabilistic models.
\newblock In \emph{Advances in Neural Information Processing Systems}, volume~33, pages 6840--6851, 2020.

\bibitem[Ho et~al.(2022)Ho, Salimans, Gritsenko, Chan, Norouzi, and Fleet]{ho2022videodiffusion}
J.~Ho, T.~Salimans, A.~Gritsenko, W.~Chan, M.~Norouzi, and D.~J. Fleet.
\newblock Video diffusion models.
\newblock In \emph{Advances in Neural Information Processing Systems}, volume~35, pages 8633--8646, 2022.

\bibitem[Hong et~al.(2024)Hong, Zhuge, Chen, Zheng, Cheng, Zhang, Wang, Wang, Yau, Lin, Zhou, Ran, Xiao, Wu, and Schmidhuber]{hong2024metagpt}
S.~Hong, M.~Zhuge, J.~P. Chen, X.~Zheng, Y.~Cheng, C.~Zhang, J.~Wang, Z.~Wang, S.~K.~S. Yau, Z.~Lin, L.~Zhou, C.~Ran, L.~Xiao, C.~Wu, and J.~Schmidhuber.
\newblock {MetaGPT}: Meta programming for a multi-agent collaborative framework.
\newblock In \emph{International Conference on Learning Representations}, 2024.

\bibitem[Hooker(2021)]{hooker2021hardware}
S.~Hooker.
\newblock The hardware lottery.
\newblock \emph{Communications of the ACM}, 64\penalty0 (12):\penalty0 58--65, 2021.

\bibitem[Hooper et~al.(2024)Hooper, Kim, Mohammadzadeh, Mahoney, Shao, Keutzer, and Gholami]{hooper2024kvquant}
C.~Hooper, S.~Kim, H.~Mohammadzadeh, M.~W. Mahoney, Y.~S. Shao, K.~Keutzer, and A.~Gholami.
\newblock {KVQuant}: Towards 10 million context length llm inference with kv cache quantization.
\newblock In \emph{Advances in Neural Information Processing Systems}, volume~37, pages 1270--1303, 2024.

\bibitem[Hu et~al.(2023)Hu, Russell, Yeo, Murez, Fedoseev, Kendall, Shotton, and Corrado]{hu2023gaia1}
A.~Hu, L.~Russell, H.~Yeo, Z.~Murez, G.~Fedoseev, A.~Kendall, J.~Shotton, and G.~Corrado.
\newblock {GAIA-1}: A generative world model for autonomous driving.
\newblock \emph{arXiv preprint arXiv:2309.17080}, 2023.

\bibitem[Hu et~al.(2025{\natexlab{a}})Hu, Liu, Wang, Zhu, Liang, Kong, Zhao, Gong, Cen, Huang, Hao, Li, Song, Li, Ma, Shen, Zhu, Tao, Liu, and Liang]{survey_vla4ad}
T.~Hu, X.~Liu, S.~Wang, Y.~Zhu, A.~Liang, L.~Kong, G.~Zhao, Z.~Gong, J.~Cen, Z.~Huang, X.~Hao, L.~Li, H.~Song, X.~Li, J.~Ma, S.~Shen, J.~Zhu, D.~Tao, Z.~Liu, and J.~Liang.
\newblock Vision-language-action models for autonomous driving: Past, present, and future.
\newblock \emph{arXiv preprint arXiv:2512.16760}, 2025{\natexlab{a}}.

\bibitem[Hu et~al.(2025{\natexlab{b}})Hu, Chen, and Lipson]{hu2025selfmodeling}
Y.~Hu, B.~Chen, and H.~Lipson.
\newblock Egocentric visual self-modeling for autonomous robot dynamics prediction and adaptation.
\newblock \emph{NPJ Robotics}, 3:\penalty0 14, 2025{\natexlab{b}}.

\bibitem[Hua et~al.(2024)Hua, Liu, Li, Amayuelas, Chen, Jiang, Jin, Fan, Sun, Wang, Wang, and Zhang]{hua2024gametheoreticllm}
W.~Hua, O.~Liu, L.~Li, A.~Amayuelas, J.~Chen, L.~Jiang, M.~Jin, L.~Fan, F.~Sun, W.~Wang, X.~Wang, and Y.~Zhang.
\newblock Game-theoretic {LLM}: Agent workflow for negotiation games.
\newblock \emph{arXiv preprint arXiv:2411.05990}, 2024.

\bibitem[Huang et~al.(2020)Huang, Zhang, Zhang, Ramsey, Sanchez-Romero, Glymour, and Sch{\"o}lkopf]{huang2020cdnod}
B.~Huang, K.~Zhang, J.~Zhang, J.~Ramsey, R.~Sanchez-Romero, C.~Glymour, and B.~Sch{\"o}lkopf.
\newblock Causal discovery from heterogeneous/nonstationary data.
\newblock \emph{Journal of Machine Learning Research}, 21\penalty0 (89):\penalty0 1--53, 2020.

\bibitem[Huang et~al.(2025{\natexlab{a}})Huang, Zhang, Wang, Qu, Lu, Roohani, Li, Qiu, Li, Zhang, Yin, Marwaha, Carter, Zhou, Wheeler, Bernstein, Wang, He, Zhou, Snyder, Cong, Regev, and Leskovec]{huang2025biomni}
K.~Huang, S.~Zhang, H.~Wang, Y.~Qu, Y.~Lu, Y.~Roohani, R.~Li, L.~Qiu, G.~Li, J.~Zhang, D.~Yin, S.~Marwaha, J.~N. Carter, X.~Zhou, M.~Wheeler, J.~A. Bernstein, M.~Wang, P.~He, J.~Zhou, M.~Snyder, L.~Cong, A.~Regev, and J.~Leskovec.
\newblock Biomni: A general-purpose biomedical {AI} agent.
\newblock bioRxiv 2025.05.30.656746, 2025{\natexlab{a}}.

\bibitem[Huang et~al.(2026{\natexlab{a}})Huang, Tang, Xu, Cao, Tu, Guo, Zheng, Liu, and Yang]{huang2026policysim}
R.~Huang, N.~Tang, J.~Xu, Y.~Cao, Q.~Tu, S.~Guo, B.~Zheng, H.~Liu, and Y.~Yang.
\newblock {PolicySim}: An {LLM}-based agent social simulation sandbox for proactive policy optimization.
\newblock \emph{arXiv preprint arXiv:2603.19649}, 2026{\natexlab{a}}.

\bibitem[Huang et~al.(2025{\natexlab{b}})Huang, Chen, Zhang, Sun, and Schwager]{huang2025particleformer}
S.~Huang, Q.~Chen, X.~Zhang, J.~Sun, and M.~Schwager.
\newblock {ParticleFormer}: A {3D} point cloud world model for multi-object, multi-material robotic manipulation.
\newblock \emph{arXiv preprint arXiv:2506.23126}, 2025{\natexlab{b}}.

\bibitem[Huang et~al.(2024{\natexlab{a}})Huang, Liao, Liu, He, Tan, Zhang, Li, Liu, and Qi]{huang2024mixture}
W.~Huang, Y.~Liao, J.~Liu, R.~He, H.~Tan, S.~Zhang, H.~Li, S.~Liu, and X.~Qi.
\newblock Mixture compressor for mixture-of-experts {LLMs} gains more.
\newblock \emph{arXiv preprint arXiv:2410.06270}, 2024{\natexlab{a}}.

\bibitem[Huang et~al.(2024{\natexlab{b}})Huang, Liu, Qin, Li, Zhang, Liu, Magno, and Qi]{huang2024billm}
W.~Huang, Y.~Liu, H.~Qin, Y.~Li, S.~Zhang, X.~Liu, M.~Magno, and X.~Qi.
\newblock {BiLLM}: Pushing the limit of post-training quantization for {LLMs}.
\newblock \emph{arXiv preprint arXiv:2402.04291}, 2024{\natexlab{b}}.

\bibitem[Huang et~al.(2026{\natexlab{b}})Huang, Chao, Mousavian, Liu, Fox, Mo, and Fei-Fei]{huang2026pointworld}
W.~Huang, Y.-W. Chao, A.~Mousavian, M.-Y. Liu, D.~Fox, K.~Mo, and L.~Fei-Fei.
\newblock {PointWorld}: Scaling {3D} world models for in-the-wild robotic manipulation.
\newblock \emph{arXiv preprint arXiv:2601.03782}, 2026{\natexlab{b}}.

\bibitem[Huang et~al.(2026{\natexlab{c}})Huang, Liao, Chen, Liu, Tan, Liu, Zhang, Yan, and Qi]{huang2026mcsharp}
W.~Huang, Y.~Liao, Y.~Chen, J.~Liu, H.~Tan, S.~Liu, S.~Zhang, S.~Yan, and X.~Qi.
\newblock Mc\#: Mixture compressor for mixture-of-experts large models.
\newblock \emph{IEEE Transactions on Pattern Analysis and Machine Intelligence}, pages 1--16, 2026{\natexlab{c}}.

\bibitem[Huang et~al.(2024{\natexlab{c}})Huang, Liu, Chen, Wang, Wang, Lian, Wang, Tang, and Chen]{huang2024planningsurvey}
X.~Huang, W.~Liu, X.~Chen, X.~Wang, H.~Wang, D.~Lian, Y.~Wang, R.~Tang, and E.~Chen.
\newblock Understanding the planning of {LLM} agents: A survey.
\newblock \emph{arXiv preprint arXiv:2402.02716}, 2024{\natexlab{c}}.

\bibitem[Huang et~al.(2025{\natexlab{c}})Huang, Chen, Fei, Li, Schwaller, and Ceder]{huang2025cascade}
X.~Huang, J.~Chen, Y.~Fei, Z.~Li, P.~Schwaller, and G.~Ceder.
\newblock Cascade: Cumulative agentic skill creation through autonomous development and evolution.
\newblock \emph{arXiv preprint arXiv:2512.23880}, 2025{\natexlab{c}}.

\bibitem[Huang et~al.(2025{\natexlab{d}})Huang, Deng, Zhong, Kaplan, Persson, and Ceder]{huang2025cross}
X.~Huang, B.~Deng, P.~Zhong, A.~D. Kaplan, K.~A. Persson, and G.~Ceder.
\newblock Cross-functional transferability in foundation machine learning interatomic potentials.
\newblock \emph{npj Computational Materials}, 11\penalty0 (1):\penalty0 313, 2025{\natexlab{d}}.

\bibitem[Huang et~al.(2025{\natexlab{e}})Huang, Li, He, Zhou, and Shechtman]{huang2025selfforcing}
X.~Huang, Z.~Li, G.~He, M.~Zhou, and E.~Shechtman.
\newblock Self forcing: Bridging the train-test gap in autoregressive video diffusion.
\newblock \emph{arXiv preprint arXiv:2506.08009}, 2025{\natexlab{e}}.

\bibitem[Huang et~al.(2024{\natexlab{d}})Huang, He, Yu, Zhang, Si, Jiang, Zhang, Wu, Jin, Chanpaisit, Wang, Chen, Wang, Lin, Qiao, and Liu]{huang2023vbench}
Z.~Huang, Y.~He, J.~Yu, F.~Zhang, C.~Si, Y.~Jiang, Y.~Zhang, T.~Wu, Q.~Jin, N.~Chanpaisit, Y.~Wang, X.~Chen, L.~Wang, D.~Lin, Y.~Qiao, and Z.~Liu.
\newblock {VBench}: Comprehensive benchmark suite for video generative models.
\newblock In \emph{IEEE/CVF Conference on Computer Vision and Pattern Recognition}, pages 21807--21818, 2024{\natexlab{d}}.

\bibitem[Huang et~al.(2025{\natexlab{f}})Huang, Yu, Chen, Qiu, Debevec, and Liu]{huang2025vchain}
Z.~Huang, N.~Yu, G.~Chen, H.~Qiu, P.~Debevec, and Z.~Liu.
\newblock {VChain}: Chain-of-visual-thought for reasoning in video generation.
\newblock \emph{arXiv preprint arXiv:2510.05094}, 2025{\natexlab{f}}.

\bibitem[Huang et~al.(2025{\natexlab{g}})Huang, Zhang, Xu, He, Yu, Dong, Ma, Chanpaisit, Si, Jiang, Wang, Chen, Chen, Wang, Lin, Qiao, and Liu]{huang2025vbench++}
Z.~Huang, F.~Zhang, X.~Xu, Y.~He, J.~Yu, Z.~Dong, Q.~Ma, N.~Chanpaisit, C.~Si, Y.~Jiang, Y.~Wang, X.~Chen, Y.-C. Chen, L.~Wang, D.~Lin, Y.~Qiao, and Z.~Liu.
\newblock {VBench++}: Comprehensive and versatile benchmark suite for video generative models.
\newblock \emph{IEEE Transactions on Pattern Analysis and Machine Intelligence}, 48\penalty0 (3):\penalty0 3268--3285, 2025{\natexlab{g}}.

\bibitem[Hume(1739)]{hume1739treatise}
D.~Hume.
\newblock \emph{A Treatise of Human Nature}.
\newblock Clarendon Press, Oxford, 1739.

\bibitem[Jacob et~al.(2024)Jacob, Shen, Farina, and Andreas]{jacob2024consensus}
A.~P. Jacob, Y.~Shen, G.~Farina, and J.~Andreas.
\newblock The consensus game: Language model generation via equilibrium search.
\newblock In \emph{International Conference on Learning Representations}, 2024.

\bibitem[James et~al.(2020)James, Ma, Arrojo, and Davison]{james2020rlbench}
S.~James, Z.~Ma, D.~R. Arrojo, and A.~J. Davison.
\newblock {RLBench}: The robot learning benchmark \& learning environment.
\newblock \emph{IEEE Robotics and Automation Letters}, 5\penalty0 (2):\penalty0 3019--3026, 2020.

\bibitem[Janner et~al.(2019)Janner, Fu, Zhang, and Levine]{janner2019mbpo}
M.~Janner, J.~Fu, M.~Zhang, and S.~Levine.
\newblock When to trust your model: Model-based policy optimization.
\newblock In \emph{Advances in Neural Information Processing Systems}, volume~32, pages 12519--12530, 2019.

\bibitem[Janner et~al.(2022)Janner, Du, Tenenbaum, and Levine]{janner2022diffuser}
M.~Janner, Y.~Du, J.~B. Tenenbaum, and S.~Levine.
\newblock Planning with diffusion for flexible behavior synthesis.
\newblock In \emph{International Conference on Machine Learning}, pages 9902--9915. PMLR, 2022.

\bibitem[Jimenez et~al.(2024)Jimenez, Yang, Wettig, Yao, Pei, Press, and Narasimhan]{jimenez2023swebench}
C.~E. Jimenez, J.~Yang, A.~Wettig, S.~Yao, K.~Pei, O.~Press, and K.~Narasimhan.
\newblock {SWE-Bench}: Can language models resolve real-world github issues?
\newblock In \emph{International Conference on Learning Representations}, 2024.

\bibitem[Jin et~al.(2025)Jin, Guo, Qu, Yang, Shang, Yang, Chao, Zhou, Xu, Xu, Zhou, Zhang, Wang, Zhang, and Cong]{jin2025biolab}
R.~Jin, Y.~Guo, Y.~Qu, M.~Yang, C.~Shang, Q.~Yang, L.~Chao, Y.~Zhou, R.~Xu, Z.~Xu, R.~Zhou, Z.~Zhang, M.~Wang, X.~Zhang, and L.~Cong.
\newblock {BioLab}: End-to-end autonomous life sciences research with multi-agents system integrating biological foundation models.
\newblock bioRxiv 2025.09.03.674085, 2025.

\bibitem[Jing et~al.(2026)Jing, Hao, Zhou, and Yu]{jing2026counterscene}
B.~Jing, R.~Hao, W.~Zhou, and H.~Yu.
\newblock {CounterScene}: Counterfactual causal reasoning in generative world models for safety-critical closed-loop evaluation.
\newblock \emph{arXiv preprint arXiv:2603.21104}, 2026.

\bibitem[Johnson-Laird(1983)]{johnson1983mental}
P.~N. Johnson-Laird.
\newblock \emph{Mental Models: Towards a Cognitive Science of Language, Inference, and Consciousness}.
\newblock Harvard University Press, 1983.

\bibitem[Jumper et~al.(2021)Jumper, Evans, Pritzel, Green, Figurnov, Ronneberger, Tunyasuvunakool, Bates, {\v{Z}}{\'\i}dek, Potapenko, Bridgland, Meyer, Kohl, Ballard, Cowie, Romera-Paredes, Nikolov, Jain, Adler, Back, Petersen, Reiman, Clancy, Zielinski, Steinegger, Pacholska, Berghammer, Bodenstein, Silver, Vinyals, Senior, Kavukcuoglu, Kohli, and Hassabis]{jumper2021alphafold}
J.~Jumper, R.~Evans, A.~Pritzel, T.~Green, M.~Figurnov, O.~Ronneberger, K.~Tunyasuvunakool, R.~Bates, A.~{\v{Z}}{\'\i}dek, A.~Potapenko, A.~Bridgland, C.~Meyer, S.~A.~A. Kohl, A.~J. Ballard, A.~Cowie, B.~Romera-Paredes, S.~Nikolov, R.~Jain, J.~Adler, T.~Back, S.~Petersen, D.~Reiman, E.~Clancy, M.~Zielinski, M.~Steinegger, M.~Pacholska, T.~Berghammer, S.~Bodenstein, D.~Silver, O.~Vinyals, A.~W. Senior, K.~Kavukcuoglu, P.~Kohli, and D.~Hassabis.
\newblock Highly accurate protein structure prediction with {AlphaFold}.
\newblock \emph{Nature}, 596:\penalty0 583--589, 2021.

\bibitem[Kaelbling et~al.(1998)Kaelbling, Littman, and Cassandra]{kaelbling1998pomdp}
L.~P. Kaelbling, M.~L. Littman, and A.~R. Cassandra.
\newblock Planning and acting in partially observable stochastic domains.
\newblock \emph{Artificial Intelligence}, 101\penalty0 (1--2):\penalty0 99--134, 1998.

\bibitem[Kaiser et~al.(2020)Kaiser, Babaeizadeh, Milos, Osinski, Campbell, Czechowski, Erhan, Finn, Kozakowski, Levine, Mohiuddin, Sepassi, Tucker, and Michalewski]{kaiser2019simple}
L.~Kaiser, M.~Babaeizadeh, P.~Milos, B.~Osinski, R.~H. Campbell, K.~Czechowski, D.~Erhan, C.~Finn, P.~Kozakowski, S.~Levine, A.~Mohiuddin, R.~Sepassi, G.~Tucker, and H.~Michalewski.
\newblock Model-based reinforcement learning for {Atari}.
\newblock In \emph{International Conference on Learning Representations}, 2020.

\bibitem[Kang et~al.(2025{\natexlab{a}})Kang, Yue, Lu, Lin, Zhao, Wang, Huang, and Feng]{kang2025howfar}
B.~Kang, Y.~Yue, R.~Lu, Z.~Lin, Y.~Zhao, K.~Wang, G.~Huang, and J.~Feng.
\newblock How far is video generation from world model: A physical law perspective.
\newblock In \emph{International Conference on Machine Learning}, pages 28991--29017. PMLR, 2025{\natexlab{a}}.

\bibitem[Kang et~al.(2025{\natexlab{b}})Kang, Gong, Yan, Xia, Wang, Wang, Ding, Cheng, Cao, Feng, He, Yan, Chen, He, Jiang, Ye, Yu, and Li]{kang2025hssbench}
Z.~Kang, J.~Gong, J.~Yan, W.~Xia, Y.~Wang, Z.~Wang, H.~Ding, Z.~Cheng, W.~Cao, Z.~Feng, S.~He, S.~Yan, J.~Chen, X.~He, C.~Jiang, W.~Ye, K.~Yu, and X.~Li.
\newblock {HSSBench}: Benchmarking humanities and social sciences ability for multimodal large language models.
\newblock \emph{arXiv preprint arXiv:2506.03922}, 2025{\natexlab{b}}.

\bibitem[Kant(1781)]{kant1781critique}
I.~Kant.
\newblock \emph{Critique of Pure Reason}.
\newblock Cambridge University Press, 1781.

\bibitem[Kara et~al.(2025)Kara, Faisal, and Nath]{xu2025warex}
S.~Kara, F.~E. Faisal, and S.~Nath.
\newblock {WAREX}: Web agent reliability evaluation on existing benchmarks.
\newblock \emph{arXiv preprint arXiv:2510.03285}, 2025.

\bibitem[Karniadakis et~al.(2021)Karniadakis, Kevrekidis, Lu, Perdikaris, Wang, and Yang]{karniadakis2021pinn}
G.~E. Karniadakis, I.~G. Kevrekidis, L.~Lu, P.~Perdikaris, S.~Wang, and L.~Yang.
\newblock Physics-informed machine learning.
\newblock \emph{Nature Reviews Physics}, 3\penalty0 (6):\penalty0 422--440, 2021.

\bibitem[Karras et~al.(2022)Karras, Aittala, Aila, and Laine]{karras2022elucidating}
T.~Karras, M.~Aittala, T.~Aila, and S.~Laine.
\newblock Elucidating the design space of diffusion-based generative models.
\newblock In \emph{Advances in Neural Information Processing Systems}, volume~35, pages 26565--26577, 2022.

\bibitem[Kelvin(1901)]{kelvin1901nineteenth}
L.~Kelvin.
\newblock I. nineteenth-century clouds over the dynamical theory of heat and light.
\newblock \emph{The London, Edinburgh, and Dublin Philosophical Magazine and Journal of Science}, 2\penalty0 (7):\penalty0 1--40, 1901.

\bibitem[Khan et~al.(2023)Khan, Gohil, Notin, van Amersfoort, Woolrich, and Gal]{khan2023dynemoc}
A.~M.~S. Khan, C.~Gohil, P.~Notin, J.~van Amersfoort, M.~Woolrich, and Y.~Gal.
\newblock {DyNeMoC}: A semi-supervised architecture for classifying time series brain data.
\newblock In \emph{International Conference on Learning Representations Workshop}, 2023.

\bibitem[Kim et~al.(2024{\natexlab{a}})Kim, Lai, Liao, Murata, Takida, Uesaka, He, Mitsufuji, and Ermon]{kim2023consistency}
D.~Kim, C.-H. Lai, W.-H. Liao, N.~Murata, Y.~Takida, T.~Uesaka, Y.~He, Y.~Mitsufuji, and S.~Ermon.
\newblock Consistency trajectory models: Learning probability flow {ODE} trajectory of diffusion.
\newblock In \emph{International Conference on Learning Representations}, 2024{\natexlab{a}}.

\bibitem[Kim et~al.(2023)Kim, Sclar, Zhou, Le~Bras, Kim, Choi, and Sap]{kim2023fantom}
H.~Kim, M.~Sclar, X.~Zhou, R.~Le~Bras, G.~Kim, Y.~Choi, and M.~Sap.
\newblock {FANToM}: A benchmark for stress-testing machine theory of mind in interactions.
\newblock In \emph{Conference on Empirical Methods in Natural Language Processing}, pages 14397--14413, 2023.

\bibitem[Kim et~al.(2025)Kim, Sclar, Zhi-Xuan, Ying, Levine, Liu, Tenenbaum, and Choi]{kim2025hypothesis}
H.~Kim, M.~Sclar, T.~Zhi-Xuan, L.~Ying, S.~Levine, Y.~Liu, J.~B. Tenenbaum, and Y.~Choi.
\newblock Hypothesis-driven theory-of-mind reasoning for large language models.
\newblock \emph{arXiv preprint arXiv:2502.11881}, 2025.

\bibitem[Kim et~al.(2024{\natexlab{b}})Kim, Hooper, Gholami, Dong, Li, Shen, Mahoney, and Keutzer]{kim2024squeezellm}
S.~Kim, C.~Hooper, A.~Gholami, Z.~Dong, X.~Li, S.~Shen, M.~W. Mahoney, and K.~Keutzer.
\newblock {SqueezeLLM}: Dense-and-sparse quantization.
\newblock In \emph{International Conference on Machine Learning}, pages 23901--23923. PMLR, 2024{\natexlab{b}}.

\bibitem[Kim et~al.(2020)Kim, Zhou, Philion, Torralba, and Fidler]{kim2020gamegan}
S.~W. Kim, Y.~Zhou, J.~Philion, A.~Torralba, and S.~Fidler.
\newblock Learning to simulate dynamic environments with {GameGAN}.
\newblock In \emph{IEEE/CVF Conference on Computer Vision and Pattern Recognition}, pages 1231--1240, 2020.

\bibitem[Kingma and Welling(2014)]{kingma2014vae}
D.~P. Kingma and M.~Welling.
\newblock Auto-encoding variational {Bayes}.
\newblock In \emph{International Conference on Learning Representations}, 2014.

\bibitem[Kleppmann and Beresford(2017)]{kleppmann2017crdt}
M.~Kleppmann and A.~R. Beresford.
\newblock A conflict-free replicated {JSON} datatype.
\newblock \emph{IEEE Transactions on Parallel and Distributed Systems}, 28\penalty0 (10):\penalty0 2733--2746, 2017.

\bibitem[Kochkov et~al.(2024)Kochkov, Yuval, Langmore, Norgaard, Smith, Mooers, Kl{\"o}wer, Lottes, Rasp, D{\"u}ben, Hatfield, Battaglia, Sanchez-Gonzalez, Willson, Brenner, and Hoyer]{kochkov2024neuralgcm}
D.~Kochkov, J.~Yuval, I.~Langmore, P.~Norgaard, J.~Smith, G.~Mooers, M.~Kl{\"o}wer, J.~Lottes, S.~Rasp, P.~D{\"u}ben, S.~Hatfield, P.~Battaglia, A.~Sanchez-Gonzalez, M.~Willson, M.~P. Brenner, and S.~Hoyer.
\newblock Neural general circulation models for weather and climate.
\newblock \emph{Nature}, 632:\penalty0 1060--1066, 2024.

\bibitem[Koh et~al.(2021)Koh, Lee, Yang, Baldridge, and Anderson]{koh2021pathdreamer}
J.~Y. Koh, H.~Lee, Y.~Yang, J.~Baldridge, and P.~Anderson.
\newblock {PathDreamer}: A world model for indoor navigation.
\newblock In \emph{IEEE/CVF International Conference on Computer Vision}, pages 14738--14748, 2021.

\bibitem[Koh et~al.(2026)Koh, Han, Lee, Yun, and Shin]{gworld2026}
W.~Koh, S.~Han, S.~Lee, S.-Y. Yun, and J.~Shin.
\newblock Generative visual code mobile world models.
\newblock \emph{arXiv preprint arXiv:2602.01576}, 2026.

\bibitem[Kondratyuk et~al.(2024)Kondratyuk, Yu, Gu, Lezama, Huang, Schindler, Hornung, Birodkar, Yan, Chiu, Somandepalli, Akbari, Alon, Cheng, Dillon, Gupta, Hahn, Hauth, Hendon, Martinez, Minnen, Sirotenko, Sohn, Yang, Adam, Yang, Essa, Wang, Ross, Seybold, and Jiang]{kondratyuk2024videopoet}
D.~Kondratyuk, L.~Yu, X.~Gu, J.~Lezama, J.~Huang, G.~Schindler, R.~Hornung, V.~Birodkar, J.~Yan, M.-C. Chiu, K.~Somandepalli, H.~Akbari, Y.~Alon, Y.~Cheng, J.~V. Dillon, A.~Gupta, M.~Hahn, A.~Hauth, D.~Hendon, A.~Martinez, D.~Minnen, M.~Sirotenko, K.~Sohn, X.~Yang, H.~Adam, M.-H. Yang, I.~Essa, H.~Wang, D.~A. Ross, B.~Seybold, and L.~Jiang.
\newblock {VideoPoet}: A large language model for zero-shot video generation.
\newblock In \emph{International Conference on Machine Learning}, pages 25105--25124. PMLR, 2024.

\bibitem[Kong et~al.(2025)Kong, Yang, Mei, Liu, Liang, Zhu, Lu, Yin, Hu, Jia, Deng, Zhang, Wu, Yan, Gao, Wang, Li, Pan, Liu, Zhu, Ooi, Hoi, and Liu]{kong2025survey_3d4d}
L.~Kong, W.~Yang, J.~Mei, Y.~Liu, A.~Liang, D.~Zhu, D.~Lu, W.~Yin, X.~Hu, M.~Jia, J.~Deng, K.~Zhang, Y.~Wu, T.~Yan, S.~Gao, S.~Wang, L.~Li, L.~Pan, Y.~Liu, J.~Zhu, W.~T. Ooi, S.~C.~H. Hoi, and Z.~Liu.
\newblock {3D} and {4D} world modeling: A survey.
\newblock \emph{arXiv preprint arXiv:2509.07996}, 2025.

\bibitem[Korovina et~al.(2020)Korovina, Xu, Kandasamy, Neiswanger, Poczos, Schneider, and Xing]{pmlr-v108-korovina20a}
K.~Korovina, S.~Xu, K.~Kandasamy, W.~Neiswanger, B.~Poczos, J.~Schneider, and E.~Xing.
\newblock {ChemBO}: Bayesian optimization of small organic molecules with synthesizable recommendations.
\newblock In \emph{International Conference on Artificial Intelligence and Statistics}, pages 3393--3403, 2020.

\bibitem[Kovachki et~al.(2023)Kovachki, Li, Liu, Azizzadenesheli, Bhattacharya, Stuart, and Anandkumar]{kovachki2023neuraloperator}
N.~Kovachki, Z.~Li, B.~Liu, K.~Azizzadenesheli, K.~Bhattacharya, A.~Stuart, and A.~Anandkumar.
\newblock Neural operator: Learning maps between function spaces with applications to {PDEs}.
\newblock \emph{Journal of Machine Learning Research}, 24\penalty0 (89):\penalty0 1--97, 2023.

\bibitem[Kripke(1963)]{kripke1963semantic}
S.~A. Kripke.
\newblock Semantical considerations on modal logic.
\newblock \emph{Acta Philosophica Fennica}, 16:\penalty0 83--94, 1963.

\bibitem[Kripke(1980)]{kripke1980naming}
S.~A. Kripke.
\newblock \emph{Naming and Necessity}.
\newblock Harvard University Press, 1980.

\bibitem[Krizhevsky et~al.(2012)Krizhevsky, Sutskever, and Hinton]{krizhevsky2012imagenet}
A.~Krizhevsky, I.~Sutskever, and G.~E. Hinton.
\newblock {ImageNet} classification with deep convolutional neural networks.
\newblock In \emph{Advances in Neural Information Processing Systems}, volume~25, pages 1097--1105, 2012.

\bibitem[Kuhn(1962)]{kuhn1962structure}
T.~S. Kuhn.
\newblock \emph{The Structure of Scientific Revolutions}.
\newblock University of Chicago Press, 4th edition, 1962.

\bibitem[Kumar et~al.(2026)Kumar, Saito, Niranjani, Yessou, and Tan]{kumar2026constitutions}
U.~Kumar, A.~Saito, H.~Niranjani, R.~Yessou, and P.~X. Tan.
\newblock Evolving interpretable constitutions for multi-agent coordination.
\newblock \emph{arXiv preprint arXiv:2602.00755}, 2026.

\bibitem[Kurenkov et~al.(2023)Kurenkov, Nikulin, Tarasov, and Kolesnikov]{kurenkov2023katakomba}
V.~Kurenkov, A.~Nikulin, D.~Tarasov, and S.~Kolesnikov.
\newblock Katakomba: Tools and benchmarks for data-driven nethack.
\newblock In \emph{Advances in Neural Information Processing Systems}, volume~36, pages 52969--52990, 2023.

\bibitem[Kusne et~al.(2020)Kusne, Yu, Wu, Zhang, Hattrick-Simpers, DeCost, Sarker, Oses, Toher, Curtarolo, Davydov, Agarwal, Bendersky, Li, Mehta, and Takeuchi]{kusne2020cameo}
A.~G. Kusne, H.~Yu, C.~Wu, H.~Zhang, J.~Hattrick-Simpers, B.~DeCost, S.~Sarker, C.~Oses, C.~Toher, S.~Curtarolo, A.~V. Davydov, R.~Agarwal, L.~A. Bendersky, M.~Li, A.~Mehta, and I.~Takeuchi.
\newblock On-the-fly closed-loop materials discovery via {Bayesian} active learning.
\newblock \emph{Nature Communications}, 11\penalty0 (1):\penalty0 5966, 2020.

\bibitem[Kwon et~al.(2023)Kwon, Li, Zhuang, Sheng, Zheng, Yu, Gonzalez, Zhang, and Stoica]{kwon2023vllm}
W.~Kwon, Z.~Li, S.~Zhuang, Y.~Sheng, L.~Zheng, C.~H. Yu, J.~E. Gonzalez, H.~Zhang, and I.~Stoica.
\newblock Efficient memory management for large language model serving with {PagedAttention}.
\newblock In \emph{ACM SIGOPS Symposium on Operating Systems Principles}, pages 611--626, 2023.

\bibitem[Lakatos(1978)]{lakatos1978methodology}
I.~Lakatos.
\newblock \emph{The Methodology of Scientific Research Programmes}, volume~1 of \emph{Philosophical Papers}.
\newblock Cambridge University Press, 1978.

\bibitem[Lake et~al.(2017)Lake, Ullman, Tenenbaum, and Gershman]{lake2017building}
B.~M. Lake, T.~D. Ullman, J.~B. Tenenbaum, and S.~J. Gershman.
\newblock Building machines that learn and think like people.
\newblock \emph{Behavioral and Brain Sciences}, 40:\penalty0 e253, 2017.

\bibitem[Lam et~al.(2023)Lam, Sanchez-Gonzalez, Willson, Wirnsberger, Fortunato, Alet, Ravuri, Ewalds, Eaton-Rosen, Hu, Merose, Hoyer, Holland, Vinyals, Stott, Pritzel, Mohamed, and Battaglia]{lam2023graphcast}
R.~Lam, A.~Sanchez-Gonzalez, M.~Willson, P.~Wirnsberger, M.~Fortunato, F.~Alet, S.~Ravuri, T.~Ewalds, Z.~Eaton-Rosen, W.~Hu, A.~Merose, S.~Hoyer, G.~Holland, O.~Vinyals, J.~Stott, A.~Pritzel, S.~Mohamed, and P.~Battaglia.
\newblock Learning skillful medium-range global weather forecasting.
\newblock \emph{Science}, 382\penalty0 (6677):\penalty0 1416--1421, 2023.

\bibitem[Lan et~al.(2024)Lan, Hu, Wang, Wang, Ye, Zhao, Lim, Xiong, and Wang]{lan2024avalonllm}
Y.~Lan, Z.~Hu, L.~Wang, Y.~Wang, D.~Ye, P.~Zhao, E.-P. Lim, H.~Xiong, and H.~Wang.
\newblock {LLM}-based agent society investigation: Collaboration and confrontation in avalon gameplay.
\newblock In \emph{Conference on Empirical Methods in Natural Language Processing}, pages 128--145, 2024.

\bibitem[Laplace(1814)]{laplace1814essai}
P.-S. Laplace.
\newblock \emph{Essai philosophique sur les probabilit{\'e}s}.
\newblock Courcier, Paris, 1814.

\bibitem[Laskin et~al.(2020)Laskin, Srinivas, and Abbeel]{srinivas2020curl}
M.~Laskin, A.~Srinivas, and P.~Abbeel.
\newblock {CURL}: Contrastive unsupervised representations for reinforcement learning.
\newblock In \emph{International Conference on Machine Learning}, pages 5639--5650. PMLR, 2020.

\bibitem[Lassig et~al.(2017)Lassig, Mustonen, and Walczak]{lassig2017predicting}
M.~Lassig, V.~Mustonen, and A.~M. Walczak.
\newblock Predicting evolution.
\newblock \emph{Nature Ecology \& Evolution}, 1\penalty0 (3):\penalty0 0077, 2017.

\bibitem[Le et~al.(2019)Le, Boureau, and Nickel]{le2019tomi}
M.~Le, Y.-L. Boureau, and M.~Nickel.
\newblock Revisiting the evaluation of theory of mind through question answering.
\newblock In \emph{Conference on Empirical Methods in Natural Language Processing and the International Joint Conference on Natural Language Processing}, pages 5872--5877, 2019.

\bibitem[LeCun(2022)]{lecun2022path}
Y.~LeCun.
\newblock A path towards autonomous machine intelligence.
\newblock \emph{OpenReview preprint}, 2022.
\newblock URL \url{https://openreview.net/pdf?id=BZ5a1r-kVsf}.

\bibitem[LeCun et~al.(1998)LeCun, Bottou, Bengio, and Haffner]{lecun1998gradient}
Y.~LeCun, L.~Bottou, Y.~Bengio, and P.~Haffner.
\newblock Gradient-based learning applied to document recognition.
\newblock \emph{Proceedings of the IEEE}, 86\penalty0 (11):\penalty0 2278--2324, 1998.

\bibitem[Lee et~al.(2019)Lee, Ajanthan, Gould, and Torr]{lee2019signal}
N.~Lee, T.~Ajanthan, S.~Gould, and P.~H. Torr.
\newblock A signal propagation perspective for pruning neural networks at initialization.
\newblock \emph{arXiv preprint arXiv:1906.06307}, 2019.

\bibitem[Lehrach et~al.(2025)Lehrach, Hennes, Lazaro-Gredilla, Lou, Wendelken, Li, Dedieu, Grau-Moya, Lanctot, Iscen, Schultz, Chiam, Gemp, Zielinski, Singh, and Murphy]{lehrach2025cwmgame}
W.~Lehrach, D.~Hennes, M.~Lazaro-Gredilla, X.~Lou, C.~Wendelken, Z.~Li, A.~Dedieu, J.~Grau-Moya, M.~Lanctot, A.~Iscen, J.~Schultz, M.~Chiam, I.~Gemp, P.~Zielinski, S.~Singh, and K.~P. Murphy.
\newblock Code world models for general game playing.
\newblock \emph{arXiv preprint arXiv:2510.04542}, 2025.

\bibitem[Leibo et~al.(2021)Leibo, Du{\'e}{\~n}ez-Guzm{\'a}n, Vezhnevets, Agapiou, Sunehag, Koster, Matyas, Beattie, Mordatch, and Graepel]{leibo2021meltingpot}
J.~Z. Leibo, E.~A. Du{\'e}{\~n}ez-Guzm{\'a}n, A.~S. Vezhnevets, J.~P. Agapiou, P.~Sunehag, R.~Koster, J.~Matyas, C.~Beattie, I.~Mordatch, and T.~Graepel.
\newblock Scalable evaluation of multi-agent reinforcement learning with melting pot.
\newblock In \emph{International Conference on Machine Learning}, pages 6187--6199. PMLR, 2021.

\bibitem[Lenat(1995)]{lenat1995cyc}
D.~B. Lenat.
\newblock {CYC}: A large-scale investment in knowledge infrastructure.
\newblock \emph{Communications of the ACM}, 38\penalty0 (11):\penalty0 33--38, 1995.

\bibitem[Lesort et~al.(2018)Lesort, Diaz-Rodr{\'i}guez, Goudou, and Filliat]{lesort2018srl}
T.~Lesort, N.~Diaz-Rodr{\'i}guez, J.-F. Goudou, and D.~Filliat.
\newblock State representation learning for control: An overview.
\newblock \emph{Neural Networks}, 108:\penalty0 379--392, 2018.

\bibitem[Lewis(1973)]{lewis1973counterfactuals}
D.~Lewis.
\newblock \emph{Counterfactuals}.
\newblock Harvard University Press, 1973.

\bibitem[Lewis et~al.(2017)Lewis, Yarats, Dauphin, Parikh, and Batra]{lewis2017dealornodeal}
M.~Lewis, D.~Yarats, Y.~Dauphin, D.~Parikh, and D.~Batra.
\newblock Deal or no deal? end-to-end learning of negotiation dialogues.
\newblock In \emph{Conference on Empirical Methods in Natural Language Processing}, pages 2443--2453, 2017.

\bibitem[Li et~al.(2021{\natexlab{a}})Li, Xia, Mart{\'i}n-Mart{\'i}n, Lingelbach, Srivastava, Shen, Vainio, Gokmen, Dharan, Jain, Kurenkov, Liu, Gweon, Wu, Fei-Fei, and Savarese]{li2021igibson2}
C.~Li, F.~Xia, R.~Mart{\'i}n-Mart{\'i}n, M.~Lingelbach, S.~Srivastava, B.~Shen, K.~Vainio, C.~Gokmen, G.~Dharan, T.~Jain, A.~Kurenkov, C.~K. Liu, H.~Gweon, J.~Wu, L.~Fei-Fei, and S.~Savarese.
\newblock {iGibson 2.0}: Object-centric simulation for robot learning of everyday household tasks.
\newblock In \emph{Conference on Robot Learning}, pages 455--465, 2021{\natexlab{a}}.

\bibitem[Li et~al.(2023{\natexlab{a}})Li, Leng, Yan, Shen, Wang, Mi, Fei, Feng, Yan, Wang, Zhan, Jia, Wu, and Sun]{li2023chatharuhi}
C.~Li, Z.~Leng, C.~Yan, J.~Shen, H.~Wang, W.~Mi, Y.~Fei, X.~Feng, S.~Yan, H.~Wang, L.~Zhan, Y.~Jia, P.~Wu, and H.~Sun.
\newblock {ChatHaruhi}: Reviving anime character in reality via large language model.
\newblock \emph{arXiv preprint arXiv:2308.09597}, 2023{\natexlab{a}}.

\bibitem[Li et~al.(2024{\natexlab{a}})Li, Zhang, Wong, Gokmen, Srivastava, Mart{\'i}n-Mart{\'i}n, Wang, Levine, Ai, Martinez, Yin, Lingelbach, Hwang, Hiranaka, Garlanka, Aydin, Lee, Sun, Anvari, Sharma, Bansal, Hunter, Kim, Lou, Matthews, Villa-Renteria, Tang, Tang, Xia, Li, Savarese, Gweon, Liu, Wu, and Fei-Fei]{li2024behavior1k}
C.~Li, R.~Zhang, J.~Wong, C.~Gokmen, S.~Srivastava, R.~Mart{\'i}n-Mart{\'i}n, C.~Wang, G.~Levine, W.~Ai, B.~Martinez, H.~Yin, M.~Lingelbach, M.~Hwang, A.~Hiranaka, S.~Garlanka, A.~Aydin, S.~Lee, J.~Sun, M.~Anvari, M.~Sharma, D.~Bansal, S.~Hunter, K.-Y. Kim, A.~Lou, C.~R. Matthews, I.~Villa-Renteria, J.~H. Tang, C.~Tang, F.~Xia, Y.~Li, S.~Savarese, H.~Gweon, C.~K. Liu, J.~Wu, and L.~Fei-Fei.
\newblock {BEHAVIOR-1K}: A human-centered, embodied {AI} benchmark with 1,000 everyday activities and realistic simulation.
\newblock \emph{arXiv preprint arXiv:2403.09227}, 2024{\natexlab{a}}.

\bibitem[Li et~al.(2025{\natexlab{a}})Li, Fang, Chen, Yang, Cao, Wong, Luo, Wang, Yin, Gonzalez, Stoica, Han, and Lu]{fan2025worldmodelbench}
D.~Li, Y.~Fang, Y.~Chen, S.~Yang, S.~Cao, J.~Wong, M.~Luo, X.~Wang, H.~Yin, J.~E. Gonzalez, I.~Stoica, S.~Han, and Y.~Lu.
\newblock {WorldModelBench}: Judging video generation models as world models.
\newblock \emph{arXiv preprint arXiv:2502.20694}, 2025{\natexlab{a}}.

\bibitem[Li et~al.(2025{\natexlab{b}})Li, Tang, Xu, Wu, Zhou, Shao, Yu, Cao, and Lu]{li2025gamecraft}
J.~Li, J.~Tang, Z.~Xu, L.~Wu, Y.~Zhou, S.~Shao, T.~Yu, Z.~Cao, and Q.~Lu.
\newblock {Hunyuan-GameCraft}: High-dynamic interactive game video generation with hybrid history condition.
\newblock \emph{arXiv preprint arXiv:2506.17201}, 2025{\natexlab{b}}.

\bibitem[Li et~al.(2024{\natexlab{b}})Li, Lin, Zhang, Cai, Li, Guo, Xie, Meng, Zhu, and Han]{li2024svdquant}
M.~Li, Y.~Lin, Z.~Zhang, T.~Cai, X.~Li, J.~Guo, E.~Xie, C.~Meng, J.-Y. Zhu, and S.~Han.
\newblock {SVDQuant}: Absorbing outliers by low-rank components for 4-bit diffusion models.
\newblock \emph{arXiv preprint arXiv:2411.05007}, 2024{\natexlab{b}}.

\bibitem[Li et~al.(2024{\natexlab{c}})Li, Jing, Han, Zhou, and Du]{li2024ldc}
R.~Li, L.~Jing, C.~Han, J.~Zhou, and X.~Du.
\newblock Learning to generate research idea with dynamic control.
\newblock In \emph{Advances in Neural Information Processing Systems Workshop}, 2024{\natexlab{c}}.

\bibitem[Li et~al.(2025{\natexlab{c}})Li, Kallidromitis, Gokul, Kato, Kozuka, and Grover]{li2025mobileworldbench}
S.~Li, K.~Kallidromitis, A.~Gokul, Y.~Kato, K.~Kozuka, and A.~Grover.
\newblock {MobileWorldBench}: Towards semantic world modeling for mobile agents.
\newblock \emph{arXiv preprint arXiv:2512.14014}, 2025{\natexlab{c}}.

\bibitem[Li et~al.(2025{\natexlab{d}})Li, Zhao, Yu, Du, Zou, Hu, and Xu]{li2025pinwm}
W.~Li, H.~Zhao, Z.~Yu, Y.~Du, Q.~Zou, R.~Hu, and K.~Xu.
\newblock {PIN-WM}: Learning physics-informed world models for non-prehensile manipulation.
\newblock \emph{arXiv preprint arXiv:2504.16693}, 2025{\natexlab{d}}.

\bibitem[Li et~al.(2023{\natexlab{b}})Li, Liu, Lian, Yang, Dong, Kang, Zhang, and Keutzer]{li2023q}
X.~Li, Y.~Liu, L.~Lian, H.~Yang, Z.~Dong, D.~Kang, S.~Zhang, and K.~Keutzer.
\newblock {Q-Diffusion}: Quantizing diffusion models.
\newblock In \emph{IEEE/CVF International Conference on Computer Vision}, pages 17535--17545, 2023{\natexlab{b}}.

\bibitem[Li et~al.(2024{\natexlab{d}})Li, Zhou, Zhang, Wei, Hou, and Cheng]{li2024sora_geometry}
X.~Li, D.~Zhou, C.~Zhang, S.~Wei, Q.~Hou, and M.-M. Cheng.
\newblock Sora generates videos with stunning geometrical consistency.
\newblock \emph{arXiv preprint arXiv:2402.17403}, 2024{\natexlab{d}}.

\bibitem[Li et~al.(2025{\natexlab{e}})Li, He, Zhang, Wu, Li, and Liu]{li2025embodied_wm_survey}
X.~Li, X.~He, L.~Zhang, M.~Wu, X.~Li, and Y.~Liu.
\newblock A comprehensive survey on world models for embodied {AI}.
\newblock \emph{arXiv preprint arXiv:2510.16732}, 2025{\natexlab{e}}.

\bibitem[Li et~al.(2026)Li, Zheng, Gao, Xia, Wang, Wang, et~al.]{li2026embodiedsafety}
X.~Li, X.~Zheng, Y.~Gao, X.~Xia, Y.~Wang, X.~Wang, et~al.
\newblock Safety in embodied {AI}: A survey of risks, attacks, and defenses.
\newblock \emph{arXiv preprint arXiv:2605.02900}, 2026.

\bibitem[Li et~al.(2023{\natexlab{c}})Li, Xu, Cao, Sun, and Zhang]{li2024q}
Y.~Li, S.~Xu, X.~Cao, X.~Sun, and B.~Zhang.
\newblock {Q-DM}: An efficient low-bit quantized diffusion model.
\newblock \emph{Advances in Neural Information Processing Systems}, 36:\penalty0 76680--76691, 2023{\natexlab{c}}.

\bibitem[Li et~al.(2025{\natexlab{f}})Li, Inan, Yue, Chen, Wutschitz, Kulkarni, Poovendran, Sim, and Rajmohan]{li2025simia}
Y.~Li, H.~A. Inan, X.~Yue, W.-N. Chen, L.~Wutschitz, J.~Kulkarni, R.~Poovendran, R.~Sim, and S.~Rajmohan.
\newblock Simulating environments with reasoning models for agent training.
\newblock \emph{arXiv preprint arXiv:2511.01824}, 2025{\natexlab{f}}.

\bibitem[Li et~al.(2025{\natexlab{g}})Li, Wang, Qiu, Yin, Zhang, Qian, Li, Ma, Chen, and Ji]{li2025wordtoworld}
Y.~Li, H.~Wang, J.~Qiu, Z.~Yin, D.~Zhang, C.~Qian, Z.~Li, P.~Ma, G.~Chen, and H.~Ji.
\newblock From word to world: Can large language models be implicit text-based world models?
\newblock \emph{arXiv preprint arXiv:2512.18832}, 2025{\natexlab{g}}.

\bibitem[Li et~al.(2021{\natexlab{b}})Li, Kovachki, Azizzadenesheli, Liu, Bhattacharya, Stuart, and Anandkumar]{li2021fno}
Z.~Li, N.~Kovachki, K.~Azizzadenesheli, B.~Liu, K.~Bhattacharya, A.~Stuart, and A.~Anandkumar.
\newblock Fourier neural operator for parametric partial differential equations.
\newblock In \emph{International Conference on Learning Representations}, 2021{\natexlab{b}}.

\bibitem[Li et~al.(2024{\natexlab{e}})Li, Zheng, Kovachki, Jin, Chen, Liu, Azizzadenesheli, and Anandkumar]{li2021pino}
Z.~Li, H.~Zheng, N.~Kovachki, D.~Jin, H.~Chen, B.~Liu, K.~Azizzadenesheli, and A.~Anandkumar.
\newblock Physics-informed neural operator for learning partial differential equations.
\newblock \emph{ACM/IMS Journal of Data Science}, 2024{\natexlab{e}}.

\bibitem[Lian et~al.(2025)Lian, Wu, Ma, Ding, Song, Chen, Zheng, and Li]{lian2025ui}
S.~Lian, Y.~Wu, J.~Ma, Y.~Ding, Z.~Song, B.~Chen, X.~Zheng, and H.~Li.
\newblock {UI-AGILE}: Advancing {GUI} agents with effective reinforcement learning and precise inference-time grounding.
\newblock \emph{arXiv preprint arXiv:2507.22025}, 2025.

\bibitem[Liang et~al.(2026{\natexlab{a}})Liang, Kong, Yan, Liu, Yang, Huang, Yin, Zuo, Hu, Zhu, Lu, Liu, Jiang, Li, Li, Zhuo, Ng, Cottereau, Gao, Pan, Ooi, and Liu]{worldlens}
A.~Liang, L.~Kong, T.~Yan, H.~Liu, W.~Yang, Z.~Huang, W.~Yin, J.~Zuo, Y.~Hu, D.~Zhu, D.~Lu, Y.~Liu, G.~Jiang, L.~Li, X.~Li, L.~Zhuo, L.~X. Ng, B.~R. Cottereau, C.~Gao, L.~Pan, W.~T. Ooi, and Z.~Liu.
\newblock {WorldLens}: Full-spectrum evaluations of driving world models in real world.
\newblock In \emph{IEEE/CVF Conference on Computer Vision and Pattern Recognition}, pages 36385--36399, 2026{\natexlab{a}}.

\bibitem[Liang et~al.(2026{\natexlab{b}})Liang, Liu, Yang, Lu, Li, Kong, Zhao, and Ooi]{liang2026lidarcrafter}
A.~Liang, Y.~Liu, Y.~Yang, D.~Lu, L.~Li, L.~Kong, H.~Zhao, and W.~T. Ooi.
\newblock {LiDARCrafter}: Dynamic {4D} world modeling from {LiDAR} sequences.
\newblock In \emph{AAAI Conference on Artificial Intelligence}, volume~40, pages 18406--18414, 2026{\natexlab{b}}.

\bibitem[Light et~al.(2023)Light, Cai, Shen, and Hu]{light2023avalonbench}
J.~Light, M.~Cai, S.~Shen, and Z.~Hu.
\newblock {AvalonBench}: Evaluating llms playing the game of avalon.
\newblock \emph{arXiv preprint arXiv:2310.05036}, 2023.

\bibitem[Lighthill(1973)]{lighthill1973artificial}
J.~Lighthill.
\newblock Artificial intelligence: A general survey.
\newblock Technical report, Science Research Council, 1973.

\bibitem[Lin et~al.(2024)Lin, Tang, Tang, Yang, Chen, Wang, Xiao, Dang, Gan, and Han]{lin2024awq}
J.~Lin, J.~Tang, H.~Tang, S.~Yang, W.-M. Chen, W.-C. Wang, G.~Xiao, X.~Dang, C.~Gan, and S.~Han.
\newblock Awq: Activation-aware weight quantization for llm compression and acceleration.
\newblock In \emph{MLSys}, 2024.

\bibitem[Lin and Shou(2025)]{lin2025vlog}
K.~Q. Lin and M.~Z. Shou.
\newblock Vlog: Video-language models by generative retrieval of narration vocabulary.
\newblock In \emph{IEEE/CVF Conference on Computer Vision and Pattern Recognition}, pages 3218--3228, 2025.

\bibitem[Lin et~al.(2025{\natexlab{a}})Lin, Hu, Li, Yang, Wang, Torr, and Shou]{lin2025computer}
K.~Q. Lin, S.~Hu, L.~Li, Z.~Yang, L.~Wang, P.~Torr, and M.~Z. Shou.
\newblock Computer-use agents as judges for generative user interface.
\newblock \emph{arXiv preprint arXiv:2511.15567}, 2025{\natexlab{a}}.

\bibitem[Lin et~al.(2025{\natexlab{b}})Lin, Li, Gao, Yang, Wu, Bai, Lei, Wang, and Shou]{lin2025showui}
K.~Q. Lin, L.~Li, D.~Gao, Z.~Yang, S.~Wu, Z.~Bai, W.~Lei, L.~Wang, and M.~Z. Shou.
\newblock {ShowUI}: One vision-language-action model for {GUI} visual agent.
\newblock In \emph{IEEE/CVF Conference on Computer Vision and Pattern Recognition}, pages 19498--19508, 2025{\natexlab{b}}.

\bibitem[Lin et~al.(2025{\natexlab{c}})Lin, Zheng, Ran, Zhu, Mao, Li, Torr, and Wang]{lin2025vcode}
K.~Q. Lin, Y.~Zheng, H.~Ran, D.~Zhu, D.~Mao, L.~Li, P.~Torr, and A.~J. Wang.
\newblock {VCode}: a multimodal coding benchmark with {SVG} as symbolic visual representation.
\newblock \emph{arXiv preprint arXiv:2511.02778}, 2025{\natexlab{c}}.

\bibitem[Lin et~al.(2025{\natexlab{d}})Lin, Xia, Ren, Yang, Xiao, and Jiang]{lin2025diffusion}
S.~Lin, X.~Xia, Y.~Ren, C.~Yang, X.~Xiao, and L.~Jiang.
\newblock Diffusion adversarial post-training for one-step video generation.
\newblock \emph{arXiv preprint arXiv:2501.08316}, 2025{\natexlab{d}}.

\bibitem[Lin et~al.(2023)Lin, Akin, Rao, Hie, Zhu, Lu, Smetanin, Verkuil, Kabeli, Shmueli, Dos Santos~Costa, Fazel-Zarandi, Sercu, Candido, and Rives]{lin2023esmfold}
Z.~Lin, H.~Akin, R.~Rao, B.~Hie, Z.~Zhu, W.~Lu, N.~Smetanin, R.~Verkuil, O.~Kabeli, Y.~Shmueli, A.~Dos Santos~Costa, M.~Fazel-Zarandi, T.~Sercu, S.~Candido, and A.~Rives.
\newblock Evolutionary-scale prediction of atomic-level protein structure with a language model.
\newblock \emph{Science}, 379\penalty0 (6637):\penalty0 1123--1130, 2023.

\bibitem[Lipman et~al.(2023)Lipman, Chen, Ben-Hamu, Nickel, and Le]{lipman2022flow}
Y.~Lipman, R.~T.~Q. Chen, H.~Ben-Hamu, M.~Nickel, and M.~Le.
\newblock Flow matching for generative modeling.
\newblock In \emph{International Conference on Learning Representations}, 2023.

\bibitem[Liu et~al.(2025)Liu, Yan, Zaharia, and Abbeel]{liu2024lwm}
H.~Liu, W.~Yan, M.~Zaharia, and P.~Abbeel.
\newblock World model on million-length video and language with blockwise ringattention.
\newblock In \emph{International Conference on Learning Representations}, 2025.

\bibitem[Liu et~al.(2026)Liu, Wang, Long, Hou, Sun, Li, Yang, Peng, Liu, Jiang, Yao, and Mu]{liu2026jailwam}
H.~Liu, S.~Wang, J.~Long, J.~Hou, J.~Sun, C.~Li, Y.~Yang, W.~Peng, X.~Liu, T.~Jiang, W.~Yao, and Y.~Mu.
\newblock {JailWAM}: Jailbreaking world action models in robot control.
\newblock \emph{arXiv preprint arXiv:2604.05498}, 2026.

\bibitem[Liu et~al.(2021)Liu, Zhang, Kuang, Zhou, Xue, Wang, Chen, Yang, Liao, and Zhang]{liu2021group}
L.~Liu, S.~Zhang, Z.~Kuang, A.~Zhou, J.-H. Xue, X.~Wang, Y.~Chen, W.~Yang, Q.~Liao, and W.~Zhang.
\newblock Group fisher pruning for practical network compression.
\newblock In \emph{International Conference on Machine Learning}, pages 7021--7032. PMLR, 2021.

\bibitem[Liu et~al.(2024{\natexlab{a}})Liu, Yu, Zhang, Xu, Lei, Lai, Gu, Ding, Men, Yang, Zhang, Deng, Zeng, Du, Zhang, Shen, Zhang, Su, Sun, Huang, Dong, and Tang]{liu2023agentbench}
X.~Liu, H.~Yu, H.~Zhang, Y.~Xu, X.~Lei, H.~Lai, Y.~Gu, H.~Ding, K.~Men, K.~Yang, S.~Zhang, X.~Deng, A.~Zeng, Z.~Du, C.~Zhang, S.~Shen, T.~Zhang, Y.~Su, H.~Sun, M.~Huang, Y.~Dong, and J.~Tang.
\newblock {AgentBench}: Evaluating {LLMs} as agents.
\newblock In \emph{International Conference on Learning Representations}, 2024{\natexlab{a}}.

\bibitem[Liu et~al.(2024{\natexlab{b}})Liu, Yuan, Jin, Zhong, Xu, Braverman, Chen, and Hu]{liu2024kivi}
Z.~Liu, J.~Yuan, H.~Jin, S.~Zhong, Z.~Xu, V.~Braverman, B.~Chen, and X.~Hu.
\newblock {KIVI}: A tuning-free asymmetric 2bit quantization for {KV} cache.
\newblock In \emph{International Conference on Machine Learning}, pages 32332--32344, 2024{\natexlab{b}}.

\bibitem[Lu and Song(2025)]{lu2024simplifying}
C.~Lu and Y.~Song.
\newblock Simplifying, stabilizing and scaling continuous-time consistency models.
\newblock In \emph{International Conference on Learning Representations}, 2025.

\bibitem[Lu et~al.(2022)Lu, Zhou, Bao, Chen, Li, and Zhu]{lu2022dpm}
C.~Lu, Y.~Zhou, F.~Bao, J.~Chen, C.~Li, and J.~Zhu.
\newblock {DPM-Solver}: A fast {ODE} solver for diffusion probabilistic model sampling in around 10 steps.
\newblock In \emph{Advances in Neural Information Processing Systems}, volume~35, pages 5775--5787, 2022.

\bibitem[Lu et~al.(2024{\natexlab{a}})Lu, Lu, Lange, Foerster, Clune, and Ha]{lu2024aiscientist}
C.~Lu, C.~Lu, R.~T. Lange, J.~Foerster, J.~Clune, and D.~Ha.
\newblock The {AI} scientist: Towards fully automated open-ended scientific discovery.
\newblock \emph{arXiv preprint arXiv:2408.06292}, 2024{\natexlab{a}}.

\bibitem[Lu et~al.(2025{\natexlab{a}})Lu, Zhou, Bao, Chen, Li, and Zhu]{lu2025dpm}
C.~Lu, Y.~Zhou, F.~Bao, J.~Chen, C.~Li, and J.~Zhu.
\newblock {DPM-Solver++}: Fast solver for guided sampling of diffusion probabilistic models.
\newblock \emph{Machine Intelligence Research}, 22\penalty0 (4):\penalty0 730--751, 2025{\natexlab{a}}.

\bibitem[Lu et~al.(2025{\natexlab{b}})Lu, Jia, Li, Chen, Wang, Tang, and Huang]{lu2025gwm}
G.~Lu, B.~Jia, P.~Li, Y.~Chen, Z.~Wang, Y.~Tang, and S.~Huang.
\newblock {GWM}: Towards scalable gaussian world models for robotic manipulation.
\newblock In \emph{IEEE/CVF International Conference on Computer Vision}, pages 9263--9274, 2025{\natexlab{b}}.

\bibitem[Lu et~al.(2024{\natexlab{b}})Lu, Zhou, Lin, Liu, Xu, Zhang, Yang, Yan, Gao, and Li]{lu2024terdit}
X.~Lu, A.~Zhou, Z.~Lin, Q.~Liu, Y.~Xu, R.~Zhang, X.~Yang, J.~Yan, P.~Gao, and H.~Li.
\newblock {TerDiT}: Ternary diffusion models with transformers.
\newblock \emph{arXiv preprint arXiv:2405.14854}, 2024{\natexlab{b}}.

\bibitem[Luo et~al.(2025)Luo, Tang, Li, Papoudakis, Song, Gong, Hao, Wang, and Shao]{luo2025vimo}
D.~Luo, B.~Tang, K.~Li, G.~Papoudakis, J.~Song, S.~Gong, J.~Hao, J.~Wang, and K.~Shao.
\newblock {ViMo}: A generative visual {GUI} world model for app agents.
\newblock \emph{arXiv preprint arXiv:2504.13936}, 2025.

\bibitem[Ma et~al.(2026)Ma, Sun, Yu, Wang, Chua, and Bian]{ma2026thinking}
W.~Ma, S.~Sun, T.~Yu, R.~Wang, T.-S. Chua, and J.~Bian.
\newblock Thinking with blueprints: Assisting vision-language models in spatial reasoning via structured object representation.
\newblock \emph{arXiv preprint arXiv:2601.01984}, 2026.

\bibitem[Magne et~al.(2026)Magne, Awadalla, Wang, Xu, Belofsky, Hu, Kim, Schmidt, Gkioxari, Kautz, Yue, Choi, Zhu, and Fan]{nitrogen2025}
L.~Magne, A.~Awadalla, G.~Wang, Y.~Xu, J.~Belofsky, F.~Hu, J.~Kim, L.~Schmidt, G.~Gkioxari, J.~Kautz, Y.~Yue, Y.~Choi, Y.~Zhu, and L.~Fan.
\newblock {NitroGen}: An open foundation model for generalist gaming agents.
\newblock \emph{arXiv preprint arXiv:2601.02427}, 2026.

\bibitem[Majumder et~al.(2025)Majumder, Surana, Agarwal, Mishra, Meena, Prakhar, Vora, Khot, Sabharwal, and Clark]{majumder2024discoverybench}
B.~P. Majumder, H.~Surana, D.~Agarwal, B.~D. Mishra, A.~Meena, A.~Prakhar, T.~Vora, T.~Khot, A.~Sabharwal, and P.~Clark.
\newblock {DiscoveryBench}: Towards data-driven discovery with large language models.
\newblock In \emph{International Conference on Learning Representations}, 2025.

\bibitem[Makoviychuk et~al.(2021)Makoviychuk, Wawrzyniak, Guo, Lu, Storey, Macklin, Hoeller, Rudin, Allshire, Handa, and State]{makoviychuk2021isaacgym}
V.~Makoviychuk, L.~Wawrzyniak, Y.~Guo, M.~Lu, K.~Storey, M.~Macklin, D.~Hoeller, N.~Rudin, A.~Allshire, A.~Handa, and G.~State.
\newblock {Isaac Gym}: High performance {GPU}-based physics simulation for robot learning.
\newblock In \emph{Advances in Neural Information Processing Systems}, 2021.

\bibitem[Mao et~al.(2025)Mao, Li, Li, Xu, Ying, He, Pang, Qiao, and Zhang]{mao2025yume}
X.~Mao, Z.~Li, C.~Li, X.~Xu, K.~Ying, T.~He, J.~Pang, Y.~Qiao, and K.~Zhang.
\newblock Yume-1.5: A text-controlled interactive world generation model.
\newblock \emph{arXiv preprint arXiv:2512.22096}, 2025.

\bibitem[Mao et~al.(2026{\natexlab{a}})Mao, Huang, Ding, Liu, He, and Zhang]{mao2026realcustompp}
Z.~Mao, M.~Huang, F.~Ding, M.~Liu, Q.~He, and Y.~Zhang.
\newblock {RealCustom++}: Representing images as real textual word for real-time customization.
\newblock \emph{IEEE Transactions on Pattern Analysis and Machine Intelligence}, 48\penalty0 (2):\penalty0 2078--2095, 2026{\natexlab{a}}.

\bibitem[Mao et~al.(2026{\natexlab{b}})Mao, Huang, Lin, Wang, Zhang, and Zhang]{mao2026dgit}
Z.~Mao, M.~Huang, Y.~Lin, Q.~Wang, L.~Zhang, and Y.~Zhang.
\newblock Toward accurate image generation via dynamic generative image transformer.
\newblock \emph{IEEE Transactions on Pattern Analysis and Machine Intelligence}, 48\penalty0 (5):\penalty0 5910--5927, 2026{\natexlab{b}}.

\bibitem[Marcus(2018)]{marcus2018deep}
G.~Marcus.
\newblock Deep learning: A critical appraisal.
\newblock \emph{arXiv preprint arXiv:1801.00631}, 2018.

\bibitem[Marra et~al.(2024)Marra, Duman{\v{c}}i{\'c}, Manhaeve, and De~Raedt]{marra2024neurosymbolic_survey}
G.~Marra, S.~Duman{\v{c}}i{\'c}, R.~Manhaeve, and L.~De~Raedt.
\newblock From statistical relational to neuro-symbolic artificial intelligence: A survey.
\newblock \emph{Artificial Intelligence}, 328:\penalty0 104062, 2024.

\bibitem[McCarthy and Hayes(1969)]{mccarthy1969some}
J.~McCarthy and P.~J. Hayes.
\newblock Some philosophical problems from the standpoint of artificial intelligence.
\newblock In \emph{Machine Intelligence}, volume~4, pages 463--502. Edinburgh University Press, 1969.

\bibitem[Mees et~al.(2022)Mees, Hermann, Rosete-Beas, and Burgard]{mees2022calvin}
O.~Mees, L.~Hermann, E.~Rosete-Beas, and W.~Burgard.
\newblock {CALVIN}: A benchmark for language-conditioned policy learning for long-horizon robot manipulation tasks.
\newblock \emph{IEEE Robotics and Automation Letters}, 7\penalty0 (3):\penalty0 7327--7334, 2022.

\bibitem[Micheli et~al.(2023)Micheli, Alonso, and Fleuret]{micheli2023iris}
V.~Micheli, E.~Alonso, and F.~Fleuret.
\newblock Transformers are sample-efficient world models.
\newblock In \emph{International Conference on Learning Representations}, 2023.

\bibitem[Micheli et~al.(2024)Micheli, Alonso, and Fleuret]{micheli2024deltairis}
V.~Micheli, E.~Alonso, and F.~Fleuret.
\newblock Efficient world models with context-aware tokenization.
\newblock In \emph{International Conference on Machine Learning}, pages 35623--35638. PMLR, 2024.

\bibitem[Min et~al.(2024)Min, Zhao, Xiao, Zhao, Xu, Zhu, Jin, Li, Guo, Xing, Jing, Nie, and Dai]{min2024driveworld}
C.~Min, D.~Zhao, L.~Xiao, J.~Zhao, X.~Xu, Z.~Zhu, L.~Jin, J.~Li, Y.~Guo, J.~Xing, L.~Jing, Y.~Nie, and B.~Dai.
\newblock {DriveWorld}: 4{D} pre-trained scene understanding via world models for autonomous driving.
\newblock In \emph{IEEE/CVF Conference on Computer Vision and Pattern Recognition}, pages 15522--15533, 2024.

\bibitem[Minami et~al.(2025)Minami, Hayashi, Wu, Fukumizu, Sugisawa, Ishii, Kuwajima, Shiratori, and Yoshida]{minami2025simtorealsciml}
S.~Minami, Y.~Hayashi, S.~Wu, K.~Fukumizu, H.~Sugisawa, M.~Ishii, I.~Kuwajima, K.~Shiratori, and R.~Yoshida.
\newblock Scaling law of {Sim2Real} transfer learning in expanding computational materials databases for real-world predictions.
\newblock \emph{NPJ Computational Materials}, 11\penalty0 (146), 2025.

\bibitem[Mittal et~al.(2025)Mittal, Roth, Tigue, Richard, Zhang, Du, Serrano-Muñoz, Yao, Zurbrügg, Rudin, Wawrzyniak, Rakhsha, Denzler, Heiden, Borovicka, Ahmed, Akinola, Anwar, Carlson, Feng, Garg, Gasoto, Gulich, Guo, Gussert, Hansen, Kulkarni, Li, Liu, Makoviychuk, Malczyk, Mazhar, Moghani, Murali, Noseworthy, Poddubny, Ratliff, Rehberg, Schwarke, Singh, Smith, Tang, Thaker, Trepte, Wyk, Yu, Millane, Ramasamy, Steiner, Subramanian, Volk, Chen, Jawale, Kuruttukulam, Lin, Mandlekar, Patzwaldt, Welsh, Zhao, Anes, Lafleche, Moënne-Loccoz, Park, Stepinski, Gelder, Amevor, Carius, Chang, Chen, de~Heras~Ciechomski, Daviet, Mohajerani, von Muralt, Reutskyy, Sauter, Schirm, Shi, Terdiman, Vilella, Widmer, Yeoman, Chen, Grizan, Li, Li, Smith, Wiltz, Alexis, Chang, Chu, Fan, Farshidian, Handa, Huang, Hutter, Narang, Pouya, Sheng, Zhu, Macklin, Moravanszky, Reist, Guo, Hoeller, and State]{nvidia2025isaaclab}
M.~Mittal, P.~Roth, J.~Tigue, A.~Richard, O.~Zhang, P.~Du, A.~Serrano-Muñoz, X.~Yao, R.~Zurbrügg, N.~Rudin, L.~Wawrzyniak, M.~Rakhsha, A.~Denzler, E.~Heiden, A.~Borovicka, O.~Ahmed, I.~Akinola, A.~Anwar, M.~T. Carlson, J.~Y. Feng, A.~Garg, R.~Gasoto, L.~Gulich, Y.~Guo, M.~Gussert, A.~Hansen, M.~Kulkarni, C.~Li, W.~Liu, V.~Makoviychuk, G.~Malczyk, H.~Mazhar, M.~Moghani, A.~Murali, M.~Noseworthy, A.~Poddubny, N.~Ratliff, W.~Rehberg, C.~Schwarke, R.~Singh, J.~L. Smith, B.~Tang, R.~Thaker, M.~Trepte, K.~V. Wyk, F.~Yu, A.~Millane, V.~Ramasamy, R.~Steiner, S.~Subramanian, C.~Volk, C.~Chen, N.~Jawale, A.~V. Kuruttukulam, M.~A. Lin, A.~Mandlekar, K.~Patzwaldt, J.~Welsh, H.~Zhao, F.~Anes, J.-F. Lafleche, N.~Moënne-Loccoz, S.~Park, R.~Stepinski, D.~V. Gelder, C.~Amevor, J.~Carius, J.~Chang, A.~H. Chen, P.~de~Heras~Ciechomski, G.~Daviet, M.~Mohajerani, J.~von Muralt, V.~Reutskyy, M.~Sauter, S.~Schirm, E.~L. Shi, P.~Terdiman, K.~Vilella, T.~Widmer, G.~Yeoman, T.~Chen, S.~Grizan, C.~Li, L.~Li, C.~Smith, R.~Wiltz,
  K.~Alexis, Y.~Chang, D.~Chu, L.~J. Fan, F.~Farshidian, A.~Handa, S.~Huang, M.~Hutter, Y.~Narang, S.~Pouya, S.~Sheng, Y.~Zhu, M.~Macklin, A.~Moravanszky, P.~Reist, Y.~Guo, D.~Hoeller, and G.~State.
\newblock {Isaac Lab}: A {GPU}-accelerated simulation framework for multi-modal robot learning.
\newblock \emph{arXiv preprint arXiv:2511.04831}, 2025.

\bibitem[Moerland et~al.(2023)Moerland, Broekens, Plaat, and Jonker]{moerland2023mbrl}
T.~M. Moerland, J.~Broekens, A.~Plaat, and C.~M. Jonker.
\newblock Model-based reinforcement learning: A survey.
\newblock \emph{Foundations and Trends in Machine Learning}, 16\penalty0 (1):\penalty0 1--118, 2023.

\bibitem[Nagabandi et~al.(2018)Nagabandi, Kahn, Fearing, and Levine]{nagabandi2018mbmf}
A.~Nagabandi, G.~Kahn, R.~S. Fearing, and S.~Levine.
\newblock Neural network dynamics for model-based deep reinforcement learning with model-free fine-tuning.
\newblock In \emph{IEEE International Conference on Robotics and Automation}, pages 7559--7566, 2018.

\bibitem[Nasiriany et~al.(2024)Nasiriany, Maddukuri, Zhang, Parikh, Lo, Joshi, Mandlekar, and Zhu]{nasiriany2024robocasa}
S.~Nasiriany, A.~Maddukuri, L.~Zhang, A.~Parikh, A.~Lo, A.~Joshi, A.~Mandlekar, and Y.~Zhu.
\newblock {RoboCasa}: Large-scale simulation of everyday tasks for generalist robots.
\newblock In \emph{Robotics: Science and Systems}, 2024.

\bibitem[Nasiriany et~al.(2026)Nasiriany, Nasiriany, Maddukuri, and Zhu]{nasiriany2026robocasa365}
S.~Nasiriany, S.~Nasiriany, A.~Maddukuri, and Y.~Zhu.
\newblock {RoboCasa365}: A large-scale simulation framework for training and benchmarking generalist robots.
\newblock \emph{arXiv preprint arXiv:2603.04356}, 2026.

\bibitem[Newton(1687)]{newton1687principia}
I.~Newton.
\newblock \emph{Philosophi{\ae} Naturalis Principia Mathematica}.
\newblock Royal Society, London, 1687.

\bibitem[Ngo et~al.(2013)Ngo, Martinetz, Born, and M{\"o}lle]{ngo2013auditory}
H.-V.~V. Ngo, T.~Martinetz, J.~Born, and M.~M{\"o}lle.
\newblock Auditory closed-loop stimulation of the sleep slow oscillation enhances memory.
\newblock \emph{Neuron}, 78\penalty0 (3):\penalty0 545--553, 2013.

\bibitem[Nguyen et~al.(2023)Nguyen, Brandstetter, Kapoor, Gupta, and Grover]{nguyen2023climax}
T.~Nguyen, J.~Brandstetter, A.~Kapoor, J.~K. Gupta, and A.~Grover.
\newblock {ClimaX}: A foundation model for weather and climate.
\newblock In \emph{International Conference on Machine Learning}, pages 25904--25938. PMLR, 2023.

\bibitem[Nie et~al.(2026)Nie, Berner, Ma, Liu, Xie, and Vahdat]{nie2026transition}
W.~Nie, J.~Berner, N.~Ma, C.~Liu, S.~Xie, and A.~Vahdat.
\newblock Transition matching distillation for fast video generation.
\newblock \emph{arXiv preprint arXiv:2601.09881}, 2026.

\bibitem[No{\'e} et~al.(2019)No{\'e}, Olsson, K{\"o}hler, and Wu]{noe2019boltzmann}
F.~No{\'e}, S.~Olsson, J.~K{\"o}hler, and H.~Wu.
\newblock Boltzmann generators: Sampling equilibrium states of many-body systems with deep learning.
\newblock \emph{Science}, 365\penalty0 (6457):\penalty0 eaaw1147, 2019.

\bibitem[No{\'e} et~al.(2020)No{\'e}, Tkatchenko, M{\"u}ller, and Clementi]{noe2020mlmolsim}
F.~No{\'e}, A.~Tkatchenko, K.-R. M{\"u}ller, and C.~Clementi.
\newblock Machine learning for molecular simulation.
\newblock \emph{Annual Review of Physical Chemistry}, 71:\penalty0 361--390, 2020.

\bibitem[Novikov et~al.(2025)Novikov, V{\~u}, Eisenberger, Dupont, Huang, Wagner, Shirobokov, Kozlovskii, Ruiz, Mehrabian, Kumar, See, Chaudhuri, Holland, Davies, Nowozin, Kohli, and Balog]{alphaevolve2025}
A.~Novikov, N.~V{\~u}, M.~Eisenberger, E.~Dupont, P.-S. Huang, A.~Z. Wagner, S.~Shirobokov, B.~Kozlovskii, F.~J.~R. Ruiz, A.~Mehrabian, M.~P. Kumar, A.~See, S.~Chaudhuri, G.~Holland, A.~Davies, S.~Nowozin, P.~Kohli, and M.~Balog.
\newblock {AlphaEvolve}: A coding agent for scientific and algorithmic discovery.
\newblock Technical report, Google DeepMind, 2025.

\bibitem[Okada and Taniguchi(2022)]{okada2022dreamingv2}
M.~Okada and T.~Taniguchi.
\newblock {DreamingV2}: Reinforcement learning with discrete world models without reconstruction.
\newblock In \emph{IEEE/RSJ International Conference on Intelligent Robots and Systems}, pages 985--991, 2022.

\bibitem[Oord et~al.(2018)Oord, Li, and Vinyals]{oord2018cpc}
A.~v.~d. Oord, Y.~Li, and O.~Vinyals.
\newblock Representation learning with contrastive predictive coding.
\newblock \emph{arXiv preprint arXiv:1807.03748}, 2018.

\bibitem[Oquab et~al.(2024)Oquab, Darcet, Moutakanni, Vo, Szafraniec, Khalidov, Fernandez, Haziza, Massa, El{-}Nouby, Assran, Ballas, Galuba, Howes, Huang, Li, Misra, Rabbat, Sharma, Synnaeve, Xu, J{\'{e}}gou, Mairal, Labatut, Joulin, and Bojanowski]{oquab2024dinov2}
M.~Oquab, T.~Darcet, T.~Moutakanni, H.~V. Vo, M.~Szafraniec, V.~Khalidov, P.~Fernandez, D.~Haziza, F.~Massa, A.~El{-}Nouby, M.~Assran, N.~Ballas, W.~Galuba, R.~Howes, P.~Huang, S.~Li, I.~Misra, M.~Rabbat, V.~Sharma, G.~Synnaeve, H.~Xu, H.~J{\'{e}}gou, J.~Mairal, P.~Labatut, A.~Joulin, and P.~Bojanowski.
\newblock {DINOv2}: Learning robust visual features without supervision.
\newblock \emph{Transactions on Machine Learning Research}, 2024.

\bibitem[Ouyang et~al.(2026)Ouyang, Hu, Lin, Ng, and Shou]{ouyang2026gameworld}
M.~Ouyang, S.~Hu, K.~Q. Lin, H.~T. Ng, and M.~Z. Shou.
\newblock {GameWorld}: Towards standardized and verifiable evaluation of multimodal game agents.
\newblock \emph{arXiv preprint arXiv:2604.07429}, 2026.

\bibitem[Pan et~al.(2026)Pan, Zou, Guo, Ni, and Zheng]{pan2026nlah}
L.~Pan, L.~Zou, S.~Guo, J.~Ni, and H.-T. Zheng.
\newblock Natural-language agent harnesses.
\newblock \emph{arXiv preprint arXiv:2603.25723}, 2026.

\bibitem[Park et~al.(2022)Park, Popowski, Cai, Morris, Liang, and Bernstein]{park2022socialsimulacra}
J.~S. Park, L.~Popowski, C.~J. Cai, M.~R. Morris, P.~Liang, and M.~S. Bernstein.
\newblock Social simulacra: Creating populated prototypes for social computing systems.
\newblock In \emph{ACM Symposium on User Interface Software and Technology}, 2022.

\bibitem[Park et~al.(2023)Park, O'Brien, Cai, Morris, Liang, and Bernstein]{park2023generative}
J.~S. Park, J.~C. O'Brien, C.~J. Cai, M.~R. Morris, P.~Liang, and M.~S. Bernstein.
\newblock Generative agents: Interactive simulacra of human behavior.
\newblock In \emph{Annual ACM Symposium on User Interface Software and Technology}, pages 1--22, 2023.

\bibitem[Park et~al.(2020)Park, Lee, Mo, and Shin]{park2020lookahead}
S.~Park, J.~Lee, S.~Mo, and J.~Shin.
\newblock Lookahead: A far-sighted alternative of magnitude-based pruning.
\newblock \emph{arXiv preprint arXiv:2002.04809}, 2020.

\bibitem[Pathak et~al.(2017)Pathak, Agrawal, Efros, and Darrell]{pathak2017curiosity}
D.~Pathak, P.~Agrawal, A.~A. Efros, and T.~Darrell.
\newblock Curiosity-driven exploration by self-supervised prediction.
\newblock In \emph{International Conference on Machine Learning}, pages 2778--2787. PMLR, 2017.

\bibitem[Pearl(2009)]{pearl2009causality}
J.~Pearl.
\newblock \emph{Causality: Models, Reasoning, and Inference}.
\newblock Cambridge University Press, 2 edition, 2009.

\bibitem[Peebles and Xie(2023)]{peebles2023dit}
W.~Peebles and S.~Xie.
\newblock Scalable diffusion models with transformers.
\newblock In \emph{IEEE/CVF International Conference on Computer Vision}, pages 4195--4205, 2023.

\bibitem[Peirce(1932)]{peirce1931collected}
C.~S. Peirce.
\newblock \emph{Collected Papers of Charles Sanders Peirce}, volume~2.
\newblock Harvard University Press, Cambridge, MA, 1932.

\bibitem[Piao et~al.(2025)Piao, Yan, Zhang, Li, Yan, Lan, Lu, Zheng, Wang, Zhou, Gao, Xu, Zhang, Rong, Su, and Li]{piao2025agentsociety}
J.~Piao, Y.~Yan, J.~Zhang, N.~Li, J.~Yan, X.~Lan, Z.~Lu, Z.~Zheng, J.~Y. Wang, D.~Zhou, C.~Gao, F.~Xu, F.~Zhang, K.~Rong, J.~Su, and Y.~Li.
\newblock {AgentSociety}: Large-scale simulation of {LLM}-driven generative agents advances understanding of human behaviors and society.
\newblock \emph{arXiv preprint arXiv:2502.08691}, 2025.

\bibitem[Piatti et~al.(2024)Piatti, Jin, Kleiman-Weiner, Sch{\"o}lkopf, Sachan, and Mihalcea]{piatti2024cooperateorcollapse}
G.~Piatti, Z.~Jin, M.~Kleiman-Weiner, B.~Sch{\"o}lkopf, M.~Sachan, and R.~Mihalcea.
\newblock Cooperate or collapse: Emergence of sustainable cooperation in a society of {LLM} agents.
\newblock In \emph{Advances in Neural Information Processing Systems}, volume~37, pages 111715--111759, 2024.

\bibitem[Plato(1992)]{plato1992republic}
Plato.
\newblock \emph{Republic}.
\newblock Hackett Publishing, 1992.

\bibitem[Popper(1959)]{popper1959logic}
K.~R. Popper.
\newblock \emph{The Logic of Scientific Discovery}.
\newblock Hutchinson, 1959.
\newblock English translation; original German 1935.

\bibitem[Price et~al.(2024)Price, Sanchez-Gonzalez, Alet, Andersson, El-Kadi, Masters, Ewalds, Stott, Mohamed, Battaglia, Lam, and Willson]{price2024gencast}
I.~Price, A.~Sanchez-Gonzalez, F.~Alet, T.~R. Andersson, A.~El-Kadi, D.~Masters, T.~Ewalds, J.~Stott, S.~Mohamed, P.~Battaglia, R.~Lam, and M.~Willson.
\newblock Probabilistic weather forecasting with machine learning.
\newblock \emph{Nature}, 637:\penalty0 84--90, 2024.

\bibitem[Puig et~al.(2018)Puig, Ra, Boben, Li, Wang, Fidler, and Torralba]{puig2018virtualhome}
X.~Puig, K.~Ra, M.~Boben, J.~Li, T.~Wang, S.~Fidler, and A.~Torralba.
\newblock {VirtualHome}: Simulating household activities via programs.
\newblock In \emph{IEEE/CVF Conference on Computer Vision and Pattern Recognition}, pages 8494--8502, 2018.

\bibitem[Puig et~al.(2024)Puig, Undersander, Szot, Cote, Yang, Partsey, Desai, Clegg, Hlavac, Min, Vondru{\v{s}}, Gervet, Berges, Turner, Maksymets, Kira, Kalakrishnan, Malik, Chaplot, Jain, Batra, Rai, and Mottaghi]{puig2023habitat3}
X.~Puig, E.~Undersander, A.~Szot, M.~D. Cote, T.-Y. Yang, R.~Partsey, R.~Desai, A.~W. Clegg, M.~Hlavac, S.~Y. Min, V.~Vondru{\v{s}}, T.~Gervet, V.-P. Berges, J.~M. Turner, O.~Maksymets, Z.~Kira, M.~Kalakrishnan, J.~Malik, D.~S. Chaplot, U.~Jain, D.~Batra, A.~Rai, and R.~Mottaghi.
\newblock Habitat 3.0: A co-habitat for humans, avatars, and robots.
\newblock In \emph{International Conference on Learning Representations}, 2024.

\bibitem[Puterman(1994)]{puterman1994mdp}
M.~L. Puterman.
\newblock \emph{Markov Decision Processes: Discrete Stochastic Dynamic Programming}.
\newblock John Wiley \& Sons, 1994.

\bibitem[Qian et~al.(2024)Qian, Liu, Liu, Chen, Dang, Li, Yang, Chen, Su, Cong, Xu, Li, Liu, and Sun]{qian2024chatdev}
C.~Qian, W.~Liu, H.~Liu, N.~Chen, Y.~Dang, J.~Li, C.~Yang, W.~Chen, Y.~Su, X.~Cong, J.~Xu, D.~Li, Z.~Liu, and M.~Sun.
\newblock {ChatDev}: Communicative agents for software development.
\newblock In \emph{Annual Meeting of the Association for Computational Linguistics}, pages 15174--15186, 2024.

\bibitem[Qian et~al.(2026)Qian, Acikgoz, Li, Chen, Zhang, He, Luo, Hakkani-T{\"u}r, Tur, Li, and Ji]{qian2026foresight}
C.~Qian, E.~C. Acikgoz, B.~Li, X.~Chen, Y.~Zhang, B.~He, Q.~Luo, D.~Hakkani-T{\"u}r, G.~Tur, Y.~Li, and H.~Ji.
\newblock Current agents fail to leverage world model as tool for foresight.
\newblock \emph{arXiv preprint arXiv:2601.03905}, 2026.

\bibitem[Qiao et~al.(2024)Qiao, Fang, Zhang, Zhu, Chen, Deng, Jiang, Xie, Huang, and Chen]{qiao2024wkm}
S.~Qiao, R.~Fang, N.~Zhang, Y.~Zhu, X.~Chen, S.~Deng, Y.~Jiang, P.~Xie, F.~Huang, and H.~Chen.
\newblock Agent planning with world knowledge model.
\newblock In \emph{Advances in Neural Information Processing Systems}, volume~37, pages 114843--114871, 2024.

\bibitem[Qin et~al.(2024)Qin, Shi, Yu, Wang, Zhou, Li, Yin, Liu, Sheng, Shao, Bai, Ouyang, and Zhang]{qin2025worldsimbench}
Y.~Qin, Z.~Shi, J.~Yu, X.~Wang, E.~Zhou, L.~Li, Z.~Yin, X.~Liu, L.~Sheng, J.~Shao, L.~Bai, W.~Ouyang, and R.~Zhang.
\newblock {WorldSimBench}: Towards video generation models as world simulators.
\newblock \emph{arXiv preprint arXiv:2410.18072}, 2024.

\bibitem[Qin et~al.(2025)Qin, Ye, Fang, Wang, Liang, Tian, Zhang, Li, Li, Huang, Zhong, Li, Yang, Miao, Lin, Liu, Jiang, Ma, Li, Xiao, Cai, Li, Zheng, Jin, Li, Zhou, Wang, Chen, Li, Yang, Liu, Lin, Peng, Liu, and Shi]{qin2025ui}
Y.~Qin, Y.~Ye, J.~Fang, H.~Wang, S.~Liang, S.~Tian, J.~Zhang, J.~Li, Y.~Li, S.~Huang, W.~Zhong, K.~Li, J.~Yang, Y.~Miao, W.~Lin, L.~Liu, X.~Jiang, Q.~Ma, J.~Li, X.~Xiao, K.~Cai, C.~Li, Y.~Zheng, C.~Jin, C.~Li, X.~Zhou, M.~Wang, H.~Chen, Z.~Li, H.~Yang, H.~Liu, F.~Lin, T.~Peng, X.~Liu, and G.~Shi.
\newblock {UI-TARS}: Pioneering automated gui interaction with native agents.
\newblock \emph{arXiv preprint arXiv:2501.12326}, 2025.

\bibitem[Quine(1951)]{quine1951two}
W.~V.~O. Quine.
\newblock Two dogmas of empiricism.
\newblock \emph{The Philosophical Review}, 60\penalty0 (1):\penalty0 20--43, 1951.

\bibitem[Rabinowitz et~al.(2018)Rabinowitz, Perbet, Song, Zhang, Eslami, and Botvinick]{rabinowitz2018tomnet}
N.~Rabinowitz, F.~Perbet, F.~Song, C.~Zhang, S.~A. Eslami, and M.~Botvinick.
\newblock Machine theory of mind.
\newblock In \emph{International Conference on Machine Learning}, pages 4218--4227. PMLR, 2018.

\bibitem[Rajasekaran(2026)]{rajasekaran2026harness}
P.~Rajasekaran.
\newblock Harness design for long-running application development.
\newblock Anthropic Engineering Blog, 2026.
\newblock URL \url{https://www.anthropic.com/engineering/harness-design-long-running-apps}.

\bibitem[Rao and Georgeff(1995)]{rao1995bdi}
A.~S. Rao and M.~P. Georgeff.
\newblock {BDI} agents: From theory to practice.
\newblock In \emph{International Conference on Multi-Agent Systems}, pages 312--319, 1995.

\bibitem[Rao and Ballard(1999)]{rao1999predictive}
R.~P.~N. Rao and D.~H. Ballard.
\newblock Predictive coding in the visual cortex: a functional interpretation of some extra-classical receptive-field effects.
\newblock \emph{Nature Neuroscience}, 2\penalty0 (1):\penalty0 79--87, 1999.

\bibitem[Rawles et~al.(2025)Rawles, Clinckemaillie, Chang, Waltz, Lau, Fair, Li, Bishop, Li, Campbell-Ajala, Toyama, Berry, Tyamagundlu, Lillicrap, and Riva]{rawles2024androidworld}
C.~Rawles, S.~Clinckemaillie, Y.~Chang, J.~Waltz, G.~Lau, M.~Fair, A.~Li, W.~Bishop, W.~Li, F.~Campbell-Ajala, D.~Toyama, R.~Berry, D.~Tyamagundlu, T.~Lillicrap, and O.~Riva.
\newblock {AndroidWorld}: A dynamic benchmarking environment for autonomous agents.
\newblock In \emph{International Conference on Learning Representations}, 2025.

\bibitem[Ren et~al.(2023)Ren, Dai, Burchfiel, and Majumdar]{ren2023adaptsim}
A.~Z. Ren, H.~Dai, B.~Burchfiel, and A.~Majumdar.
\newblock {AdaptSim}: Task-driven simulation adaptation for sim-to-real transfer.
\newblock In \emph{Conference on Robot Learning}, 2023.

\bibitem[Ren et~al.(2026)Ren, Yao, Sun, Qiao, Zhang, and Chen]{ren2026aligning}
B.~Ren, Y.~Yao, R.~Sun, S.~Qiao, N.~Zhang, and H.~Chen.
\newblock Aligning agentic world models via knowledgeable experience learning.
\newblock \emph{arXiv preprint arXiv:2601.13247}, 2026.

\bibitem[Ren et~al.(2024)Ren, Cui, Song, Wang, and Hu]{ren2024norms}
S.~Ren, Z.~Cui, R.~Song, Z.~Wang, and S.~Hu.
\newblock Emergence of social norms in large language model-based agent societies.
\newblock \emph{arXiv preprint arXiv:2403.08251}, 2024.

\bibitem[Ren et~al.(2025)Ren, Zhang, Qian, Gao, Shi, Zheng, and He]{ren2025gtm}
Z.~Ren, X.~Zhang, Z.~Qian, Y.~Gao, Y.~Shi, S.~Zheng, and J.~He.
\newblock {GTM}: Simulating the world of tools for {AI} agents.
\newblock \emph{arXiv preprint arXiv:2512.04535}, 2025.

\bibitem[Rezende et~al.(2014)Rezende, Mohamed, and Wierstra]{rezende2014stochastic}
D.~J. Rezende, S.~Mohamed, and D.~Wierstra.
\newblock Stochastic backpropagation and approximate inference in deep generative models.
\newblock In \emph{International Conference on Machine Learning}, pages 1278--1286. PMLR, 2014.

\bibitem[Rivard et~al.(2025)Rivard, Sun, Guo, Chen, and Deng]{rivard2025neuralos}
L.~Rivard, S.~Sun, H.~Guo, W.~Chen, and Y.~Deng.
\newblock {NeuralOS}: Towards simulating operating systems via neural generative models.
\newblock \emph{arXiv preprint arXiv:2507.08800}, 2025.

\bibitem[Rombach et~al.(2022)Rombach, Blattmann, Lorenz, Esser, and Ommer]{rombach2022latentdiffusion}
R.~Rombach, A.~Blattmann, D.~Lorenz, P.~Esser, and B.~Ommer.
\newblock High-resolution image synthesis with latent diffusion models.
\newblock In \emph{IEEE/CVF Conference on Computer Vision and Pattern Recognition}, pages 10684--10695, 2022.

\bibitem[Romera-Paredes et~al.(2024)Romera-Paredes, Barekatain, Novikov, Balog, Kumar, Dupont, Ruiz, Ellenberg, Wang, Fawzi, Kohli, and Fawzi]{romera2024funsearch}
B.~Romera-Paredes, M.~Barekatain, A.~Novikov, M.~Balog, M.~P. Kumar, E.~Dupont, F.~J.~R. Ruiz, J.~S. Ellenberg, P.~Wang, O.~Fawzi, P.~Kohli, and A.~Fawzi.
\newblock Mathematical discoveries from program search with large language models.
\newblock \emph{Nature}, 625:\penalty0 468--475, 2024.

\bibitem[Rumelhart et~al.(1986)Rumelhart, Hinton, and Williams]{rumelhart1986learning}
D.~E. Rumelhart, G.~E. Hinton, and R.~J. Williams.
\newblock Learning representations by back-propagating errors.
\newblock \emph{Nature}, 323\penalty0 (6088):\penalty0 533--536, 1986.

\bibitem[Russell et~al.(2025)Russell, Hu, Bertoni, Fedoseev, Shotton, Arani, and Corrado]{nvidia2025gaia2}
L.~Russell, A.~Hu, L.~Bertoni, G.~Fedoseev, J.~Shotton, E.~Arani, and G.~Corrado.
\newblock {GAIA-2}: A controllable multi-view generative world model for autonomous driving.
\newblock \emph{arXiv preprint arXiv:2503.20523}, 2025.

\bibitem[Sabour et~al.(2024)Sabour, Fidler, and Kreis]{sabour2024align}
A.~Sabour, S.~Fidler, and K.~Kreis.
\newblock Align your steps: Optimizing sampling schedules in diffusion models.
\newblock \emph{arXiv preprint arXiv:2404.14507}, 2024.

\bibitem[Salemi et~al.(2024)Salemi, Mysore, Bendersky, and Zamani]{salemi2023lamp}
A.~Salemi, S.~Mysore, M.~Bendersky, and H.~Zamani.
\newblock {LaMP}: When large language models meet personalization.
\newblock In \emph{Annual Meeting of the Association for Computational Linguistics}, pages 7370--7392, 2024.

\bibitem[Salimans and Ho(2022)]{salimans2022progressive}
T.~Salimans and J.~Ho.
\newblock Progressive distillation for fast sampling of diffusion models.
\newblock In \emph{International Conference on Learning Representations}, 2022.

\bibitem[Salimans et~al.(2024)Salimans, Mensink, Heek, and Hoogeboom]{salimans2024multistep}
T.~Salimans, T.~Mensink, J.~Heek, and E.~Hoogeboom.
\newblock Multistep distillation of diffusion models via moment matching.
\newblock In \emph{Advances in Neural Information Processing Systems}, volume~37, pages 36046--36070, 2024.

\bibitem[Samuel et~al.(2025)Samuel, Zou, Zhou, Chaudhari, Kalyan, Rajpurohit, Deshpande, Narasimhan, and Murahari]{samuel2024personagym}
V.~Samuel, H.~P. Zou, Y.~Zhou, S.~Chaudhari, A.~Kalyan, T.~Rajpurohit, A.~Deshpande, K.~Narasimhan, and V.~Murahari.
\newblock {PersonaGym}: Evaluating persona agents and {LLMs}.
\newblock In \emph{Conference on Empirical Methods in Natural Language Processing}, page 6999–7022, 2025.

\bibitem[Sanchez-Gonzalez et~al.(2020)Sanchez-Gonzalez, Godwin, Pfaff, Ying, Leskovec, and Battaglia]{sanchez2020gns}
A.~Sanchez-Gonzalez, J.~Godwin, T.~Pfaff, R.~Ying, J.~Leskovec, and P.~Battaglia.
\newblock Learning to simulate complex physics with graph networks.
\newblock In \emph{International Conference on Machine Learning}, pages 8459--8468. PMLR, 2020.

\bibitem[Sanh et~al.(2020)Sanh, Wolf, and Rush]{sanh2020movement}
V.~Sanh, T.~Wolf, and A.~Rush.
\newblock Movement pruning: Adaptive sparsity by fine-tuning.
\newblock In \emph{Advances in Neural Information Processing Systems}, volume~33, pages 20378--20389, 2020.

\bibitem[Sauer et~al.(2024{\natexlab{a}})Sauer, Boesel, Dockhorn, Blattmann, Esser, and Rombach]{sauer2024fast}
A.~Sauer, F.~Boesel, T.~Dockhorn, A.~Blattmann, P.~Esser, and R.~Rombach.
\newblock Fast high-resolution image synthesis with latent adversarial diffusion distillation.
\newblock In \emph{SIGGRAPH Asia}, pages 1--11, 2024{\natexlab{a}}.

\bibitem[Sauer et~al.(2024{\natexlab{b}})Sauer, Lorenz, Blattmann, and Rombach]{sauer2024adversarial}
A.~Sauer, D.~Lorenz, A.~Blattmann, and R.~Rombach.
\newblock Adversarial diffusion distillation.
\newblock In \emph{European Conference on Computer Vision}, pages 87--103. Springer, 2024{\natexlab{b}}.

\bibitem[Savva et~al.(2019)Savva, Kadian, Maksymets, Zhao, Wijmans, Jain, Straub, Liu, Koltun, Malik, Parikh, and Batra]{savva2019habitat}
M.~Savva, A.~Kadian, O.~Maksymets, Y.~Zhao, E.~Wijmans, B.~Jain, J.~Straub, J.~Liu, V.~Koltun, J.~Malik, D.~Parikh, and D.~Batra.
\newblock Habitat: A platform for embodied {AI} research.
\newblock In \emph{IEEE/CVF International Conference on Computer Vision}, pages 9339--9347, 2019.

\bibitem[Schrittwieser et~al.(2020)Schrittwieser, Antonoglou, Hubert, Simonyan, Sifre, Schmitt, Guez, Lockhart, Hassabis, Graepel, Lillicrap, and Silver]{schrittwieser2020muzero}
J.~Schrittwieser, I.~Antonoglou, T.~Hubert, K.~Simonyan, L.~Sifre, S.~Schmitt, A.~Guez, E.~Lockhart, D.~Hassabis, T.~Graepel, T.~Lillicrap, and D.~Silver.
\newblock Mastering {A}tari, {G}o, chess and shogi by planning with a learned model.
\newblock \emph{Nature}, 588\penalty0 (7839):\penalty0 604--609, 2020.

\bibitem[Sch{\"u}tt et~al.(2017)Sch{\"u}tt, Kindermans, Sauceda, Chmiela, Tkatchenko, and M{\"u}ller]{schutt2017schnet}
K.~T. Sch{\"u}tt, P.-J. Kindermans, H.~E. Sauceda, S.~Chmiela, A.~Tkatchenko, and K.-R. M{\"u}ller.
\newblock {SchNet}: A continuous-filter convolutional neural network for modeling quantum interactions.
\newblock In \emph{Advances in Neural Information Processing Systems}, volume~30, pages 992--1002, 2017.

\bibitem[Schwarzer et~al.(2021)Schwarzer, Anand, Goel, Hjelm, Courville, and Bachman]{schwarzer2021spr}
M.~Schwarzer, A.~Anand, R.~Goel, R.~D. Hjelm, A.~Courville, and P.~Bachman.
\newblock Data-efficient reinforcement learning with self-predictive representations.
\newblock In \emph{International Conference on Learning Representations}, 2021.

\bibitem[Sclar et~al.(2023)Sclar, Kumar, West, Suhr, Choi, and Tsvetkov]{sclar2023symbolictom}
M.~Sclar, S.~Kumar, P.~West, A.~Suhr, Y.~Choi, and Y.~Tsvetkov.
\newblock Minding language models’(lack of) theory of mind: A plug-and-play multi-character belief tracker.
\newblock In \emph{Annual Meeting of the Association for Computational Linguistics}, pages 13960--13980, 2023.

\bibitem[Sclar et~al.(2024)Sclar, Yu, Fazel-Zarandi, Tsvetkov, Bisk, Choi, and Celikyilmaz]{sclar2024exploretom}
M.~Sclar, J.~Yu, M.~Fazel-Zarandi, Y.~Tsvetkov, Y.~Bisk, Y.~Choi, and A.~Celikyilmaz.
\newblock Explore theory of mind: Program-guided adversarial data generation for theory of mind reasoning.
\newblock \emph{arXiv preprint arXiv:2412.12175}, 2024.

\bibitem[Sekar et~al.(2020)Sekar, Rybkin, Daniilidis, Abbeel, Hafner, and Pathak]{sekar2020plan2explore}
R.~Sekar, O.~Rybkin, K.~Daniilidis, P.~Abbeel, D.~Hafner, and D.~Pathak.
\newblock Planning to explore via self-supervised world models.
\newblock In \emph{International Conference on Machine Learning}, pages 8583--8592. PMLR, 2020.

\bibitem[Senior et~al.(2020)Senior, Evans, Jumper, Kirkpatrick, Sifre, Green, Qin, {\v{Z}}{\'\i}dek, Nelson, Bridgland, Penedones, Petersen, Simonyan, Crossan, Kohli, Jones, Silver, Kavukcuoglu, and Hassabis]{senior2020alphafold}
A.~W. Senior, R.~Evans, J.~Jumper, J.~Kirkpatrick, L.~Sifre, T.~Green, C.~Qin, A.~{\v{Z}}{\'\i}dek, A.~W.~R. Nelson, A.~Bridgland, H.~Penedones, S.~Petersen, K.~Simonyan, S.~Crossan, P.~Kohli, D.~T. Jones, D.~Silver, K.~Kavukcuoglu, and D.~Hassabis.
\newblock Improved protein structure prediction using potentials from deep learning.
\newblock \emph{Nature}, 577:\penalty0 706--710, 2020.

\bibitem[Shaj et~al.(2023)Shaj, Zadeh, Demir, Douat, and Neumann]{shaj2023multitimescale}
V.~Shaj, S.~G. Zadeh, O.~Demir, L.~R. Douat, and G.~Neumann.
\newblock Multi time scale world models.
\newblock In \emph{Advances in Neural Information Processing Systems}, volume~36, pages 26764--26775, 2023.

\bibitem[Shanahan(1997)]{shanahan1997frame}
M.~Shanahan.
\newblock \emph{Solving the Frame Problem: A Mathematical Investigation of the Common Sense Law of Inertia}.
\newblock MIT Press, 1997.

\bibitem[Shanahan et~al.(2023)Shanahan, McDonell, and Reynolds]{shanahan2023roleplay}
M.~Shanahan, K.~McDonell, and L.~Reynolds.
\newblock Role play with large language models.
\newblock \emph{Nature}, 623:\penalty0 493--498, 2023.

\bibitem[Shang et~al.(2023)Shang, Yuan, Xie, Wu, and Yan]{shang2023post}
Y.~Shang, Z.~Yuan, B.~Xie, B.~Wu, and Y.~Yan.
\newblock Post-training quantization on diffusion models.
\newblock In \emph{IEEE/CVF Conference on Computer Vision and Pattern Recognition}, pages 1972--1981, 2023.

\bibitem[Shang et~al.(2025)Shang, Zhang, Tang, Jin, Gao, Wu, and Li]{chen2025roboscape}
Y.~Shang, X.~Zhang, Y.~Tang, L.~Jin, C.~Gao, W.~Wu, and Y.~Li.
\newblock {RoboScape}: Physics-informed embodied world model.
\newblock \emph{arXiv preprint arXiv:2506.23135}, 2025.

\bibitem[Shao et~al.(2023)Shao, Li, Dai, and Qiu]{shao2023characterllm}
Y.~Shao, L.~Li, J.~Dai, and X.~Qiu.
\newblock {Character-LLM}: A trainable agent for role-playing.
\newblock In \emph{Conference on Empirical Methods in Natural Language Processing}, pages 13153--13187, 2023.

\bibitem[Shen et~al.(2026)Shen, Hu, Li, Fang, Li, and Zhang]{wac2026}
Z.~Shen, X.~Hu, X.~Li, T.~Fang, J.~Li, and S.~Zhang.
\newblock World-model-augmented web agents with action correction.
\newblock \emph{arXiv preprint arXiv:2602.15384}, 2026.

\bibitem[Shi et~al.(2017)Shi, Karpathy, Fan, Hernandez, and Liang]{shi2017worldofbits}
T.~Shi, A.~Karpathy, L.~Fan, J.~Hernandez, and P.~Liang.
\newblock World of bits: An open-domain platform for web-based agents.
\newblock In \emph{International Conference on Machine Learning}, pages 3135--3144, 2017.

\bibitem[Singh et~al.(2024)Singh, Farrell-Maupin, and Faghihi]{singh2024opal}
P.~K. Singh, K.~A. Farrell-Maupin, and D.~Faghihi.
\newblock A framework for strategic discovery of credible neural network surrogate models under uncertainty.
\newblock \emph{Computer Methods in Applied Mechanics and Engineering}, 427:\penalty0 117061, 2024.

\bibitem[Song et~al.(2023{\natexlab{a}})Song, Yao, Fan, Dong, Chen, Niebles, Xing, and Zhang]{song2023nctrl}
X.~Song, W.~Yao, Y.~Fan, X.~Dong, G.~Chen, J.~C. Niebles, E.~Xing, and K.~Zhang.
\newblock Temporally disentangled representation learning under unknown nonstationarity.
\newblock In \emph{Advances in Neural Information Processing Systems}, volume~36, pages 8092--8113, 2023{\natexlab{a}}.

\bibitem[Song and Dhariwal(2024)]{song2023improved}
Y.~Song and P.~Dhariwal.
\newblock Improved techniques for training consistency models.
\newblock In \emph{International Conference on Learning Representations}, 2024.

\bibitem[Song et~al.(2023{\natexlab{b}})Song, Dhariwal, Chen, and Sutskever]{song2023consistency}
Y.~Song, P.~Dhariwal, M.~Chen, and I.~Sutskever.
\newblock Consistency models.
\newblock In \emph{International Conference on Machine Learning}, volume 202, pages 32211--32252. PMLR, 2023{\natexlab{b}}.

\bibitem[Sparkes et~al.(2010)Sparkes, Aubrey, Byrne, Clare, Khan, Liakata, Markham, Rowland, Soldatova, Whelan, Young, and King]{sparkes2010robotscientist}
A.~Sparkes, W.~Aubrey, E.~Byrne, A.~Clare, M.~N. Khan, M.~Liakata, M.~Markham, J.~Rowland, L.~N. Soldatova, K.~E. Whelan, M.~Young, and R.~D. King.
\newblock Towards robot scientists for autonomous scientific discovery.
\newblock \emph{Automated Experimentation}, 2\penalty0 (1):\penalty0 1, 2010.

\bibitem[Stalnaker(1968)]{stalnaker1968conditionals}
R.~C. Stalnaker.
\newblock A theory of conditionals.
\newblock In \emph{Studies in Logical Theory}, volume~2 of \emph{American Philosophical Quarterly Monograph Series}, pages 98--112. Blackwell, 1968.

\bibitem[Stani{\'c} et~al.(2023)Stani{\'c}, Tang, Ha, and Schmidhuber]{stanic2023learning}
A.~Stani{\'c}, Y.~Tang, D.~Ha, and J.~Schmidhuber.
\newblock Learning to generalize with object-centric agents in the open world survival game crafter.
\newblock \emph{IEEE Transactions on Games}, 16\penalty0 (2):\penalty0 384--395, 2023.

\bibitem[Strieth-Kalthoff et~al.(2024)Strieth-Kalthoff, Hao, Rathore, Derasp, Gaudin, Angello, Seifrid, Trushina, Guy, Liu, Tang, Mamada, et~al.]{striethkalthoff2024sdl}
F.~Strieth-Kalthoff, H.~Hao, V.~Rathore, J.~Derasp, T.~Gaudin, N.~H. Angello, M.~Seifrid, E.~Trushina, M.~Guy, J.~Liu, X.~Tang, M.~Mamada, et~al.
\newblock Delocalized, asynchronous, closed-loop discovery of organic laser emitters.
\newblock \emph{Science}, 384\penalty0 (6697):\penalty0 eadk9227, 2024.

\bibitem[Su et~al.(2025{\natexlab{a}})Su, Wang, Ren, Lin, and Chen]{wang2025pixelreasoner}
A.~Su, H.~Wang, W.~Ren, F.~Lin, and W.~Chen.
\newblock Pixel reasoner: Incentivizing pixel-space reasoning with curiosity-driven reinforcement learning.
\newblock \emph{arXiv preprint arXiv:2505.15966}, 2025{\natexlab{a}}.

\bibitem[Su et~al.(2025{\natexlab{b}})Su, Chen, Shen, Wei, Li, Yu, and Yuan]{su2025rotatekv}
Z.~Su, Z.~Chen, W.~Shen, H.~Wei, L.~Li, H.~Yu, and K.~Yuan.
\newblock {RotateKV}: Accurate and robust 2-bit {KV} cache quantization for {LLMs} via outlier-aware adaptive rotations.
\newblock \emph{arXiv preprint arXiv:2501.16383}, 2025{\natexlab{b}}.

\bibitem[Sumers et~al.(2024)Sumers, Yao, Narasimhan, and Griffiths]{sumers2024coala}
T.~R. Sumers, S.~Yao, K.~Narasimhan, and T.~L. Griffiths.
\newblock Cognitive architectures for language agents.
\newblock \emph{Transactions on Machine Learning Research}, 2024.

\bibitem[Sun et~al.(2025{\natexlab{a}})Sun, Yang, Tang, Huang, Xu, Chen, Liu, Yang, Zhu, Wang, He, Chen, Dai, Ye, and Gu]{sun2025learningprimitiveembodiedworld}
Q.~Sun, L.~Yang, W.~Tang, W.~Huang, K.~Xu, Y.~Chen, M.~Liu, J.~Yang, H.~Zhu, Y.~Wang, T.~He, Y.~Chen, X.~Dai, N.~Ye, and Q.~Gu.
\newblock Learning primitive embodied world models: Towards scalable robotic learning.
\newblock \emph{arXiv preprint arXiv:2508.20840}, 2025{\natexlab{a}}.

\bibitem[Sun et~al.(2026)Sun, Song, Huang, Jiang, Le, Lv, Chen, Hu, Luo, Zhao, et~al.]{sweworld}
S.~Sun, H.~Song, L.~Huang, J.~Jiang, R.~Le, Z.~Lv, Z.~Chen, Y.~Hu, W.~Luo, W.~X. Zhao, et~al.
\newblock Swe-world: Building software engineering agents in docker-free environments.
\newblock \emph{arXiv preprint arXiv:2602.03419}, 2026.

\bibitem[Sun et~al.(2025{\natexlab{b}})Sun, Zhang, Wang, Wu, Wang, Wang, Wang, Zhang, Wang, and Guo]{sun2025worldplay}
W.~Sun, H.~Zhang, H.~Wang, J.~Wu, Z.~Wang, Z.~Wang, Y.~Wang, J.~Zhang, T.~Wang, and C.~Guo.
\newblock {WorldPlay}: Towards long-term geometric consistency for real-time interactive world modeling.
\newblock \emph{arXiv preprint arXiv:2512.14614}, 2025{\natexlab{b}}.

\bibitem[Sutton(1991)]{sutton1991dyna}
R.~S. Sutton.
\newblock Dyna, an integrated architecture for learning, planning, and reacting.
\newblock \emph{ACM SIGART Bulletin}, 2\penalty0 (4):\penalty0 160--163, 1991.

\bibitem[Szot et~al.(2021)Szot, Clegg, Undersander, Wijmans, Zhao, Turner, Maestre, Mukadam, Chaplot, Maksymets, Gokaslan, Vondrus, Dharur, Meier, Galuba, Chang, Kira, Koltun, Malik, Savva, and Batra]{szot2021habitat20}
A.~Szot, A.~Clegg, E.~Undersander, E.~Wijmans, Y.~Zhao, J.~Turner, N.~Maestre, M.~Mukadam, D.~Chaplot, O.~Maksymets, A.~Gokaslan, V.~Vondrus, S.~Dharur, F.~Meier, W.~Galuba, A.~Chang, Z.~Kira, V.~Koltun, J.~Malik, M.~Savva, and D.~Batra.
\newblock Habitat 2.0: Training home assistants to rearrange their habitat.
\newblock In \emph{Advances in Neural Information Processing Systems}, volume~34, pages 251--266, 2021.

\bibitem[Szymanski et~al.(2023)Szymanski, Rendy, Fei, Kumar, He, Milsted, McDermott, Gallant, Cubuk, Merchant, Kim, Jain, Bartel, Persson, Zeng, and Ceder]{szymanski2023alab}
N.~J. Szymanski, B.~Rendy, Y.~Fei, R.~E. Kumar, T.~He, D.~Milsted, M.~J. McDermott, M.~Gallant, E.~D. Cubuk, A.~Merchant, H.~Kim, A.~Jain, C.~J. Bartel, K.~Persson, Y.~Zeng, and G.~Ceder.
\newblock An autonomous laboratory for the accelerated synthesis of inorganic materials.
\newblock \emph{Nature}, 624:\penalty0 86--91, 2023.

\bibitem[Tang et~al.(2024)Tang, Key, and Ellis]{tang2024worldcoder}
H.~Tang, D.~Key, and K.~Ellis.
\newblock {WorldCoder}, a model-based {LLM} agent: Building world models by writing code and interacting with the environment.
\newblock In \emph{Advances in Neural Information Processing Systems}, volume~37, pages 70148--70212, 2024.

\bibitem[Tao et~al.(2024)Tao, Xiang, Shukla, Qin, Hinrichsen, Yuan, Bao, Lin, Liu, kai Chan, Gao, Li, Mu, Xiao, Gurha, Rajesh, Choi, Chen, Huang, Calandra, Chen, Luo, and Su]{tao2024maniskill3}
S.~Tao, F.~Xiang, A.~Shukla, Y.~Qin, X.~Hinrichsen, X.~Yuan, C.~Bao, X.~Lin, Y.~Liu, T.~kai Chan, Y.~Gao, X.~Li, T.~Mu, N.~Xiao, A.~Gurha, V.~N. Rajesh, Y.~W. Choi, Y.-R. Chen, Z.~Huang, R.~Calandra, R.~Chen, S.~Luo, and H.~Su.
\newblock {ManiSkill3}: {GPU} parallelized robotics simulation and rendering for generalizable embodied ai.
\newblock \emph{arXiv preprint arXiv:2410.00425}, 2024.

\bibitem[Tassa et~al.(2020)Tassa, Tunyasuvunakool, Muldal, Doron, Trochim, Liu, Bohez, Merel, Erez, Lillicrap, and Heess]{tassa2020dmcontrol}
Y.~Tassa, S.~Tunyasuvunakool, A.~Muldal, Y.~Doron, P.~Trochim, S.~Liu, S.~Bohez, J.~Merel, T.~Erez, T.~Lillicrap, and N.~Heess.
\newblock dm\_control: Software and tasks for continuous control.
\newblock \emph{Software Impacts}, 6:\penalty0 100022, 2020.

\bibitem[Taubenfeld et~al.(2024)Taubenfeld, Dover, Reichart, and Goldstein]{taubenfeld2024biases}
A.~Taubenfeld, Y.~Dover, R.~Reichart, and A.~Goldstein.
\newblock Systematic biases in {LLM} simulations of debates.
\newblock In \emph{Conference on Empirical Methods in Natural Language Processing}, pages 251--267, 2024.

\bibitem[Telang et~al.(2021)Telang, Singh, and Yorke-Smith]{telang2023commitments}
P.~R. Telang, M.~P. Singh, and N.~Yorke-Smith.
\newblock Maintenance of social commitments in multiagent systems.
\newblock In \emph{AAAI Conference on Artificial Intelligence}, volume~35, pages 11369--11377, 2021.

\bibitem[Tobin et~al.(2017)Tobin, Fong, Ray, Schneider, Zaremba, and Abbeel]{tobin2017domain}
J.~Tobin, R.~Fong, A.~Ray, J.~Schneider, W.~Zaremba, and P.~Abbeel.
\newblock Domain randomization for transferring deep neural networks from simulation to the real world.
\newblock In \emph{IEEE/RSJ International Conference on Intelligent Robots and Systems}, 2017.

\bibitem[Todorov et~al.(2012)Todorov, Erez, and Tassa]{todorov2012mujoco}
E.~Todorov, T.~Erez, and Y.~Tassa.
\newblock {MuJoCo}: A physics engine for model-based control.
\newblock In \emph{IEEE/RSJ International Conference on Intelligent Robots and Systems}, pages 5026--5033, 2012.

\bibitem[Tu et~al.(2025)Tu, Zhou, Liang, Jiang, Zhang, Li, and Bai]{tu2025wm_ad_survey}
S.~Tu, X.~Zhou, D.~Liang, X.~Jiang, Y.~Zhang, X.~Li, and X.~Bai.
\newblock The role of world models in shaping autonomous driving: A comprehensive survey.
\newblock \emph{arXiv preprint arXiv:2502.10498}, 2025.

\bibitem[Turing(1950)]{turing1950computing}
A.~M. Turing.
\newblock Computing machinery and intelligence.
\newblock \emph{Mind}, 59\penalty0 (236):\penalty0 433--460, 1950.

\bibitem[Unterthiner et~al.(2018)Unterthiner, van Steenkiste, Kurach, Marinier, Michalski, and Gelly]{unterthiner2018fvd}
T.~Unterthiner, S.~van Steenkiste, K.~Kurach, R.~Marinier, M.~Michalski, and S.~Gelly.
\newblock Towards accurate generative models of video: A new metric \& challenges.
\newblock \emph{arXiv preprint arXiv:1812.01717}, 2018.

\bibitem[Vafa et~al.(2024)Vafa, Chen, Rambachan, Kleinberg, and Mullainathan]{vafa2024evaluating}
K.~Vafa, J.~Y. Chen, A.~Rambachan, J.~Kleinberg, and S.~Mullainathan.
\newblock Evaluating the world model implicit in a generative model.
\newblock In \emph{Advances in Neural Information Processing Systems}, volume~37, pages 26941--26975, 2024.

\bibitem[Valevski et~al.(2025)Valevski, Leviathan, Arar, and Fruchter]{valevski2025gamengin}
D.~Valevski, Y.~Leviathan, M.~Arar, and S.~Fruchter.
\newblock Diffusion models are real-time game engines.
\newblock In \emph{International Conference on Learning Representations}, 2025.

\bibitem[Vallinder and Hughes(2025)]{vallinder2024culturalevolution}
A.~Vallinder and E.~Hughes.
\newblock Cultural evolution of cooperation among {LLM} agents.
\newblock In \emph{International Conference on Autonomous Agents and Multiagent Systems}, pages 2771--2773, 2025.

\bibitem[van~de Ven et~al.(2024)van~de Ven, Soures, and Kudithipudi]{vandeven2024continuallearning}
G.~M. van~de Ven, N.~Soures, and D.~Kudithipudi.
\newblock Continual learning and catastrophic forgetting.
\newblock \emph{arXiv preprint arXiv:2403.05175}, 2024.

\bibitem[van~den Oord et~al.(2017)van~den Oord, Vinyals, and Kavukcuoglu]{oord2017vqvae}
A.~van~den Oord, O.~Vinyals, and K.~Kavukcuoglu.
\newblock Neural discrete representation learning.
\newblock In \emph{Advances in Neural Information Processing Systems}, 2017.

\bibitem[van Es et~al.(2025)van Es, Higgins, Gohil, Quinn, Vidaurre, and Woolrich]{van2025large}
M.~W. van Es, C.~Higgins, C.~Gohil, A.~J. Quinn, D.~Vidaurre, and M.~W. Woolrich.
\newblock Large-scale cortical functional networks are organized in structured cycles.
\newblock \emph{Nature Neuroscience}, 28\penalty0 (10):\penalty0 2118--2128, 2025.

\bibitem[Vaswani et~al.(2017)Vaswani, Shazeer, Parmar, Uszkoreit, Jones, Gomez, Kaiser, and Polosukhin]{vaswani2017attention}
A.~Vaswani, N.~Shazeer, N.~Parmar, J.~Uszkoreit, L.~Jones, A.~N. Gomez, L.~Kaiser, and I.~Polosukhin.
\newblock Attention is all you need.
\newblock In \emph{Advances in Neural Information Processing Systems}, 2017.

\bibitem[Vidaurre et~al.(2018)Vidaurre, Hunt, Quinn, Hunt, Brookes, Nobre, and Woolrich]{vidaurre2018spontaneous}
D.~Vidaurre, L.~T. Hunt, A.~J. Quinn, B.~A. Hunt, M.~J. Brookes, A.~C. Nobre, and M.~W. Woolrich.
\newblock Spontaneous cortical activity transiently organises into frequency specific phase-coupling networks.
\newblock \emph{Nature Communications}, 9\penalty0 (1):\penalty0 2987, 2018.

\bibitem[Wang et~al.(2025{\natexlab{a}})Wang, Wang, Chen, Liu, Xue, Peng, Qi, Lin, and Yan]{wang2025virl}
C.~Wang, H.~Wang, X.~Chen, J.~Liu, T.~Xue, C.~Peng, D.~Qi, F.~Lin, and Y.~Yan.
\newblock From illusion to intention: Visual rationale learning for vision-language reasoning.
\newblock \emph{arXiv preprint arXiv:2511.23031}, 2025{\natexlab{a}}.

\bibitem[Wang et~al.(2024{\natexlab{a}})Wang, Huang, Bergman, Shen, Gao, Lingelbach, Sun, Bian, Song, Liu, Wang, and Li]{wang2024phased}
F.-Y. Wang, Z.~Huang, A.~W. Bergman, D.~Shen, P.~Gao, M.~Lingelbach, K.~Sun, W.~Bian, G.~Song, Y.~Liu, X.~Wang, and H.~Li.
\newblock Phased consistency models.
\newblock In \emph{Advances in Neural Information Processing Systems}, volume~37, pages 83951--84009, 2024{\natexlab{a}}.

\bibitem[Wang et~al.(2024{\natexlab{b}})Wang, Xie, Jiang, Mandlekar, Xiao, Zhu, Fan, and Anandkumar]{wang2023voyager}
G.~Wang, Y.~Xie, Y.~Jiang, A.~Mandlekar, C.~Xiao, Y.~Zhu, L.~Fan, and A.~Anandkumar.
\newblock Voyager: An open-ended embodied agent with large language models.
\newblock \emph{Transactions on Machine Learning Research}, 2024{\natexlab{b}}.

\bibitem[Wang et~al.(2025{\natexlab{b}})Wang, Li, Qu, Xu, Zhu, Chu, and Lin]{wang2025tocode}
H.~Wang, L.~Li, C.~Qu, W.~Xu, F.~Zhu, W.~Chu, and F.~Lin.
\newblock To code or not to code? adaptive tool integration for math language models via expectation-maximization.
\newblock In \emph{Annual Meeting of the Association for Computational Linguistics}, pages 3060--3075, 2025{\natexlab{b}}.

\bibitem[Wang et~al.(2025{\natexlab{c}})Wang, Qu, Huang, Chu, Lin, and Chen]{wang2025vlrethinker}
H.~Wang, C.~Qu, Z.~Huang, W.~Chu, F.~Lin, and W.~Chen.
\newblock {VL-Rethinker}: Incentivizing self-reflection of vision-language models with reinforcement learning.
\newblock \emph{arXiv preprint arXiv:2504.08837}, 2025{\natexlab{c}}.

\bibitem[Wang et~al.(2025{\natexlab{d}})Wang, Ye, Tao, Pan, Mallik, Yaman, Ren, and Zhang]{zhao2025adawm}
H.~Wang, X.~Ye, F.~Tao, C.~Pan, A.~Mallik, B.~Yaman, L.~Ren, and J.~Zhang.
\newblock {AdaWM}: Adaptive world model based planning for autonomous driving.
\newblock In \emph{International Conference on Learning Representations}, 2025{\natexlab{d}}.

\bibitem[Wang et~al.(2026{\natexlab{a}})Wang, Jiang, He, Sun, Zhang, He, Cao, Gan, Sun, Shao, and Yue]{wang2026mvista4d}
J.~Wang, Y.~Jiang, T.~He, J.~Sun, Q.~Zhang, J.~He, J.~Cao, Z.~Gan, M.~Sun, Q.~Shao, and X.~Yue.
\newblock {MVISTA-4D}: View-consistent {4D} world model with test-time action inference for robotic manipulation.
\newblock \emph{arXiv preprint arXiv:2602.09878}, 2026{\natexlab{a}}.

\bibitem[Wang et~al.(2025{\natexlab{e}})Wang, Zhang, Wang, Gao, Li, Wang, Chen, Wan, Lu, Yang, et~al.]{wang2025vagen}
K.~Wang, P.~Zhang, Z.~Wang, Y.~Gao, L.~Li, Q.~Wang, H.~Chen, C.~Wan, Y.~Lu, Z.~Yang, et~al.
\newblock Vagen: Reinforcing world model reasoning for multi-turn vlm agents.
\newblock \emph{arXiv preprint arXiv:2510.16907}, 2025{\natexlab{e}}.

\bibitem[Wang et~al.(2024{\natexlab{c}})Wang, Yang, Wang, Jin, Zeng, and Yang]{wang2024coworld}
Q.~Wang, J.~Yang, Y.~Wang, X.~Jin, W.~Zeng, and X.~Yang.
\newblock Making offline {RL} online: Collaborative world models for offline visual reinforcement learning.
\newblock In \emph{Advances in Neural Information Processing Systems}, volume~37, pages 97203--97230, 2024{\natexlab{c}}.

\bibitem[Wang et~al.(2025{\natexlab{f}})Wang, Yin, Zhang, Zhang, Wang, Wang, Zhang, Chandrasegaran, Krishna, Xie, et~al.]{mindcube}
Q.~Wang, B.~Yin, P.~Zhang, J.~Zhang, K.~Wang, Z.~Wang, J.~Zhang, K.~Chandrasegaran, H.~L.~R. Krishna, S.~Xie, et~al.
\newblock Spatial mental modeling from limited views.
\newblock In \emph{The Fourteenth International Conference on Learning Representations}, 2025{\natexlab{f}}.

\bibitem[Wang et~al.(2022)Wang, Jansen, C{\^o}t{\'e}, and Ammanabrolu]{wang2022scienceworld}
R.~Wang, P.~Jansen, M.-A. C{\^o}t{\'e}, and P.~Ammanabrolu.
\newblock {ScienceWorld}: Is your agent smarter than a 5th grader?
\newblock In \emph{Conference on Empirical Methods in Natural Language Processing}, pages 11279--11298, 2022.

\bibitem[Wang et~al.(2024{\natexlab{d}})Wang, Todd, Xiao, Yuan, C{\^o}t{\'e}, Clark, and Jansen]{wang2024worldsim}
R.~Wang, G.~Todd, Z.~Xiao, X.~Yuan, M.-A. C{\^o}t{\'e}, P.~Clark, and P.~Jansen.
\newblock Can language models serve as text-based world simulators?
\newblock In \emph{Annual Meeting of the Association for Computational Linguistics}, 2024{\natexlab{d}}.

\bibitem[Wang et~al.(2024{\natexlab{e}})Wang, Yu, Zhang, Qi, Sap, Bisk, Neubig, and Zhu]{wang2024sotopia_pi}
R.~Wang, H.~Yu, W.~Zhang, Z.~Qi, M.~Sap, Y.~Bisk, G.~Neubig, and H.~Zhu.
\newblock Sotopia-$\pi$: Interactive learning of socially intelligent language agents.
\newblock In \emph{Annual Meeting of the Association for Computational Linguistics}, pages 12912--12940, 2024{\natexlab{e}}.

\bibitem[Wang et~al.(2024{\natexlab{f}})Wang, Liu, Zheng, Qi, Chen, Yang, Zhao, Wang, Song, and Huang]{wang2024recon}
S.~Wang, C.~Liu, Z.~Zheng, S.~Qi, S.~Chen, Q.~Yang, A.~Zhao, C.~Wang, S.~Song, and G.~Huang.
\newblock Boosting {LLM} agents with recursive contemplation for effective deception handling.
\newblock In \emph{Annual Meeting of the Association for Computational Linguistics}, pages 9909--9953, 2024{\natexlab{f}}.

\bibitem[Wang et~al.(2024{\natexlab{g}})Wang, Dong, Jiang, Parkes, and Tambe]{wang2025decpomdp}
T.~Wang, H.~Dong, Y.~Jiang, D.~C. Parkes, and M.~Tambe.
\newblock On diffusion models for multi-agent partial observability: Shared attractors, error bounds, and composite flow.
\newblock \emph{arXiv preprint arXiv:2410.13953}, 2024{\natexlab{g}}.

\bibitem[Wang et~al.(2019)Wang, Shi, Kim, Oh, Yang, Zhang, and Yu]{wang2019persuasionforgood}
X.~Wang, W.~Shi, R.~Kim, Y.~Oh, S.~Yang, J.~Zhang, and Z.~Yu.
\newblock Persuasion for good: Towards a personalized persuasive dialogue system for social good.
\newblock In \emph{Annual Meeting of the Association for Computational Linguistics}, pages 5635--5649, 2019.

\bibitem[Wang et~al.(2024{\natexlab{h}})Wang, Zhu, Huang, Chen, Zhu, and Lu]{wang2024drivedreamer}
X.~Wang, Z.~Zhu, G.~Huang, X.~Chen, J.~Zhu, and J.~Lu.
\newblock {DriveDreamer}: Towards real-world-driven world models for autonomous driving.
\newblock In \emph{European Conference on Computer Vision}, pages 55--72. Springer, 2024{\natexlab{h}}.

\bibitem[Wang et~al.(2025{\natexlab{g}})Wang, Luo, Bai, Cao, Che, Chen, Chen, Diamond, Ding, Ding, Feng, Heinrich, Huang, Karkus, Li, Li, Lin, Liu, Liu, Liu, Liu, Lu, Mao, Molchanov, Pavao, Peng, Ranzinger, Schmerling, Shen, Shi, Tariq, Tian, Wekel, Weng, Xiao, Yang, Yang, You, Zeng, Zhang, Ivanovic, and Pavone]{wang2025alpamayo}
Y.~Wang, W.~Luo, J.~Bai, Y.~Cao, T.~Che, K.~Chen, Y.~Chen, J.~Diamond, Y.~Ding, W.~Ding, L.~Feng, G.~Heinrich, J.~Huang, P.~Karkus, B.~Li, P.~Li, T.-Y. Lin, D.~Liu, M.-Y. Liu, L.~Liu, Z.~Liu, J.~Lu, Y.~Mao, P.~Molchanov, L.~Pavao, Z.~Peng, M.~Ranzinger, E.~Schmerling, S.~Shen, Y.~Shi, S.~Tariq, R.~Tian, T.~Wekel, X.~Weng, T.~Xiao, E.~Yang, X.~Yang, Y.~You, X.~Zeng, W.~Zhang, B.~Ivanovic, and M.~Pavone.
\newblock {Alpamayo-R1}: Bridging reasoning and action prediction for generalizable autonomous driving in the long tail.
\newblock \emph{arXiv preprint arXiv:2511.00088}, 2025{\natexlab{g}}.

\bibitem[Wang et~al.(2025{\natexlab{h}})Wang, Li, Ye, Fang, Wang, Liu, Liang, Lu, Wu, Feng, et~al.]{wang2025game}
Z.~Wang, X.~Li, Y.~Ye, J.~Fang, H.~Wang, L.~Liu, S.~Liang, J.~Lu, Z.~Wu, J.~Feng, et~al.
\newblock Game-tars: Pretrained foundation models for scalable generalist multimodal game agents.
\newblock \emph{arXiv preprint arXiv:2510.23691}, 2025{\natexlab{h}}.

\bibitem[Wang et~al.(2025{\natexlab{i}})Wang, Wang, Wang, Zhang, Li, Yang, Jin, Yu, Nguyen, Liu, et~al.]{wang2025ragen}
Z.~Wang, K.~Wang, Q.~Wang, P.~Zhang, L.~Li, Z.~Yang, X.~Jin, K.~Yu, M.~N. Nguyen, L.~Liu, et~al.
\newblock Ragen: Understanding self-evolution in llm agents via multi-turn reinforcement learning.
\newblock \emph{arXiv preprint arXiv:2504.20073}, 2025{\natexlab{i}}.

\bibitem[Wang et~al.(2025{\natexlab{j}})Wang, Zhang, Yue, Yue, Li, Ouyang, and Bai]{wang2025transition}
Z.~Wang, Y.~Zhang, X.~Yue, X.~Yue, Y.~Li, W.~Ouyang, and L.~Bai.
\newblock Transition models: Rethinking the generative learning objective.
\newblock \emph{arXiv preprint arXiv:2509.04394}, 2025{\natexlab{j}}.

\bibitem[Wang et~al.(2026{\natexlab{b}})Wang, Gui, Jin, Wang, Liu, Wang, Chen, Li, Yang, Zhang, et~al.]{wang2026ragen}
Z.~Wang, C.~Gui, X.~Jin, Q.~Wang, L.~Liu, K.~Wang, S.~Chen, L.~Li, Z.~Yang, P.~Zhang, et~al.
\newblock Ragen-2: Reasoning collapse in agentic rl.
\newblock \emph{arXiv preprint arXiv:2604.06268}, 2026{\natexlab{b}}.

\bibitem[Wang et~al.(2026{\natexlab{c}})Wang, Liu, Li, Huang, Xu, Kang, An, Wang, Jiang, Wei, Xietian, Pei, Hu, Jiang, Xue, Wang, Sun, Li, Ouyang, He, Liu, Li, and Zhou]{matrixgame3}
Z.~Wang, Z.~Liu, J.~Li, K.~Huang, B.~Xu, F.~Kang, M.~An, P.~Wang, B.~Jiang, Y.~Wei, Y.~Xietian, J.~Pei, L.~Hu, B.~Jiang, H.~Xue, Z.~Wang, H.~Sun, W.~Li, W.~Ouyang, X.~He, Y.~Liu, Y.~Li, and Y.~Zhou.
\newblock {Matrix-Game 3.0}: Real-time and streaming interactive world model with long-horizon memory.
\newblock \emph{arXiv preprint arXiv:2604.08995}, 2026{\natexlab{c}}.

\bibitem[Wang et~al.(2026{\natexlab{d}})Wang, Xu, Liu, Wang, Han, Yao, Yao, and He]{wang2026awm}
Z.~Wang, C.~Xu, B.~Liu, Y.~Wang, S.~Han, Z.~Yao, H.~Yao, and Y.~He.
\newblock Agent world model: Infinity synthetic environments for agentic reinforcement learning.
\newblock \emph{arXiv preprint arXiv:2602.10090}, 2026{\natexlab{d}}.

\bibitem[Wang et~al.(2024{\natexlab{i}})Wang, Peng, Que, Liu, Zhou, Wu, Guo, Gan, Ni, Yang, Zhang, Zhang, Ouyang, Xu, Huang, Fu, and Peng]{wang2024rolellm}
Z.~M. Wang, Z.~Peng, H.~Que, J.~Liu, W.~Zhou, Y.~Wu, H.~Guo, R.~Gan, Z.~Ni, J.~Yang, M.~Zhang, Z.~Zhang, W.~Ouyang, K.~Xu, S.~W. Huang, J.~Fu, and J.~Peng.
\newblock {RoleLLM}: Benchmarking, eliciting, and enhancing role-playing abilities of large language models.
\newblock In \emph{Annual Meeting of the Association for Computational Linguistics}, pages 14743--14777, 2024{\natexlab{i}}.

\bibitem[Watter et~al.(2015)Watter, Springenberg, Boedecker, and Riedmiller]{watter2015e2c}
M.~Watter, J.~T. Springenberg, J.~Boedecker, and M.~Riedmiller.
\newblock Embed to control: A locally linear latent dynamics model for control from raw images.
\newblock In \emph{Advances in Neural Information Processing Systems}, volume~28, pages 2746--2754, 2015.

\bibitem[Wei et~al.(2025{\natexlab{a}})Wei, Zhang, He, Xia, Pan, and Liu]{wei2025plangenllms}
H.~Wei, Z.~Zhang, S.~He, T.~Xia, S.~Pan, and F.~Liu.
\newblock {PlanGenLLMs}: A modern survey of {LLM} planning capabilities.
\newblock \emph{arXiv preprint arXiv:2502.11221}, 2025{\natexlab{a}}.

\bibitem[Wei et~al.(2025{\natexlab{b}})Wei, Yang, Zhang, Chen, Zhuang, Gao, Zhou, Wang, Gao, Cao, Qiu, Hu, Ma, Tang, He, Song, He, Zhang, You, Zheng, Ding, Ouyang, Dong, Cheng, Sun, Bai, and Zhou]{wei2025agenticscience}
J.~Wei, Y.~Yang, X.~Zhang, Y.~Chen, X.~Zhuang, Z.~Gao, D.~Zhou, G.~Wang, Z.~Gao, J.~Cao, Z.~Qiu, M.~Hu, C.~Ma, S.~Tang, J.~He, C.~Song, X.~He, Q.~Zhang, C.~You, S.~Zheng, N.~Ding, W.~Ouyang, N.~Dong, Y.~Cheng, S.~Sun, L.~Bai, and B.~Zhou.
\newblock From {AI} for science to agentic science: A survey on autonomous scientific discovery.
\newblock \emph{arXiv preprint arXiv:2508.14111}, 2025{\natexlab{b}}.

\bibitem[Wilf et~al.(2024)Wilf, Lee, Liang, and Morency]{wilf2024simtom}
A.~Wilf, S.~Lee, P.~P. Liang, and L.-P. Morency.
\newblock Think twice: Perspective-taking improves large language models’ theory-of-mind capabilities.
\newblock In \emph{Annual Meeting of the Association for Computational Linguistics}, pages 8292--8308, 2024.

\bibitem[Wolpert(1996)]{wolpert1996nofreelunch}
D.~H. Wolpert.
\newblock The lack of a priori distinctions between learning algorithms.
\newblock \emph{Neural Computation}, 8\penalty0 (7):\penalty0 1341--1390, 1996.

\bibitem[{World Labs team}(2025{\natexlab{a}})]{worldlabs2025marble}
{World Labs team}.
\newblock Marble: A multimodal world model.
\newblock {World Labs Technical Post}, 2025{\natexlab{a}}.
\newblock URL \url{https://www.worldlabs.ai/blog/marble-world-model}.

\bibitem[{World Labs team}(2025{\natexlab{b}})]{worldlabs2025rtfm}
{World Labs team}.
\newblock {RTFM}: A real-time frame model.
\newblock {World Labs Research Preview}, 2025{\natexlab{b}}.
\newblock URL \url{https://www.worldlabs.ai/blog/rtfm}.

\bibitem[Wu et~al.(2024{\natexlab{a}})Wu, Jing, Cheang, Chen, Xu, Li, Liu, Li, and Kong]{wu2023gr1}
H.~Wu, Y.~Jing, C.~Cheang, G.~Chen, J.~Xu, X.~Li, M.~Liu, H.~Li, and T.~Kong.
\newblock Unleashing large-scale video generative pre-training for visual robot manipulation.
\newblock In \emph{International Conference on Learning Representations}, 2024{\natexlab{a}}.

\bibitem[Wu et~al.(2024{\natexlab{b}})Wu, Wang, Shang, Shah, and Yan]{wu2024ptq4dit}
J.~Wu, H.~Wang, Y.~Shang, M.~Shah, and Y.~Yan.
\newblock {PTQ4DiT}: Post-training quantization for diffusion transformers.
\newblock In \emph{Advances in Neural Information Processing Systems}, volume~37, pages 62732--62755, 2024{\natexlab{b}}.

\bibitem[Wu et~al.(2024{\natexlab{c}})Wu, Yin, Feng, He, Li, Hao, and Long]{wu2024ivideogpt}
J.~Wu, S.~Yin, N.~Feng, X.~He, D.~Li, J.~Hao, and M.~Long.
\newblock {iVideoGPT}: Interactive {VideoGPTs} are scalable world models.
\newblock In \emph{Advances in Neural Information Processing Systems}, volume~37, pages 68082--68119, 2024{\natexlab{c}}.

\bibitem[Wu et~al.(2023{\natexlab{a}})Wu, Escontrela, Hafner, Goldberg, and Abbeel]{wu2023daydreamer}
P.~Wu, A.~Escontrela, D.~Hafner, K.~Goldberg, and P.~Abbeel.
\newblock {DayDreamer}: World models for physical robot learning.
\newblock In \emph{Conference on Robot Learning}. PMLR, 2023{\natexlab{a}}.

\bibitem[Wu et~al.(2023{\natexlab{b}})Wu, He, Jia, Mihalcea, Chen, and Deng]{wu2023hitom}
Y.~Wu, Y.~He, Y.~Jia, R.~Mihalcea, Y.~Chen, and N.~Deng.
\newblock {Hi-ToM}: A benchmark for evaluating higher-order theory of mind reasoning in large language models.
\newblock In \emph{Conference on Empirical Methods in Natural Language Processing}, pages 10691--10706, 2023{\natexlab{b}}.

\bibitem[Xia et~al.(2024)Xia, Lin, Ma, and Wang]{xia2024video2game}
H.~Xia, Z.-H. Lin, W.-C. Ma, and S.~Wang.
\newblock {Video2Game}: Real-time, interactive, realistic and browser-compatible environment from a single video.
\newblock In \emph{IEEE/CVF Conference on Computer Vision and Pattern Recognition}, pages 4578--4588, 2024.

\bibitem[Xiang et~al.(2020)Xiang, Qin, Mo, Xia, Zhu, Liu, Liu, Jiang, Yuan, Wang, Yi, Chang, Guibas, and Su]{xiang2020sapien}
F.~Xiang, Y.~Qin, K.~Mo, Y.~Xia, H.~Zhu, F.~Liu, M.~Liu, H.~Jiang, Y.~Yuan, H.~Wang, L.~Yi, A.~X. Chang, L.~J. Guibas, and H.~Su.
\newblock {SAPIEN}: A simulated part-based interactive environment.
\newblock In \emph{IEEE/CVF Conference on Computer Vision and Pattern Recognition}, pages 11097--11107, 2020.

\bibitem[Xiao et~al.(2024)Xiao, Tian, Chen, Han, and Lewis]{xiao2024streamingllm}
G.~Xiao, Y.~Tian, B.~Chen, S.~Han, and M.~Lewis.
\newblock Efficient streaming language models with attention sinks.
\newblock In \emph{International Conference on Learning Representations}, 2024.

\bibitem[Xiao et~al.(2026)Xiao, Tu, Zou, Zuo, Li, Wang, Yu, Huang, Lin, and Liu]{xiao2026webworld}
Z.~Xiao, J.~Tu, C.~Zou, Y.~Zuo, Z.~Li, P.~Wang, B.~Yu, F.~Huang, J.~Lin, and Z.~Liu.
\newblock {WebWorld}: A large-scale world model for web agent training.
\newblock \emph{arXiv preprint arXiv:2602.14721}, 2026.

\bibitem[Xie et~al.(2024)Xie, Zhang, Chen, Li, Zhao, Cao, Hua, Cheng, Shin, Lei, Liu, Xu, Zhou, Savarese, Xiong, Zhong, and Yu]{xie2024osworld}
T.~Xie, D.~Zhang, J.~Chen, X.~Li, S.~Zhao, R.~Cao, T.~J. Hua, Z.~Cheng, D.~Shin, F.~Lei, Y.~Liu, Y.~Xu, S.~Zhou, S.~Savarese, C.~Xiong, V.~Zhong, and T.~Yu.
\newblock {OSWorld}: Benchmarking multimodal agents for open-ended tasks in real computer environments.
\newblock In \emph{Advances in Neural Information Processing Systems}, volume~37, pages 52040--52094, 2024.

\bibitem[Xing et~al.(2024)Xing, Xia, Zhang, Chen, Yu, Liu, Wang, Wong, and Shan]{xing2023dynamicrafter}
J.~Xing, M.~Xia, Y.~Zhang, H.~Chen, W.~Yu, H.~Liu, X.~Wang, T.-T. Wong, and Y.~Shan.
\newblock {DynamiCrafter}: Animating open-domain images with video diffusion priors.
\newblock In \emph{European Conference on Computer Vision}, pages 399--417. Springer, 2024.

\bibitem[Xu et~al.(2024{\natexlab{a}})Xu, Zhao, Zhu, Du, and He]{xu2024opentom}
H.~Xu, R.~Zhao, L.~Zhu, J.~Du, and Y.~He.
\newblock {OpenToM}: A comprehensive benchmark for evaluating theory-of-mind reasoning capabilities of large language models.
\newblock In \emph{Annual Meeting of the Association for Computational Linguistics}, pages 8593--8623, 2024{\natexlab{a}}.

\bibitem[Xu et~al.(2026{\natexlab{a}})Xu, Liang, Zheng, Luo, Hu, Zhang, and Tao]{xu2026ctrlattack}
S.~Xu, S.~Liang, H.~Zheng, Y.~Luo, H.~Hu, L.~Zhang, and D.~Tao.
\newblock {CtrlAttack}: A unified attack on world-model control in diffusion models.
\newblock \emph{arXiv preprint arXiv:2603.13435}, 2026{\natexlab{a}}.

\bibitem[Xu et~al.(2026{\natexlab{b}})Xu, Li, Ye, Chen, Zeng, Chen, Xu, Lin, Li, and Pang]{xu2026futurevla}
X.~Xu, H.~Li, J.~Ye, Y.~Chen, J.~Zeng, X.~Chen, L.~Xu, D.~Lin, W.~Li, and J.~Pang.
\newblock {FutureVLA}: Joint visuomotor prediction for vision-language-action model.
\newblock \emph{arXiv preprint arXiv:2603.10712}, 2026{\natexlab{b}}.

\bibitem[Xu et~al.(2026{\natexlab{c}})Xu, Liang, Liu, Li, Kong, Liu, and Liu]{xu2025u4d}
X.~Xu, A.~Liang, Y.~Liu, L.~Li, L.~Kong, Z.~Liu, and Q.~Liu.
\newblock {U4D}: Uncertainty-aware {4D} world modeling from {LiDAR} sequences.
\newblock In \emph{IEEE/CVF Conference on Computer Vision and Pattern Recognition}, 2026{\natexlab{c}}.

\bibitem[Xu et~al.(2024{\natexlab{b}})Xu, Wang, Wang, Lu, Xie, Saha, Sahoo, Yu, and Xiong]{xu2024aguvis}
Y.~Xu, Z.~Wang, J.~Wang, D.~Lu, T.~Xie, A.~Saha, D.~Sahoo, T.~Yu, and C.~Xiong.
\newblock Aguvis: Unified pure vision agents for autonomous {GUI} interaction.
\newblock \emph{arXiv preprint arXiv:2412.04454}, 2024{\natexlab{b}}.

\bibitem[Xu et~al.(2023)Xu, Yu, Fang, Wang, and Wu]{xu2024werewolf}
Z.~Xu, C.~Yu, F.~Fang, Y.~Wang, and Y.~Wu.
\newblock Language agents with reinforcement learning for strategic play in the werewolf game.
\newblock \emph{arXiv preprint arXiv:2310.18940}, 2023.

\bibitem[Yamada et~al.(2025)Yamada, Lange, Lu, Hu, Lu, Foerster, Clune, and Ha]{yamada2025aiscientistv2}
Y.~Yamada, R.~T. Lange, C.~Lu, S.~Hu, C.~Lu, J.~Foerster, J.~Clune, and D.~Ha.
\newblock The {AI Scientist}-v2: Workshop-level automated scientific discovery via agentic tree search.
\newblock \emph{arXiv preprint arXiv:2504.08066}, 2025.

\bibitem[Yan et~al.(2026{\natexlab{a}})Yan, Lin, Zhu, and Wang]{yan2026safedream}
B.~Yan, W.~Lin, Y.~Zhu, and S.~Wang.
\newblock {SafeDream}: Safety world model for proactive early jailbreak detection.
\newblock \emph{arXiv preprint arXiv:2604.16824}, 2026{\natexlab{a}}.

\bibitem[Yan et~al.(2026{\natexlab{b}})Yan, Tang, Gui, Li, Zhesng, Huang, Kong, Han, Zhou, Zhang, Zhan, Zhan, zhong Xu, and Shen]{yan2025ad-r1}
T.~Yan, T.~Tang, X.~Gui, Y.~Li, J.~Zhesng, W.~Huang, L.~Kong, W.~Han, X.~Zhou, X.~Zhang, Y.~Zhan, K.~Zhan, C.~zhong Xu, and J.~Shen.
\newblock Ad-r1: Closed-loop reinforcement learning for end-to-end autonomous driving with impartial world models.
\newblock In \emph{IEEE/CVF Conference on Computer Vision and Pattern Recognition}, 2026{\natexlab{b}}.

\bibitem[Yang et~al.(2026)Yang, Lin, Li, Wang, Guo, Feng, and Chua]{yang2026llmknowledge}
C.~Yang, X.~Lin, S.~Li, W.~Wang, R.~Guo, F.~Feng, and T.-S. Chua.
\newblock Can large language models derive new knowledge? {A} dynamic benchmark for biological knowledge discovery.
\newblock \emph{arXiv preprint arXiv:2603.03322}, 2026.

\bibitem[Yang et~al.(2024{\natexlab{a}})Yang, Jimenez, Wettig, Lieret, Yao, Narasimhan, and Press]{yang2024sweagent}
J.~Yang, C.~E. Jimenez, A.~Wettig, K.~Lieret, S.~Yao, K.~Narasimhan, and O.~Press.
\newblock {SWE-Agent}: Agent-computer interfaces enable automated software engineering.
\newblock In \emph{Advances in Neural Information Processing Systems}, volume~37, pages 50528--50652, 2024{\natexlab{a}}.

\bibitem[Yang et~al.(2025{\natexlab{a}})Yang, Ci, and Shou]{yang2025macosworld}
P.~Yang, H.~Ci, and M.~Z. Shou.
\newblock {macOSWorld}: A multilingual interactive benchmark for gui agents.
\newblock \emph{arXiv preprint arXiv:2506.04135}, 2025{\natexlab{a}}.

\bibitem[Yang et~al.(2024{\natexlab{b}})Yang, Du, Ghasemipour, Tompson, Kaelbling, Schuurmans, and Abbeel]{yang2024unisim}
S.~Yang, Y.~Du, S.~K.~S. Ghasemipour, J.~Tompson, L.~P. Kaelbling, D.~Schuurmans, and P.~Abbeel.
\newblock Learning interactive real-world simulators.
\newblock In \emph{International Conference on Learning Representations}, 2024{\natexlab{b}}.

\bibitem[Yang et~al.(2025{\natexlab{b}})Yang, Huang, Chu, Xiao, Zhao, Wang, Li, Xie, Chen, Lu, Han, and Chen]{yang2025longlive}
S.~Yang, W.~Huang, R.~Chu, Y.~Xiao, Y.~Zhao, X.~Wang, M.~Li, E.~Xie, Y.~Chen, Y.~Lu, S.~Han, and Y.~Chen.
\newblock {LongLive}: Real-time interactive long video generation.
\newblock \emph{arXiv preprint arXiv:2509.22622}, 2025{\natexlab{b}}.

\bibitem[Yang et~al.(2025{\natexlab{c}})Yang, Xi, Zhao, Li, Zhang, Cai, Lin, Li, Xu, Peng, Chen, Han, Keutzer, and Stoica]{yang2025sparse}
S.~Yang, H.~Xi, Y.~Zhao, M.~Li, J.~Zhang, H.~Cai, Y.~Lin, X.~Li, C.~Xu, K.~Peng, J.~Chen, S.~Han, K.~Keutzer, and I.~Stoica.
\newblock {Sparse VideoGen2}: Accelerate video generation with sparse attention via semantic-aware permutation.
\newblock \emph{arXiv preprint arXiv:2505.18875}, 2025{\natexlab{c}}.

\bibitem[Yang et~al.(2024{\natexlab{c}})Yang, Du, Li, Zheng, Poria, and Cambria]{yang2024moose}
Z.~Yang, X.~Du, J.~Li, J.~Zheng, S.~Poria, and E.~Cambria.
\newblock Large language models for automated open-domain scientific hypotheses discovery.
\newblock In \emph{Annual Meeting of the Association for Computational Linguistics}, pages 13545--13565, 2024{\natexlab{c}}.

\bibitem[Yang et~al.(2024{\natexlab{d}})Yang, Zhang, Zheng, Jiang, Gan, Wang, Ling, Chen, Ma, Dong, Gupta, Hu, Yin, Li, Jia, Wang, Ghanem, Lu, Lu, Ouyang, Qiao, Torr, and Shao]{yang2024oasis}
Z.~Yang, Z.~Zhang, Z.~Zheng, Y.~Jiang, Z.~Gan, Z.~Wang, Z.~Ling, J.~Chen, M.~Ma, B.~Dong, P.~Gupta, S.~Hu, Z.~Yin, G.~Li, X.~Jia, L.~Wang, B.~Ghanem, H.~Lu, C.~Lu, W.~Ouyang, Y.~Qiao, P.~Torr, and J.~Shao.
\newblock Oasis: Open agent social interaction simulations with one million agents.
\newblock \emph{arXiv preprint arXiv:2411.11581}, 2024{\natexlab{d}}.

\bibitem[Yang et~al.(2025{\natexlab{d}})Yang, Liu, Gao, Liu, Li, Xie, Bing, Ouyang, Cambria, and Zhou]{yang2025moosechem2}
Z.~Yang, W.~Liu, B.~Gao, Y.~Liu, W.~Li, T.~Xie, L.~Bing, W.~Ouyang, E.~Cambria, and D.~Zhou.
\newblock {MOOSE}-chem2: Exploring {LLM} limits in fine-grained scientific hypothesis discovery via hierarchical search.
\newblock In \emph{The Thirty-ninth Annual Conference on Neural Information Processing Systems}, 2025{\natexlab{d}}.

\bibitem[Yang et~al.(2025{\natexlab{e}})Yang, Liu, Gao, Xie, Li, Ouyang, Poria, Cambria, and Zhou]{yang2025moosechem}
Z.~Yang, W.~Liu, B.~Gao, T.~Xie, Y.~Li, W.~Ouyang, S.~Poria, E.~Cambria, and D.~Zhou.
\newblock {MOOSE-Chem}: Large language models for rediscovering unseen chemistry scientific hypotheses.
\newblock In \emph{International Conference on Learning Representations}, 2025{\natexlab{e}}.

\bibitem[Yao et~al.(2022)Yao, Chen, Yang, and Narasimhan]{yao2022webshop}
S.~Yao, H.~Chen, J.~Yang, and K.~Narasimhan.
\newblock {WebShop}: Towards scalable real-world web interaction with grounded language agents.
\newblock In \emph{Advances in Neural Information Processing Systems}, volume~35, pages 20744--20757, 2022.

\bibitem[Ye et~al.(2025)Ye, Wang, Sun, Chandrasegaran, Durante, Eyzaguirre, Bisk, Niebles, Adeli, Fei-Fei, Wu, and Li]{ye2025re}
J.~Ye, Z.~Wang, H.~Sun, K.~Chandrasegaran, Z.~Durante, C.~Eyzaguirre, Y.~Bisk, J.~C. Niebles, E.~Adeli, L.~Fei-Fei, J.~Wu, and M.~Li.
\newblock Re-thinking temporal search for long-form video understanding.
\newblock In \emph{IEEE/CVF Conference on Computer Vision and Pattern Recognition}, pages 8579--8591, 2025.

\bibitem[Ye et~al.(2026{\natexlab{a}})Ye, Wang, Gao, Yu, Zhu, Wang, Zhang, Jin, Fu, Zheng, Chen, and Pang]{ye2026st4vla}
J.~Ye, F.~Wang, N.~Gao, J.~Yu, Y.~Zhu, B.~Wang, J.~Zhang, W.~Jin, Y.~Fu, F.~Zheng, Y.~Chen, and J.~Pang.
\newblock {ST4VLA}: Spatially guided training for vision-language-action models.
\newblock \emph{arXiv preprint arXiv:2602.10109}, 2026{\natexlab{a}}.

\bibitem[Ye et~al.(2026{\natexlab{b}})Ye, Ge, Zheng, Gao, Yu, Kurian, Indupuru, Tan, Zhu, Xiang, Malik, Lee, Liang, Ranawaka, Gu, Xu, Wang, Hu, Narayan, Bjorck, Wang, Kim, Niu, Zheng, Xie, Wu, Wang, Julian, Xu, Du, Chebotar, Reed, Kautz, Zhu, Fan, and Jang]{ye2026dreamzero}
S.~Ye, Y.~Ge, K.~Zheng, S.~Gao, S.~Yu, G.~Kurian, S.~Indupuru, Y.~L. Tan, C.~Zhu, J.~Xiang, A.~Malik, K.~Lee, W.~Liang, N.~Ranawaka, J.~Gu, Y.~Xu, G.~Wang, F.~Hu, A.~Narayan, J.~Bjorck, J.~Wang, G.~Kim, D.~Niu, R.~Zheng, Y.~Xie, J.~Wu, Q.~Wang, R.~Julian, D.~Xu, Y.~Du, Y.~Chebotar, S.~Reed, J.~Kautz, Y.~Zhu, L.~Fan, and J.~Jang.
\newblock World action models are zero-shot policies.
\newblock \emph{arXiv preprint arXiv:2602.15922}, 2026{\natexlab{b}}.

\bibitem[Ye et~al.(2021)Ye, Liu, Kurutach, Abbeel, and Gao]{ye2021efficientzero}
W.~Ye, S.~Liu, T.~Kurutach, P.~Abbeel, and Y.~Gao.
\newblock Mastering atari games with limited data.
\newblock In \emph{Advances in Neural Information Processing Systems}, volume~34, 2021.

\bibitem[Yin et~al.(2026)Yin, Ge, Wang, Li, Black, Darrell, Kanazawa, and Feng]{yin2026vision}
S.~Yin, J.~Ge, Z.~Z. Wang, X.~Li, M.~J. Black, T.~Darrell, A.~Kanazawa, and H.~Feng.
\newblock Vision-as-inverse-graphics agent via interleaved multimodal reasoning.
\newblock \emph{arXiv preprint arXiv:2601.11109}, 2026.

\bibitem[Yin et~al.(2024{\natexlab{a}})Yin, Gharbi, Park, Zhang, Shechtman, Durand, and Freeman]{yin2024improved}
T.~Yin, M.~Gharbi, T.~Park, R.~Zhang, E.~Shechtman, F.~Durand, and W.~T. Freeman.
\newblock Improved distribution matching distillation for fast image synthesis.
\newblock In \emph{Advances in Neural Information Processing Systems}, volume~37, pages 47455--47487, 2024{\natexlab{a}}.

\bibitem[Yin et~al.(2024{\natexlab{b}})Yin, Zhang, Zhang, Freeman, Durand, Shechtman, and Huang]{yin2025slow}
T.~Yin, Q.~Zhang, R.~Zhang, W.~T. Freeman, F.~Durand, E.~Shechtman, and X.~Huang.
\newblock From slow bidirectional to fast autoregressive video diffusion models.
\newblock \emph{arXiv preprint arXiv:2412.07772}, 2024{\natexlab{b}}.

\bibitem[Ying et~al.(2025)Ying, Zhi-Xuan, Wong, Mansinghka, and Tenenbaum]{ying2025labtom}
L.~Ying, T.~Zhi-Xuan, L.~Wong, V.~Mansinghka, and J.~B. Tenenbaum.
\newblock Understanding epistemic language with a language-augmented bayesian theory of mind.
\newblock \emph{Transactions of the Association for Computational Linguistics}, 13:\penalty0 613--637, 2025.

\bibitem[Yokoyama et~al.(2024)Yokoyama, Ramrakhya, Das, Batra, and Ha]{yokoyama2024hm3d}
N.~Yokoyama, R.~Ramrakhya, A.~Das, D.~Batra, and S.~Ha.
\newblock {HM3D-OVON}: A dataset and benchmark for open-vocabulary object goal navigation.
\newblock In \emph{IEEE/RSJ International Conference on Intelligent Robots and Systems}, pages 5543--5550, 2024.

\bibitem[Yu et~al.(2025{\natexlab{a}})Yu, Qin, Wang, Wan, Zhang, and Liu]{yu2025gamefactory}
J.~Yu, Y.~Qin, X.~Wang, P.~Wan, D.~Zhang, and X.~Liu.
\newblock {GameFactory}: Creating new games with generative interactive videos.
\newblock In \emph{IEEE/CVF International Conference on Computer Vision}, pages 11590--11599, 2025{\natexlab{a}}.

\bibitem[Yu et~al.(2020)Yu, Quillen, He, Julian, Hausman, Finn, and Levine]{yu2020metaworld}
T.~Yu, D.~Quillen, Z.~He, R.~Julian, K.~Hausman, C.~Finn, and S.~Levine.
\newblock {Meta-World}: A benchmark and evaluation for multi-task and meta reinforcement learning.
\newblock In \emph{Conference on Robot Learning}, pages 1094--1100, 2020.

\bibitem[Yu et~al.(2025{\natexlab{b}})Yu, Qi, Li, Zhang, Zhang, Lin, Shechtman, Wang, and Nitzan]{yu2025self}
X.~Yu, X.~Qi, Z.~Li, K.~Zhang, R.~Zhang, Z.~Lin, E.~Shechtman, T.~Wang, and Y.~Nitzan.
\newblock Self-evaluation unlocks any-step text-to-image generation.
\newblock \emph{arXiv preprint arXiv:2512.22374}, 2025{\natexlab{b}}.

\bibitem[Yu et~al.(2026)Yu, Peng, Xu, Shen, He, Nath, Singh, Gao, and Yu]{yu2026rwml}
X.~Yu, B.~Peng, R.~Xu, Y.~Shen, P.~He, S.~Nath, N.~Singh, J.~Gao, and Z.~Yu.
\newblock Reinforcement world model learning for {LLM}-based agents.
\newblock \emph{arXiv preprint arXiv:2602.05842}, 2026.

\bibitem[Yue et~al.(2025)Yue, Huang, Chen, Wang, Wan, and Liu]{yue2025simulating}
J.~Yue, Z.~Huang, Z.~Chen, X.~Wang, P.~Wan, and Z.~Liu.
\newblock Simulating the visual world with artificial intelligence: A roadmap.
\newblock \emph{arXiv preprint arXiv:2511.08585}, 2025.

\bibitem[Zeng et~al.(2024)Zeng, Zhang, Liu, Sifakis, Zhang, Liu, and Wang]{zeng2024wmsafety}
Z.~Zeng, C.~Zhang, F.~Liu, J.~Sifakis, Q.~Zhang, S.~Liu, and P.~Wang.
\newblock World models: The safety perspective.
\newblock \emph{arXiv preprint arXiv:2411.07690}, 2024.

\bibitem[Zeng et~al.(2025)Zeng, Liu, Chen, He, Liao, Tian, Wang, Wang, Yang, Yin, Yin, Zhu, Cai, Chen, Chen, Du, Gao, Guo, Hu, Jiao, Li, Liu, Ni, Wen, Zhang, Zhang, Zhou, Blanchet, Qiu, Wang, and Huang]{zeng2025futurex}
Z.~Zeng, J.~Liu, S.~Chen, T.~He, Y.~Liao, Y.~Tian, J.~Wang, Z.~Wang, Y.~Yang, L.~Yin, M.~Yin, Z.~Zhu, T.~Cai, Z.~Chen, J.~Chen, Y.~Du, X.~Gao, J.~Guo, L.~Hu, J.~Jiao, X.~Li, J.~Liu, S.~Ni, Z.~Wen, G.~Zhang, K.~Zhang, X.~Zhou, J.~Blanchet, X.~Qiu, M.~Wang, and W.~Huang.
\newblock {FutureX}: An advanced live benchmark for {LLM} agents in future prediction.
\newblock \emph{arXiv preprint arXiv:2508.11987}, 2025.

\bibitem[Zhan et~al.(2024)Zhan, Xu, Fang, Wei, Hu, and Wang]{zhan2024bundle}
Z.~Zhan, H.~Xu, Z.~Fang, X.~Wei, Y.~Hu, and C.~Wang.
\newblock Bundle adjustment in the eager mode.
\newblock \emph{arXiv preprint arXiv:2409.12190}, 2024.

\bibitem[Zhang et~al.(2025{\natexlab{a}})Zhang, Yang, Liu, Han, Chen, Huang, Fu, and Yu]{zhang2025appagent}
C.~Zhang, Z.~Yang, J.~Liu, Y.~Han, X.~Chen, Z.~Huang, B.~Fu, and G.~Yu.
\newblock {AppAgent}: Multimodal agents as smartphone users.
\newblock In \emph{CHI Conference on Human Factors in Computing Systems}, 2025{\natexlab{a}}.

\bibitem[Zhang et~al.(2025{\natexlab{b}})Zhang, Lei, Li, Wang, Liu, Yang, Li, Wang, Yang, Wu, Ye, Ouyang, and Zhou]{zhang2025critic}
D.~Zhang, J.~Lei, J.~Li, X.~Wang, Y.~Liu, Z.~Yang, J.~Li, W.~Wang, S.~Yang, J.~Wu, P.~Ye, W.~Ouyang, and D.~Zhou.
\newblock {Critic-V}: {VLM} critics help catch vlm errors in multimodal reasoning.
\newblock In \emph{IEEE/CVF Conference on Computer Vision and Pattern Recognition}, pages 9050--9061, 2025{\natexlab{b}}.

\bibitem[Zhang et~al.(2026{\natexlab{a}})Zhang, Yuan, Yuan, Xu, Bian, Cheng, Huang, Zhao, and Rong]{zhang2026lingshucell}
H.~Zhang, G.-H. Yuan, C.~Yuan, T.~Xu, T.~Bian, H.~Cheng, W.~Huang, D.~Zhao, and Y.~Rong.
\newblock {Lingshu-Cell}: A generative cellular world model for transcriptome modeling toward virtual cells.
\newblock \emph{arXiv preprint arXiv:2603.25240}, 2026{\natexlab{a}}.

\bibitem[Zhang et~al.(2025{\natexlab{c}})Zhang, Chen, Liu, Xue, Liao, Liu, Wang, Ning, Chen, Fu, et~al.]{earlyexp}
K.~Zhang, X.~Chen, B.~Liu, T.~Xue, Z.~Liao, Z.~Liu, X.~Wang, Y.~Ning, Z.~Chen, X.~Fu, et~al.
\newblock Agent learning via early experience.
\newblock \emph{arXiv preprint arXiv:2510.08558}, 2025{\natexlab{c}}.

\bibitem[Zhang et~al.(2024{\natexlab{a}})Zhang, Xiong, Yang, Casas, Hu, and Urtasun]{zhang2024copilot4d}
L.~Zhang, Y.~Xiong, Z.~Yang, S.~Casas, R.~Hu, and R.~Urtasun.
\newblock {Copilot4D}: Learning unsupervised world models for autonomous driving via discrete diffusion.
\newblock In \emph{International Conference on Learning Representations}, 2024{\natexlab{a}}.

\bibitem[Zhang et~al.(2025{\natexlab{d}})Zhang, Cai, Li, Wetzstein, and Agrawala]{zhang2025framepack}
L.~Zhang, S.~Cai, M.~Li, G.~Wetzstein, and M.~Agrawala.
\newblock Frame context packing and drift prevention in next-frame-prediction video diffusion models.
\newblock \emph{arXiv preprint arXiv:2504.12626}, 2025{\natexlab{d}}.

\bibitem[Zhang et~al.(2026{\natexlab{b}})Zhang, Huang, Wang, Zhang, Xue, Wang, Wang, Chandrasegaran, Zhang, Choi, et~al.]{tos}
P.~Zhang, Z.~Huang, Y.~Wang, J.~Zhang, L.~Xue, Z.~Wang, Q.~Wang, K.~Chandrasegaran, R.~Zhang, Y.~Choi, et~al.
\newblock Theory of space: Can foundation models construct spatial beliefs through active exploration?
\newblock In \emph{The Fourteenth International Conference on Learning Representations}, 2026{\natexlab{b}}.

\bibitem[Zhang et~al.(2025{\natexlab{e}})Zhang, Cheng, Sun, Wang, Li, Zhu, and Shen]{zhang2025wm_manipulation_survey}
P.-F. Zhang, Y.~Cheng, X.~Sun, S.~Wang, F.~Li, L.~Zhu, and H.~T. Shen.
\newblock A step toward world models: A survey on robotic manipulation.
\newblock \emph{arXiv preprint arXiv:2511.02097}, 2025{\natexlab{e}}.

\bibitem[Zhang and Chen(2023)]{zhang2022fast}
Q.~Zhang and Y.~Chen.
\newblock Fast sampling of diffusion models with exponential integrator.
\newblock In \emph{International Conference on Learning Representations}, 2023.

\bibitem[Zhang et~al.(2023{\natexlab{a}})Zhang, Wang, Sun, Yuan, and Huang]{zhang2023storm}
W.~Zhang, G.~Wang, J.~Sun, Y.~Yuan, and G.~Huang.
\newblock {STORM}: Efficient stochastic transformer based world models for reinforcement learning.
\newblock In \emph{Advances in Neural Information Processing Systems}, volume~36, 2023{\natexlab{a}}.

\bibitem[Zhang et~al.(2026{\natexlab{c}})Zhang, Terver, Zholus, Chitnis, Sutaria, Assran, Bar, Balestriero, Bardes, LeCun, and Ballas]{zhang2026hwm}
W.~Zhang, B.~Terver, A.~Zholus, S.~Chitnis, H.~Sutaria, M.~Assran, A.~Bar, R.~Balestriero, A.~Bardes, Y.~LeCun, and N.~Ballas.
\newblock Hierarchical planning with latent world models.
\newblock \emph{arXiv preprint arXiv:2604.03208}, 2026{\natexlab{c}}.

\bibitem[Zhang et~al.(2025{\natexlab{f}})Zhang, Huang, Ma, Chen, Ma, Du, Zhu, Yang, and Feng]{zhang2025swmap}
X.~Zhang, Y.~Huang, C.~Ma, Z.~Chen, L.~Ma, Y.~Du, S.-C. Zhu, Y.~Yang, and X.~Feng.
\newblock Social world model-augmented mechanism design policy learning.
\newblock \emph{arXiv preprint arXiv:2510.19270}, 2025{\natexlab{f}}.

\bibitem[Zhang et~al.(2025{\natexlab{g}})Zhang, Lin, Mou, Yang, Liu, Sun, Lyu, Yang, Qi, Chen, Li, Yan, Hu, Chen, Wang, Huang, Luo, Tang, Wu, Zhou, and Wei]{zhang2025socioverse}
X.~Zhang, J.~Lin, X.~Mou, S.~Yang, X.~Liu, L.~Sun, H.~Lyu, Y.~Yang, W.~Qi, Y.~Chen, G.~Li, L.~Yan, Y.~Hu, S.~Chen, Y.~Wang, X.~Huang, J.~Luo, S.~Tang, L.~Wu, B.~Zhou, and Z.~Wei.
\newblock {SocioVerse}: A world model for social simulation powered by {LLM} agents and a pool of 10 million real-world users.
\newblock \emph{arXiv preprint arXiv:2504.10157}, 2025{\natexlab{g}}.

\bibitem[Zhang et~al.(2025{\natexlab{h}})Zhang, Zhang, Wang, Ng, and Deng]{zhang2025masim}
X.~Zhang, W.~Zhang, A.~Wang, S.-K. Ng, and Y.~Deng.
\newblock {MASim}: Multilingual agent-based simulation for social science.
\newblock \emph{arXiv preprint arXiv:2512.07195}, 2025{\natexlab{h}}.

\bibitem[Zhang et~al.(2026{\natexlab{d}})Zhang, He, Zhu, Wu, Yu, Chu, Zhang, Tan, and Jia]{zhang2026searchgym}
X.~Zhang, Z.~He, Y.~Zhu, S.~Wu, S.~Yu, M.~Chu, W.~Zhang, H.~Tan, and J.~Jia.
\newblock {SearchGym}: Bootstrapping real-world search agents via cost-effective and high-fidelity environment simulation.
\newblock \emph{arXiv preprint arXiv:2601.14615}, 2026{\natexlab{d}}.

\bibitem[Zhang et~al.(2026{\natexlab{e}})Zhang, Wu, Zhu, Tan, Yu, He, and Jia]{zhang2026scafgrpo}
X.~Zhang, S.~Wu, Y.~Zhu, H.~Tan, S.~Yu, Z.~He, and J.~Jia.
\newblock Scaf-{GRPO}: Scaffolded group relative policy optimization for enhancing {LLM} reasoning.
\newblock In \emph{International Conference on Learning Representations}, 2026{\natexlab{e}}.

\bibitem[Zhang et~al.(2025{\natexlab{i}})Zhang, Mao, Ge, Wang, Xia, Lan, and Wei]{zhang2025k}
Y.~Zhang, S.~Mao, T.~Ge, X.~Wang, Y.~Xia, M.~Lan, and F.~Wei.
\newblock {K-Level} reasoning: Establishing higher order beliefs in large language models for strategic reasoning.
\newblock In \emph{Findings of the Association for Computational Linguistics: NAACL}, pages 7212--7234, 2025{\natexlab{i}}.

\bibitem[Zhang et~al.(2023{\natexlab{b}})Zhang, Sheng, Zhou, Chen, Zheng, Cai, Song, Tian, R{\'e}, Barrett, Wang, and Chen]{zhang2023h2o}
Z.~Zhang, Y.~Sheng, T.~Zhou, T.~Chen, L.~Zheng, R.~Cai, Z.~Song, Y.~Tian, C.~R{\'e}, C.~Barrett, Z.~Wang, and B.~Chen.
\newblock {H2O}: Heavy-hitter oracle for efficient generative inference of large language models.
\newblock In \emph{Advances in Neural Information Processing Systems}, volume~36, pages 34661--34710, 2023{\natexlab{b}}.

\bibitem[Zhang et~al.(2024{\natexlab{b}})Zhang, Li, Wu, Xu, Kag, Skorokhodov, Menapace, Siarohin, Cao, Metaxas, Tulyakov, and Ren]{zhang2024sf}
Z.~Zhang, Y.~Li, Y.~Wu, Y.~Xu, A.~Kag, I.~Skorokhodov, W.~Menapace, A.~Siarohin, J.~Cao, D.~Metaxas, S.~Tulyakov, and J.~Ren.
\newblock {SF-V}: Single forward video generation model.
\newblock In \emph{Advances in Neural Information Processing Systems}, volume~37, pages 103599--103618, 2024{\natexlab{b}}.

\bibitem[Zhang et~al.(2025{\natexlab{j}})Zhang, Qiu, Wu, Li, Wang, Zhou, An, Chen, Li, Wang, Ou, Wang, Chen, Zhang, Hu, Zhang, Wei, Ma, Liu, Dong, He, Feng, Bai, Gao, Sun, and Zheng]{zhang2025origene}
Z.~Zhang, Z.~Qiu, Y.~Wu, S.~Li, D.~Wang, Z.~Zhou, D.~An, Y.~Chen, Y.~Li, Y.~Wang, C.~Ou, Z.~Wang, J.~X. Chen, B.~Zhang, Y.~Hu, W.~Zhang, Z.~Wei, R.~Ma, Q.~Liu, B.~Dong, Y.~He, Q.~Feng, L.~Bai, Q.~Gao, S.~Sun, and S.~Zheng.
\newblock {OriGene}: A self-evolving virtual disease biologist automating therapeutic target discovery.
\newblock bioRxiv 2025.06.03.657658, 2025{\natexlab{j}}.

\bibitem[Zhang et~al.(2025{\natexlab{k}})Zhang, Zhang, Cui, Shi, Guo, Han, Zhao, Sun, Cao, Wang, Cheng, Ju, Che, Xu, and Tang]{zhang2025robooccworld}
Z.~Zhang, Q.~Zhang, W.~Cui, S.~Shi, Y.~Guo, G.~Han, W.~Zhao, J.~Sun, J.~Cao, J.~Wang, H.~Cheng, X.~Ju, Z.~Che, R.~Xu, and J.~Tang.
\newblock Occupancy world model for robots.
\newblock \emph{arXiv preprint arXiv:2505.05512}, 2025{\natexlab{k}}.

\bibitem[Zhao et~al.(2025)Zhao, Foo, Hu, Theobalt, Rahmani, and Liu]{zhao2025agenticreasoning}
B.~Zhao, L.~G. Foo, P.~Hu, C.~Theobalt, H.~Rahmani, and J.~Liu.
\newblock {LLM}-based agentic reasoning frameworks: A survey from methods to scenarios.
\newblock \emph{arXiv preprint arXiv:2508.17692}, 2025.

\bibitem[Zhao et~al.(2026)Zhao, Zhou, Yang, Qin, and Zhou]{zhao2026neurosymbolic_synergy}
H.~Zhao, S.~Zhou, H.~Yang, Z.~Qin, and T.~Zhou.
\newblock Neuro-symbolic synergy for interactive world modeling.
\newblock \emph{arXiv preprint arXiv:2602.10480}, 2026.

\bibitem[Zhao et~al.(2024)Zhao, Fang, Huang, Liu, Wan, Soedarmadji, Li, Lin, Dai, Yan, Yang, Ning, and Wang]{zhao2024vidit}
T.~Zhao, T.~Fang, H.~Huang, E.~Liu, R.~Wan, W.~Soedarmadji, S.~Li, Z.~Lin, G.~Dai, S.~Yan, H.~Yang, X.~Ning, and Y.~Wang.
\newblock {ViDiT-Q}: Efficient and accurate quantization of diffusion transformers for image and video generation.
\newblock \emph{arXiv preprint arXiv:2406.02540}, 2024.

\bibitem[Zhen et~al.(2025)Zhen, Sun, Zhang, Li, Zhou, Du, and Gan]{zhen2025tesseract}
H.~Zhen, Q.~Sun, H.~Zhang, J.~Li, S.~Zhou, Y.~Du, and C.~Gan.
\newblock {TesserAct}: Learning {4D} embodied world models.
\newblock \emph{arXiv preprint arXiv:2504.20995}, 2025.

\bibitem[Zheng et~al.(2025{\natexlab{a}})Zheng, Huang, Liu, Zou, He, Zhang, Zhang, He, Zheng, Qiao, and Liu]{zheng2025vbench2}
D.~Zheng, Z.~Huang, H.~Liu, K.~Zou, Y.~He, F.~Zhang, Y.~Zhang, J.~He, W.-S. Zheng, Y.~Qiao, and Z.~Liu.
\newblock {VBench-2.0}: Advancing video generation benchmark suite for intrinsic faithfulness.
\newblock \emph{arXiv preprint arXiv:2503.21755}, 2025{\natexlab{a}}.

\bibitem[Zheng et~al.(2023)Zheng, Lu, Chen, and Zhu]{zheng2023dpm}
K.~Zheng, C.~Lu, J.~Chen, and J.~Zhu.
\newblock {DPM-Solver-v3}: Improved diffusion ode solver with empirical model statistics.
\newblock In \emph{Advances in Neural Information Processing Systems}, volume~36, pages 55502--55542, 2023.

\bibitem[Zheng et~al.(2025{\natexlab{b}})Zheng, Wang, Ma, Chen, Zhang, Balaji, Chen, Liu, Zhu, and Zhang]{zheng2025large}
K.~Zheng, Y.~Wang, Q.~Ma, H.~Chen, J.~Zhang, Y.~Balaji, J.~Chen, M.-Y. Liu, J.~Zhu, and Q.~Zhang.
\newblock Large scale diffusion distillation via score-regularized continuous-time consistency.
\newblock \emph{arXiv preprint arXiv:2510.08431}, 2025{\natexlab{b}}.

\bibitem[Zheng et~al.(2024)Zheng, Chen, Huang, Zhang, Duan, and Lu]{zheng2024occworld}
W.~Zheng, W.~Chen, Y.~Huang, B.~Zhang, Y.~Duan, and J.~Lu.
\newblock {OccWorld}: Learning a 3{D} occupancy world model for autonomous driving.
\newblock In \emph{European Conference on Computer Vision}, pages 55--72. Springer, 2024.

\bibitem[Zheng et~al.(2025{\natexlab{c}})Zheng, Lin, He, Wang, Fu, Fu, Zheng, and Liang]{zheng2023mcu}
X.~Zheng, H.~Lin, K.~He, Z.~Wang, Q.~Fu, H.~Fu, Z.~Zheng, and Y.~Liang.
\newblock {MCU}: An evaluation framework for open-ended game agents.
\newblock In \emph{International Conference on Machine Learning}, pages 78221--78259. PMLR, 2025{\natexlab{c}}.

\bibitem[Zheng et~al.(2026)Zheng, Zhong, Wang, Dai, Liu, Chu, Lv, Torr, and Lin]{zheng2026code2world}
Y.~Zheng, L.~Zhong, Y.~Wang, R.~Dai, K.~Liu, X.~Chu, L.~Lv, P.~Torr, and K.~Q. Lin.
\newblock {Code2World}: A {GUI} world model via renderable code generation.
\newblock \emph{arXiv preprint arXiv:2602.09856}, 2026.

\bibitem[Zhou et~al.(2024{\natexlab{a}})Zhou, Zheng, Wang, Yin, and Huang]{zhou2024score}
M.~Zhou, H.~Zheng, Z.~Wang, M.~Yin, and H.~Huang.
\newblock Score identity distillation: Exponentially fast distillation of pretrained diffusion models for one-step generation.
\newblock In \emph{International Conference on Machine Learning}, pages 62307--62331. PMLR, 2024{\natexlab{a}}.

\bibitem[Zhou et~al.(2024{\natexlab{b}})Zhou, Xu, Zhu, Zhou, Lo, Sridhar, Cheng, Ou, Bisk, Fried, Alon, and Neubig]{zhou2023webarena}
S.~Zhou, F.~F. Xu, H.~Zhu, X.~Zhou, R.~Lo, A.~Sridhar, X.~Cheng, T.~Ou, Y.~Bisk, D.~Fried, U.~Alon, and G.~Neubig.
\newblock {WebArena}: A realistic web environment for building autonomous agents.
\newblock In \emph{International Conference on Learning Representations}, 2024{\natexlab{b}}.

\bibitem[Zhou et~al.(2024{\natexlab{c}})Zhou, Zhu, Mathur, Zhang, Yu, Qi, Morency, Bisk, Fried, Neubig, and Sap]{zhou2024sotopia}
X.~Zhou, H.~Zhu, L.~Mathur, R.~Zhang, H.~Yu, Z.~Qi, L.-P. Morency, Y.~Bisk, D.~Fried, G.~Neubig, and M.~Sap.
\newblock {SOTOPIA}: Interactive evaluation for social intelligence in language agents.
\newblock In \emph{International Conference on Learning Representations}, 2024{\natexlab{c}}.

\bibitem[Zhou et~al.(2025{\natexlab{a}})Zhou, Liang, Tu, Chen, Ding, Zhang, Tan, Zhao, and Bai]{zhou2025hermes}
X.~Zhou, D.~Liang, S.~Tu, X.~Chen, Y.~Ding, D.~Zhang, F.~Tan, H.~Zhao, and X.~Bai.
\newblock {HERMES}: A unified self-driving world model for simultaneous {3D} scene understanding and generation.
\newblock In \emph{IEEE/CVF International Conference on Computer Vision}, pages 27817--27827, 2025{\natexlab{a}}.

\bibitem[Zhou et~al.(2025{\natexlab{b}})Zhou, Liu, Yerukola, Kim, and Sap]{zhou2025socialwm}
X.~Zhou, J.~Liu, A.~Yerukola, H.~Kim, and M.~Sap.
\newblock Social world models.
\newblock \emph{arXiv preprint arXiv:2509.00559}, 2025{\natexlab{b}}.

\bibitem[Zhu et~al.(2025)Zhu, Wang, Zhou, Chang, Zhou, Li, Chen, Shen, Pang, and He]{zhu2025aether}
H.~Zhu, Y.~Wang, J.~Zhou, W.~Chang, Y.~Zhou, Z.~Li, J.~Chen, C.~Shen, J.~Pang, and T.~He.
\newblock Aether: Geometric-aware unified world modeling.
\newblock In \emph{IEEE/CVF International Conference on Computer Vision}, pages 8535--8546, 2025.

\bibitem[Zhu et~al.(2024)Zhu, Wang, Zhao, Min, Li, Deng, Dou, Wang, Shi, Wang, Zhang, You, Zhang, Zhao, Xiao, Zhao, Lu, and Huang]{zhu2024sora_survey}
Z.~Zhu, X.~Wang, W.~Zhao, C.~Min, B.~Li, N.~Deng, M.~Dou, Y.~Wang, B.~Shi, K.~Wang, C.~Zhang, Y.~You, Z.~Zhang, D.~Zhao, L.~Xiao, J.~Zhao, J.~Lu, and G.~Huang.
\newblock Is {Sora} a world simulator? a comprehensive survey on general world models and beyond.
\newblock \emph{arXiv preprint arXiv:2405.03520}, 2024.

\end{thebibliography}
\makeatletter
\if@github
  \bibliographystyle{abbrvnat}
\else
  \bibliographystyle{tmlr}
\fi
\makeatother
\endgroup

\clearpage
\appendix
\section{Philosophical Motivations for Hierarchical World Modeling}
\label{app:philosophy_extended}

This appendix expands the philosophical motivations behind the L1/L2/L3 hierarchy of Section~\ref{sec:philosophy}.

\subsection{L1: Inductive Priors and the No-Free-Lunch Theorem}

The i.i.d.~ premise in supervised learning mirrors Hume's uniformity-of-nature assumption: without it, induction has no logical warrant.
\citet{wolpert1996nofreelunch} formalizes a complementary point: absent structural constraints, no learner is universally better than any other.
Architectures and training curricula therefore serve as \emph{inductive priors} analogous to what \citet{kant1781critique} called \emph{synthetic a priori} structures.
In practice, mechanisms such as
convolutions, equivariances, attention patterns, and JEPA-style prediction targets~\citep{lecun2022path} constrain which regularities are even expressible.
LeCun's JEPA-centric architecture (perception, world model, actor, cost, short-term memory) can be viewed as one particular choice of such priors; our taxonomy classifies the \textbf{capabilities} these design choices enable, regardless of module count or naming convention.

This perspective is supported by cognitive science.
The predictive coding framework in cognitive science~\citep{rao1999predictive} posits that the brain continuously generates top-down predictions of incoming sensory signals and updates its internal model by minimizing prediction error.
Friston's active inference framework~\citep{friston2010free} unifies perception, planning, and exploration under variational free energy, blurring the boundary between prediction and action.
The ``Bayesian brain'' hypothesis~\citep{clark2015surfing} holds that perception itself is a form of probabilistic inference over latent causes of sensory input, suggesting that one-step latent forecasting is a primitive from which richer world-modeling capabilities emerge~\citep{lake2017building}.

Contemporary systems that embody L1's Humean empiricism include Dreamer-style latent prediction~\citep{hafner2019dreamer, hafner2020dreamerv2, hafner2023dreamerv3} and large sequence models for short-horizon forecasting.
These systems extract temporal regularities from trajectories and assume persistence, wagering next-step accuracy.

\subsection{L2: Modal Semantics, Epistemic Drift, and Active Inference}

Modal semantics (``possible worlds,'' ``closest'' counterfactuals)~\citep{kripke1963semantic, kripke1980naming, stalnaker1968conditionals} supplies helpful \textbf{vocabulary} (latent states index alternative futures; actions carve navigable branches), but the core engineering content is interventional rather than purely modal.

Lewis's theory of ``closest possible worlds''~\citep{lewis1973counterfactuals} provides an operational core of counterfactual reasoning: effective reasoning does not involve exploring arbitrary possibilities but analyzing worlds that are maximally similar to our own, where a minimal intervention yields a coherent trajectory.
This ``near-factual'' heuristic lets L2 simulators remain tractable while still supporting intervention-structured imagination.

MuZero's search over learned dynamics~\citep{schrittwieser2020muzero} is a concrete instance: it uses Monte Carlo Tree Search to explore action sequences in a learned model, demonstrating how Lewis's ``closest possible worlds'' theory translates into practical AI systems that enable counterfactual reasoning for decision-making.

Plato's cave offers a diagnostic image (shadows projected from an incomplete generator~\citep{plato1992republic}) as a reminder that visual fluency does not guarantee fidelity.
An L2 simulator that excels at predicting shadows on a wall may remain fundamentally bounded by the wall's dimensions, unable to access the fire that casts those shadows.
This metaphor captures \textbf{epistemic drift}: internally coherent trajectories that leave the training manifold.
When the data distribution used to train the simulator does not align with the true ``fire'' of reality, the L2 agent becomes trapped within its own simulated shadows.
Its ``modal stability'' thus becomes its greatest liability: the more a model relies on its internal assumptions to fill in the gaps of missing data, the more brittle it becomes when confronted with the irreducible complexity of the real world.

Friston's active inference framework~\citep{friston2017active} addresses a similar concern from the opposite direction: by unifying perception, planning, and exploration under variational free energy, it blurs the L1/L2 boundary into a continuum.
Our stage boundary remains useful as an engineering diagnostic even when the theory is continuous: the question is whether rollouts are reliable enough downstream.

\subsection{L3: Falsifiability, Paradigm Shifts, and Abduction}

Popper's emphasis on \textbf{risky, falsifiable predictions}~\citep{popper1959logic} resonates with governed L3 loops: proposed revisions should yield measurable improvements on held-out probes, regression suites, or experimental outcomes, not merely post-hoc accommodation.
\citet{kuhn1962structure}'s ``paradigm shift'' provides a useful \textbf{contrast}: Kuhn stresses non-cumulative revolutionary breaks, whereas most engineered L3 systems today perform \textbf{incremental} scaffold updates, closer to Lakatos's progressive problem-shifts than to Kuhnian revolution.

Peircean \textbf{abduction} (inference to the best explanation)~\citep{peirce1931collected} loosely motivates the hypothesis-generation step when monitors flag anomaly, but the analogy should not be pressed too far: contemporary systems typically search within \emph{structured} spaces (program sketches, simulator hooks, experiment templates)~\citep{lu2024aiscientist, yamada2025aiscientistv2} rather than inventing wholly new ontologies.

Lakatos's methodology of scientific research programmes~\citep{lakatos1978methodology} offers a precise framework for understanding model revision.
Systems have a ``hard core'' of fundamental principles (the model's architecture and inductive biases) and a ``protective belt'' of auxiliary hypotheses (the learned parameters).
In conventional training (L1 and L2), errors are absorbed by the protective belt via gradient descent.
When persistent structured errors occur, however, an L3 system identifies that the crisis lies within the hard core itself and performs what Kuhn called a ``paradigm shift'', reorganizing its internal ontology.

The Duhem--Quine thesis of confirmation holism~\citep{duhem1954aim, quine1951two} explains why blame-assignment is non-trivial: errors redistribute across modules until diagnostics isolate the brittle component.
This holism complicates the transition from parameter adjustment to structural revision, making evidence-driven diagnosis via held-out probes and targeted ablations a central capability for L3 systems.

Moerland et al.'s model-based RL taxonomy~\citep{moerland2023mbrl} organizes methods along axes such as how dynamics integrate with policy optimization; those axes are \textbf{orthogonal complements} to ours; two algorithms with identical integration strategy can sit at different capability levels depending on simulation depth and whether the model structure itself changes under evidence.

\section{Conceptual Boundaries}
\label{app:boundaries}

\subsection{World Modeling versus Generic Prediction}
Unlike standard machine learning prediction (e.g., classification, recommendation), world modeling targets \emph{stateful dynamics}: how environments evolve over time under actions or interventions.
The three L2 boundary conditions defined above (long-horizon coherence, intervention sensitivity, and constraint consistency) already capture the core of this distinction at the simulation level.
A fourth capability further separates world modeling but is \textbf{orthogonal} to the L1/L2/L3 hierarchy:
\begin{itemize}[leftmargin=*]
  \item \textbf{Closed-loop use:} supporting planning, acting, and self-improvement through interaction with the modeled environment.  This capability is essential for embodied agents but is \textbf{not} part of the L1/L2/L3 definition: a weather emulator or video generator can be an L2 world model with no embedded planner.  Conversely, strong closed-loop performance additionally requires exploration, reward specification, safety, and search; failure modes that are not always traceable to dynamics error alone~\citep{kaelbling1998pomdp, schrittwieser2020muzero}.
\end{itemize}
A useful practical distinction is that world models are organized around action-conditioned transition queries: by conditioning on actions, they compress the simulation problem into decision-relevant futures rather than attempting to model all observable variation indiscriminately.

\subsection{World Model versus Planner}
\label{subsubsec:wm_vs_planner}
The \emph{world model} is \textbf{descriptive}: it approximates how states and observations evolve under actions or interventions.
The \emph{planner} is \textbf{normative}: it chooses actions to optimize an objective given those predictions~\citep{kaelbling1998pomdp, puterman1994mdp}.
Conflating the two obscures where failures originate (wrong dynamics vs.\ wrong objectives or search) and blocks reuse of one model with many planners.

The world model corresponds to \(T\), \(O\), and learned \((q_\phi, p_\theta, p_\psi)\); the planner is \(\pi\), a value function, or a search procedure consuming rollouts (MCTS, imagined trajectories, etc.)~\citep{schrittwieser2020muzero, hafner2023dreamerv3}.
Different planners can sit atop the same dynamics; different dynamics can be swapped into the same planner for ablation.
Forecasting simulators and video models illustrate world models without planners~\citep{brooks2024sora, ding2024survey_wm}; model-based RL typically \emph{co-trains} dynamics and policy (Dreamer-style), yet the \textbf{roles} remain distinct~\citep{hafner2019dreamer, hafner2023dreamerv3}.

The planner issues \emph{queries} (one-step forward, multi-step rollout, counterfactual edits); the world model answers them.
L1/L2/L3 classify \textbf{query depth and reliability}, not the planner's existence.

\subsection{World Modeling versus Commonsense Modeling}
World modeling targets \textbf{how} states evolve; commonsense supplies \textbf{what usually persists, what is relevant, and what must not break} across transitions~\citep{johnson1983mental, lecun2022path, marcus2018deep}.
However, neither alone suffices a comprehensive modelling of the real world: fluent rollouts can violate invariants, while static ``facts'' without dynamics cannot support control. For example, a world model may generate a visually smooth rollout of a cup falling, yet without commonsense or physical invariants it may let the cup pass through a solid table; conversely, knowing that solid objects do not interpenetrate does not by itself predict the exact trajectory after a push.

World modeling supports predictions \(z_{t-1} \rightarrow z_t\), \((z_{t-1},a_t)\rightarrow z_t\), and \(z_t\rightarrow o_t\), including L1 operators and L2 trajectories.
Commonsense encodes persistence, default invariants, and normative structure, classically linked to the \textbf{frame problem} (specifying what does \emph{not} change)~\citep{mccarthy1969some}.
In L2, \textbf{constraint consistency} operationalizes a \textbf{testable subset} of commonsense: violations are measurable against explicit rules or domain simulators, even though full commonsense remains open-ended.

\subsection{Taxonomy Comparison}
\label{subsec:taxonomy_comparison}

Our L1/L2/L3 hierarchy is not the only capability decomposition for intelligent systems. We situate it against four influential frameworks below. Correspondences are \emph{partial and analogical}, not one-to-one: each comparison framework was designed with a different scope and computational commitment, and the levels do not map cleanly onto L1/L2/L3 in every case. The key distinction is that our taxonomy is \emph{architecture-agnostic} and \emph{cross-domain}: it applies uniformly across physical, digital, social, and scientific regimes.

\begin{enumerate}
    \item \textbf{Pearl's causal hierarchy}~\citep{pearl2009causality}: L1 $\approx$ Association; L2 $\approx$ Intervention. The counterfactual rung reasons within a fixed model, whereas L3 revises the model itself.
    \item \textbf{LeCun's autonomous-machine architecture}~\citep{lecun2022path}: L1 $\approx$ Reactive; L2 $\approx$ Planning via world model. This framework lacks an explicit model-revision stage comparable to L3.
    \item \textbf{Friston's active inference}~\citep{friston2010free}: L1 $\approx$ Perception; L2 $\approx$ Planning. Provides a unified Bayesian principle but includes no discrete revision stage comparable to L3.
    \item \textbf{Moerland et al.'s MBRL survey}~\citep{moerland2023mbrl}: L1 $\approx$ Model learning; L2 $\approx$ Planning with learned model. Organizes by integration strategy rather than capability level, and is RL-specific.
\end{enumerate}

\section{Detailed L2 Systems by Different Domains}
\label{app:l2_extended}

This appendix provides extended details for methods summarized in Section~\ref{sec:l2}.

\subsection{Embodied and 3D Systems}
Two recurring dimensions help compare these systems: the \textbf{representation carrier} (occupancy, point/particle, Gaussian, or asset-based scene state) and the
\textbf{degree of action-coupling}, ranging from passive continuation to
action-conditioned forecasting and simulator-ready planning support.

\paragraph{3D-structured world models.}
A visible trend in learned physical-world modeling is the shift from
appearance-first continuation toward geometry-carrying simulation.
Rather than merely extending pixels frame by frame, these systems increasingly
maintain explicit or semi-explicit 3D scene structure that can support
action-conditioned forecasting, collision reasoning, and planning~\citep{zhan2024bundle}.

\emph{Industrial previews.}
As illustrative industrial signals rather than peer-reviewed systems, World Labs' 2025 previews point in this direction: RTFM emphasizes persistent real-time interaction, while Marble lifts text, images, and coarse 3D layouts into explorable 3D worlds~\citep{worldlabs2025rtfm,worldlabs2025marble}.

\emph{Geometry and volumetric forecasting.}
Among academic systems, Aether couples reconstruction, action-conditioned
prediction, and visual planning in a geometry-aware framework
\citep{zhu2025aether}, while TesserAct formulates embodied forecasting as learning
4D scenes with spatial and temporal consistency \citep{zhen2025tesseract}.
RoboOccWorld makes geometric commitment explicit at the volumetric level by
forecasting scene evolution directly in occupied 3D space, so that whether a 
structure is occupied becomes a first-class output for collision checking, motion planning,
and multi-step spatial reasoning \citep{zhang2025robooccworld}.

\emph{Fine-grained physical representations.}
Other work moves toward more detailed simulation substrates. GAF represents robotic interaction with 4D Gaussian fields \citep{chai2025gaf}, ParticleFormer models point-cloud dynamics for multi-material manipulation \citep{huang2025particleformer}, and GWM treats Gaussian primitive propagation as both a neural simulator and a representation-learning substrate \citep{lu2025gwm}. LiDARCrafter \citep{liang2026lidarcrafter}, DynamicCity \citep{bian2025dynamiccity}, and U4D \citep{xu2025u4d} simulate 4D worlds from native 3D representations, such as point clouds and occupancy grids. PointWorld further unifies state and action as 3D point flows, targeting cross-embodiment manipulation \citep{huang2026pointworld}. Taken together, these methods suggest a shift from generic video continuation toward representations that expose finer operational structure for contact-rich, action-sensitive dynamics.

\emph{Simulation-ready extensions.}
A related direction focuses less on forecasting alone and more on producing
assets and views that can directly support downstream interaction.
PhysX-Anything generates articulated, physical 3D assets from a single
in-the-wild image for direct use in simulation \citep{cao2025physxanything},
while MVISTA-4D extends toward single-view-to-arbitrary-view RGBD imagination
with test-time action optimization \citep{wang2026mvista4d}. Across these
systems, the world model becomes less like a passive renderer and more like a
spatially queryable latent scene machine. Broadly, occupancy-based methods favor global free-space reasoning and collision
checking but remain relatively coarse; point- and particle-based methods better
capture contact-rich local dynamics but are harder to scale; Gaussian-style
representations improve visual and spatial fidelity but often require additional
structure for strict physical interaction.

\paragraph{Autonomous driving world models.}
Autonomous driving is a particularly clear L2 setting because useful rollouts must jointly preserve geometric accuracy (lane structure, free space), dynamic consistency (vehicle kinematics, traffic flow), and counterfactual sensitivity: if the ego vehicle brakes earlier or changes lanes, surrounding trajectories and occupancy should update coherently rather than merely continuing the same scene \citep{survey_vla4ad,wang2025alpamayo,worldlens}.
Earlier systems like GAIA-1~\citep{hu2023gaia1} and DriveWorld~\citep{min2024driveworld} established scene generation conditioned on control signals.

Subsequent work has branched along two main axes.
Along the \emph{representation} axis, Copilot4D~\citep{zhang2024copilot4d}
introduced unsupervised 4D modeling via discrete diffusion on LiDAR point
clouds, OccWorld~\citep{zheng2024occworld} moved to 3D occupancy with a
GPT-like spatial-temporal transformer, and
Hermes~\citep{zhou2025hermes} unified BEV scene understanding with future
generation.
Along the \emph{fidelity--controllability} axis,
VISTA~\citep{gao2024vista} demonstrated 576$\times$1024 resolution at 10~Hz
with 15-second coherent rollouts, while
DriveDreamer~\citep{wang2024drivedreamer} built world models entirely from
naturalistic driving data using a diffusion backbone. AD-R1 \citep{yan2025ad-r1} builds the first closed-loop simulator by combining impartial world modeling with a rich curriculum of plausible collisions and off-road events.

A further line of work concerns \emph{policy alignment under fine-tuning}
rather than base representation alone:
AdaWM~\citep{zhao2025adawm} addresses representation degradation during RL
fine-tuning via low-rank alignment that preserves pre-trained structure while
adapting to new driving policies. This progression also marks a shift from
open-loop scene generation to closed-loop control support, where actions are
not merely conditioning variables but candidate interventions whose
consequences must be compared before execution.

\subsection{Software, Web, and Game Systems}

\paragraph{Game world models.}
Game worlds occupy a distinctive position at the intersection of physical and digital intelligence: visual dynamics follow physics-like rules (rendering, object motion, collision), yet transitions are ultimately governed by deterministic game logic (score updates, level triggers, inventory changes). This overlap makes games a natural testbed for world models that must integrate perceptual prediction with rule-based reasoning. NitroGen~\citep{nitrogen2025}, NVIDIA's open vision-action foundation model trained on 40K hours of gameplay across 1000+ games, achieves 52\% improvement on unseen games via large-scale behavior cloning. Earlier work at L1, including DIAMOND~\citep{alonso2024diamond} and Genie~\citep{bruce2024genie} (Section~\ref{sec:l1}), established frame-by-frame prediction; the L2 challenge is long-horizon, action-conditioned simulation respecting both visual dynamics and underlying game rules. GameNGen~\citep{valevski2025gamengin} demonstrated that a diffusion model trained on DOOM gameplay can serve as a real-time neural game engine at 20~FPS, generating interactive frames indistinguishable from the original engine. Video2Game~\citep{xia2024video2game} converts a single video into an interactive 3D game-like environment with real-time physics and rendering, bridging passive video understanding with interactive world simulation. Across these domains, state includes DOM structure, focus, file system, and application state machines; evaluable tasks span OS~\citep{xie2024osworld,yang2025macosworld}, web~\citep{zhou2023webarena,deng2023mind2web,yao2022webshop}, and software debugging workflows~\citep{jimenez2023swebench,yang2024sweagent,shi2017worldofbits}.

\subsection{Social Simulation and Multi-Agent Systems}

\paragraph{ToM prompting and reasoning.}
Structured prompting strategies suggest the bottleneck in social reasoning is reasoning structure rather than knowledge. SymbolicToM~\citep{sclar2023symbolictom} constructs explicit per-character belief graphs after each story event, supporting up to third-order beliefs through graph traversal (ACL 2023 Outstanding Paper). SimToM~\citep{wilf2024simtom} implements perspective-taking as a two-stage process inspired by Simulation Theory from cognitive science: first filtering context to what the target character knows, then answering from that filtered view. K-Level Reasoning~\citep{zhang2025k} implements the behavioral economics Level-K framework recursively in LLMs for negotiation. Thought-Tracing~\citep{kim2025hypothesis} implements approximate Bayesian inference via Sequential Monte Carlo-like hypothesis generation, significantly outperforming reasoning models like o3-mini, suggesting social reasoning may require fundamentally different computational mechanisms than mathematical deduction.

\paragraph{Sandbox architectures and scale.}
Project Sid~\citep{altera2024projectsid} deployed up to 1,000 agents across six towns in Minecraft using the PIANO architecture (Parallel Information Aggregation via Neural Orchestration), a brain-inspired modular design with separate concurrent modules for cognition, planning, motor execution, and speech. Emergent phenomena included autonomous professional specialization, personality-driven social network formation, democratic governance, and cultural transmission including spontaneous religious proselytization. Sotopia extensions include Sotopia-$\pi$~\citep{wang2024sotopia_pi} (interactive self-reinforcement learning for social skills) and Lifelong-Sotopia (multi-episode long-term consistency evaluation). AgentSociety~\citep{piao2025agentsociety} simulated 10,000+ agents generating 5 million interactions in an integrated urban-social-economic environment with emotion and cognitive modeling inspired by Maslow's hierarchy. Deployed platforms such as Moltbook\footnote{\url{https://www.moltbook.com/}} provide persistent social environments where AI agents autonomously post, discuss, and form community norms, bridging the gap between simulation and real-world agent societies.

\paragraph{Emergent social phenomena.}
Only 2 of 15 LLMs achieve sustainable cooperation in commons dilemma scenarios~\citep{piatti2024cooperateorcollapse}, and cooperation evolution across generations of LLM agents proves strongly model-dependent~\citep{vallinder2024culturalevolution}. Yet norms and conventions do emerge: \citet{ren2024norms} document norm formation in LLM societies, and \citet{ashery2025conventions} find social conventions with critical mass tipping points, where collective biases appear at the group level that do not exist in individual agents. Melting Pot~\citep{leibo2021meltingpot} provides 50+ substrates covering cooperation, competition, deception, and coordination for systematic evaluation of such dynamics. Role-playing systems such as RoleLLM~\citep{wang2024rolellm}, CharacterLLM~\citep{shao2023characterllm}, and ChatHaruhi~\citep{li2023chatharuhi} probe character-consistency through persona fine-tuning and memory-based maintenance. \citet{shanahan2023roleplay} argue that LLMs maintain implicit world models of character situations through distributional representations. Werewolf and Avalon serve as concentrated testbeds for deception and trust: comprehensive Avalon investigation~\citep{lan2024avalonllm} documented emergent leadership and camouflage strategies, ReCon~\citep{wang2024recon} introduced recursive perspective transitions for deception handling, and The Traitors~\citep{curvo2025traitors} found that deceivers consistently prevail by exploiting the cognitive limitations of honest participants.

\paragraph{Digital twin societies.}
S$^3$~\citep{gao2023s3} simulates information propagation, emotion contagion, and attitude polarization on social media platforms; an extended version successfully predicted 2024 US presidential election results, demonstrating predictive validity for real-world phenomena. SocioVerse~\citep{zhang2025socioverse} validates social simulation against a pool of 10 million real-world users, enabling election prediction, breaking-news response, and economic survey replication at unprecedented scale. PersuasionForGood~\citep{wang2019persuasionforgood} modeled persuasion as a social state transition process, tracking how 10 distinct strategies shift attitudes, establishing that social dynamics are personalized rather than universal.

\paragraph{Institutional and formal approaches.}
As~\citet{dignum2026agentifying} argue, current LLM-based agents exhibit behavioral autonomy without explicit reasoning structures. The BDI (Belief--Desire--Intention) architecture~\citep{rao1995bdi}, normative multi-agent systems~\citep{boella2007norms}, electronic institutions~\citep{esteva2001electronic}, and formal commitment models~\citep{telang2023commitments} provide the missing machinery: explicit, inspectable representations of mental states, social obligations, and institutional roles. MetaGPT~\citep{hong2024metagpt} encodes organizational knowledge through Standardized Operating Procedures, and ChatDev~\citep{qian2024chatdev} implements chat-chain architectures with communicative dehallucination, both showing that explicit institutional constraints outperform individual agent prompting for organizational coherence. Strategic dialogue systems further test social dynamics: CraigslistBargain~\citep{he2018craigslist} decoupled strategy from generation, NegotiationArena~\citep{bianchi2024negotiationarena} quantifies irrational behaviors, the Consensus Game~\citep{jacob2024consensus} formalizes LM decoding as equilibrium search, and the Game-theoretic LLM framework~\citep{hua2024gametheoreticllm} incorporates backward induction into agent workflows.

\subsection{AI-for-Science Systems}

\paragraph{Neural dynamics and interpretability.}
DyNeMo~\citep{gohil2022dynemo,khan2023dynemoc} combines an encoder mapping observations to latent network modes with a memory model capturing their temporal evolution, forming a generative dynamical system. With this structure, DyNeMo supports forward simulation of future latent states and prediction of neural responses to external interventions~\citep{helfrich2014entrainment,ngo2013auditory} via in-silico simulation rather than direct experimentation. However, unlike physical systems where governing laws are well established, the dynamics of large-scale neural activity remain largely unknown, shifting the primary scientific objective toward interpretable mechanism discovery. DyNeMo facilitates this by learning structured and interpretable latent representations that capture spatial patterns of functional brain networks, whose temporal statistics reveal higher-level organizational principles including structured cycles in network activations~\citep{van2025large}. This highlights a distinct role of scientific world models: not only simulating known dynamics, but discovering the state space and transition structure themselves through interpretable representations and their statistical regularities.

\paragraph{Operator learning and molecular surrogates.}
The neural operator framework~\citep{kovachki2023neuraloperator} provides a unified theoretical foundation for learning maps between infinite-dimensional function spaces, establishing approximation theory and error bounds that underpin FNO, DeepONet, and subsequent architectures. PINO~\citep{li2021pino} combines the neural operator architecture with physics-informed PDE residual losses, enabling zero-shot super-resolution and improved generalization under sparse data. PI-DeepONet~\citep{goswami2022pideepone} extends the DeepONet framework with physics-informed training, embedding governing PDE residuals directly into the operator learning objective. SchNet~\citep{schutt2017schnet} introduced continuous-filter convolutions for molecular graphs, enabling end-to-end learning of quantum chemical properties without handcrafted features and serving as the architectural precursor to equivariant GNN potentials. For a comprehensive treatment of ML approaches to molecular simulation, including neural potentials, coarse-grained models, and generative sampling, see \citet{noe2020mlmolsim}. Boltzmann Generators~\citep{noe2019boltzmann} pioneered deep generative models for sampling thermodynamic equilibrium states of molecular systems, bypassing the sequential bottleneck of traditional molecular dynamics. ClimaX~\citep{nguyen2023climax} introduced the foundation-model paradigm for weather and climate, pretraining on CMIP6 reanalysis data with self-supervised learning and fine-tuning to both forecasting and climate projection tasks.

\section{Illustrative L3 Evolution Loops}
\label{app:l3_examples}

This appendix provides illustrative worked examples showing how the L3 evolution loop operates in each governing-law regime. These are constructive illustrations, not reports of deployed system behavior.

\subsection{Dynamics Model Revision from Grasp Failure}

A robot repeatedly drops an object despite the planner predicting success.
\begin{enumerate}
\item \emph{Anomaly}: force/torque sensor logs and contact timestamps reveal systematic deviation from predicted friction.
\item \emph{Attribution}: the dynamics model underestimates friction for smooth cylindrical objects.
\item \emph{Revision}: update the friction prior; add a pre-grasp diagnostic squeeze to estimate friction online; create a regression test with 5 material variants.
\item \emph{Validation}: replay across RoboCasa~\citep{nasiriany2024robocasa} tasks to confirm the update does not degrade other material classes.
\end{enumerate}

\subsection{Strategy Revision in a Service Environment}

A conversational agent in a simulated service environment consistently fails to retain users who threaten to cancel, despite the societal world model predicting that a discount would resolve the conversation.
\begin{enumerate}
\item \emph{Anomaly}: across 50 cancellation dialogues, the discount offer succeeds only 20\% of the time, far below the model's predicted 70\% retention rate.
\item \emph{Attribution}: the social model assumes users who threaten cancellation are price-sensitive, but conversation analysis reveals that most are frustrated by service quality, not price. The model's belief attribution is systematically wrong.
\item \emph{Revision}: update the user-intent classifier to distinguish price-sensitive from quality-frustrated users; for quality-frustrated users, replace the discount strategy with an acknowledgment-and-escalation strategy; add a regression test ensuring price-sensitive users still receive discounts.
\item \emph{Validation}: A/B test the revised strategy on the next 100 cancellation interactions, measuring retention rate, user satisfaction, and whether the revision introduces new failure modes (e.g., unnecessary escalations for genuinely price-sensitive users).
\end{enumerate}

\subsection{Installation Failure Producing Reusable Skill}

An agent runs ``install dependencies,'' encounters an error, and retries fruitlessly.
\begin{enumerate}
\item \emph{Anomaly}: error codes and permission status reveal a \texttt{permission denied} failure class.
\item \emph{Attribution}: insufficient privileges, not a typo or network error.
\item \emph{Revision}: create a skill template (on permission errors: check privileges $\rightarrow$ choose alternative install path $\rightarrow$ re-verify), plus a regression test in a restricted-permission environment.
\item \emph{Validation}: replay across OSWorld and macOSWorld variants~\citep{xie2024osworld,yang2025macosworld} to confirm that the extracted skill transfers beyond the original failure case.
\end{enumerate}

\subsection{Closed-Loop Materials Discovery}

A Bayesian surrogate predicts a specific crystal phase, but synthesis yields an unexpected mixed phase.
\begin{enumerate}
\item \emph{Anomaly}: XRD diffraction pattern deviates from the predicted single-phase output.
\item \emph{Attribution}: the surrogate underweights the effect of synthesis temperature on phase stability.
\item \emph{Revision}: update the Bayesian model with the new data point; expand the hypothesis space to include temperature-dependent phase boundaries; design a follow-up experiment at an intermediate temperature.
\item \emph{Validation}: the next synthesis confirms the revised prediction, and calibration improves on held-out compositions~\citep{kusne2020cameo,szymanski2023alab}.
\end{enumerate}

\section{Additional Details of Evaluation and Benchmarks}
\label{app:eval_extended}

This appendix provides detailed evaluation protocols for the three L2 boundary conditions, along with world-model-specific benchmarks, capability coverage analysis, and the Minimal Reproducible Evaluation Package (MREP) introduced in Section~\ref{sec:evaluation} of the main text.

\subsection{Long-Horizon Coherence}

Long-horizon coherence asks whether composed predictions $\hat p(\tau\mid z_0,a_{1:H},c)$ remain usable as the rollout horizon $H$ grows. The signature failure is \textbf{compounding error}: small per-step deviations amplify over time, pushing imagined trajectories into unreachable branches (Section~\ref{subsec:l2_failure_modes}, failure mode~1). A secondary failure is \textbf{state aliasing} (failure mode~2), where distinct real states collapse into similar representations, causing the rollout to silently diverge from reality.

Operationally, coherence is measured by tracking a task-relevant metric as a function of horizon. For physical manipulation, RoboCasa~\citep{nasiriany2024robocasa} and ManiSkill3~\citep{tao2024maniskill3} offer multi-step tasks where success rate degrades predictably with horizon; the degradation curve itself is the diagnostic. SWE-bench~\citep{jimenez2023swebench} poses an analogous challenge for code: multi-file resolution rate reveals whether the model maintains a coherent codebase state across $k$ interdependent editing steps. Social settings introduce a different flavour of drift. Sotopia~\citep{zhou2024sotopia} tracks whether commitments, alliances, and relational variables remain stable across multi-turn interactions rather than silently eroding. Scientific reasoning demands another form of coherence: in ScienceWorld~\citep{wang2022scienceworld}, sequences of lab actions must preserve causal ordering (e.g., heating a substance before measuring its temperature, not after).

A significant gap in current practice is that most benchmarks report fixed-horizon success rates without degradation curves: a system is tested at horizon $H$ but the relationship between performance and $H$ is not characterized. Techniques from long-form video understanding, such as temporal search methods that selectively retrieve and reason over distant segments~\citep{ye2025re}, suggest a promising direction for diagnosing coherence at scale. Open-world environments provide the strongest existing tests of long-horizon coherence precisely because they demand skill composition across hundreds of steps. Minecraft tasks (via MCU~\citep{zheng2023mcu} or Voyager~\citep{wang2023voyager}) require gathering resources, crafting tools, and navigating terrain in sequences that can exceed a thousand actions. Crafter~\citep{stanic2023learning} compresses similar compositional demands into a more controlled setting, while NetHack~\citep{kurenkov2023katakomba} adds procedural generation and symbolic complexity. In model-based RL, MBPO-style horizon truncation~\citep{janner2019mbpo} implicitly acknowledges coherence limits by restricting rollout length to a range where the model remains accurate, but this adaptive truncation is rarely evaluated as a coherence metric in its own right.

\subsection{Intervention Sensitivity}

Intervention sensitivity asks whether the rollout responds meaningfully to changes in the action sequence $a_{1:H}$ or the initial condition $z_0$. A model that produces the same trajectory regardless of the action taken is useless for planning, even if each individual frame is perceptually realistic. This directly tests \textbf{controllability} (Section~\ref{subsec:l2_failure_modes}, failure mode~3)~\citep{wu2024ivideogpt,brooks2024sora}.

The core protocol is \emph{counterfactual divergence testing}: from the same $z_0$, execute two action sequences $a_{1:H}$ and $a'_{1:H}$ that differ at a single step, and measure whether the resulting trajectories $\tau$ and $\tau'$ diverge in a task-relevant way. Two complementary metrics capture this: the \emph{action sensitivity ratio} (fraction of action perturbations that produce a detectable outcome change) and the \emph{counterfactual outcome divergence} (magnitude of difference in task-relevant variables between $\tau$ and $\tau'$, normalized by the action change magnitude). In OSWorld~\citep{xie2024osworld}, this corresponds to injecting pop-up interruptions or network failures and testing whether the agent replans rather than clicking blindly. In RoboCasa, it corresponds to perturbing object placement and verifying that the manipulation strategy adapts. In Sotopia, it corresponds to changing one agent's opening move and checking whether the negotiation outcome shifts accordingly.

In current evaluation practice, intervention sensitivity receives markedly less attention than the other two boundary conditions. Most benchmarks measure output quality (success rate, perceptual fidelity) but do not explicitly test whether model predictions change appropriately with actions. Closing this gap requires evaluation protocols that explicitly vary actions and measure outcome divergence beyond output quality.

\subsection{Constraint Consistency}

Constraint consistency asks whether rollouts respect the governing laws $c(\tau)$ of the target regime (Section~\ref{subsec:l2_requirements}, condition~3). Because $c(\tau)$ depends on entire trajectory, violations are often invisible to per-step metrics but catastrophic for planning. This condition is also the primary surface for \textbf{exploitability} (failure mode~4)~\citep{xie2024osworld,zheng2023mcu} and \textbf{calibration failure under distribution shift} (failure mode~5).

Verification methods vary with the governing law. For physics, the key signals are penetration depth, energy conservation violation, and support-relation consistency; VBench~\citep{huang2023vbench} decomposes video generation quality into fine-grained physical compliance checks, while BuilderBench~\citep{ghugare2025builderbench} tests structural stability under physical load. Software environments foreground receipt match rate, type-constraint satisfaction, and API contract adherence, with OSWorld and macOSWorld~\citep{yang2025macosworld} providing loggable receipt streams. Social simulation raises norm violation detection rate, commitment consistency, and Theory of Mind accuracy, assessed through Sotopia's seven-dimensional framework~\citep{zhou2024sotopia} and ExploreToM's adversarial probing~\citep{sclar2024exploretom}. Scientific domains use conservation law satisfaction, causal graph consistency, and evidence-chain validity; DiscoveryBench~\citep{majumder2024discoverybench} and FutureX~\citep{zeng2025futurex} test evidence grounding. Mobile and cross-platform environments (AppAgent~\citep{zhang2025appagent}, AndroidWorld~\citep{rawles2024androidworld}) extend digital-world constraint testing, while AgentBench~\citep{liu2023agentbench} provides cross-domain breadth. ChemCrow~\citep{bran2024augmenting} evaluates chemical synthesis under strict validity constraints where a single violation renders the entire plan invalid.

\subsection{World-Model-Specific Evaluation}

WorldSimBench~\citep{qin2025worldsimbench} introduces a dual evaluation (perceptual quality and manipulative capability) for video generation models used as world simulators, revealing that perceptual realism and action-conditioned fidelity can diverge sharply. WorldModelBench~\citep{fan2025worldmodelbench} spans 7 domains, 56 subdomains, and 350 prompts with 67K human labels, finding widespread constraint consistency gaps across all 14 frontier models tested. \citet{vafa2024evaluating} apply the Myhill-Nerode theorem to show that passing per-step tests does not guarantee long-horizon coherence: models can maintain highly incoherent internal world states while performing well on standard diagnostics. \citet{kang2025howfar} find that video generation models perform ``case-based'' mimicry rather than learning generalizable physical principles, even at scale.

\subsection{Capability Coverage Matrix}

No single benchmark covers all three boundary conditions and four governing-law regimes; researchers should explicitly map their chosen evaluation suites to these axes to avoid over-claiming generalization.

\subsection{Minimal Reproducible Evaluation Package (MREP)}

The lack of standardization in agent evaluation leads to non-comparable results and ``leaderboard hacking''~\citep{henderson2018deep}. We propose the MREP as a community standard:

\begin{enumerate}
    \item \textbf{Version Locking:} Exact commit hashes for the environment and task set definition.
    \item \textbf{Trace Logs:} Full logs of intermediate steps, including observations, actions, and receipts, sufficient to replay and attribute failures post hoc.
    \item \textbf{Failure Taxonomy:} Automated classification aligned with our five L2 failure categories.
    \item \textbf{Tail Statistics:} Stratified bootstrap confidence intervals, IQM, and performance profiles rather than point estimates~\citep{agarwal2021statistical}.
    \item \textbf{Boundary Condition Mapping:} Explicit declaration of the boundary conditions tested.
\end{enumerate}

Several MREP components already have tooling support (trace logging via LangSmith/W\&B Weave; version locking via containerization; tail statistics per~\citealt{agarwal2021statistical}). The components requiring new infrastructure are automated failure taxonomy and boundary condition mapping. The MREP framework connects directly to L3 governed validation (Section~\ref{subsec:l3_definition}): the evaluation assets MREP requires are precisely what an L3 gatekeeper needs to decide whether a model update should be promoted or rolled back.

\section{Implementation Details and Efficient Deployment}
\label{app:impl_extended}

This appendix provides extended details on the practical implementation considerations and efficiency techniques for world-model systems that were summarized in Section~\ref{sec:implementation} of the main text.

\subsection{Practical Implementation Considerations}

\paragraph{Training paradigms: end-to-end vs.\ modular.}
Two dominant strategies exist. In \emph{end-to-end} training, the encoder, dynamics model, decoder, and policy are optimized jointly through a shared objective such as the ELBO or a value-aware loss. The Dreamer family~\citep{hafner2019dreamer,hafner2020dreamerv2,hafner2023dreamerv3} exemplifies this: all components share gradients, and the policy trains entirely on imagined latent rollouts. In \emph{modular} training, each component is trained with its own objective: representation via self-supervised learning~\citep{assran2023ijepa,bardes2024vjepa}, dynamics via maximum likelihood or TD objectives~\citep{hansen2024tdmpc2}, and policy via model-free RL or planning. Modular development frameworks such as StarVLA~\citep{starvla2025} provide composable codebases for systematic ablation and recombination. End-to-end avoids cross-module error propagation while unstable; modular systems are easier to debug but risk mismatched interfaces.

\paragraph{Latency, compute, and the O(1) argument.}
The choice of dynamics model is tightly constrained by the latency budget. Robotics control loops typically require sub-100\,ms inference, favoring lightweight latent dynamics with short-horizon rollouts~\citep{hansen2024tdmpc2,chua2018pets}. Web and OS agents tolerate seconds-scale latency, allowing richer search and receipt parsing. High-fidelity generative models~\citep{brooks2024sora} may take minutes per rollout, restricting them to offline planning or data augmentation.

A fundamental computational argument underlies the appeal of learned world models: a neural forward pass runs in $O(1)$ time with respect to the complexity of the simulated system, whereas explicit simulation scales as $O(N)$ or worse. This $O(1)$ property makes learned world models viable where analytic simulation is intractable. However, the constant factor matters: the $O(1)$ forward pass of a large diffusion model may still be slower than the $O(N)$ pass of a lightweight physics engine for modest $N$.

World-model scaling laws also differ from LLM scaling laws. Language models must memorize vast factual knowledge; world models primarily need to capture transition structure. This filtering-and-organizing role suggests a different compute-optimality regime where architectural inductive biases may substitute for raw parameter count more effectively than in language modeling.

\paragraph{Sim-to-real transfer.}
For embodied systems, the gap between the training simulator and deployment is a persistent bottleneck. Domain randomization~\citep{tobin2017domain} remains widely used. Complementary strategies include system identification, progressive transfer, and hybrid approaches combining learned residual dynamics with analytic physics models~\citep{nagabandi2018mbmf}. DayDreamer~\citep{wu2023daydreamer} demonstrated that Dreamer-style latent imagination can transfer to physical robots by training on real sensor data collected online, sidestepping the sim-to-real gap at the cost of slower data collection.

\paragraph{Error handling and graceful degradation.}
A mature world-model system must degrade gracefully when predictions become unreliable. MBPO~\citep{janner2019mbpo} limits rollout length to the accurate regime, falling back to real data for longer horizons. Ensemble disagreement~\citep{chua2018pets} provides a practical signal for triggering replanning or escalation. In software environments, receipt parsing serves an analogous role: unexpected error codes signal that assumptions may be violated~\citep{xie2024osworld}.

\subsection{Few-Step Distillation for Generative Dynamics}

Generative simulation systems driven by diffusion and flow-matching frameworks are inherently constrained by iterative denoising latency. Initial efforts relied on training-free, high-order ODE solvers~\citep{dockhorn2022genie, karras2022elucidating, lu2022dpm, lu2025dpm, sabour2024align, zhang2022fast, zheng2023dpm}, but these fall short of the low-step budgets demanded by real-time agents.

The field has pivoted toward few-step distillation. Early strategies focused on compressing teacher model trajectories by matching long-stride transitions~\citep{lipman2022flow, salimans2022progressive}. This paradigm evolved into Consistency Models~\citep{geng2024consistency, lu2024simplifying, song2023improved, song2023consistency}, which circumvent iterative sampling by learning a direct PF-ODE mapping from noise to clean data. Flow-map models further generalize this concept~\citep{boffi2024flow, frans2024one, heek2024multistep, kim2023consistency, wang2024phased}. Large-scale pre-training initiatives like TiM~\citep{wang2025transition} and MeanFlow~\citep{geng2025mean} have advanced this approach. Distribution-matching distillation~\citep{salimans2024multistep, sauer2024adversarial,sauer2024fast,yin2024improved,zhou2024score, yu2025self} has emerged as an alternative, aligning student output with teacher target distributions.

As world models increasingly rely on video generation, the higher sampling costs of video have catalyzed adaptation of acceleration techniques to the spatiotemporal domain~\citep{ding2025dollar, lin2025diffusion, zhang2024sf,zheng2025large, nie2026transition, yang2025longlive}. These efficiency gains transform generative models from offline simulators into viable engines for real-time planning.

\subsection{Model Compression: Quantization and Pruning}

Quantization maps high-precision weights into lower-bit formats (e.g., INT/FP-8, INT/FP-4), reducing memory bandwidth constraints. Pruning removes redundant parameters or structural blocks. For world modeling, the primary challenge is mitigating compounding errors: even minor quantization noise or pruning-induced degradation can lead to severe semantic drift over long horizons.

Foundational post-training quantization techniques~\citep{frantar2023gptq,lin2024awq,huang2026mcsharp,huang2024billm,dettmers2022gpt3int8,huang2024mixture} were designed for LLMs but are applicable to both LLM transformers and DiTs. SqueezeLLM~\citep{kim2024squeezellm} combines sensitivity-based non-uniform quantization with dense-and-sparse decomposition for ultra-low precision. On the serving side, efficient memory management is equally critical: vLLM~\citep{kwon2023vllm} introduced PagedAttention, which manages KV cache memory in non-contiguous blocks analogous to OS virtual memory, dramatically improving throughput for large-model inference. For diffusion models, QAT methods like QDM~\citep{li2024q} and TerDiT~\citep{lu2024terdit} maintain performance at 1-2 bit precision but require substantial training overhead. PTQ approaches for UNet-based diffusion models include QDiffusion~\citep{li2023q}, PTQ4DM~\citep{shang2023post}, and EfficientDM~\citep{he2023efficientdm}. For transformer backbones, Q-DiT~\citep{chen2025q}, PTQ4DiT~\citep{wu2024ptq4dit}, SVDQuant~\citep{li2024svdquant}, and ViDiTQ~\citep{zhao2024vidit} account for the unique activation distributions of diffusion transformers through attention-aware calibration.

Network pruning includes unstructured approaches~\citep{dong2017learning,park2020lookahead,sanh2020movement,lee2019signal} and structured pruning~\citep{ding2019centripetal,liu2021group}. Token merging~\citep{bolya2023token, bolya2022token} provides training-free alternatives. For world models, rollout-aware pruning is a promising frontier: pruning criteria must preserve the parameters that are critical for long-horizon coherence, aligning with the L2 boundary condition of temporal consistency defined in Section~\ref{subsec:l2_requirements}.

\subsection{Memory and KV Cache Compression}

Autoregressive token dynamics are severely memory-bound during long-horizon rollouts as the KV cache grows linearly. Key compression strategies include:
\begin{enumerate}[leftmargin=*,nosep]
  \item \textbf{Token eviction:} heavy-hitter retention~\citep{zhang2023h2o} and attention-sink preservation~\citep{xiao2024streamingllm} discard low-salience entries to bound cache size.
  \item \textbf{Chunk-level autoregressive generation:} modern video models generate in chunks~\citep{yin2025slow, huang2025selfforcing, feng2025streamdiffusionv2}, though hardware constraints often limit output to ${\sim}60$ seconds.
  \item \textbf{KV quantization:} schemes such as KIVI~\citep{liu2024kivi}, KVQuant~\citep{hooper2024kvquant}, QuaRot~\citep{ashkboos2024quarot}, and RotateKV~\citep{su2025rotatekv} are mature for LLM serving, but porting them to video diffusion causes severe quality loss due to different activation statistics.
  \item \textbf{Spatiotemporal-aware compression:} effective video KV compression requires frameworks that explicitly leverage video-specific spatiotemporal redundancy~\citep{yang2025sparse}.
\end{enumerate}

\end{document}